\documentclass{article}

\PassOptionsToPackage{numbers, compress}{natbib}

\usepackage[final]{neurips_2022}

\usepackage[utf8]{inputenc} % allow utf-8 input
\usepackage[T1]{fontenc}    % use 8-bit T1 fonts
\usepackage{hyperref}       % hyperlinks
\usepackage{url}            % simple URL typesetting
\usepackage{booktabs}       % professional-quality tables
\usepackage{amsfonts}       % blackboard math symbols
\usepackage{nicefrac}       % compact symbols for 1/2, etc.
\usepackage{microtype}      % microtypography
\usepackage{xcolor}         % colors

\usepackage{graphicx}
\usepackage{amsmath}
\usepackage{amsfonts}
\usepackage[capitalise]{cleveref}
\usepackage[english]{babel}
\usepackage[autostyle,english=british]{csquotes}
\usepackage{bm}

\usepackage{mathtools}

\usepackage{float}
\usepackage{capt-of}
\usepackage{calc}

\usepackage{subcaption}

\usepackage{multirow}
\usepackage{multicol}
\usepackage{array}

\hypersetup{
    colorlinks,
    linkcolor={blue},
    citecolor={blue},
    urlcolor={blue}
}

\usepackage{footmisc}
\usepackage{rotating}
\usepackage{siunitx}
\usepackage{makecell}

\Crefname{appendixfigure}{Figure}{Figures}
\Crefname{appendixtable}{Table}{Tables}
\Crefname{appendixname}{Appendix Information}{Appendix Information}
\Crefname{appendixsubfigure}{Figure}{Figures}

\DeclareCaptionLabelFormat{andtable}{#1~#2  \&  \tablename~\thetable}

\title{Graph Neural Networks with Adaptive Readouts}
\author{David Buterez~\thanks{Correspondence to: David Buterez <db804@cam.ac.uk>.}~~$^1$\\%Deptartment of Computer Science\\University of Cambridge, UK\\
\And
Jon Paul Janet~$^2$\\%CVRM, BioPharmaceuticals R\&D\\AstraZeneca, Sweden
\And
Steven J. Kiddle~$^3$\\%DS\&AI, BioPharmaceuticals R\&D\\AstraZeneca, UK
\And
Dino Oglic~$^3$\\%DS\&AI, BioPharmaceuticals R\&D\\AstraZeneca, UK\\
\And
Pietro Liò~$^1$\\%Department of Computer Science\\University of Cambridge, UK\\
\And~\\
\begin{minipage}{0.85\textwidth}\vspace{-5ex}
$^1$~Department of Computer Science and Technology, University of Cambridge, UK\\
$^2$~CVRM, BioPharmaceuticals R\&D, AstraZeneca, Sweden\\
$^3$~DS\&AI, BioPharmaceuticals R\&D, AstraZeneca, UK   
\end{minipage}
}
\begin{document}

\maketitle

%\vspace{-2ex}
\begin{abstract}
An effective aggregation of node features into a graph-level representation via readout functions is an essential step in numerous learning tasks involving graph neural networks. Typically, readouts are simple and non-adaptive functions designed such that the resulting hypothesis space is permutation invariant. Prior work on deep sets indicates that such readouts might require complex node embeddings that can be difficult to learn via standard neighborhood aggregation schemes. Motivated by this, we investigate the potential of adaptive readouts given by neural networks that do not necessarily give rise to permutation invariant hypothesis spaces. We argue that in some problems such as binding affinity prediction where molecules are typically presented in a canonical form it might be possible to relax the constraints on permutation invariance of the hypothesis space and learn a more effective model of the affinity by employing an adaptive readout function. Our empirical results demonstrate the effectiveness of neural readouts on more than 40 datasets spanning different domains and graph characteristics. Moreover, we observe a consistent improvement over standard readouts (i.e., sum, max, and mean) relative to the number of neighborhood aggregation iterations and different convolutional operators. %The source code is available at \href{https://github.com/davidbuterez/gnn-neural-readouts}{https://github.com/davidbuterez/gnn-neural-readouts}.
\end{abstract}

\section{Introduction}
\label{sec: intro}
We investigate empirically the potential of adaptive and differentiable readout functions for learning an effective representation of graph structured data (e.g., molecular, social, biological, and other relational data) using graph neural networks (\textsc{gnn}s). Recently, there has been a surge of interest in developing neural architectures from this class~\citep[][]{DBLP:journals/corr/KipfW16,HamiltonYL17,velickovic2018graph,DBLP:journals/corr/abs-2104-13478,DBLP:journals/corr/abs-1810-00826}. Graph neural networks typically employ a permutation invariant neighborhood aggregation scheme %(or a message passing algorithm) 
that is repeated for several iterations, where in each iteration node representations are updated by aggregating the feature vectors corresponding to their neighbors. The process is repeated for a pre-specified number of iterations and the resulting node representations capture the information contained in the respective neighborhoods given by vertex rooted sub-trees. The final step is aggregation of node features into a graph-level representation using a readout function. The readouts are typically selected such that the resulting hypothesis space is permutation invariant. For instance, simple functions such as sum, mean, and max, all satisfy this requirement~\cite{deepsets,Wagstaff19}. \textcolor{black}{Graph neural networks can, thus, be seen as a special case of representation learning over (node) sets.} Zaheer et al.~\cite{deepsets} have studied learning on sets and demonstrated that a permutation invariant hypothesis over such domains admits a decomposition as a sum of individual set items represented in a latent space given by a suitable embedding function. In a follow up work, Wagstaff et al.~\cite{Wagstaff19} have demonstrated that simple pooling/readout functions such as sum, mean, or max might require complex item/node embedding functions that might be difficult to learn using standard neural networks. The expressiveness of graph neural networks specifically has also been studied in~\cite{DBLP:journals/corr/abs-1810-00826}, where it has been recommended to use an injective neighborhood aggregation scheme. For such schemes, it can be demonstrated that graph neural networks can be as expressive as the Weisfeiler--Lehman isomorphism test which is known to be an effective and computationally efficient approximation scheme for differentiating between a large number of graph isomorphism classes~\cite{Laszlo,DBLP:journals/corr/abs-2112-09992}.

As it can be challenging to learn a permutation invariant hypothesis over graphs using simple readouts, we empirically investigate possible extensions and relaxations for problems where graphs might be presented in a canonical form (i.e., with an identical ordering of vertices). In such cases, it might be possible to relax the constraint on permutation invariance of the hypothesis space. For instance, in problems such as binding affinity prediction, molecular graphs are typically generated from a canonical \textsc{smiles} representation and, thus, inputs to graph neural networks are graphs with a fixed ordering of nodes. The latter is sufficient to ensure consistent predictions over molecular graphs for graph neural networks with readouts that do not give rise to permutation invariant hypotheses.

We start with a review of graph neural networks and then focus on introducing different classes of adaptive and differentiable readout functions. The first class of such readouts is based on set transformers~\cite{DBLP:journals/corr/abs-1810-00825} and it gives rise to permutation invariant hypotheses. We then slightly relax the constraint on permutation invariance of graph-level representations by introducing readouts inspired by a neighborhood aggregation scheme known as Janossy pooling~\cite{DBLP:journals/corr/abs-1811-01900}. These approximately permutation invariant readouts are based on multi-layer perceptrons (\textsc{mlp}) and recurrent neural architectures known as \textsc{gru}s. Finally, we consider neural readouts based on plain \textsc{mlp} and \textsc{gru} architectures, thus completely lifting the constraint on permutation invariance of the hypothesis space.

Our empirical study is extensive and covers more than 40 datasets across different domains and graph characteristics. The ultimate goal of the study is to explore the potential of learning hypotheses over graph structured data via adaptive and differentiable readouts. To this end, we first consider the most frequently used neighborhood aggregation schemes or convolutional operators and fix the number of iterations to two. Our empirical results demonstrate a significant improvement as a result of employing neural readouts, irrespective of the convolutional operator and dataset/domain. Following this, we then compare our neural readouts to the standard readouts (sum, max, mean) while varying the number of neighborhood aggregation iterations. The results indicate that neural readout functions are again more effective than the standard readouts, with a significant difference in performance between different neighborhood aggregation schemes. We hypothesize that the latter might be due to the expressiveness of different neighborhood aggregation operators. More specifically, in drug design and lead optimization it is typical that through a change in sub-structure of a parent compound that one can improve the potency. These changes are local and we hypothesize that they would be reflected in a small number of node representations, whose signal could be consumed by a large number of noisy nodes within the standard readouts. We also present results of an ablation study on the influence of node permutations on the hypotheses learned by \textsc{gnn}s with neural readouts that do not enforce permutation invariance. The results indicate that there might be some node permutations that are detrimental to the predictions and this might be an interesting avenue for future work.

We conclude with an experiment involving multi-million scale proprietary datasets from \textcolor{black}{AstraZeneca that have been collected by primary screening assays}. Our results again demonstrate that the avenue of neural readouts merits further exploration from both theoretical and empirical perspectives. \textcolor{black}{More specifically, we observe that only plain \textsc{mlp} readouts significantly improve the performance on these challenging tasks and they do not give rise to permutation invariant hypotheses.} Extensive results, including additional datasets and neural architectures (variational graph autoencoders and visualizations of latent spaces that correspond to different readouts) have been provided in \Cref{section:vgae-latent}. An analysis involving computational and memory costs and trade-offs can be found in \Cref{sec:time-mem}.

\section{Graph Neural Networks with Adaptive and Differentiable Readouts}
\label{sec:gnn-readouts}

Let $\mathcal{G} = (\mathcal{V}, \mathcal{E})$ be a graph, where $\mathcal{V}$ is the set of \textit{nodes} or \textit{vertices} and $\mathcal{E} \subseteq \mathcal{V} \times \mathcal{V}$ is the set of \textit{edges}. Suppose the nodes are associated with $d$-dimensional feature vectors $\mathbf{x}_u$ for all $u \in \mathcal{V}$. Let $A$ be the adjacency matrix of a graph $G$ such that $A_{uv} = 1$ if $(u,v) \in \mathcal{E}$ and $A_{uv}=0$ otherwise. For a vertex $u \in \mathcal{V}$ denote the set of neighboring nodes with $\mathcal{N}_u = \{v \mid (u,v) \in \mathcal{E} \lor (v,u) \in \mathcal{E}\}$. Suppose also that a set of graphs with corresponding labels $\{(G_i, y_i)\}_{i=1}^n$ has been sampled independently from some target probability measure defined over $\mathcal{G} \times \mathcal{Y}$, where $\mathcal{G}$ is a space of graphs and $\mathcal{Y} \subset \mathbb{R}$ is the set of labels. We are interested in the problem of learning a graph neural network that can approximate well the target label $y \in \mathcal{Y}$ for a given graph $G \in \mathcal{G}$.

Henceforth, we will assume that a graph $G$ is represented with a tuple $(X_G, A_G)$, with $X_G$ denoting the matrix with node features as rows and $A_G$ the adjacency matrix. Graph neural networks take such tuples as inputs and generate predictions over the label space. A function $f$ defined over a graph $G$ is called permutation invariant if there exists a permutation matrix $P$ such that $f(PX_G, PA_GP^{\top}) = f(X_G, A_G)$. In general, graph neural networks aim at learning permutation invariant hypotheses to have consistent predictions for the same graph when presented with permuted vertices/nodes. This property is achieved through neighborhood aggregation schemes and readouts that give rise to permutation invariant hypotheses. More specifically, the node features $X_G$ and the graph structure (adjacency matrix) $A_G$ are used to first learn representations of nodes $h_v$, for all $v \in \mathcal{V}$. The neighborhood aggregation schemes enforce permutation invariance by employing standard pooling functions --- sum, mean, or max. This step is followed by a readout function that aggregates the node features $h_v$ into a graph representation $h_G$. As succinctly described in~\cite{DBLP:journals/corr/abs-1810-00826}, typical neighborhood aggregation schemes characteristic of graph neural networks can be described by two steps:
\begin{align}
a_v^{(k)}=\textsc{aggregate}(\{h_u^{(k-1)} \mid u \in \mathcal{N}_v\}) \quad \text{and} \quad h_v^{(k)}=\textsc{combine}(h_v^{(k-1)},a_v^{(k-1)})
\end{align}
where $h_u^{(k)}$ is a representation of node $u\in \mathcal{V}$ at the output of the $k^{\text{th}}$ iteration. For example, in graph convolutional networks the two steps are realized via mean pooling and a linear transformation~\cite{DBLP:journals/corr/KipfW16}:
\begin{align*}
	h_v^{(k)} = \sigma \left( \frac{1}{|\mathcal{N}_v^*|} \sum_{u \in \mathcal{N}_v^*} W^{(k)} h_{u}^{(k-1)} \right) \quad \text{with} \quad \mathcal{N}_v^* = \mathcal{N}_v \cup \{v\}
\end{align*}
where $\sigma$ is an activation function and $W^{(k)}$ is a weight matrix for the $k^{\text{th}}$ iteration/layer.

After $k$ iterations the representation of a node captures the information contained in its $k$-hop neighborhood ~\cite[e.g., see the illustration of a vertex rooted sub-tree in][Figure 1]{DBLP:journals/corr/abs-1810-00826}. The node features at the output of the last iteration are aggregated into a graph-level representation using a \emph{readout} function. To enforce permutation invariant hypotheses, it is common to employ the standard pooling functions as readouts --- sum, mean, or max. In the next section, we consider possible extensions that would allow for learning readout functions jointly with other parameters of graph neural networks.

\subsection{Neural Readouts}
\label{subsec:adaptive-readouts}

Suppose that after completing a pre-specified number of neighborhood aggregation iterations, the resulting node features are collected into a matrix $H \in \mathbb{R}^{M \times D}$, where $M$ is the maximal number of nodes that a graph can have in the dataset and $D$ is the dimension of the output node embedding. For graphs with less than $ M$ vertices the padded values in $H$ are set to zero. We also denote with a vector $h \in \mathbb{R}^{M \cdot D}$ the flattened (i.e., concatenated rows) version of the node feature matrix $H$.

\textbf{Set Transformer Readouts.} Recently, an attention-based neural architecture for learning on sets has been proposed in~\cite[][]{DBLP:journals/corr/abs-1810-00825}. The main difference compared to the classical attention model proposed by Vaswani et al.~\cite{DBLP:journals/corr/VaswaniSPUJGKP17} is the absence of positional encoding and dropout layers. The approach can be motivated by the desire to exploit dependencies between set items when learning permutation invariant hypotheses on that domain. More specifically, other approaches within the deep sets framework typically embed set items independently into a latent space and then generate a permutation invariant hypothesis by standard pooling operators (sum, max, or mean). As graphs can be seen as sets of nodes, we propose to exploit this architecture as a readout function in graph neural networks. For the sake of brevity, classical attention models are described in~\Cref{sec:attention-recapitulation} and here we summarize the adaptation to sets. The set transformers take as input matrices with items/nodes as rows and generate graph representations by composing attention-based encoder and decoder modules:
\begin{align}
	\textsc{st}(H) = \frac{1}{K} \sum_{k=1}^K \left[ \textsc{decoder} \left( \textsc{encoder} \left( H \right) \right) \right]_k
\end{align}
where $\left[\cdot\right]_k$ refers to computation specific to head $k$. The encoder-decoder modules are given by~\cite[][]{DBLP:journals/corr/abs-1810-00825}:\vspace{0.25cm}

\begingroup
\makeatletter
\@fleqntrue
\makeatother
\addtolength{\jot}{0.2em}
$
\begin{aligned}
& \textsc{encoder}\left(H\right) \coloneqq \textsc{mab}^n \left(H, H\right) \quad \ \text{and} \quad \textsc{decoder}(Z) \coloneqq  \textsc{ff}\left( \textsc{mab}^m \left( \textsc{pma}(Z), \textsc{pma}(Z) \right) \right) & \\
& \text{where} \quad \textsc{pma}(Z) \coloneqq  \textsc{mab}(s, \textsc{ff}(Z)) \quad \text{and} \quad \textsc{mab}(X, Y)  \coloneqq  A + \textsc{ff}(A) &
\end{aligned}
\vspace{0.2em}\\\\
\begin{aligned}
& \text{with} \quad A \coloneqq X + \textsc{multi-head}(X, Y, Y) \ . \vspace{0.5em}
\end{aligned}
$
\raisetag{12.5pt}
\endgroup

Here, $H$ denotes the node features after neighborhood aggregation and $Z$ is the encoder output. The encoder is a chain of $n$ classical multi-head attention blocks (\textsc{mab}) without positional encoding and dropouts. The decoder component employs a seed vector $s$ within a multi-head attention block to create an initial readout vector that is further processed via a chain of $m$ self-attention modules and a feedforward projection block (\textsc{ff}).

\textbf{Janossy Readouts.} Janossy pooling was proposed in~\cite{DBLP:journals/corr/abs-1811-01900} with the goal of providing means for learning flexible permutation invariant hypotheses that in their core employ classical neural architectures such as recurrent and/or convolutional neural networks. The main idea is to process each permutation of set elements with such an architecture and then average the resulting latent representations. Additionally, one could also add a further block of feedforward or recurrent layers to process the permutation invariant latent embedding of a set. Motivated by this pooling function initially designed for node aggregation, we design a readout that is approximately permutation invariant. More specifically, we consider \textsc{mlp} and \textsc{gru} as base architectures and sample $p$ permutations of graph nodes. The Janossy readout then averages the latent representations of permuted graphs as follows:
\begin{align}
	\textsc{janossy-mlp}(H) \coloneqq \frac{1}{p}\sum_{i=1}^{p}\textsc{mlp}(h_{\pi_i}) \quad \text{or} \quad
	\textsc{janossy-gru}(H) \coloneqq \frac{1}{p}\sum_{i=1}^{p}\textsc{gru}(H_{\pi_i}),
\end{align}
where $\pi_i$ is a permutation of graph nodes and $h_{\pi_i}$ is the permuted and then flattened matrix $H$.

\textbf{Plain Feedforward/Recurrent Readouts.} Having proposed (approximate) permutation invariant readouts, we consider standard feedforward and recurrent neural architectures as well.  Our \textsc{mlp} neural readout consists of a two-layer fully connected neural network (i.e., multi-layer perceptron) applied to the flattened node feature matrix $H$ denoted with $h$:
\begin{align}
	\textsc{mlp}(H) \coloneqq \textsc{relu}(\textsc{bn}_2(W_2z_1 + b_2)) \quad \text{with} \quad	z_1 = \textsc{relu}(\textsc{bn}_1(W_1h + b_1))
\end{align}
where $W_1 \in \mathbb{R}^{(M \cdot D) \times d_1}$, $b_1 \in \mathbb{R}^{ d_1}$, $z_1$ is the output of the first layer, $W_2 \in \mathbb{R}^{d_1 \times d_{\text{out}}}$, $b_2 \in \mathbb{R}^{d_{\text{out}}}$, $d_1$ and $d_{\text{out}}$ are hyperparameters, $\textsc{bn}_i$ is a batch normalization layer, and \textsc{relu} is the rectified linear unit. In our experiments, we also apply Bernoulli dropout with rate $p=0.4$ as the last operation within \textsc{mlp}. The \textsc{gru} neural readout is composed of a single-layer, unidirectional gated recurrent unit (\textsc{gru}, \cite{DBLP:journals/corr/ChoMBB14}), taking sequences with shape $(M, D)$. We accept the input order on graph nodes as the order within the sequence that is passed to a \textsc{gru} module. This recurrent module outputs a sequence with shape $(M, d_\text{out})$, as well as a tensor of hidden states. The graph-level representation created by this readout is given by the last element of the output sequence. %We refer to CITE for a full description of \textsc{gru}s.

\textcolor{black}{In contrast to typical set-based neural architectures that process individual items in isolation (e.g., deep sets), the presented adaptive readouts account for interactions between all the node representations generated by the neighborhood aggregation scheme. In addition, the dimension of the graph-level representation can now be disentangled from the node output dimension and the aggregation scheme.}

\section{Experiments}
\label{sec:exps}

\setcounter{footnote}{0}
We perform a series of experiments\footnote{The source code is available at \href{https://github.com/davidbuterez/gnn-neural-readouts}{https://github.com/davidbuterez/gnn-neural-readouts}.} to evaluate the effectiveness of the adaptive and differentiable neural readouts presented in Section~\ref{subsec:adaptive-readouts} relative to the standard pooling functions (i.e., sum, max, mean) used for the aggregation of node features into a graph-level representation. \textcolor{black}{In our experiments, we rely on two performance metrics: $\text{R}^2$ for regression tasks and Matthews correlation coefficient (MCC) for classifications tasks. As outlined in prior work~\cite{Chicco2020,Chicco2021}, these metrics can be better at quantifying performance improvements than typical measures of effectiveness such as mean absolute/squared error, accuracy, and $F_1$ score.}
To showcase the potential for learning effective representation models over graph structured data, we use in excess of $40$ datasets originating from different domains such as quantum mechanics, biophysics, bioinformatics, computer vision, social networks, synthetic graphs, and function call graphs.
We focus on quantifying the difference in performance between the readouts relative to various factors such as: \emph{i}) most frequently used neighborhood aggregations schemes, \emph{ii}) the number of neighborhood aggregation iterations that correspond to layers in graph neural networks, \emph{iii}) convergence rates measured in the number of epochs required for training an effective model, \emph{iv}) different graph characteristics and domains from which the structured data originates, and \textcolor{black}{\emph{v)} the parameter budget employed by each of the neural readouts}. Moreover, we also evaluate a variety of neighborhood aggregation and readout schemes using large scale proprietary datasets with millions of compounds collected by primary screening assays. This is one of the first studies of a kind that demonstrates the behavior of standard graph neural networks on large scale datasets and illustrates some of the potential shortcomings that might be obfuscated by the small scale benchmarks that are typically used for evaluating their effectiveness in drug design tasks. \textcolor{black}{In all of our experiments, we have used a node output dimension of $50$, which gives rise to a graph embedding dimension of that size for the sum, mean, and max readouts. For the adaptive readouts, the dimensionality of the graph representation is a hyperparameter of the model, which was fixed to $64$}. For the sake of brevity, we present only a part of our analysis in this section and refer to the appendix for more detailed description of experiments and additional plots. 

\begin{figure}[t]
    \centering
    \captionsetup{labelfont=bf, font=small, skip=8pt}
    \caption{The performance of the best neural relative to the best standard readout on a collection of representative datasets from different domains. We use the ratio between the effectiveness scores ($\text{R}^2$ for \textsc{qm9} and Matthew correlation coefficient otherwise), computed by averaging over five random splits of the data.
    }\vspace{-2ex}
    \label{figure:summary-performance}
    \includegraphics[width=0.65\textwidth]{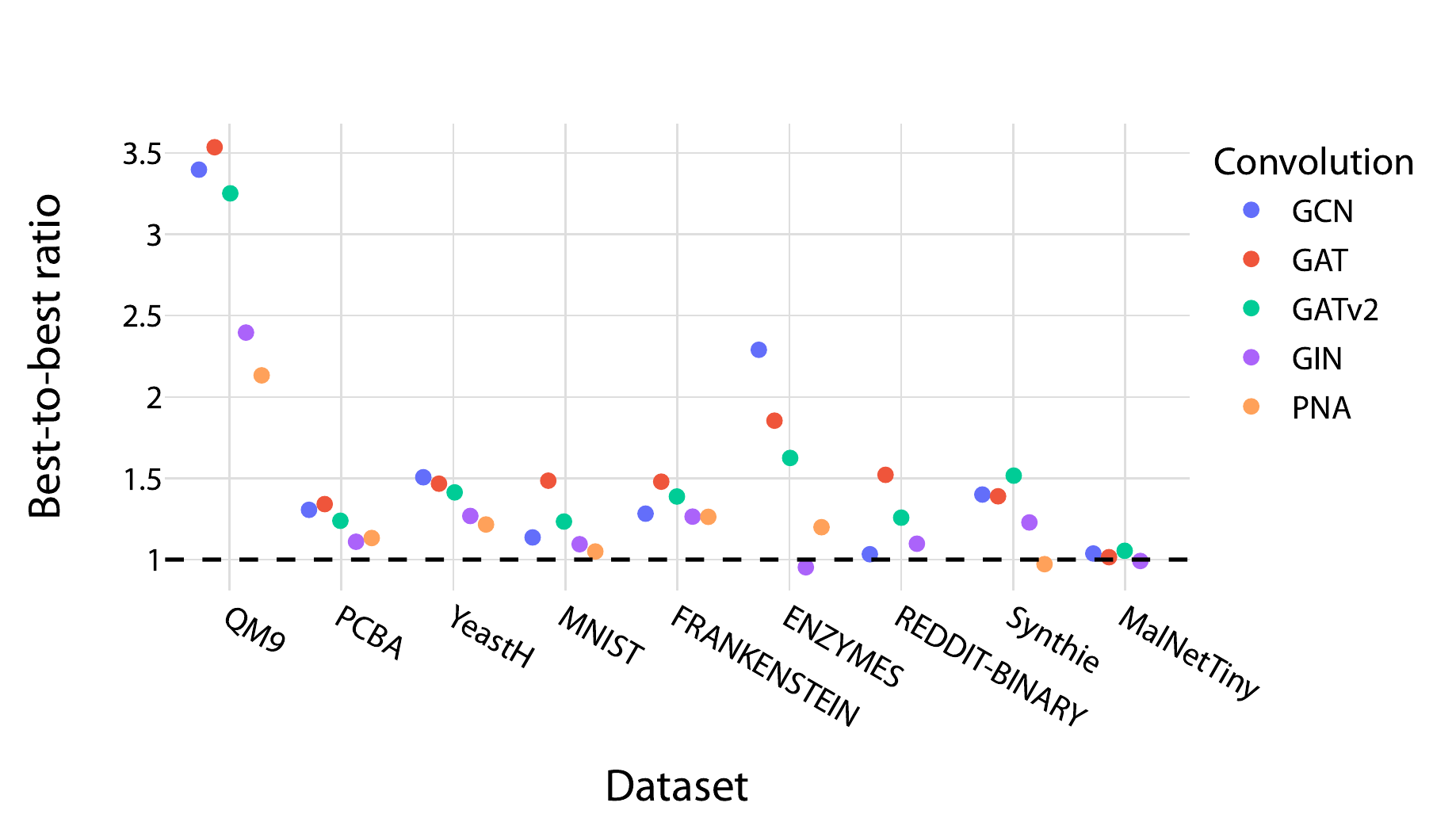}
        \vspace{-4ex}
\end{figure}

\subsubsection*{Neural vs standard readouts across different datasets and neighborhood aggregation schemes}
\label{subsec:benchmarks}
\vspace{-1.5ex}
The goal of this experiment is to evaluate the effectiveness of neural readouts relative to different datasets ($39$ in total, \Cref{section:datasets-summary}), graph characteristics, and neighborhood aggregation schemes or convolutional operators. To this end, we opt for graph neural networks with two layers and compare neural to standard readouts across different graph convolutional operators: \textsc{gcn}~\cite{DBLP:journals/corr/KipfW16}, \textsc{gat}~\cite{velickovic2018graph}, \textsc{gatv2}~\cite{DBLP:journals/corr/abs-2105-14491}, \textsc{gin}~\cite{DBLP:journals/corr/abs-1810-00826}, and \textsc{pna}~\cite{DBLP:journals/corr/abs-2004-05718}. A detailed experimental setup (including loss functions, reporting metrics, and other details relevant for reproducibility) has been provided in~\Cref{apx:sec:exp-design,section:nn-agg-hyperparams}.

\Cref{figure:summary-performance} summarizes the result of this experiment over 9 representative datasets (please see \Cref{section:ratios-all-datasets}, \Cref{figure:best-to-best-regr,figure:best-to-best-cls-1,figure:best-to-best-cls-2} for the results on the remaining $30$ datasets, including other metrics). The figure depicts the ratio between the best neural and best standard readouts for each evaluated configuration (i.e., pair of dataset, convolution). We observe a considerable uplift in the performance on the majority of datasets. More specifically, in regression tasks we measure the performance using the $\text{R}^2$ score and on $36$ configurations out of the possible $45$ (i.e., $80\%$ of time) there is an improvement (with $\text{ratio} > 1$) as a result of employing neural readouts. In datasets where standard readouts fare better than the neural ones, the relative degradation is minor (i.e., less than $5\%$).
In classification tasks, we use the \textsc{mcc} to measure the performance of graph neural networks and again observe an improvement as a result of employing neural readouts on $93$ out of $147$ (dataset, convolution) configurations ($\approx 63\%$ of time). We also observe that on $54$ configurations where standard readouts are more effective than the neural ones that the relative degradation is below $10\%$ relative. We note that three models also failed to complete due to memory issues when using \textsc{pna}, leading to a total of 147 configurations for the classification tasks (\Cref{section:all-benchmarks-metrics}). The minimum observed ratio between neural and standard readouts was $0.79$.

\begin{figure}
    \captionsetup{labelfont=bf, font=small}
    \caption{\textcolor{black}{The panel on the left illustrates the parameter budget of \textsc{gnn}s that does not account for the readouts, while varying the layer type and depth (\textsc{qm9} dataset). The number of parameters for the \textsc{mlp} readout is represented using dashed lines parallel to the $x$-axis (slightly higher when using \textsc{pna} as the output node dimension must be divisible by the tower hyperparameter). The panel on the right compares the performance of graph neural networks with set transformers (\textsc{st}) and plain \textsc{mlp}s as readouts, while varying the number of parameters in the readout layer. This illustrative example has been obtained using the \textsc{enzymes} dataset and demonstrates that the effectiveness is not aligned with the number of trainable parameters in the readout layer but the type of architecture realizing it. For example, the \textsc{\textsc{st} 1--\textsc{mab}} model employs a single \textsc{mab}-block encoder and decoder, being simpler in terms of the number of parameters than \textsc{st complex} and more complex than \textsc{st minimal} (\Cref{sec:st-architectures}). The \textsc{mlp} configurations are reported using the format \textsc{mlp} ($d_1$, $d_{\text{out}}$) (\Cref{section:nn-agg-hyperparams}).}}
    \label{fig:par-budget}
    \begin{minipage}[t]{0.5\textwidth}\vspace{0pt}
      \centering
      \includegraphics[width=0.9\textwidth]{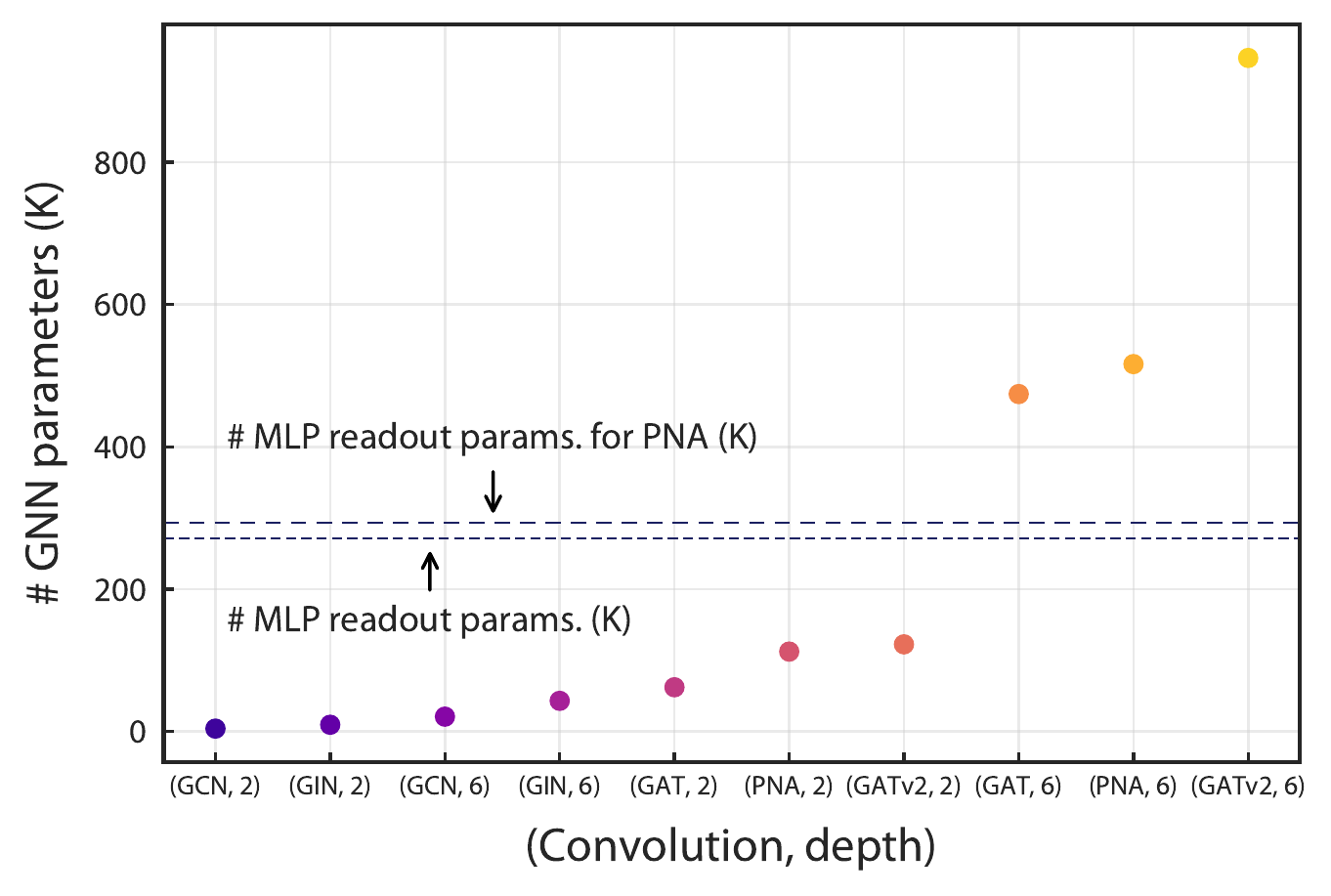}
    \end{minipage}\hfill
    \begin{minipage}[t]{0.5\textwidth}\vspace{3.3pt}
      \centering
        \begingroup
            \renewcommand{\arraystretch}{1.05}
            \small
            \begin{tabular}{@{}lcS[table-format=2.2]@{}}
            \toprule
            \textbf{Readout} & {\textbf{\# Params.}} & {\textbf{Avg. MCC}} \\ \midrule
            \textsc{st minimal} & 365K & 0.352892 \\
            \textsc{st} 1--\textsc{mab} & 628K & 0.338341 \\
            \textsc{st complex} & 1154K & 0.360129 \\ \midrule
            \textsc{mlp} (32, 32) & 130K & 0.406878 \\
            \textsc{mlp} (64, 32) & 260K & 0.436712 \\
            \textsc{mlp} (128, 64) & 525K & 0.403322 \\
            \textsc{mlp} (64, 128) & 267K & 0.472538 \\ \bottomrule
            \end{tabular}
        \endgroup
        \captionsetup{skip=15pt}
    \end{minipage}\vspace{-3ex}
\end{figure}
\setcounter{figure}{2}

It is also worth noting that neural readouts come with hyperparameters that were not tuned/cross-validated in our experiments. \textcolor{black}{For example, set transformer readouts can be configured by specifying the number of attention heads and latent/hidden dimensions. We have, throughout our experiments, followed standard practices and selected such hyperparameters to be powers of two, tailored to the dataset size (more details can be found in \Cref{section:nn-agg-hyperparams}, \Cref{table:neural-agg-hyperparams}). This was also, in part, motivated by previous knowledge from bio-affinity prediction tasks with graph neural networks}. Thus, it is likely that the performance can be further improved by hyperparameter tuning.
\textcolor{black}{We also emphasize that for this experiment the architecture is kept fixed across datasets (number of graph layers, hidden dimensions, etc.), while only varying the graph layer type and readout layer. In addition, over-smoothing is a known problem for vanilla graph neural networks. To avoid any hidden contribution from over-smoothing correction techniques, we opted for shallower architectures. We also note that two-layer graph neural networks performed well on tasks such as bio-affinity prediction on the $1+$ million scale datasets (i.e., can be sufficiently expressive). To validate that additional expressiveness due to more layers can be successfully exploited by adaptive readouts, we performed a separate suite of experiments where we varied the depth (see the next experiment/section).} 
\textcolor{black}{We conclude the discussion of this experiment with an insight into the trainable parameter budgets for a selection of neural readouts. For \textsc{mlp}, with the exception of \textsc{gcn} and \textsc{gin} which use an extremely small number of trainable weights, the parameter budget/count is on the same scale as the rest of the graph neural networks (see the left panel in Figure~\ref{fig:par-budget}). Furthermore, simply increasing the number of parameters does not necessarily improve the performance (see the right panel in Figure~\ref{fig:par-budget}, \textsc{st} vs \textsc{mlp} rows}).

\subsubsection*{Neural vs standard readouts relative to the number of neighborhood aggregations iterations}
\begin{figure}[!tbp]
    \centering
    \captionsetup{labelfont=bf, font=small}
    \caption{Increasing the number of neighborhood aggregation iterations or convolutional layers has different effects on \textsc{qm9}. The trends observed for the standard readouts are mirrored for the neural ones, particularly for the most powerful one on this dataset (\textsc{set transformer}). For \textsc{gcn}, \textsc{gat}, and \textsc{gatv2}, the performance improves as the depth is increased to $6$ layers and drops afterwards. \textsc{gin} is generally stable relative to the number of layers, while \textsc{pna} has an initial performance improvement (up to $3$, $4$ layers) and then plateaus.}
    \label{figure:qm9-7-layers}
        \includegraphics[width=1\textwidth]{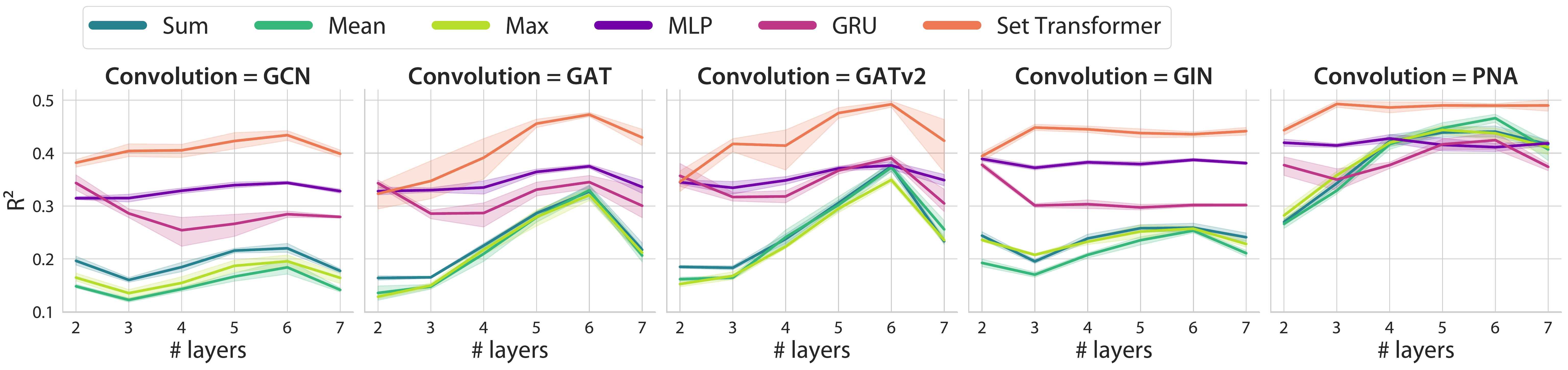}\vspace{-2ex}
\end{figure}

\vspace{-1.5ex}
The goal of this experiment is two-fold: \emph{i}) assessment of the effectiveness of neural readouts relative to the depth of graph neural networks, and \emph{ii}) validation of the observations in the experiments with two layer graph neural networks, i.e., the improvements characteristic of neural readouts are not due to underfitting that can be caused by employing a small number of neighborhood aggregation iterations along with standard readouts. We perform the experiment with various graph convolutional operators using datasets with different number of instances, ranging from $600$ to $132,480$ graphs on \textsc{enzymes} and \textsc{qm9}, respectively. \textcolor{black}{In these experiments, the only variable graph neural network hyperparameter is the number of neighborhood aggregation iterations (i.e,. depth)}. We have also performed this experiment on one of the proprietary bio-affinity datasets with $1.5$ million graphs (see \Cref{section:train-loss-deeper}, \Cref{figure:1.5mil-GNN-VGAE-deeper}). \Cref{figure:qm9-7-layers} summarizes the results for the \textsc{qm9} experiment (see \Cref{section:deep-enzymes}, \Cref{figure:enzymes-6-layers} for \textsc{enzymes}). The trends seen for standard readouts are mirrored for the neural ones as the number of neighborhood iterations increases, i.e., deeper graph neural networks can lead to more expressive models, irrespective of the readout type.

% \vspace{-2ex}
\begin{table}[!tbp]
    \captionsetup{justification=centering, skip=4pt, labelfont=bf, width=1.03\linewidth, font=small}
    \centering
    \small
    \caption{$\text{R}^2$ (mean $\pm$ standard deviation) on \textsc{qm9}, using the \textsc{st minimal} and \textsc{st complex} architectures.}
    \label{table:st_heads_qm9}
    \begin{tabular}{rcccccc}
    \toprule
    \multicolumn{1}{l}{\textbf{Aggregator}} & \multicolumn{1}{l}{\textbf{\# Heads}} & \textbf{GCN}         & \textbf{GAT}         & \textbf{GATv2}       & \textbf{GIN}         & \textbf{PNA}          \\ \toprule
    \multirow{4}{*}{\textsc{st minimal}}    & 1                            & 0.20 $\pm$ 0.02 & 0.17 $\pm$ 0.01 & 0.20 $\pm$ 0.01 & 0.22 $\pm$ 0.01 & 0.27 $\pm$ 0.01  \\
                                   & 4                            & 0.27 $\pm$ 0.01 & 0.23 $\pm$ 0.02 & 0.27 $\pm$ 0.02 & 0.32 $\pm$ 0.01 & 0.38 $\pm$ 0.02  \\
                                   & 8                            & 0.35 $\pm$ 0.01 & 0.24 $\pm$ 0.04 & 0.29 $\pm$ 0.01 & 0.35 $\pm$ 0.01 & 0.43 $\pm$ 0.01  \\
                                   & 12                           & 0.37 $\pm$ 0.01 & 0.26 $\pm$ 0.03 & 0.27 $\pm$ 0.03 & 0.37 $\pm$ 0.01 & 0.43 $\pm$ 0.01  \\ \cmidrule(){1-7}
    \multirow{4}{*}{\textsc{st complex}}    & 1                            & 0.18 $\pm$ 0.02 & 0.16 $\pm$ 0.02 & 0.20 $\pm$ 0.01 & 0.24 $\pm$ 0.01 & 0.28 $\pm$ 0.01  \\
                                   & 4                            & 0.35 $\pm$ 0.01 & 0.27 $\pm$ 0.02 & 0.29 $\pm$ 0.02 & 0.35 $\pm$ 0.00 & 0.42 $\pm$ 0.01  \\
                                   & 8                            & 0.38 $\pm$ 0.01 & 0.30 $\pm$ 0.04 & 0.32 $\pm$ 0.02 & 0.39 $\pm$ 0.01 & 0.44 $\pm$ 0.01  \\
                                   & 12                           & 0.37 $\pm$ 0.02 & 0.32 $\pm$ 0.02 & 0.33 $\pm$ 0.02 & 0.40 $\pm$ 0.01 & 0.45 $\pm$ 0.01 \\ \bottomrule
    \end{tabular}\vspace{-2ex}
\end{table}

\subsubsection*{Convergence of training algorithms for graph neural networks relative to readouts}
\vspace{-1.5ex}
As outlined in Section~\ref{sec: intro}, it might be challenging to learn a permutation invariant hypothesis over graphs using simple and non-adaptive readout functions such as sum, mean, or max. Here, we argue that such functions might be creating a tight link between the resulting graph-level representations and the computed node features, in the sense that: \emph{i}) it takes a long time for the graph-level representation to adjust to the prediction task and \emph{ii}) it is difficult for the graph-level representation to diverge from the learned node features and adapt to the target property. To validate this hypothesis, we recorded the graph representations for a random molecule from the \textsc{qm9} dataset in each training epoch (for multiple convolutions and readouts). 
We computed the Euclidean distances between the initial graph embedding (i.e., the first epoch) and all subsequent epochs (\Cref{section:graph-emb-distance}, \Cref{figure:qm9-dist-initial}), as well as between consecutive epochs (\Cref{section:graph-emb-distance}, \Cref{figure:qm9-dist-e-by-e}). Our empirical results indicate that \textsc{gnn}s with standard readouts take hundreds of epochs to converge ($500$ to $1,000$), with minor changes in the graph representation from one epoch to another. In contrast to this, the models employing neural readouts converge quickly, typically in under $100$ epochs. Moreover, the learned representations can span a larger volume, as shown by the initial and converged representations, which are separated by distances that are orders of magnitude larger than for the standard readouts.

\subsubsection*{The importance of (approximate) permutation invariance in adaptive readouts}
\vspace{-1.5ex}
The goal of this experiment is to obtain an insight into the effects of adaptive readouts that give rise to permutation invariant hypothesis spaces on the network's ability to learn a target concept. \textcolor{black}{To this end, we exploit the modularity of \textsc{set transformers} and consider architectures with $1$ to $12$ attention heads, as well as a different number of attention blocks: an \textsc{st minimal} model with one \textsc{mab} in the encoder and no \textsc{mab}s in the decoder, and \textsc{st complex} with two \textsc{mab}s in both the encoder and decoder (for 2-layer \textsc{gnn}s)}.
\Cref{table:st_heads_qm9} provides a summary of the results for this type of readouts on the \textsc{qm9} dataset (detailed results can be found in the appendix). 
\textcolor{black}{With a small number of attention heads ($1$ and $4$), all models with \textsc{set transformer} readouts are able to outperform the ones with standard pooling functions. % --- sum, max, and mean. 
However, over different convolutions the models with few attention heads are outperformed by the ones with \textsc{mlp} and \textsc{gru} readouts that do not enforce permutation invariance. Increasing the number of heads to $8$ or $12$ leads to the best performance on this dataset for all graph convolutions. However, the relative improvement gained by increasing the number of attention heads beyond $4$ is generally minor, as is the uplift gained by adding a self-attention block}. We also evaluated the impact of enforcing approximate permutation invariance by Janossy readouts on the \textsc{qm9} dataset (\Cref{section:all-benchmarks-metrics}, \Cref{table:QM9-Janossy}). The Janossy variants presented in Section~\ref{subsec:adaptive-readouts} outperform the three standard readouts in both mean absolute error and $\text{R}^2$, but they score lower than the other neural readouts.

\begin{figure}[t]
    \centering
    \captionsetup{labelfont=bf, font=small}
    \caption{A summary of the error distributions for predictions made on random permutations of $50$ randomly selected molecules from the \textsc{qm9} dataset. The error is computed as the absolute difference between the predicted and target labels. The models are fully trained, two-layer graph neural networks. Due to permutation invariance, the predictions made for a given molecule are identical for sum/mean/max, regardless of the permutation. The variance for these readouts is a result of the differences in the predicted labels for the considered $50$ molecules. Panels \textbf{(a)} and \textbf{(b)} reflect different strategies of generating node permutations.}
    \label{figure:random-perm}
    \begin{subfigure}[b]{0.49\textwidth}
        \centering
        \captionsetup{labelfont=bf, font=small, skip=-1pt}
        \caption{For each molecule, we generate $50$ different graphs using random permutations of the nodes originating from the canonical \textsc{smiles} representation.} \label{subfigure:random-perm-nodes}
        \includegraphics[width=1\textwidth]{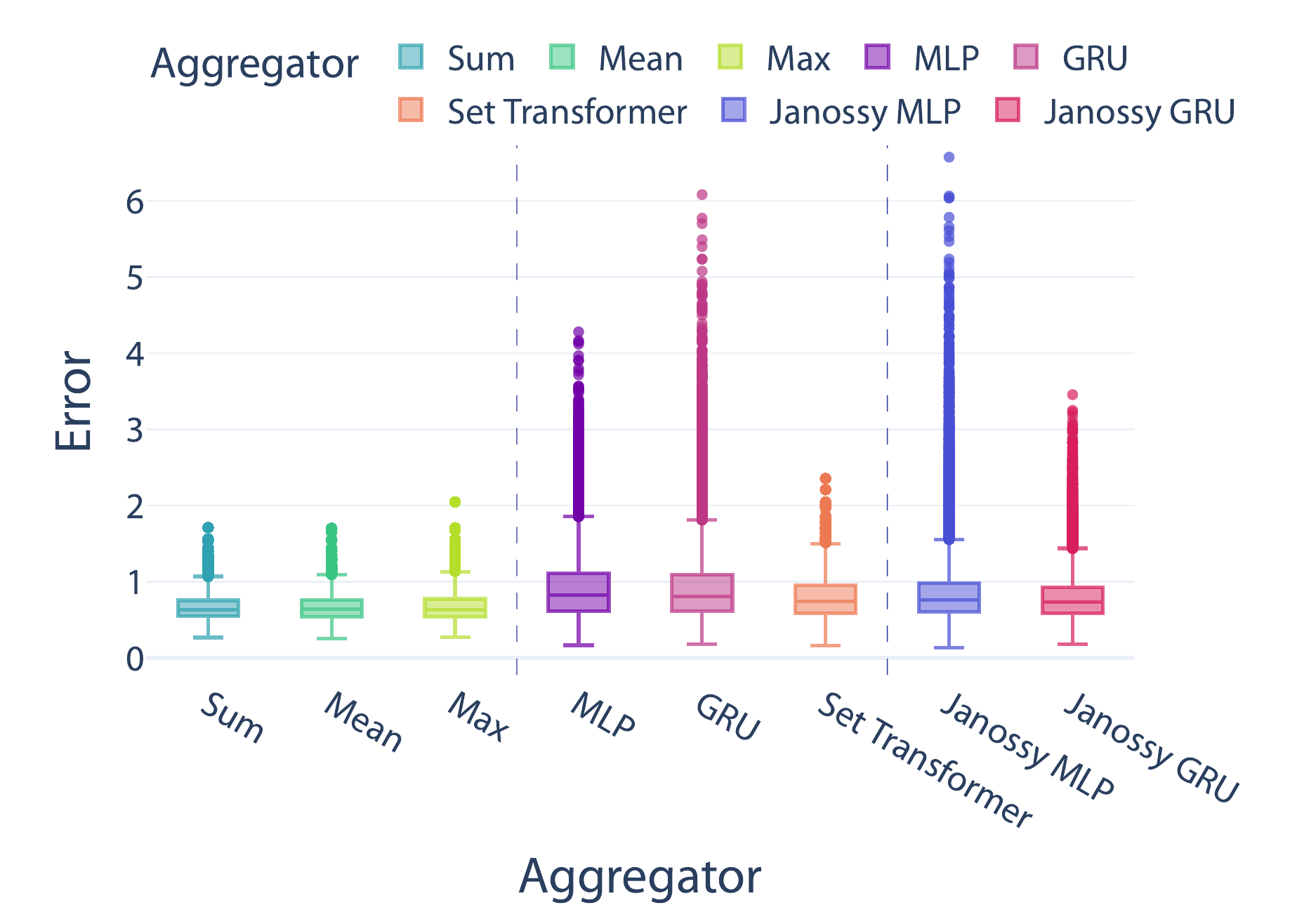}
    \end{subfigure}\vspace{-3ex}
    \hfill
    \begin{subfigure}[b]{0.49\textwidth}
        \centering
        \captionsetup{labelfont=bf, font=small, skip=-1pt}
        \caption{For each molecule, we generate graphs corresponding to different non-canonical \textsc{smiles}. The number varies from $20$ to $1,000$ (per-molecule).} \label{subfigure:random-perm-smiles}
        \includegraphics[width=1\textwidth]{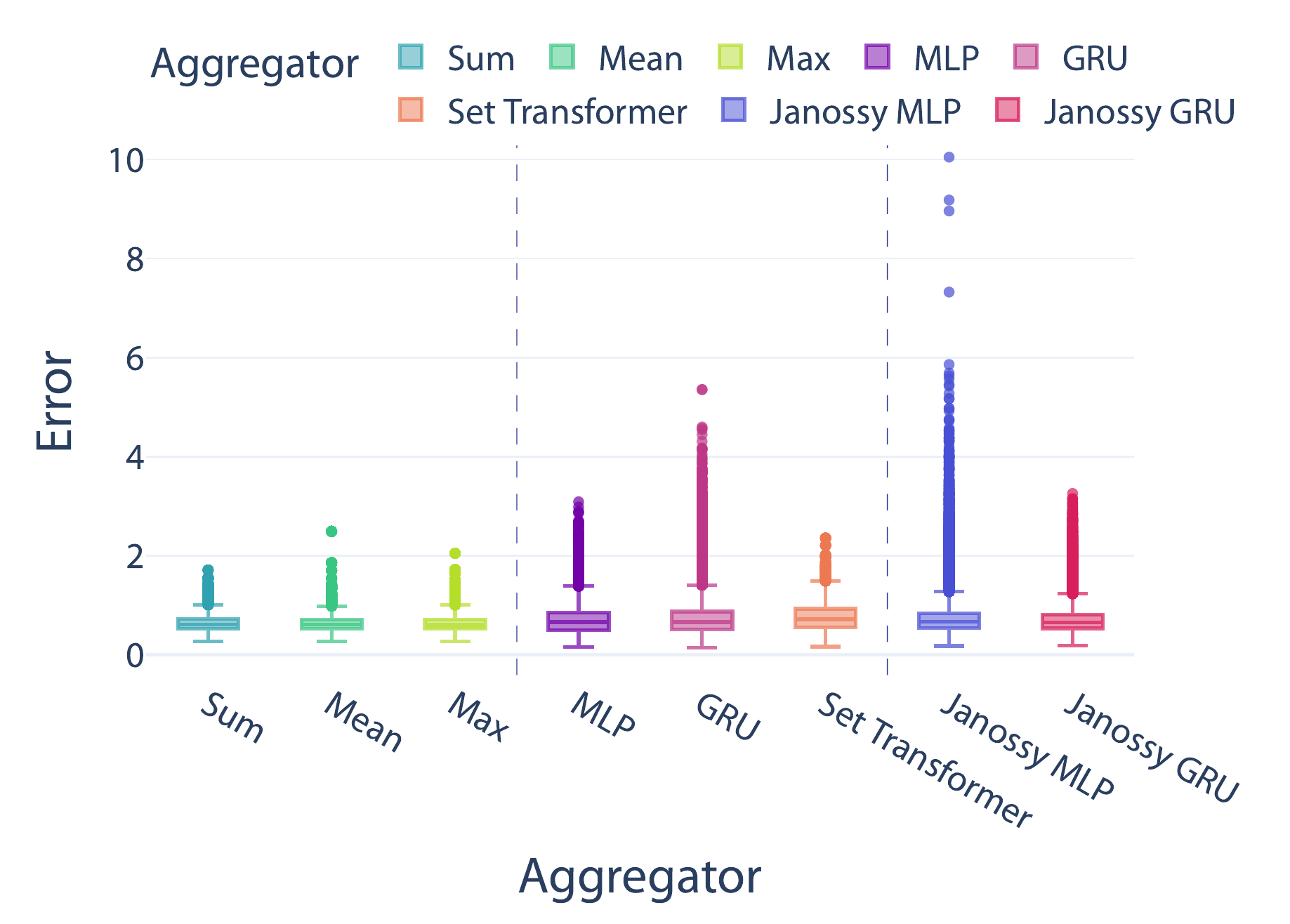}
    \end{subfigure}
\end{figure}

\subsubsection*{Robustness to node permutations for non-permutation invariant readouts}
\vspace{-1.5ex}
The objective of this experiment is to evaluate the stability of readouts that either do not enforce permutation invariance of the hypothesis space (i.e., \textsc{mlp} and \textsc{gru} readouts) or do so only approximately (i.e., \textsc{janossy} readouts). To this end, we generate $50$ random node permutations (relative to the order induced by the graphs generated from the canonical \textsc{smiles}) for $50$ randomly selected molecules from the \textsc{qm9} dataset and run these through the previously trained models with two layers (see above for details, neural vs standard readouts). As all graph molecular representations originate from a canonical \textsc{smiles} string (simplified molecular-input line-entry system, \cite{doi:10.1021/ci00057a005}), we also generate permuted graphs by using non-canonical \textsc{smiles}. For the latter, we have applied repeated sampling of permuted \textsc{smiles} using RDKit until the total number converged, resulting in permuted molecular graphs. Depending on the molecule, it is possible to generate as little as $20$ different representations or as many as $1,000$ (all the resulting graphs are used in the analysis). Permutations due to non-canonical \textsc{smiles} represent a subset of arbitrary permutations that are specific to the chemistry domain.
For each molecule and random permutation, we computed the error as the absolute difference between the predicted and target labels. \Cref{subfigure:random-perm-nodes,subfigure:random-perm-smiles} summarize the results of this experiment and show that the plain readouts (i.e., \textsc{mlp} and \textsc{gru}) can be negatively affected by some random permutations. Interestingly, the error for \textsc{mlp} is greatly accentuated on certain \textsc{qm9} prediction tasks, such as \textsc{u0}, \textsc{u298}, \textsc{h298}, and \textsc{g298}, while the error is very similar to sum, mean, and max on tasks such as \textsc{cv}, \textsc{zpve}, and \textsc{r2} (see \Cref{sec:QM9-perms-plots}, \Cref{figure:qm9-perm-nodes-individual} for more details). The Janossy readouts are trained not only on the original fixed-order graphs but also on $25$ random node permutations computed during training (i.e., a different stage compared to the evaluation presented here). We observe that the \textsc{janossy gru} readout improves upon the plain \textsc{gru}, leading to a distribution that is more similar to the \textsc{set transformer} readout, with a reduced number of outliers. In contrast, the \textsc{janossy-mlp} readout appears to be performing the worst in terms of robustness to node permutations. A breakdown of results over different neighborhood aggregation schemes and convolutions can be found in~\Cref{sec:QM9-perms-plots}, \Cref{figure:qm9-perm-nodes-conv,figure:qm9-perm-smiles-conv}. In summary, all neural readouts exhibit significantly reduced errors for the permutations originating from randomly generated non-canonical \textsc{smiles} compared to the arbitrary node permutations, despite being trained only using the canonical representations.  Moreover, we encounter many different (permuted) molecular representations that attain the minimal error values for \textsc{mlp} readouts (see \Cref{sec:QM9-perms-plots}, \Cref{figure:qm9-perm-smiles-best}).

\subsubsection*{The effectiveness of graph neural networks and readouts on multi-million graph datasets}
\begin{figure}[!t]
    \captionsetup{justification=centering, skip=4pt, font=small, labelfont=bf}
    \caption{Train loss for the \textsc{vgae} models trained on a proprietary dataset with $\approx1$ million molecular graphs.}
    \label{figure:1mil-VGAE-train-loss}
    \centering
    \includegraphics[width=1\textwidth]{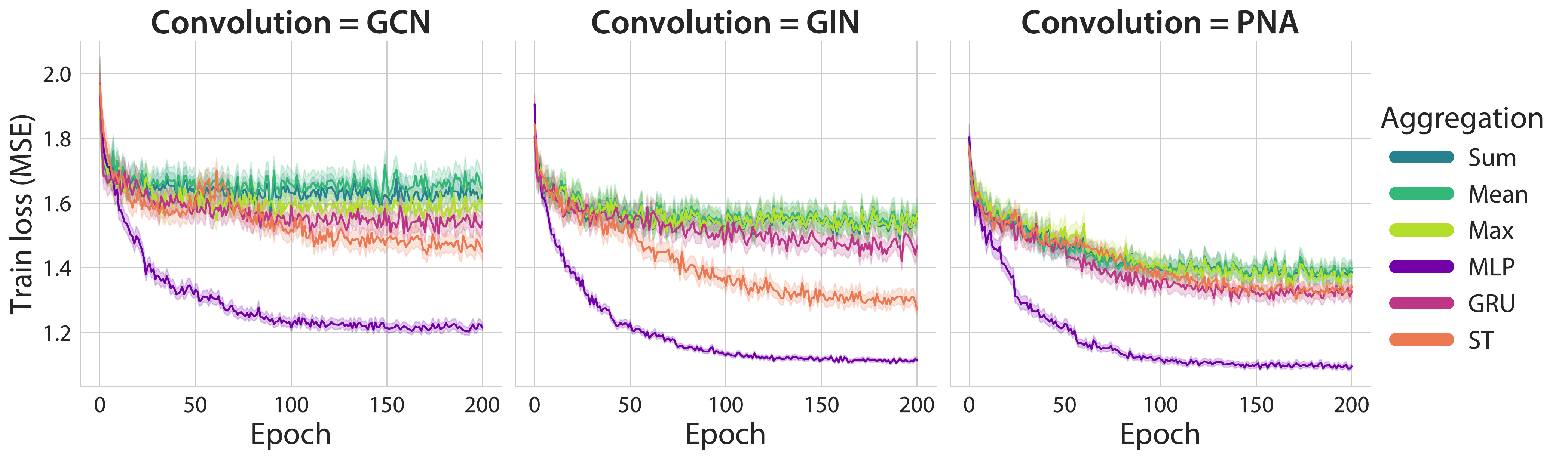}\vspace{-3ex}
\end{figure}

\vspace{-1.5ex}
In this experiment, we extend our empirical analysis with three AstraZeneca high-throughput screening assays against varied protein targets. Each dataset consists of $1$ to $2$ million molecular measurements, such that each molecule is associated with a single label (scalar) corresponding to its activity value (see \Cref{section:datasets-summary}, \Cref{table:hts-datasets} for details). We trained both non-variational and guided variational graph autoencoder~\cite{kipf2016variational,multifidelity} models (denoted by \textsc{gnn} and \textsc{vgae}, respectively) for $200$ epochs on the entirety of the available data for each dataset, with the goal of assessing the ability of graph neural networks to learn expressive graph representations at this scale (\Cref{figure:1mil-VGAE-train-loss}, with the other $5$ models in \Cref{sec:AZ-train-losses-plots}). Our analysis considers the performance relative to different neighborhood aggregation schemes or convolutional operators as well as the total number of such iterations that correspond to depth in graph neural networks (separately, see \Cref{section:train-loss-deeper}, \Cref{figure:1.5mil-GNN-VGAE-deeper}). \Cref{figure:1mil-VGAE-train-loss} summarizes the results of this experiment and indicates that the models using standard non-adaptive readouts (sum, mean, or max) generally struggle to model molecular data at the $1+$ million scale, as reflected in the plateaued training losses for each of the models. Depending on the dataset and the neighborhood aggregation scheme, the \textsc{mlp} readout significantly outperforms the other readout functions, with a train loss that rapidly decreases in the first $50$ epochs. The \textsc{set transformer} readout is the second-best choice, converging with a slightly a slower rate and occasionally diverging. The \textsc{gru} readouts offer only a slight improvement compared to the standard readout functions.
When training both variational and non-variational models with deeper architectures ($2$, $3$, and $4$ layers, excluding $\mu$- and $\sigma$-layers for the \textsc{vgae}), we do not observe significant benefits introduced by additional iterations of neighborhood aggregation schemes. This is in line with the study that introduced these datasets \cite{multifidelity}, which also evaluated multiple convolutional operators and numbers of such iterations/layers. Instead, as indicated previously, the largest benefits are generally associated with more powerful neighborhood aggregation schemes.
The results are further supported by the training metrics (\textsc{mae}, $\text{R}^2$) after $200$ epochs (see also \Cref{section:train-metric-az-datasets}, \Cref{table:2mil-GNN-train-metrics,table:2mil-VGAE-train-metrics,table:1mil-GNN-train-metrics,table:1mil-VGAE-train-metrics,table:1.5mil-GNN-train-metrics,table:1.5mil-VGAE-train-metrics}), which provide an insight into the ability of graph neural networks to fit signals on $1+$ million scale datasets. For example, our results for the \textsc{vgae gcn} model trained on the proprietary dataset with $\approx1$ million graphs show that neural readouts lead to an improvement in the $\text{R}^2$ score from $0.33$ to $0.78$, with $\approx1.5$ million graphs from $0.07$ to $0.64$, and with $\approx2$ million graphs from $0.06$ to $0.52$.

\section{Discussion}
\label{sec:discussion}
We have presented an extensive evaluation of graph neural networks with adaptive and differentiable readout functions and various neighborhood aggregation schemes. Our empirical results demonstrate that the proposed readouts can be beneficial on a wide variety of domains with graph structured data, as well as different data scales and graph characteristics. Overall, we observed improvements in over two thirds of the evaluated configurations (given by pairs consisting of datasets and neighborhood aggregation schemes), while performing competitively on the remaining ones. Moreover, we have empirically captured and quantified different aspects and trade-offs specific to adaptive readouts. For instance, the effectiveness of adaptive readouts that do not enforce permutation invariance of hypothesis spaces (\textsc{mlp} and \textsc{gru}) indicates that it might be possible to relax this constraint for certain tasks. A primary candidate for relaxation are molecular tasks, which are also one of the most popular application domains for graph neural networks. Molecules are typically presented in the canonical form, a strategy also adopted by popular open-source frameworks such as RDKit. Thus, the graph that corresponds to any given molecule comes with a fixed vertex ordering when generated from the canonical \textsc{smiles}. Our analysis suggests that neural readouts trained on canonical representations can learn chemical motifs that are applicable even to non-canonical inputs, or in other words, generally applicable chemical knowledge. It should be noted here that the canonical representations differ greatly even for extremely similar molecules (see also \Cref{section:similar-molecules-diff-repr}), such that it is improbable that the graph neural networks are learning simple associations based on position or presence of certain atoms. Instead, it might be the case that the networks can learn certain higher-level chemical patterns that are strictly relevant to the task at hand.

We have also discussed possible domain-specific interpretations for the effectiveness of models with adaptive readouts on some tasks. For instance, certain molecular properties tend to be approximated well by neural readouts, while others remain more amenable to standard pooling functions such as sum. Chemically, properties such as the internal energy, enthalpy, or free energy are generally considered additive (e.g., can be approximated by a sum of pairwise bond energies) and extensive (increasing almost linearly with the number of atoms). Such properties are a good fit for standard readouts. Other properties, such as the highest occupied molecular orbital and lowest unoccupied molecular orbital (\textsc{homo} and \textsc{lumo}, respectively) tend to be localized and are considered non-additive, such that a single atom can potentially completely alter the property, or not influence it at all. Popular problems such as bio-affinity prediction are also regarded as highly non-linear. Overall, this interplay suggests hybrid readouts for future research, where certain properties would be learned by functions such as sum, while others are left to more flexible neural readouts.

Regarding practical details such as the choice of the most suitable adaptive readout function for a given dataset, our empirical results indicate that larger and more complex (relative to the number of nodes per graph and dimension of node features) regression tasks see more pronounced performance improvements with adaptive readouts (based on statistically significant results from linear regression models detailed in \Cref{sec:stat-analysis}). We were, however, unable to observe a similar pattern for the considered classification tasks. Thanks to its potential for composing highly expressive neural architectures, \textsc{set transformer} is likely better suited for larger datasets. However, graph neural networks with that readout function tend to occasionally experience divergence on very large datasets or deep architectures ($6+$ layers), which can most likely be fully resolved with parameter tuning, especially the latent/hidden dimension of the attention mechanism. An avenue that might be promising for further study is pre-training readout networks, such that they can be quickly deployed on related tasks and fine-tuned. One starting point could be pre-training on large molecular databases, such as subsets of \textsc{gdb-17}~\cite{doi:10.1021/ci300415d} with inexpensive to compute molecular tasks (generated with RDKit, for example) as prediction targets, or unsupervised variations.

When it comes to related approaches, the majority of recent efforts have been focused on neighborhood aggregation schemes. This step also requires permutation invariance and it is interesting that a related work by Hamilton et al.~\cite{hamiltonlstm} has considered relaxation to that constraint and employed an \textsc{lstm} neural network to aggregate neighborhoods and produce node features. Along these lines, Murphy et al.~\cite{DBLP:journals/corr/abs-1811-01900} introduced Janossy pooling, a permutation-invariant pooling technique for neighborhood aggregation, designed for node classification tasks. Perhaps the most related to our direction and readouts is the concurrently developed work by Baek et al.~\cite{DBLP:journals/corr/abs-2102-11533} on graph multi-set transformers, i.e., a multi-head attention model based on a global pooling layer that models the relationship between nodes by exploiting their structural dependencies. For the purpose of measuring and fixing the over-squashing problem in graph neural networks, Alon and Yahav~\cite{DBLP:journals/corr/abs-2006-05205} proposed a fully-adjacent layer (each pair of nodes is connected by an edge), which greatly improved performance and resembles our use of the \textsc{mlp} readout. Prior work has also considered tunable $\ell_p$ pooling~\cite{lpaggfuncs} and more restrictive universal readouts based on deep sets \cite{navarin}. However, neither of these approaches offers a comprehensive empirical evaluation at the scale provided here.

\textcolor{black}{As adaptive readouts introduce a new differentiable component to graph neural networks, future studies might focus on analyzing properties such as transferability and interpretability. 
In our empirical study, we did not consider such experiments due to conceptual and practical differences. Conceptually, one of the main motivating factors for studying transferability is a scenario where the graph (network) size changes over time. This is typically encountered in recommendation systems or knowledge graphs which are not considered in our paper. Regarding the graph-to-graph transferability, there are domain-specific particularities that need to be considered. For example, learning on small graphs and transferring to larger graphs is not often required in chemical tasks, as most chemical regression benchmarks and real-world applications use only very small organic molecules (e.g., $<30$ atoms or nodes for \textsc{qm9}). There is also the requirement of selectivity, where an active molecule should bind only to a selected target and possible issues can arise with transferring a notion of similarity over the space of molecules that encodes activity to a completely different target. Moreover, there have been reports where the impact of learning (with graph neural networks) certain transferable chemical substructures (scaffolds) was not beneficial \cite{Sakai2021}. Practically, transferability has been most often studied with node-level tasks \cite{graphon}, while here we focus on graph-level predictions. Overall, we believe that studying the influence of adaptive readouts on transferability is interesting for future studies.
Regarding the interpretability, we have in this work focused on allowing for more flexibility in neural readouts (\Cref{section:graph-emb-distance}) and the structuring effect on the learned latent space, possibly making it amenable to clustering and other downstream tasks (\Cref{section:vgae-latent}).
}

\textbf{Acknowledgments}: We are grateful for access to the AstraZeneca Scientific Computing Platform. David Buterez has been supported by a fully funded PhD grant from AstraZeneca.

\clearpage

% \printbibliography
\bibliography{references}

\begin{thebibliography}{29}
\providecommand{\natexlab}[1]{#1}
\providecommand{\url}[1]{\texttt{#1}}
\expandafter\ifx\csname urlstyle\endcsname\relax
  \providecommand{\doi}[1]{doi: #1}\else
  \providecommand{\doi}{doi: \begingroup \urlstyle{rm}\Url}\fi

\bibitem[Kipf and Welling(2017)]{DBLP:journals/corr/KipfW16}
Thomas~N. Kipf and Max Welling.
\newblock Semi-supervised classification with graph convolutional networks.
\newblock In \emph{International Conference on Learning Representations
  (ICLR)}, 2017.

\bibitem[Hamilton et~al.(2017{\natexlab{a}})Hamilton, Ying, and
  Leskovec]{HamiltonYL17}
William~L. Hamilton, Rex Ying, and Jure Leskovec.
\newblock Representation learning on graphs: Methods and applications.
\newblock \emph{{IEEE} Data Engineering Bulletin}, 40\penalty0 (3):\penalty0
  52--74, 2017{\natexlab{a}}.

\bibitem[Veličković et~al.(2018)Veličković, Cucurull, Casanova, Romero,
  Liò, and Bengio]{velickovic2018graph}
Petar Veličković, Guillem Cucurull, Arantxa Casanova, Adriana Romero, Pietro
  Liò, and Yoshua Bengio.
\newblock Graph attention networks.
\newblock In \emph{International Conference on Learning Representations}, 2018.

\bibitem[Bronstein et~al.(2021)Bronstein, Bruna, Cohen, and
  Velickovic]{DBLP:journals/corr/abs-2104-13478}
Michael~M. Bronstein, Joan Bruna, Taco Cohen, and Petar Velickovic.
\newblock Geometric deep learning: Grids, groups, graphs, geodesics, and
  gauges.
\newblock \emph{CoRR}, abs/2104.13478, 2021.

\bibitem[Xu et~al.(2019)Xu, Hu, Leskovec, and
  Jegelka]{DBLP:journals/corr/abs-1810-00826}
Keyulu Xu, Weihua Hu, Jure Leskovec, and Stefanie Jegelka.
\newblock How powerful are graph neural networks?
\newblock In \emph{International Conference on Learning Representations}, 2019.

\bibitem[Zaheer et~al.(2017)Zaheer, Kottur, Ravanbakhsh, Poczos, Salakhutdinov,
  and Smola]{deepsets}
Manzil Zaheer, Satwik Kottur, Siamak Ravanbakhsh, Barnabas Poczos, Russ~R
  Salakhutdinov, and Alexander~J Smola.
\newblock Deep sets.
\newblock In \emph{Advances in Neural Information Processing Systems},
  volume~30. Curran Associates, Inc., 2017.

\bibitem[Wagstaff et~al.(2019)Wagstaff, Fuchs, Engelcke, Posner, and
  Osborne]{Wagstaff19}
Edward Wagstaff, Fabian Fuchs, Martin Engelcke, Ingmar Posner, and Michael~A.
  Osborne.
\newblock {On the Limitations of Representing Functions on Sets}.
\newblock In \emph{{Proceedings of the 36th International Conference on Machine
  Learning}}, pages 6487--6494. {PMLR}, 2019.

\bibitem[Babai and Kucera(1979)]{Laszlo}
Laszlo Babai and Ludik Kucera.
\newblock Canonical labelling of graphs in linear average time.
\newblock In \emph{20th Annual Symposium on Foundations of Computer Science
  (sfcs 1979)}, pages 39--46, 1979.

\bibitem[Morris et~al.(2021)Morris, Lipman, Maron, Rieck, Kriege, Grohe, Fey,
  and Borgwardt]{DBLP:journals/corr/abs-2112-09992}
Christopher Morris, Yaron Lipman, Haggai Maron, Bastian Rieck, Nils~M. Kriege,
  Martin Grohe, Matthias Fey, and Karsten~M. Borgwardt.
\newblock Weisfeiler and leman go machine learning: The story so far.
\newblock \emph{CoRR}, abs/2112.09992, 2021.

\bibitem[Lee et~al.(2019)Lee, Lee, Kim, Kosiorek, Choi, and
  Teh]{DBLP:journals/corr/abs-1810-00825}
Juho Lee, Yoonho Lee, Jungtaek Kim, Adam Kosiorek, Seungjin Choi, and Yee~Whye
  Teh.
\newblock Set transformer: A framework for attention-based
  permutation-invariant neural networks.
\newblock In \emph{Proceedings of the 36th International Conference on Machine
  Learning}, pages 3744--3753, 2019.

\bibitem[Murphy et~al.(2019)Murphy, Srinivasan, Rao, and
  Ribeiro]{DBLP:journals/corr/abs-1811-01900}
Ryan~L. Murphy, Balasubramaniam Srinivasan, Vinayak Rao, and Bruno Ribeiro.
\newblock Janossy pooling: Learning deep permutation-invariant functions for
  variable-size inputs.
\newblock In \emph{International Conference on Learning Representations}, 2019.

\bibitem[Vaswani et~al.(2017)Vaswani, Shazeer, Parmar, Uszkoreit, Jones, Gomez,
  Kaiser, and Polosukhin]{DBLP:journals/corr/VaswaniSPUJGKP17}
Ashish Vaswani, Noam Shazeer, Niki Parmar, Jakob Uszkoreit, Llion Jones,
  Aidan~N Gomez, Łukasz Kaiser, and Illia Polosukhin.
\newblock Attention is all you need.
\newblock In I.~Guyon, U.~Von Luxburg, S.~Bengio, H.~Wallach, R.~Fergus,
  S.~Vishwanathan, and R.~Garnett, editors, \emph{Advances in Neural
  Information Processing Systems}, volume~30. Curran Associates, Inc., 2017.

\bibitem[Cho et~al.(2014)Cho, van Merri{\"e}nboer, Bahdanau, and
  Bengio]{DBLP:journals/corr/ChoMBB14}
Kyunghyun Cho, Bart van Merri{\"e}nboer, Dzmitry Bahdanau, and Yoshua Bengio.
\newblock On the properties of neural machine translation: Encoder{--}decoder
  approaches.
\newblock In \emph{Proceedings of {SSST}-8, Eighth Workshop on Syntax,
  Semantics and Structure in Statistical Translation}, pages 103--111, Doha,
  Qatar, October 2014. Association for Computational Linguistics.
\newblock \doi{10.3115/v1/W14-4012}.

\bibitem[Chicco and Jurman(2020)]{Chicco2020}
Davide Chicco and Giuseppe Jurman.
\newblock The advantages of the matthews correlation coefficient (mcc) over f1
  score and accuracy in binary classification evaluation.
\newblock \emph{BMC Genomics}, 21\penalty0 (1):\penalty0 6, Jan 2020.
\newblock ISSN 1471-2164.

\bibitem[Chicco et~al.(2021)Chicco, Warrens, and Jurman]{Chicco2021}
Davide Chicco, Matthijs~J. Warrens, and Giuseppe Jurman.
\newblock The coefficient of determination r-squared is more informative than
  smape, mae, mape, mse and rmse in regression analysis evaluation.
\newblock \emph{PeerJ. Computer science}, 7:\penalty0 e623--e623, Jul 2021.
\newblock ISSN 2376-5992.

\bibitem[Brody et~al.(2022)Brody, Alon, and
  Yahav]{DBLP:journals/corr/abs-2105-14491}
Shaked Brody, Uri Alon, and Eran Yahav.
\newblock How attentive are graph attention networks?
\newblock In \emph{International Conference on Learning Representations}, 2022.

\bibitem[Corso et~al.(2020)Corso, Cavalleri, Beaini, Li\`{o}, and
  Veli\v{c}kovi\'{c}]{DBLP:journals/corr/abs-2004-05718}
Gabriele Corso, Luca Cavalleri, Dominique Beaini, Pietro Li\`{o}, and Petar
  Veli\v{c}kovi\'{c}.
\newblock Principal neighbourhood aggregation for graph nets.
\newblock In H.~Larochelle, M.~Ranzato, R.~Hadsell, M.F. Balcan, and H.~Lin,
  editors, \emph{Advances in Neural Information Processing Systems}, volume~33,
  pages 13260--13271. Curran Associates, Inc., 2020.

\bibitem[Weininger(1988)]{doi:10.1021/ci00057a005}
David Weininger.
\newblock Smiles, a chemical language and information system. 1. introduction
  to methodology and encoding rules.
\newblock \emph{Journal of Chemical Information and Computer Sciences},
  28\penalty0 (1):\penalty0 31--36, 1988.
\newblock \doi{10.1021/ci00057a005}.

\bibitem[Kipf and Welling(2016)]{kipf2016variational}
Thomas~N Kipf and Max Welling.
\newblock Variational graph auto-encoders.
\newblock \emph{NIPS Workshop on Bayesian Deep Learning}, 2016.

\bibitem[Buterez et~al.(2022)Buterez, Janet, Kiddle, and Liò]{multifidelity}
David Buterez, Jon~Paul Janet, Steven Kiddle, and Pietro Liò.
\newblock Multi-fidelity machine learning models for improved high-throughput
  screening predictions.
\newblock \emph{ChemRxiv}, 2022.

\bibitem[Ruddigkeit et~al.(2012)Ruddigkeit, van Deursen, Blum, and
  Reymond]{doi:10.1021/ci300415d}
Lars Ruddigkeit, Ruud van Deursen, Lorenz~C. Blum, and Jean-Louis Reymond.
\newblock Enumeration of 166 billion organic small molecules in the chemical
  universe database gdb-17.
\newblock \emph{Journal of Chemical Information and Modeling}, 52\penalty0
  (11):\penalty0 2864--2875, 2012.

\bibitem[Hamilton et~al.(2017{\natexlab{b}})Hamilton, Ying, and
  Leskovec]{hamiltonlstm}
William~L. Hamilton, Rex Ying, and Jure Leskovec.
\newblock Inductive representation learning on large graphs.
\newblock In \emph{Proceedings of the 31st International Conference on Neural
  Information Processing Systems}. Curran Associates Inc., 2017{\natexlab{b}}.

\bibitem[Baek et~al.(2021)Baek, Kang, and
  Hwang]{DBLP:journals/corr/abs-2102-11533}
Jinheon Baek, Minki Kang, and Sung~Ju Hwang.
\newblock Accurate learning of graph representations with graph multiset
  pooling.
\newblock In \emph{ICLR}, 2021.

\bibitem[Alon and Yahav(2021)]{DBLP:journals/corr/abs-2006-05205}
Uri Alon and Eran Yahav.
\newblock On the bottleneck of graph neural networks and its practical
  implications.
\newblock In \emph{International Conference on Learning Representations}, 2021.

\bibitem[Pellegrini et~al.(2021)Pellegrini, Tibo, Frasconi, Passerini, and
  Jaeger]{lpaggfuncs}
Giovanni Pellegrini, Alessandro Tibo, Paolo Frasconi, Andrea Passerini, and
  Manfred Jaeger.
\newblock Learning aggregation functions.
\newblock In \emph{Proceedings of the Thirtieth International Joint Conference
  on Artificial Intelligence}. International Joint Conferences on Artificial
  Intelligence Organization, 2021.

\bibitem[Navarin et~al.(2019)Navarin, Tran, and Sperduti]{navarin}
Nicolò Navarin, Dinh~Van Tran, and Alessandro Sperduti.
\newblock Universal readout for graph convolutional neural networks.
\newblock In \emph{2019 International Joint Conference on Neural Networks},
  2019.

\bibitem[Sakai et~al.(2021)Sakai, Nagayasu, Shibui, Andoh, Takayama, Shirakawa,
  and Kaneko]{Sakai2021}
Miyuki Sakai, Kazuki Nagayasu, Norihiro Shibui, Chihiro Andoh, Kaito Takayama,
  Hisashi Shirakawa, and Shuji Kaneko.
\newblock Prediction of pharmacological activities from chemical structures
  with graph convolutional neural networks.
\newblock \emph{Scientific Reports}, 11\penalty0 (1):\penalty0 525, Jan 2021.
\newblock ISSN 2045-2322.
\newblock \doi{10.1038/s41598-020-80113-7}.

\bibitem[Ruiz et~al.(2020)Ruiz, Chamon, and Ribeiro]{graphon}
Luana Ruiz, Luiz Chamon, and Alejandro Ribeiro.
\newblock Graphon neural networks and the transferability of graph neural
  networks.
\newblock In H.~Larochelle, M.~Ranzato, R.~Hadsell, M.F. Balcan, and H.~Lin,
  editors, \emph{Advances in Neural Information Processing Systems}, volume~33,
  pages 1702--1712. Curran Associates, Inc., 2020.

\bibitem[Wu et~al.(2018)Wu, Ramsundar, Feinberg, Gomes, Geniesse, Pappu,
  Leswing, and Pande]{C7SC02664A}
Zhenqin Wu, Bharath Ramsundar, Evan~N. Feinberg, Joseph Gomes, Caleb Geniesse,
  Aneesh~S. Pappu, Karl Leswing, and Vijay Pande.
\newblock Moleculenet: a benchmark for molecular machine learning.
\newblock \emph{Chem. Sci.}, 9:\penalty0 513--530, 2018.

\end{thebibliography}

\clearpage
\section*{Checklist}

%%% BEGIN INSTRUCTIONS %%%
% The checklist follows the references.  Please
% read the checklist guidelines carefully for information on how to answer these
% questions.  For each question, change the default \answerTODO{} to \answerYes{},
% \answerNo{}, or \answerNA{}.  You are strongly encouraged to include a {\bf
% justification to your answer}, either by referencing the appropriate section of
% your paper or providing a brief inline description.  For example:
% \begin{itemize}
%   \item Did you include the license to the code and datasets? \answerYes{where available. See \Cref{table:summary-of-datasets} and \Cref{table:summary-of-datasets-2}. Old or public domain datasets (PubChem) are exceptions. \textquote{No} for the three proprietary molecular datasets.}
% \end{itemize}
% Please do not modify the questions and only use the provided macros for your
% answers.  Note that the Checklist section does not count towards the page
% limit.  In your paper, please delete this instructions block and only keep the
% Checklist section heading above along with the questions/answers below.
%%% END INSTRUCTIONS %%%

\begin{enumerate}

\item For all authors...
\begin{enumerate}
  \item Do the main claims made in the abstract and introduction accurately reflect the paper's contributions and scope?
    \answerYes{Please see \Cref{sec:exps,sec:discussion}.}
  \item Did you describe the limitations of your work?
    \answerYes{Covered in \Cref{sec: intro,sec:discussion}.}
  \item Did you discuss any potential negative societal impacts of your work?
    \answerYes{Please see \Cref{sec:discussion} and \Cref{section:datasets-summary}.}
  \item Have you read the ethics review guidelines and ensured that your paper conforms to them?
    \answerYes{Detailed in \Cref{section:datasets-summary}.}
\end{enumerate}

\item If you are including theoretical results...
\begin{enumerate}
  \item Did you state the full set of assumptions of all theoretical results?
    \answerNA{}
        \item Did you include complete proofs of all theoretical results?
    \answerNA{}
\end{enumerate}

\item If you ran experiments...
\begin{enumerate}
  \item Did you include the code, data, and instructions needed to reproduce the main experimental results (either in the supplemental material or as a URL)?
    \answerYes{The code and instructions are available on GitHub.}
  \item Did you specify all the training details (e.g., data splits, hyperparameters, how they were chosen)?
    \answerYes{In \Cref{apx:sec:exp-design,sec:st-architectures,section:nn-agg-hyperparams} and the associated code.}
        \item Did you report error bars (e.g., with respect to the random seed after running experiments multiple times)?
    \answerYes{}
        \item Did you include the total amount of compute and the type of resources used (e.g., type of GPUs, internal cluster, or cloud provider)?
    \answerYes{Please see \Cref{sec:exp-platform}.}
\end{enumerate}

\item If you are using existing assets (e.g., code, data, models) or curating/releasing new assets...
\begin{enumerate}
  \item If your work uses existing assets, did you cite the creators?
    \answerYes{}
  \item Did you mention the license of the assets?
    \answerYes{where available. Please see \Cref{section:datasets-summary}, \Cref{table:summary-of-datasets} and \Cref{table:summary-of-datasets-2}. Old or public domain datasets (originating from PubChem) are exceptions. \answerNo{for the three proprietary molecular datasets.}}
  \item Did you include any new assets either in the supplemental material or as a URL?
    \answerNo{}
  \item Did you discuss whether and how consent was obtained from people whose data you're using/curating?
    \answerNo{for the established open source datasets.} \answerYes{Consent was given to publish results on the proprietary data.}
  \item Did you discuss whether the data you are using/curating contains personally identifiable information or offensive content?
    \answerYes{Please see \Cref{section:datasets-summary}.}
\end{enumerate}

\item If you used crowdsourcing or conducted research with human subjects...
\begin{enumerate}
  \item Did you include the full text of instructions given to participants and screenshots, if applicable?
    \answerNA{}
  \item Did you describe any potential participant risks, with links to Institutional Review Board (IRB) approvals, if applicable?
    \answerNA{}
  \item Did you include the estimated hourly wage paid to participants and the total amount spent on participant compensation?
    \answerNA{}
\end{enumerate}

\end{enumerate}

\clearpage
\appendix

\renewcommand{\figurename}{Appendix Figure}
\renewcommand{\tablename}{Appendix Table}
\crefalias{figure}{appendixfigure}
\crefalias{table}{appendixtable}
\crefalias{subfigure}{appendixsubfigure}
\setcounter{figure}{0}
\setcounter{table}{0}

\section{Summary of the used datasets}
\label{section:datasets-summary}

All of the public datasets were previously published, either as graph representation learning benchmarks or new datasets for specific graph tasks. The datasets cover a multitude of domains: quantum mechanics, biophysics, bioinformatics, computer vision, social networks, synthetic graphs, and function call graphs. As such, the datasets do not include any personally identifiable information or offensive content. This claim is based both on manual inspection and previous peer-review of the associated publications. Furthermore, no research was performed with human subjects as part of our study. The public benchmarks are listed in \Cref{section:datasets-summary}, \Cref{table:summary-of-datasets,table:summary-of-datasets-2}. The public and proprietary bio-affinity (high-throughput screening) datasets are listed in \Cref{section:datasets-summary}, \Cref{table:hts-datasets}.

\renewcommand{\sfdefault}{lmss}
\renewcommand{\familydefault}{\sfdefault}

\begin{table}[!h]
    \renewcommand{\arraystretch}{1.1}
    \centering
    \captionsetup{justification=centering, skip=4pt, labelfont=bf}
    \caption{Summary of the bio-affinity high-throughput screening datasets.}
    \label{table:hts-datasets}
    \begin{tabular}{cclS[table-format=7.0]ccc}
    \toprule
    \multicolumn{1}{c}{\textbf{Availability}} & \multicolumn{1}{c}{\textbf{Source}} & \multicolumn{1}{c}{\textbf{Dataset}} & \multicolumn{1}{c}{\textbf{Size}} & \multicolumn{1}{c}{\textbf{Splits}} & \multicolumn{1}{c}{\textbf{Task type}} & \multicolumn{1}{c}{\textbf{\# Tasks}} \\ \midrule
    \multirow{3}{*}{Public} & \multirow{3}{*}{PubChem} & AID1949 & 98472 & \multirow{3}{*}{No} & \multirow{3}{*}{Regression} & \multirow{3}{*}{1} \\
    &  & AID449762 & 311910 &  &  &  \\
    &  & AID602261 & 343811 &  &  &  \\ \cmidrule[0.75pt]{1-7}
    \multirow{3}{*}{Private} & \multirow{3}{*}{\makecell{Pharmaceutical\\company}} & & 1013581 & \multirow{3}{*}{No} & \multirow{3}{*}{Regression} & \multirow{3}{*}{1} \\
    &  & & 1482258 &  &  &  \\
    &  & & 1962638 &  &  &  \\ \bottomrule
    \end{tabular}
\end{table}

\begin{sidewaystable}
    \centering
    \small
    \captionsetup{skip=4pt, labelfont=bf}
    \caption{Summary of all the used benchmarks (datasets), including domain, datasets statistics, random splitting procedures, and source. DeepChem (DC) and PyTorch Geometric (PyG) are released under the MIT license. Several molecular datasets originate from PubChem (public domain). ENZYMES uses CC BY 4.0. reddit\_threads and twitch\_egos use GPL-3.0. COLORS and TRIANGLES use ECL-2.0. CIFAR10, MNIST, and ZINC use MIT.}
    \label{table:summary-of-datasets}
    \begin{tabular}{cclcrS[table-format=6.0]rrrcc}
      \toprule
      \textbf{Collection} & \textbf{Domain} & \multicolumn{1}{c}{\textbf{Dataset}} & \textbf{Type} & \multicolumn{1}{c}{\textbf{\# Tasks}} & \multicolumn{1}{c}{\textbf{Size}} & \multicolumn{1}{c}{\textbf{Avg. nodes}} & \multicolumn{1}{c}{\textbf{Avg. edges}} & \multicolumn{1}{c}{\textbf{Node attr.}} & \textbf{Splits} & \textbf{Source} \\ \midrule
      \multirow{12}{*}{MoleculeNet} & \multirow{3}{*}{\makecell{Quantum\\Mechanics}} & QM9 & Regr. & 12 & 132480 & 17.99 & 37.15 & 30 & 5 random & DC \\
       &  & QM8 & Regr. & 12 & 21747 & 16.09 & 32.81 & 30 & 5 random & DC \\
       &  & QM7 & Regr. & 1 & 6834 & 15.54 & 30.39 & 30 & 5 random & DC \\ \cmidrule(l){2-11}
       & \multirow{3}{*}{\makecell{Physical\\Chemistry}} & ESOL & Regr. & 1 & 1127 & 13.30 & 27.38 & 30 & 5 random & DC \\
       &  & FreeSolv & Regr. & 1 & 639 & 8.76 & 16.85 & 30 & 5 random & DC \\
       &  & Lipophilicity & Regr. & 1 & 4200 & 27.04 & 59.00 & 30 & 5 random & DC \\ \cmidrule(l){2-11}
       & \multirow{4}{*}{Biophysics} & PCBA & Cls. & 128 & 437918 & 25.97 & 56.22 & 30 & 5 random & DC \\
       &  & HIV & Cls. & 1 & 41127 & 25.51 & 54.94 & 30 & 5 random & DC \\
       &  & BACE & Cls. & 1 & 1513 & 34.09 & 73.72 & 30 & 5 random & DC \\
       &  & BACE & Regr. & 1 & 1513 & 34.09 & 73.72 & 30 & 5 random & DC \\ \cmidrule(l){2-11}
       & \multirow{2}{*}{Physiology} & BBBP & Cls. & 1 & 2039 & 24.06 & 51.91 & 30 & 5 random & DC \\
       &  & SIDER & Cls. & 27 & 1396 & 34.36 & 72.29 & 30 & 5 random & DC \\ \cmidrule[0.75pt]{1-11}
      \multirow{17}{*}{TUDataset} & \multirow{2}{*}{Bioinformatics} & ENZYMES & Cls. & 6 & 600 & 32.63 & 62.14 & 18 & 5 random & PyG \\
       &  & PROTEINS\_full & Cls. & 2 & 1113 & 39.06 & 72.82 & 29 & 5 random & PyG \\ \cmidrule(l){2-11}
       & \multirow{3}{*}{\makecell{Computer\\Vision}} & COIL-DEL & Cls. & 100 & 3900 & 21.54 & 54.24 & 2 & 5 random & PyG \\
       &  & COIL-RAG & Cls. & 100 & 3900 & 3.01 & 3.02 & 64 & 5 random & PyG \\
       &  & Cuneiform & Cls. & 30 & 267 & 21.27 & 44.80 & 3 & 5 random & PyG \\ \cmidrule(l){2-11}
       & \multirow{7}{*}{\makecell{Social\\Networks}} & github\_stargazers & Cls. & 2 & 12725 & 113.79 & 234.64 & - & 5 random & PyG \\
       &  & IMDB-BINARY & Cls. & 2 & 1000 & 19.77 & 96.53 & - & 5 random & PyG \\
       &  & REDDIT-BINARY & Cls. & 2 & 2000 & 429.63 & 497.75 & - & 5 random & PyG \\
       &  & REDDIT-MULTI-12K & Cls. & 11 & 11929 & 391.41 & 456.89 & - & 5 random & PyG \\
       &  & reddit\_threads & Cls. & 2 & 203088 & 23.93 & 24.99 & - & 5 random & PyG \\
       &  & twitch\_egos & Cls. & 2 & 127094 & 29.67 & 86.59 & - & 5 random & PyG \\
       &  & TWITTER-Real-Graph-Partial & Cls. & 2 & 144033 & 4.03 & 4.98 & - & 5 random & PyG \\ \cmidrule(l){2-11}
       & \multirow{5}{*}{Synthetic} & COLORS-3 & Cls. & 11 & 10500 & 61.31 & 91.03 & 4 & 5 random & PyG \\
       &  & SYNTHETIC & Cls. & 2 & 300 & 100.00 & 196.00 & 1 & 5 random & PyG \\
       &  & SYNTHETICnew & Cls. & 2 & 300 & 100.00 & 196.25 & 1 & 5 random & PyG \\
       &  & Synthie & Cls. & 4 & 400 & 95.00 & 172.93 & 15 & 5 random & PyG \\
       &  & TRIANGLES & Cls. & 10 & 45000 & 20.85 & 32.74 & - & 5 random & PyG \\ \cmidrule[0.75pt]{1-11}
      \multirow{2}{*}{GNNBenchmarkDataset} & \multirow{2}{*}{\makecell{Computer\\Vision}} & MNIST & Cls. & 10 & 55000 & 70.60 & 564.50 & 3 & Provided & PyG \\
       &  & CIFAR10 & Cls. & 10 & 45000 & 117.60 & 941.20 & 5 & Provided & PyG \\ \cmidrule[0.75pt]{1-11}
      ZINC & \makecell{Drug-like\\molecules} & ZINC & Regr. & 1 & 249456 & 23.15 & 49.80 & 1 & Provided & PyG \\
      \bottomrule
    \end{tabular}
\end{sidewaystable}
\renewcommand{\familydefault}{\rmdefault}

\renewcommand{\sfdefault}{lmss}
\renewcommand{\familydefault}{\sfdefault}
\begin{sidewaystable}
    \centering
    \small
    \captionsetup{skip=4pt, labelfont=bf}
    \caption{Summary of all the used benchmarks (datasets), including domain, datasets statistics, random splitting procedures, and source. PyG, PyTorch Geometric.}
    \label{table:summary-of-datasets-2}
    \begin{tabular}{cclcrS[table-format=6.0]rrrcc}
      \toprule
      \textbf{Collection} & \textbf{Domain} & \multicolumn{1}{c}{\textbf{Dataset}} & \textbf{Type} & \multicolumn{1}{c}{\textbf{\# Tasks}} & \multicolumn{1}{c}{\textbf{Size}} & \multicolumn{1}{c}{\textbf{Avg. nodes}} & \multicolumn{1}{c}{\textbf{Avg. edges}} & \multicolumn{1}{c}{\textbf{Node attr.}} & \textbf{Splits} & \textbf{Source} \\ \midrule
      \multirow{6}{*}{TUDataset} & \multirow{2}{*}{Small molecules} & AIDS & Cls. & 2 & 2000 & 15.69 & 16.20 & 4 & 5 random & PyG \\
       &  & alchemy\_full & Regr. & 12 & 202579 & 10.10 & 10.44 & 3 & 5 random & PyG \\
       &  & FRANKENSTEIN & Cls. & 2 & 4337 & 16.90 & 17.88 & 780 & 5 random & PyG \\
       &  & Mutagenicity & Cls. & 2 & 4337 & 30.32 & 30.77 & - & 5 random & PyG \\
       &  & MUTAG & Cls. & 2 & 188 & 17.93 & 19.79 & - & 5 random & PyG \\
       &  & YeastH & Cls. & 2 & 79601 & 39.44 & 40.74 & - & 5 random & PyG \\ \cmidrule[0.75pt]{1-11}
       MalNetTiny & Function call graphs & MalNetTiny & Cls. & 5 & 5000 & 1410.31 & 2859.94 & - & 5 random & PyG \\ \bottomrule
    \end{tabular}
\end{sidewaystable}
\renewcommand{\familydefault}{\rmdefault}

\section{Experimental design and reporting}
\label{apx:sec:exp-design}
Before training each model, a fixed random seed at the beginning of training is set (\texttt{pytorch\_lightning.seed\_everything(0)}). Any convolution-specific hyperparameters (such as the number of attention heads and dropout values for \textsc{gat} and \textsc{gatv2}) were chosen as reasonable defaults and frozen. As the neural readouts introduce new hyperparameters for the overall \textsc{gnn} architecture, these are also set to reasonable defaults depending on the dataset type and size (\Cref{apx:sec:exp-design}, \Cref{table:neural-agg-hyperparams}), but are not part of any hyperparameter optimization procedure.

For datasets that did not explicitly provide train, validation, and test sets, we applied an $80\%$/$10\%$/$10\%$ split, for five different times on random permutations of the datasets. This procedure was applied on all MoleculeNet datasets, as well as the TUDataset benchmarks. We did not apply any custom splits on \textsc{mnist}, \textsc{cifar10}, and \textsc{zinc}. The seeds used for the five random permutations are available in \textbf{Supplementary File 1} (available on GitHub) and can be directly used with the provided source code. All models are trained with an early stopping mechanism set to a patience value of $30$.

Almost all datasets are completely loaded in memory, with the exception of a few complex datasets (such as the \textsc{reddit} datasets). If the default batch size of $32$ was too large for such datasets, the maximum batch size that allowed the models to be trained on a GPU with $24$GB of VRAM was used.

The loss functions used within the deep learning models are set to standard choices, such as mean squared error (\textsc{mse}) for regression datasets, binary cross-entropy for binary classification datasets and cross-entropy for multi-class datasets. We also generally tried to use the recommended loss functions for the MoleculeNet datasets according to the original publication, in particular using the mean absolute error (\textsc{mae}) function for certain datasets \cite{C7SC02664A}.

The metrics chosen to report the model performance depend on the task type (regression or classification) and the number of tasks (or classes for classification). For binary classification tasks, we report the area under the receiver operating characteristic curve (\textsc{auroc}) and the Matthews correlation coefficient (\textsc{mcc}), while for classification tasks with more than $2$ classes only the \textsc{mcc} is reported. The \textsc{mcc} was recently reported to be a more helpful metric compared to popular choices such as the accuracy or the $\text{F}_1$ score and is considered one of the best summaries of the confusion matrix \cite{Chicco2020}.

For regression tasks, we report the \textsc{mae} and the coefficient of determination ($\text{R}^2$) for all datasets. $\text{R}^2$ was also recently reported as a regression metric that is more informative compared to the traditional choices of \textsc{mae}, \textsc{mse}, \textsc{rmse}, and others \cite{Chicco2021}.

To simplify reporting the results, for multi-label datasets the appropriate metrics are computed between flattened representations of the predictions and the ground truth values.

\section{Set Transformer aggregator architectures}
\label{sec:st-architectures}
Our default Set Transformer architecture, simply referred to as \textsc{set transformer} for the majority of the paper, and \textsc{st complex} in \Cref{fig:par-budget} and \Cref{section:all-benchmarks-metrics}, \Cref{table:st_heads_qm9}, uses multiple \textsc{sab} blocks, where a \textsc{sab} block is defined as $\textsc{sab}(A) = \textsc{mab}(A, A)$:
\begin{align}
    \textsc{encoder}(A) \coloneqq \textsc{sab}^2(A) \quad \text{and} \quad \textsc{decoder}(C) &\coloneqq \textsc{ff}(\textsc{sab}^2(\textsc{pma}_k(C)))
\end{align}
We also evaluate a variation termed \textsc{st minimal}, with an architecture reduced to:
\begin{align}
    \textsc{encoder}(A) \coloneqq \textsc{sab}(A) \quad \text{and} \quad \textsc{decoder}(C) &\coloneqq \textsc{ff}(\textsc{pma}_k(C))
\end{align}

\section{Summary of multi-head attention}
\label{sec:attention-recapitulation}
First of all, we recapitulate the original definition of attention, which works by initializing three learnable matrices, often called $Q$ for \textit{queries}, $K$ for \textit{keys} and $V$ for \textit{values}. For self attention, $Q = K = V$. The attention computation is then defined as
\begin{align}
    \textsc{attention}(Q, K, V) \coloneqq \text{softmax}(\frac{QK^\top}{\sqrt{d_k}})V
\end{align}
where $\text{softmax}$ is defined element-wise as $\text{softmax}(\mathbf{x})_i \coloneqq \frac{\text{exp}(\mathbf{x}_i)}{\sum_{j}\text{exp}(\mathbf{x}_j)}$

It is often beneficial to perform multiple attention computations concurrently, using different parameters (weights $W$), called multi-head attention with $h$ heads
\begin{align}
    \textsc{multi-head}(Q, K, V) &\coloneqq \text{Concatenate}(\text{head}_1, \ldots, \text{head}_h)W^O\\
    \text{head}_i &\coloneqq \textsc{attention}(QW_i^Q, KW_i^K, VW_i^V)
\end{align}

\section{Set Transformer for variable-sized inputs}
\label{sec:st-variable-input}
The \textsc{set transformer} readout supports variable-sized inputs, i.e. it is theoretically possible to start from a non-rectangular (ragged) tensor, such that zero-padding is avoided in the flattened representation. However, as our chosen deep learning library (PyTorch) does not support ragged tensors at the time of writing, our default implementation uses the already defined representations (denoted by $H$ and $h$ in the main text). We did, however, experiment with a computationally inefficient implementation relying on a for-loop instead of batching, without observing notable performance differences (not shown).

\clearpage

\section{Hyperparameters specific to neural readouts}
\label{section:nn-agg-hyperparams}
\renewcommand{\sfdefault}{lmss}
\renewcommand{\familydefault}{\sfdefault}
\begin{table}[htp]
    \captionsetup{skip=5pt, labelfont=bf}
    \centering
    \caption{Summary of the hyperparameters used for neural aggregators for each dataset. The hyperparameter names follow the same notation as introduced in the main text. Dim., dimension.}
    \label{table:neural-agg-hyperparams}
    \renewcommand{\arraystretch}{1.3}
    \begin{tabular}{l@{\hskip 1cm}rr@{\hskip 1cm}rrr}
    \toprule
    \textbf{Dataset} & \multicolumn{2}{{c@{\hspace{1cm}}}}{\textbf{MLP}} & \multicolumn{3}{c}{\textbf{Set Transformer}} \\ \cmidrule(r{0.9cm}){1-1} \cmidrule(lr{1cm}){2-3} \cmidrule(lr){4-6}
     & \multicolumn{1}{l}{$d_1$} & \multicolumn{1}{l}{$d_{\text{out}}$} & \multicolumn{1}{l}{$k$} & \multicolumn{1}{l}{Hidden dim.} & \multicolumn{1}{l}{$n_h$} \\ \cmidrule(lr){2-2} \cmidrule(lr{0.9cm}){3-3} \cmidrule(lr){4-4} \cmidrule(lr){5-5} \cmidrule(lr){6-6}
    ESOL & 64 & 32 & 8 & 32 & 4 \\
    FreeSolv & 64 & 32 & 8 & 32 & 4 \\
    Lipo & 64 & 32 & 8 & 32 & 4 \\
    BACE\_REGR & 64 & 32 & 8 & 32 & 4 \\
    BACE\_CLS & 64 & 32 & 8 & 32 & 4 \\
    BBBP & 64 & 32 & 8 & 32 & 4 \\
    SIDER & 64 & 32 & 8 & 32 & 4 \\
    QM7 & 128 & 64 & 8 & 64 & 8 \\
    QM8 & 128 & 64 & 8 & 64 & 8 \\
    QM9 & 256 & 128 & 8 & 512 & 8 \\
    PCBA & 256 & 128 & 8 & 512 & 8 \\
    HIV & 256 & 128 & 8 & 512 & 4 \\
    ENZYMES & 256 & 128 & 8 & 64 & 8 \\
    PROTEINS\_full & 256 & 128 & 8 & 64 & 8 \\
    COIL-DEL & 256 & 128 & 8 & 64 & 8 \\
    COIL-RAG & 256 & 128 & 8 & 64 & 8 \\
    Cuneiform & 256 & 128 & 8 & 64 & 8 \\
    github\_stargazers & 256 & 128 & 8 & 64 & 8 \\
    IMDB-BINARY & 256 & 128 & 8 & 64 & 8 \\
    REDDIT-BINARY & 256 & 128 & 8 & 64 & 8 \\
    REDDIT-MULTI-12K & 256 & 128 & 8 & 64 & 8 \\
    reddit\_threads & 256 & 128 & 8 & 64 & 8 \\
    twitch\_egos & 256 & 128 & 8 & 64 & 8 \\
    \makecell[l]{TWITTER-Real-\\Graph-Partial} & 256 & 128 & 8 & 64 & 8 \\
    COLORS-3 & 256 & 128 & 8 & 64 & 8 \\
    SYNTHETIC & 256 & 128 & 8 & 64 & 8 \\
    SYNTHETICnew & 256 & 128 & 8 & 64 & 8 \\
    Synthie & 256 & 128 & 8 & 64 & 8 \\
    TRIANGLES & 256 & 128 & 8 & 64 & 8 \\
    MNIST & 256 & 128 & 8 & 512 & 4 \\
    CIFAR10 & 256 & 128 & 8 & 512 & 4 \\
    ZINC & 256 & 128 & 8 & 512 & 4 \\
    \bottomrule
    \end{tabular}
\end{table}
\renewcommand{\familydefault}{\rmdefault}
\clearpage

\renewcommand{\sfdefault}{lmss}
\renewcommand{\familydefault}{\sfdefault}
\begin{table}[H]
    \captionsetup{skip=5pt, labelfont=bf}
    \centering
    \caption{(continued from \Cref{section:nn-agg-hyperparams}, \Cref{table:neural-agg-hyperparams}) Summary of the hyperparameters used for neural aggregators for each dataset. The hyperparameter names follow the same notation as introduced in the main text. Dim., dimension.}
    \label{table:neural-agg-hyperparams-2}
    \renewcommand{\arraystretch}{1.3}
    \begin{tabular}{l@{\hskip 1cm}rr@{\hskip 1cm}rrr}
    \toprule
    \textbf{Dataset} & \multicolumn{2}{{c@{\hspace{1cm}}}}{\textbf{MLP}} & \multicolumn{3}{c}{\textbf{Set Transformer}} \\ \cmidrule(r{0.9cm}){1-1} \cmidrule(lr{1cm}){2-3} \cmidrule(lr){4-6}
     & \multicolumn{1}{l}{$d_1$} & \multicolumn{1}{l}{$d_{\text{out}}$} & \multicolumn{1}{l}{$k$} & \multicolumn{1}{l}{Hidden dim.} & \multicolumn{1}{l}{$n_h$} \\ \cmidrule(lr){2-2} \cmidrule(lr{0.9cm}){3-3} \cmidrule(lr){4-4} \cmidrule(lr){5-5} \cmidrule(lr){6-6}
    AIDS & 256 & 128 & 8 & 256 & 8 \\
    FRANKENSTEIN & 256 & 128 & 8 & 256 & 8 \\
    Mutagenicity & 256 & 128 & 8 & 256 & 8 \\
    MUTAG & 256 & 128 & 8 & 256 & 8 \\
    YeastH & 256 & 128 & 8 & 256 & 8 \\
    alchemy\_full & 256 & 128 & 8 & 256 & 8 \\
    MalNetTiny & 256 & 128 & 8 & 192 & 12 \\ \bottomrule
    \end{tabular}
\end{table}
\renewcommand{\familydefault}{\rmdefault}

\section{Best-to-best ratios for all datasets}
\label{section:ratios-all-datasets}
\begin{figure}[!h]
    \captionsetup{skip=4pt, labelfont=bf}
    \caption{The performance of the best neural relative to the best standard readout on all regression benchmarks. We use the ratio between the effectiveness scores ($\text{R}^2$), computed by averaging over five random splits of the data. The differences are best appreciated by studying the associated tables (\Cref{section:all-benchmarks-metrics}, \Cref{table:MOLNET-REGR,table:ZINC,table:alchemy}).}
    \label{figure:best-to-best-regr}
    \centering
    \includegraphics[width=1\textwidth]{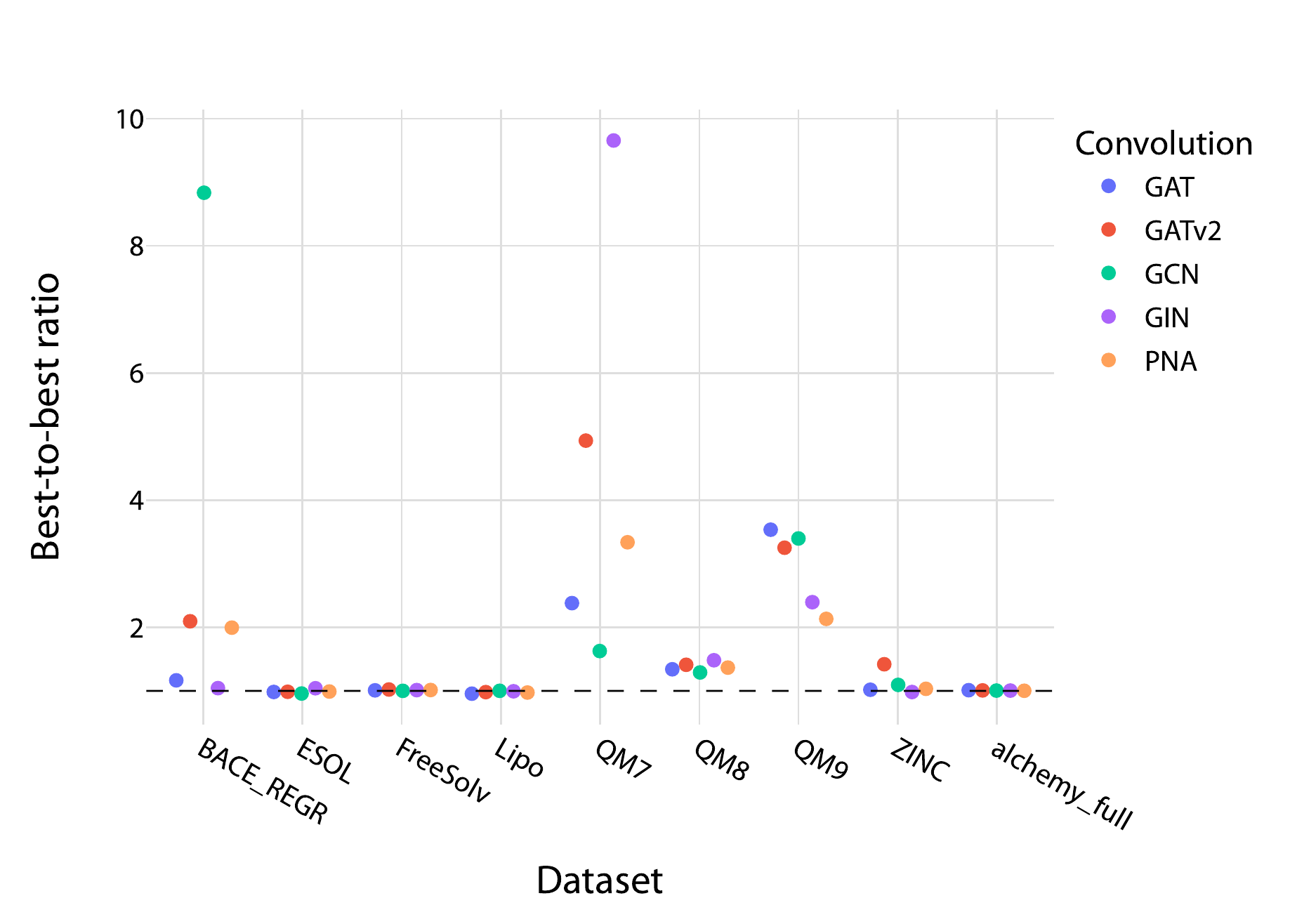}
\end{figure}
\renewcommand{\familydefault}{\rmdefault}

\begin{figure}[!h]
    \captionsetup{skip=4pt, labelfont=bf}
    \caption{The performance of the best neural relative to the best standard readout on several classification benchmarks. We use the ratio between the effectiveness scores (\textsc{mcc}), computed by averaging over five random splits of the data. The differences are best appreciated by studying the associated tables (\Cref{section:all-benchmarks-metrics}, \Cref{table:tud-extra,table:PROTEINS_full,table:ENZYMES-MCC,table:CV-TUD-MCC,table:SOCIAL,table:REDDIT-12-MCC}).}
    \label{figure:best-to-best-cls-1}
    \centering
    \includegraphics[width=0.95\textwidth]{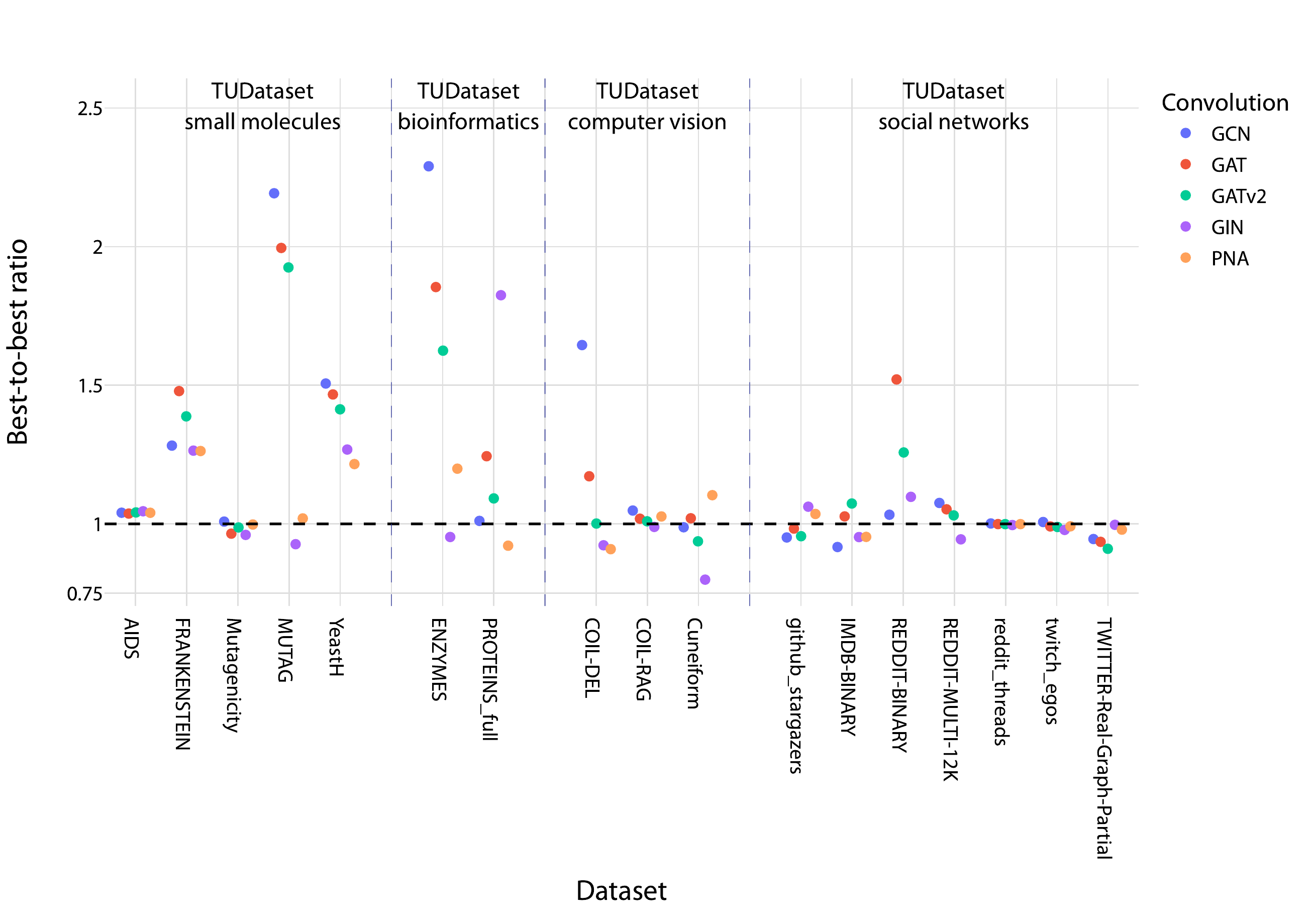}
\end{figure}
\renewcommand{\familydefault}{\rmdefault}

\begin{figure}[!h]
    \captionsetup{skip=4pt, labelfont=bf}
    \caption{The performance of the best neural relative to the best standard readout on the rest of the classification benchmarks. We use the ratio between the effectiveness scores (\textsc{mcc}), computed by averaging over five random splits of the data. The differences are best appreciated by studying the associated tables (\Cref{section:all-benchmarks-metrics}, \Cref{table:SYNTHETIC-BINARY,table:SYNTHETIC-MULTI,table:MOLNET-CLS,table:MNIST-CIFAR10-MCC,table:MalNetTiny-MCC}).}
    \label{figure:best-to-best-cls-2}
    \centering
    \includegraphics[width=0.95\textwidth]{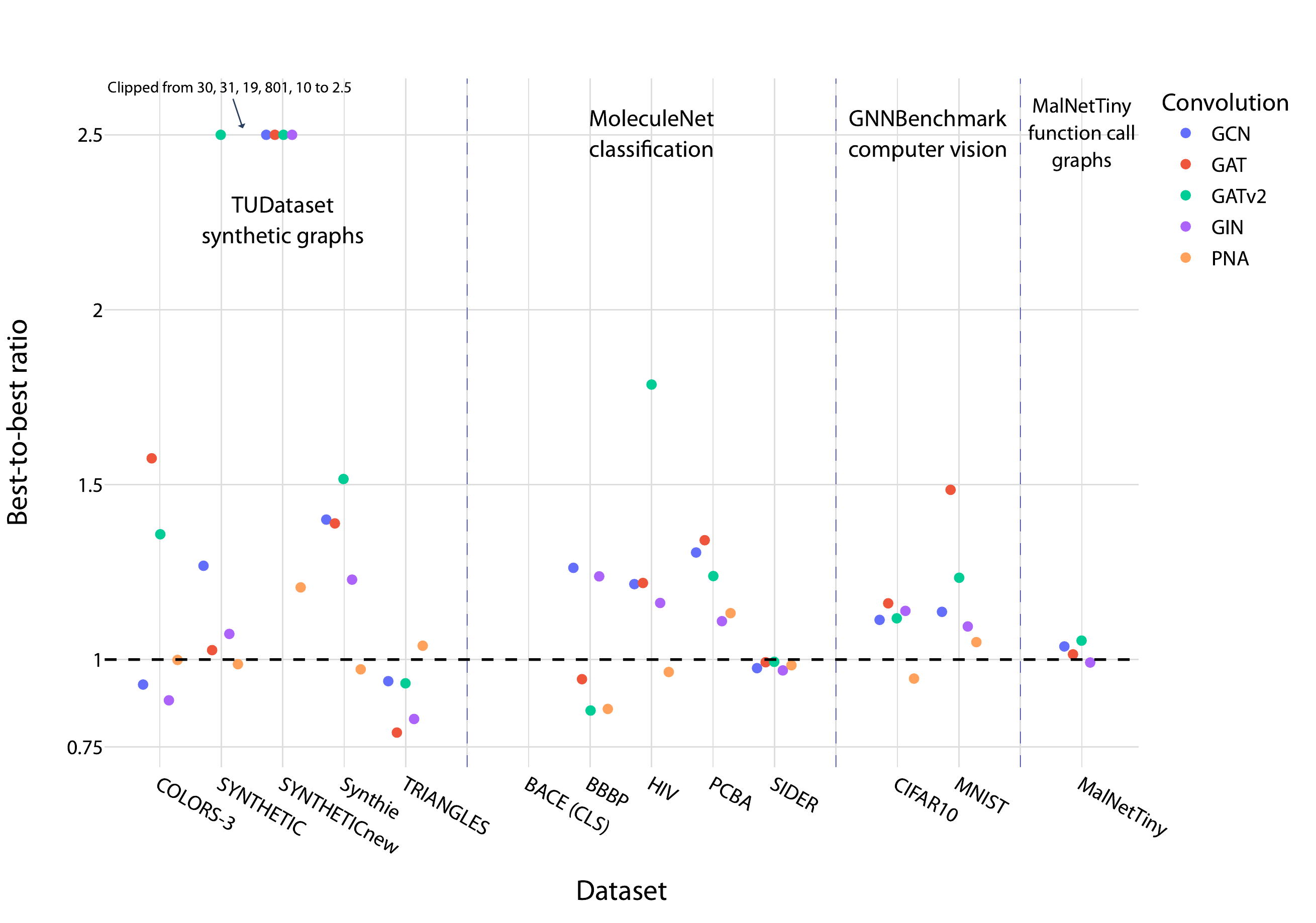}
\end{figure}
\renewcommand{\familydefault}{\rmdefault}

\clearpage
\section{Deeper GNN models for ENZYMES}
\label{section:deep-enzymes}

\begin{figure}[!h]
    \captionsetup{skip=4pt, labelfont=bf}
    \caption{Increasing the number of neighborhood aggregation iterations or convolutional layers does not produce large differences on \textsc{enzymes}. For \textsc{gcn}, the \textsc{mlp} readout improves with deeper networks, whereas most readout are relatively stable regarding the number of layers on \textsc{gat}, \textsc{gatv2}, and \textsc{gin}. For \textsc{pna}, the performance decreases drastically with more than $3$ layers for the majority of readouts.}
    \label{figure:enzymes-6-layers}
    \centering
    \includegraphics[width=1\textwidth]{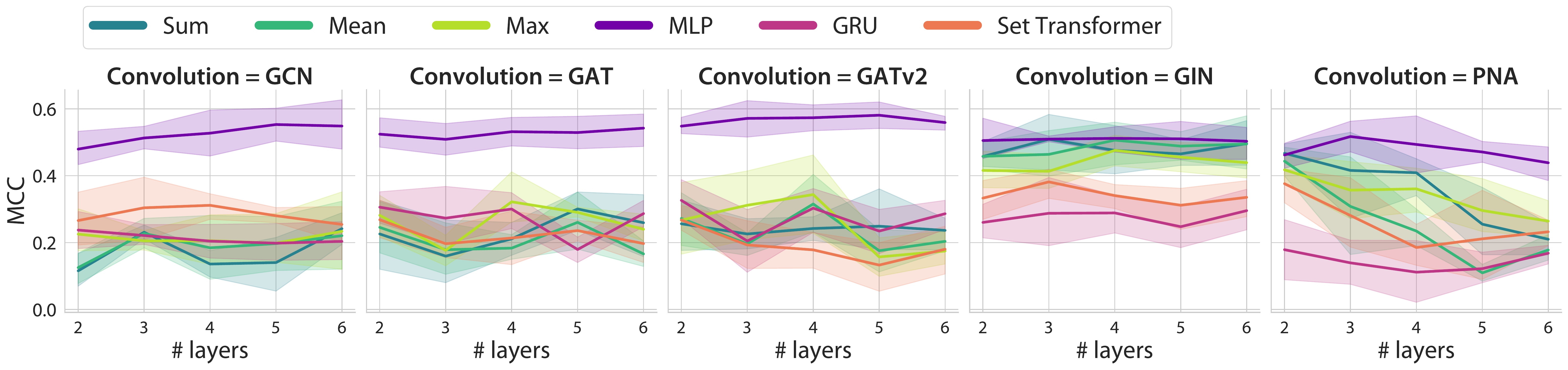}
\end{figure}
\renewcommand{\familydefault}{\rmdefault}

\section{Time and memory analysis}
\label{sec:time-mem}
We benchmarked the elapsed training time and memory utilization on the \textsc{qm9} dataset ($132,480$ data points) for one epoch on a modern high-end GPU (Nvidia RTX 3090, also see \Cref{sec:exp-platform}), averaged from $5$ epochs for each model (differences too small to plot error bars). As expected, the \textsc{janossy} variants are not competitive in terms of training time. However, the \textsc{mlp} and \textsc{gru} aggregators lead to a minimal increase of a few seconds per single epoch compared to the simple classical functions. The full \textsc{set transformer} architecture (also referred to as \textsc{st complex}) incurs an increase close to 50\%, which is, however, in line with the cost of transitioning from \textsc{gcn} to \textsc{pna}. The training cost can be minimized by adopting \textsc{isab} blocks (trading off performance). The \textsc{set transformer} readout is applicable to large scale datasets, as exemplified by our evaluation which includes \textsc{set transformer} + \textsc{pna} models with a maximum of $4$ \textsc{gnn} layers for datasets of up to $2$ million data points (also tested for $5$ \textsc{pna} layers).

\begin{figure}[H]
    \captionsetup{justification=centering, skip=4pt, labelfont=bf}
    \caption{Training times for all aggregators and convolutions on \textsc{qm9} (seconds). The models are $2$-layer \textsc{GNN}s. J., Janossy.}
    \label{figure:QM9-training-time-benchmark}
    \centering
    \includegraphics[width=1\textwidth]{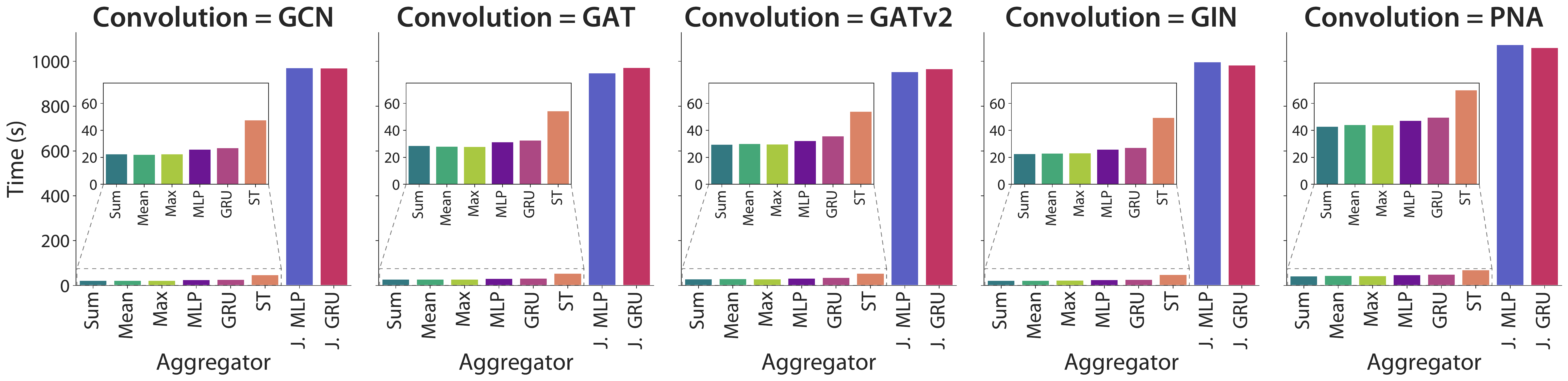}
\end{figure}

In terms of memory usage, the maximum amount of reserved (higher than allocated) memory as reported by the PyTorch profiler (version 1.10.1) was just under $149$MB for the \textsc{gru} + \textsc{pna} model, a $27$MB increase from the most efficient non-neural aggregator for \textsc{pna} (mean). It should be noted that it is common for deep learning frameworks such as PyTorch and TensorFlow to automatically reserve or prepare large amounts of memory even if only a portion is allocated during training.

% \clearpage

% \clearpage
\section{Robustness to random node permutations of QM9 molecules}
\label{sec:QM9-perms-plots}

\subsection{Random permutations of nodes}
\begin{figure}[H]
    \captionsetup{skip=4pt, labelfont=bf}
    \caption{A summary of the error distributions for predictions made on random permutations of $50$ randomly selected molecules from the \textsc{qm9} dataset, presented per-convolution for the arbitrary random permutations strategy.}
    \label{figure:qm9-perm-nodes-conv}
    \centering
    \includegraphics[width=0.8\textwidth]{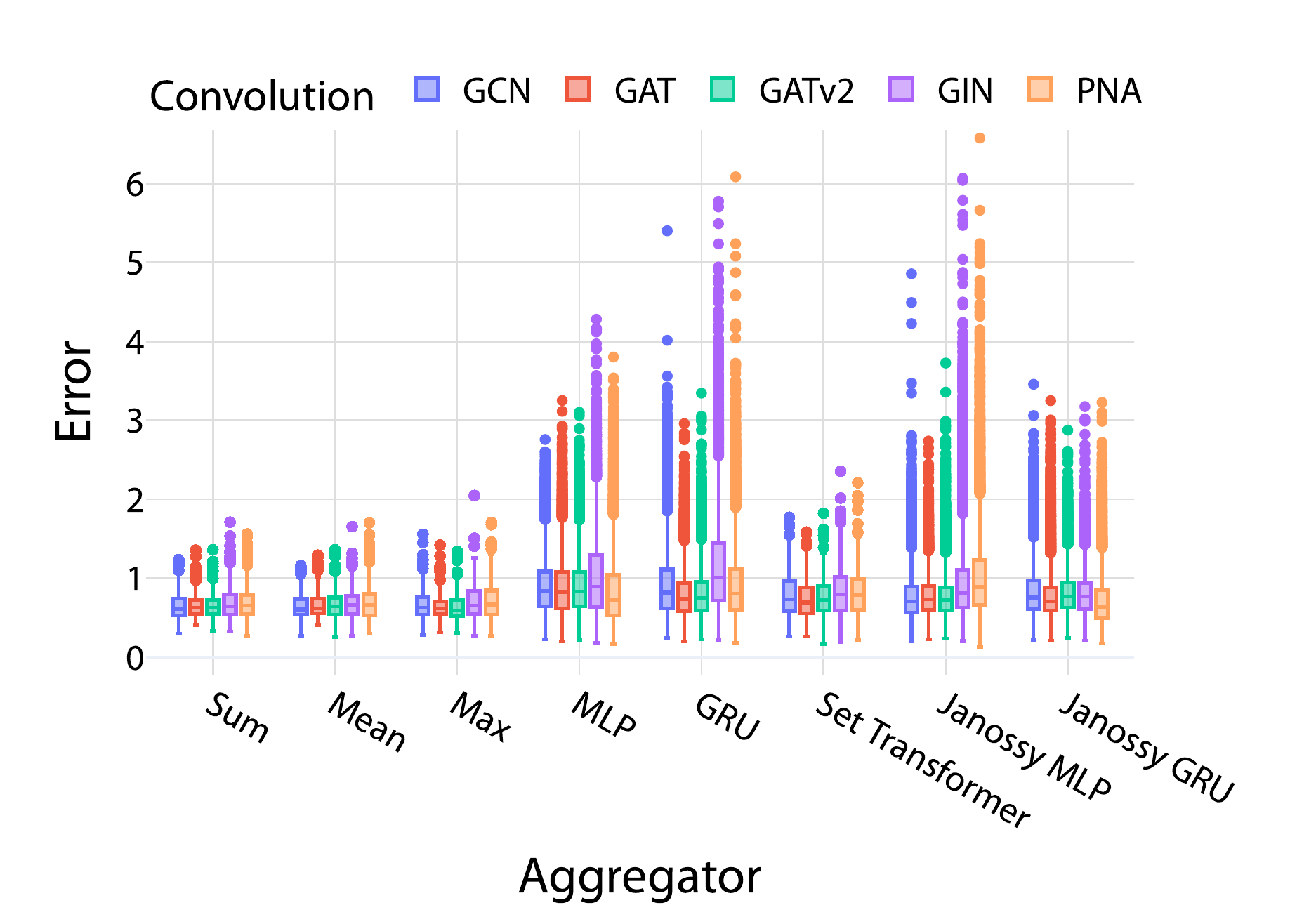}
\end{figure}
\renewcommand{\familydefault}{\rmdefault}

\subsection{Random, non-canonical \textsc{smiles}}
\begin{figure}[H]
    \captionsetup{skip=4pt, labelfont=bf}
    \caption{A summary of the error distributions for predictions made on random permutations of $50$ randomly selected molecules from the \textsc{qm9} dataset, presented per-convolution for the random non-canonical \textsc{smiles} strategy.}
    \label{figure:qm9-perm-smiles-conv}
    \centering
    \includegraphics[width=0.8\textwidth]{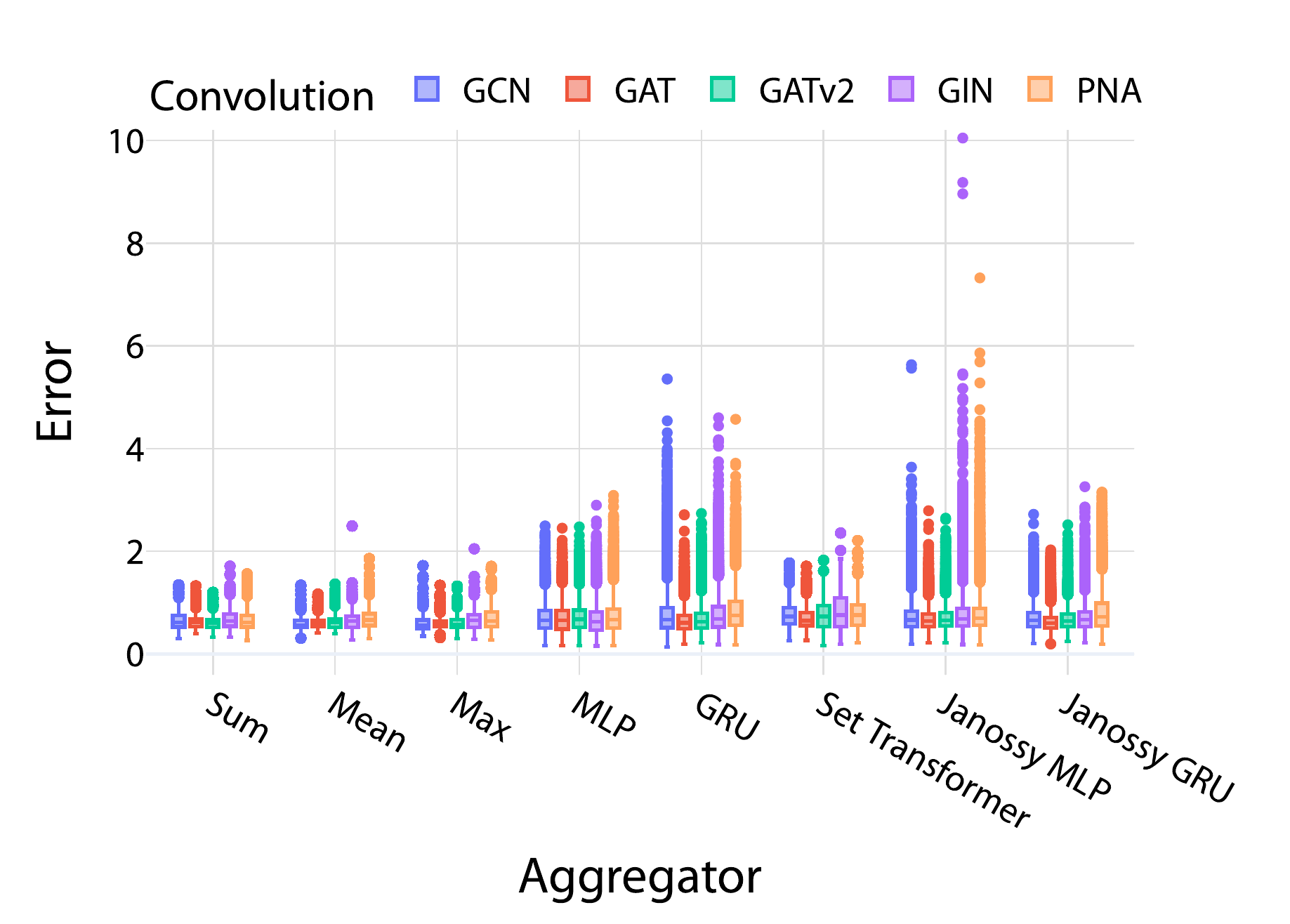}
\end{figure}
\renewcommand{\familydefault}{\rmdefault}

\begin{figure}[H]
    \captionsetup{skip=4pt, labelfont=bf}
    \caption{A summary of the error distributions for predictions made on random permutations of $50$ randomly selected molecules from the \textsc{qm9} dataset for the random non-canonical \textsc{smiles} strategy, where we selected the top $50$ lowest errors for each molecule from the multitude of non-canonical \textsc{smiles} inputs.}
    \label{figure:qm9-perm-smiles-best}
    \centering
    \includegraphics[width=0.9\textwidth]{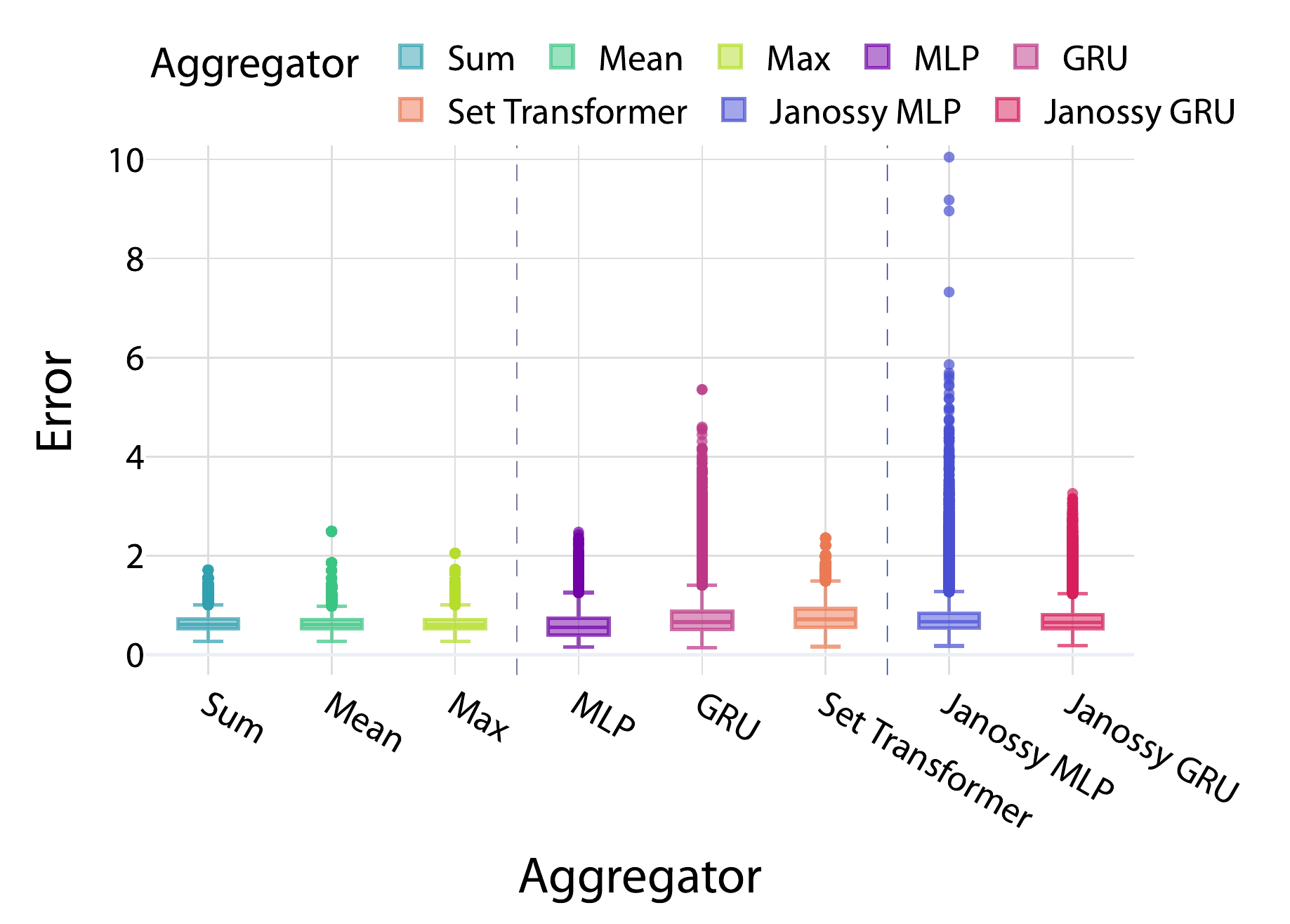}
\end{figure}
\renewcommand{\familydefault}{\rmdefault}

\subsection{Error for each QM9 prediction task for random permutations of nodes}
\begin{figure}[H]
    \captionsetup{skip=4pt, labelfont=bf}
    \caption{A summary of the error distributions for predictions made on random permutations of $50$ randomly selected molecules from the \textsc{qm9} dataset, presented per \textsc{qm9} task ($12$ in total) for the arbitrary random permutations strategy.}
    \label{figure:qm9-perm-nodes-individual}
    \centering

    \begin{subfigure}[b]{0.24\textwidth}
        \centering
        % \captionsetup{labelfont=bf, font=small, skip=-1pt}
        % \caption{...} \label{subfigure:random-perm-smiles}
        \includegraphics[width=0.9\textwidth]{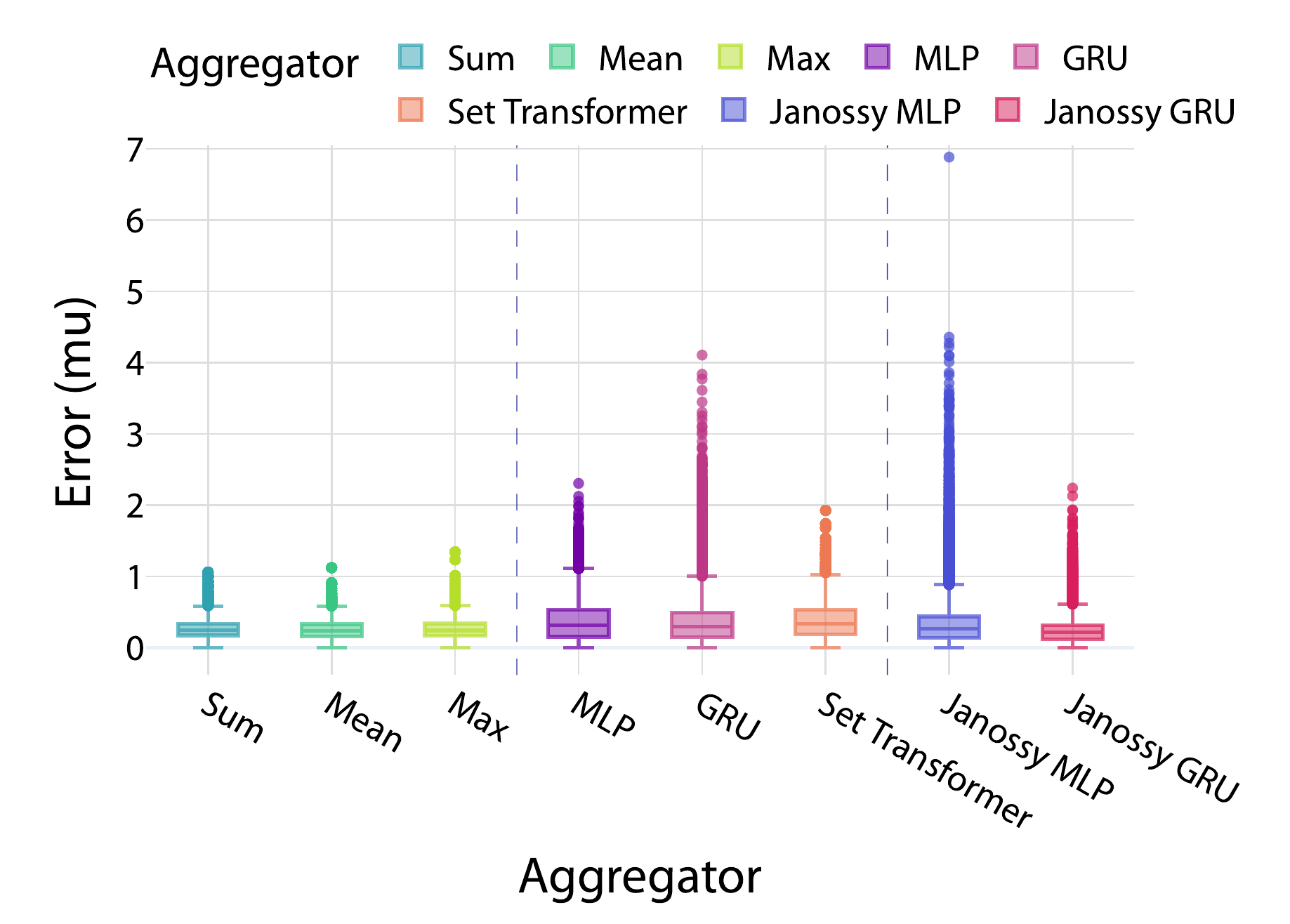}
    \end{subfigure}
    \hfill
    \begin{subfigure}[b]{0.24\textwidth}
        \centering
        % \captionsetup{labelfont=bf, font=small, skip=-1pt}
        % \caption{...} \label{subfigure:random-perm-smiles}
        \includegraphics[width=0.9\textwidth]{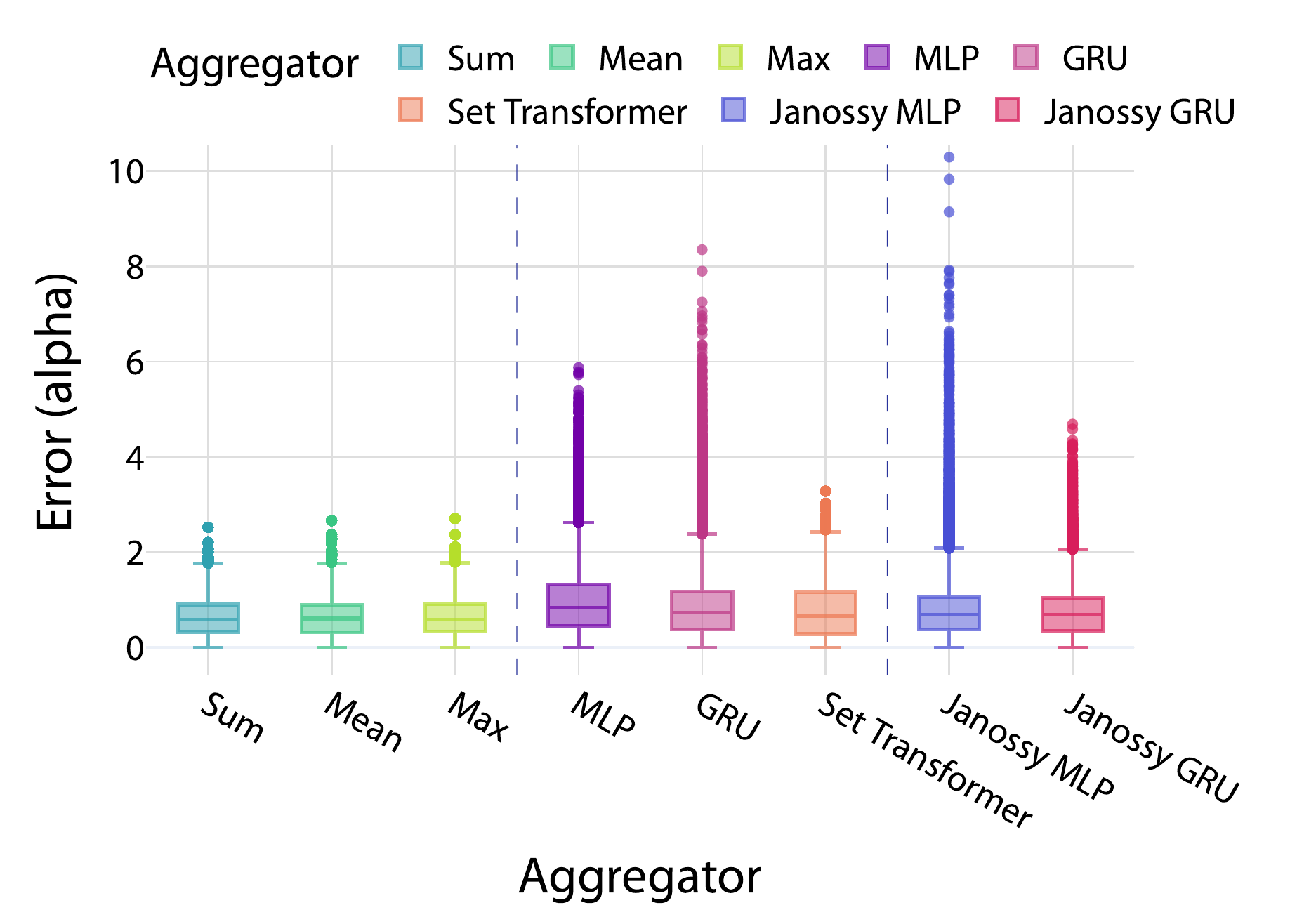}
    \end{subfigure}
    \hfill
    \begin{subfigure}[b]{0.24\textwidth}
        \centering
        % \captionsetup{labelfont=bf, font=small, skip=-1pt}
        % \caption{...} \label{subfigure:random-perm-smiles}
        \includegraphics[width=0.9\textwidth]{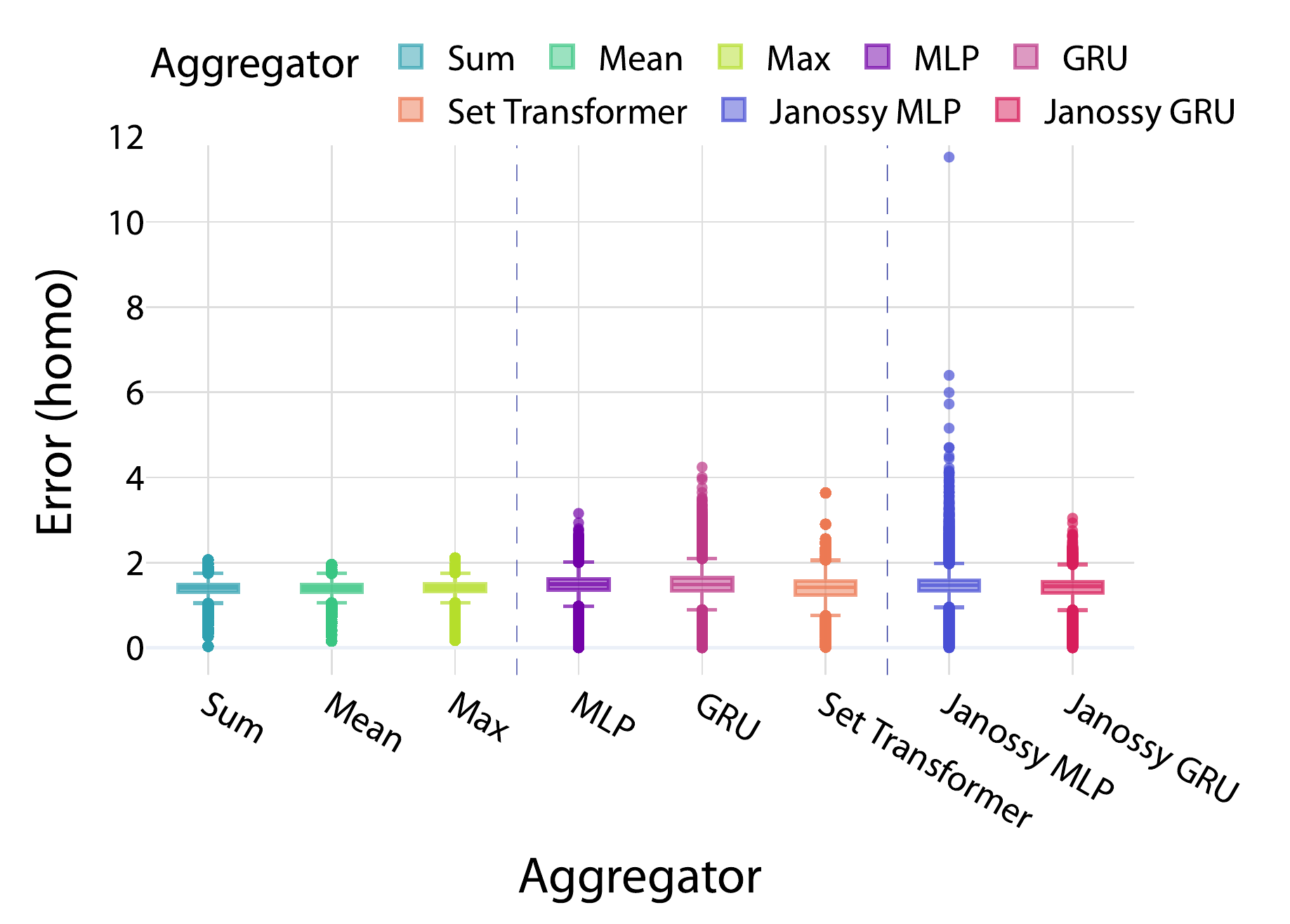}
    \end{subfigure}
    \hfill
    \begin{subfigure}[b]{0.24\textwidth}
        \centering
        % \captionsetup{labelfont=bf, font=small, skip=-1pt}
        % \caption{...} \label{subfigure:random-perm-smiles}
        \includegraphics[width=0.9\textwidth]{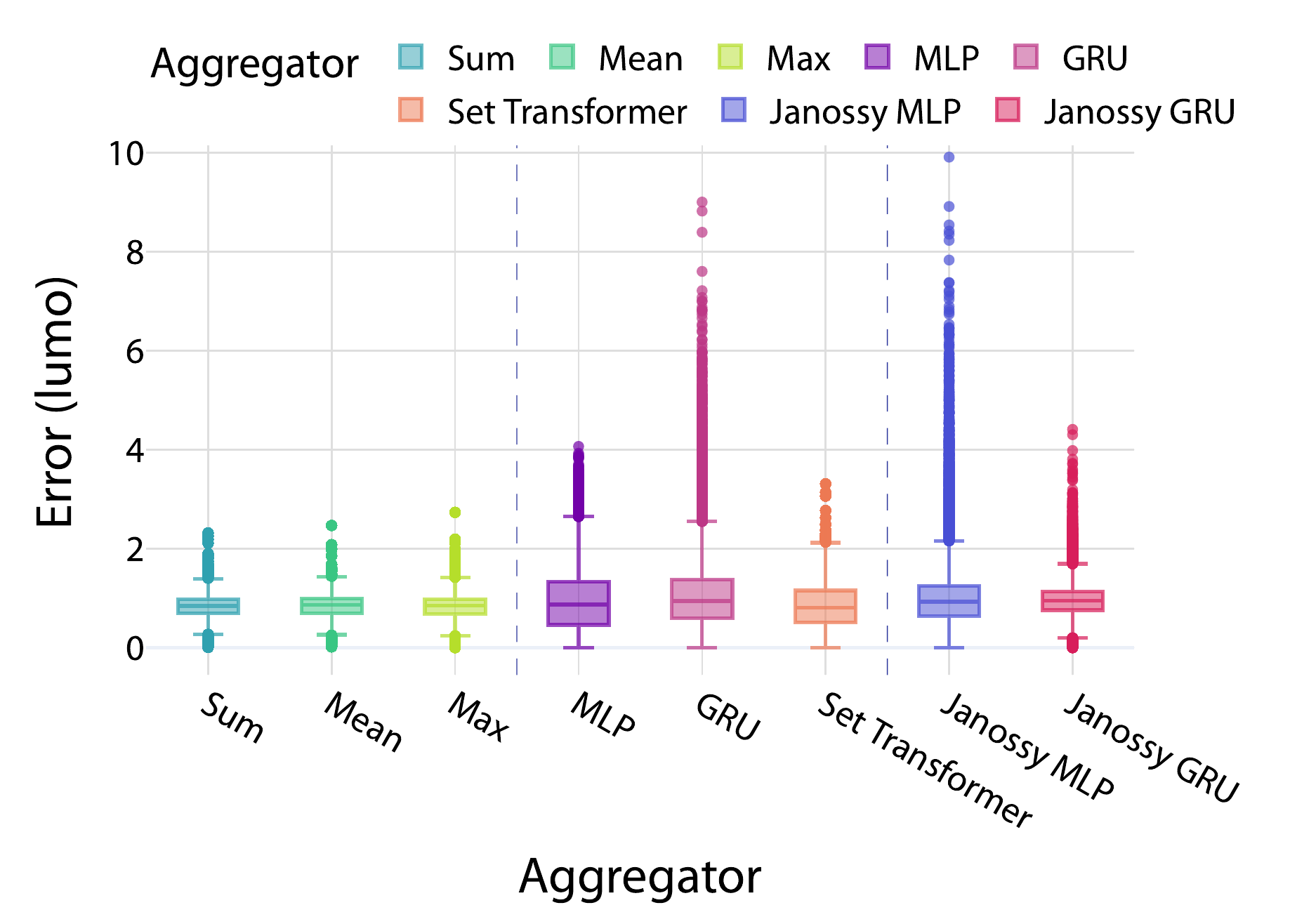}
    \end{subfigure}

    \begin{subfigure}[b]{0.24\textwidth}
        \centering
        % \captionsetup{labelfont=bf, font=small, skip=-1pt}
        % \caption{...} \label{subfigure:random-perm-smiles}
        \includegraphics[width=0.9\textwidth]{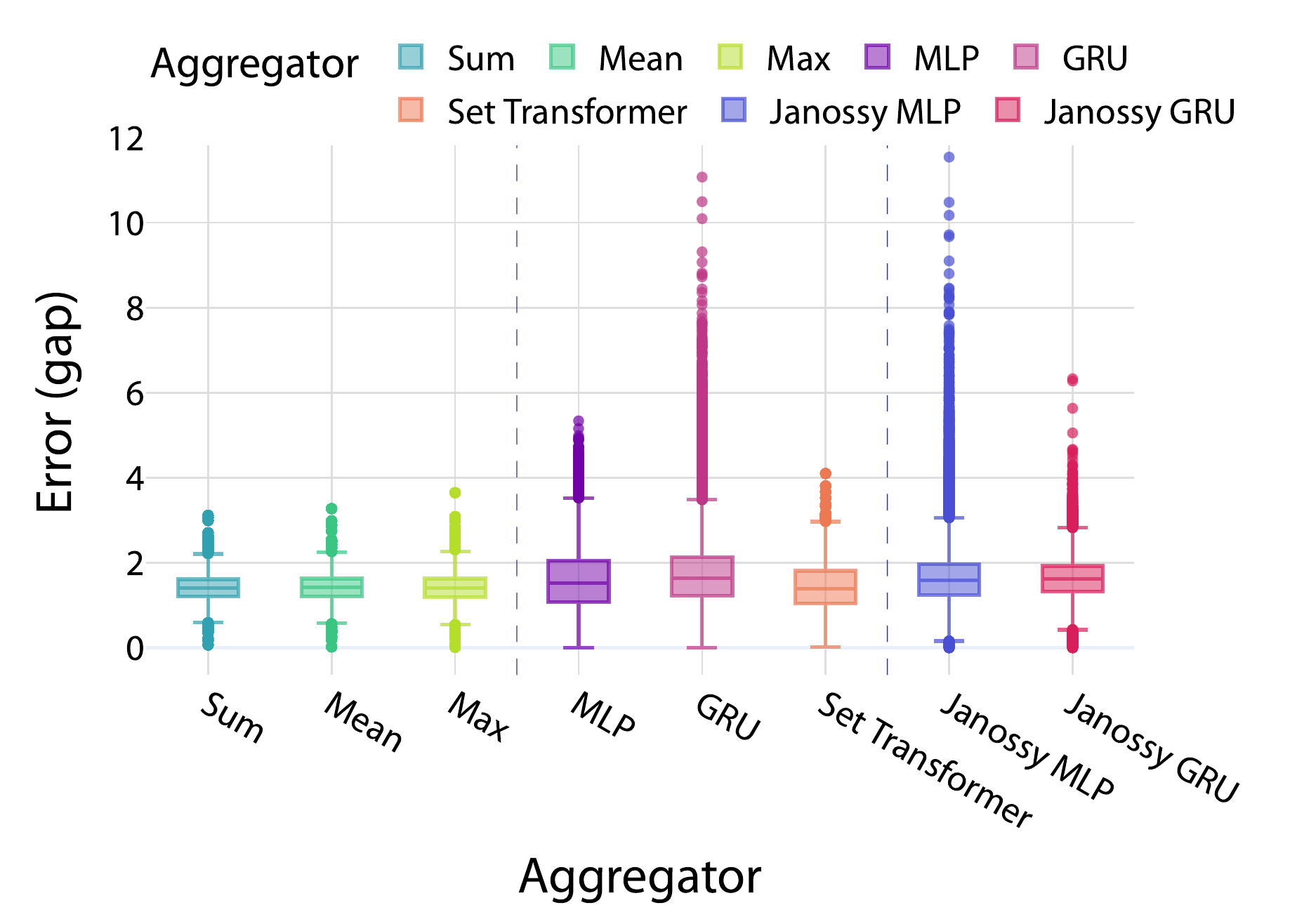}
    \end{subfigure}
    \hfill
    \begin{subfigure}[b]{0.24\textwidth}
        \centering
        % \captionsetup{labelfont=bf, font=small, skip=-1pt}
        % \caption{...} \label{subfigure:random-perm-smiles}
        \includegraphics[width=0.9\textwidth]{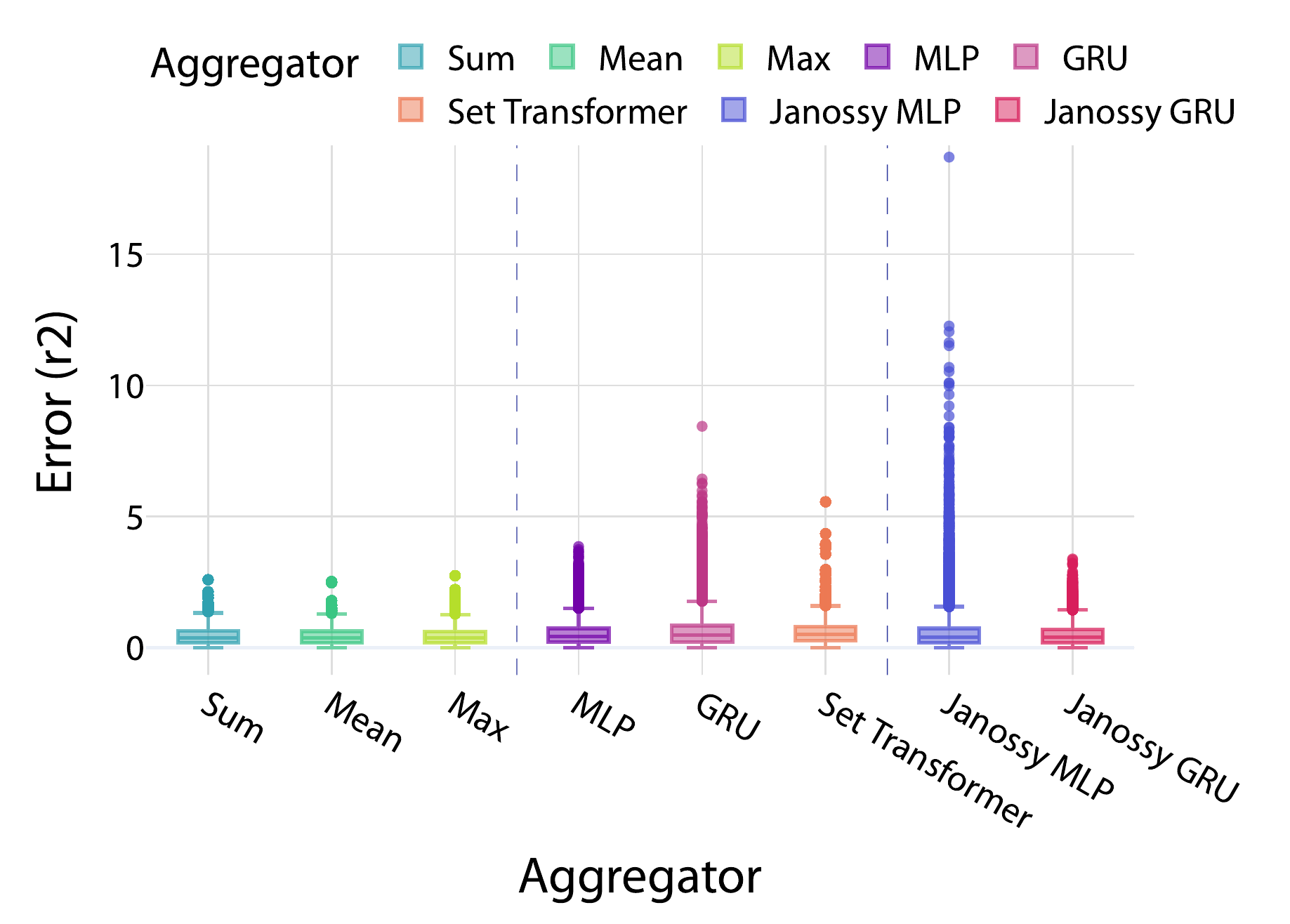}
    \end{subfigure}
    \hfill
    \begin{subfigure}[b]{0.24\textwidth}
        \centering
        % \captionsetup{labelfont=bf, font=small, skip=-1pt}
        % \caption{...} \label{subfigure:random-perm-smiles}
        \includegraphics[width=0.9\textwidth]{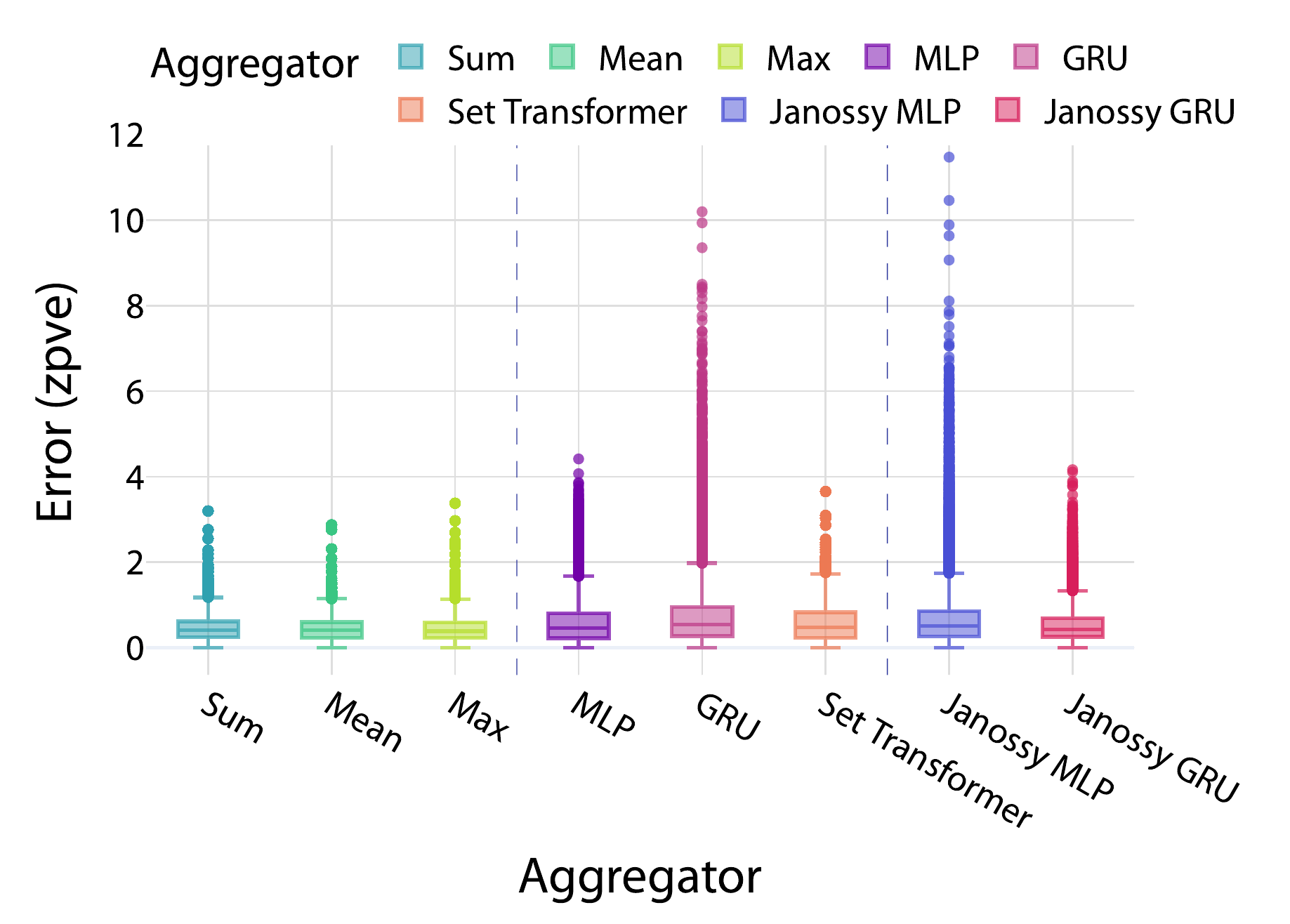}
    \end{subfigure}
    \hfill
    \begin{subfigure}[b]{0.24\textwidth}
        \centering
        % \captionsetup{labelfont=bf, font=small, skip=-1pt}
        % \caption{...} \label{subfigure:random-perm-smiles}
        \includegraphics[width=0.9\textwidth]{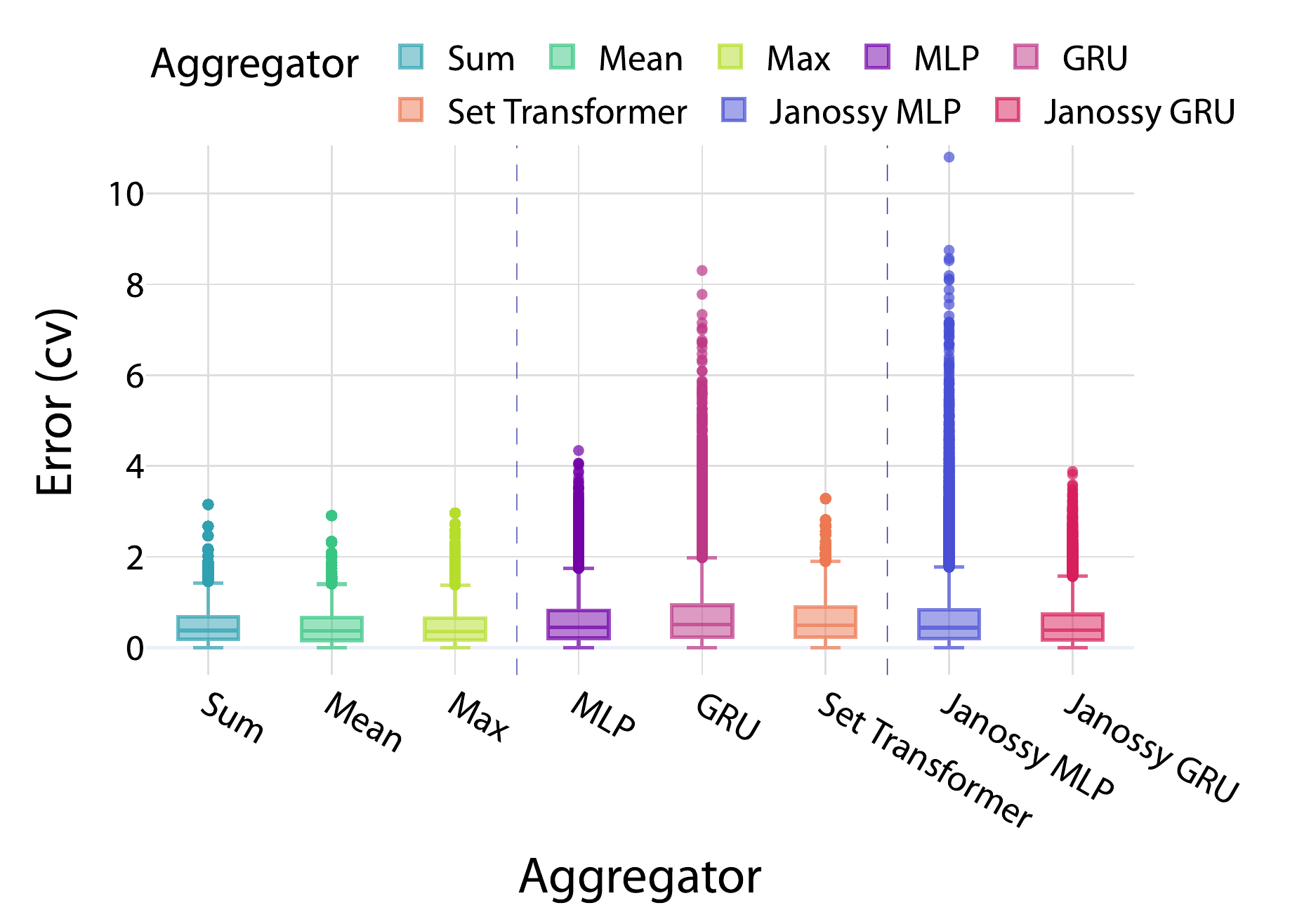}
    \end{subfigure}

    \begin{subfigure}[b]{0.24\textwidth}
        \centering
        % \captionsetup{labelfont=bf, font=small, skip=-1pt}
        % \caption{...} \label{subfigure:random-perm-smiles}
        \includegraphics[width=0.9\textwidth]{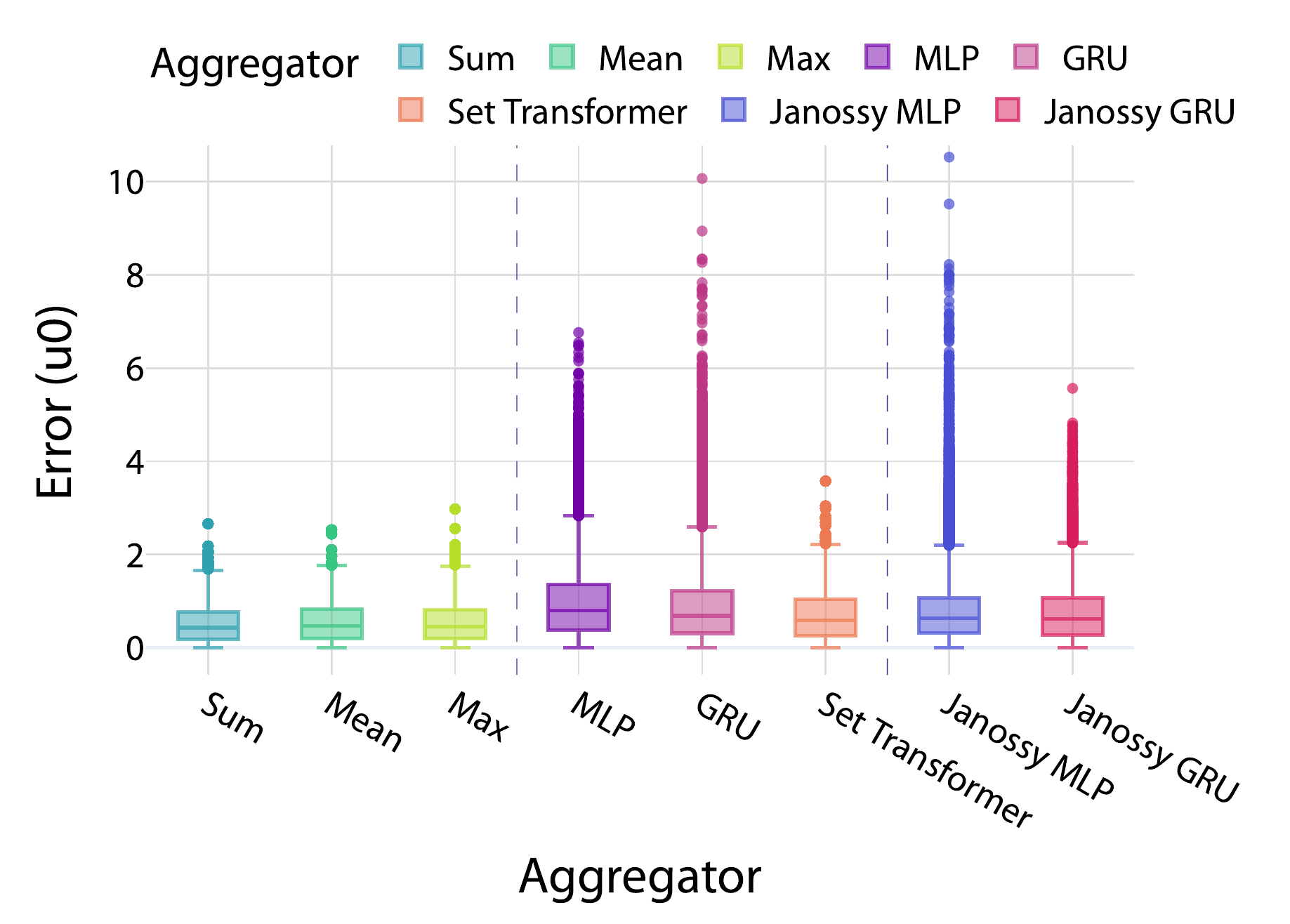}
    \end{subfigure}
    \hfill
    \begin{subfigure}[b]{0.24\textwidth}
        \centering
        % \captionsetup{labelfont=bf, font=small, skip=-1pt}
        % \caption{...} \label{subfigure:random-perm-smiles}
        \includegraphics[width=0.9\textwidth]{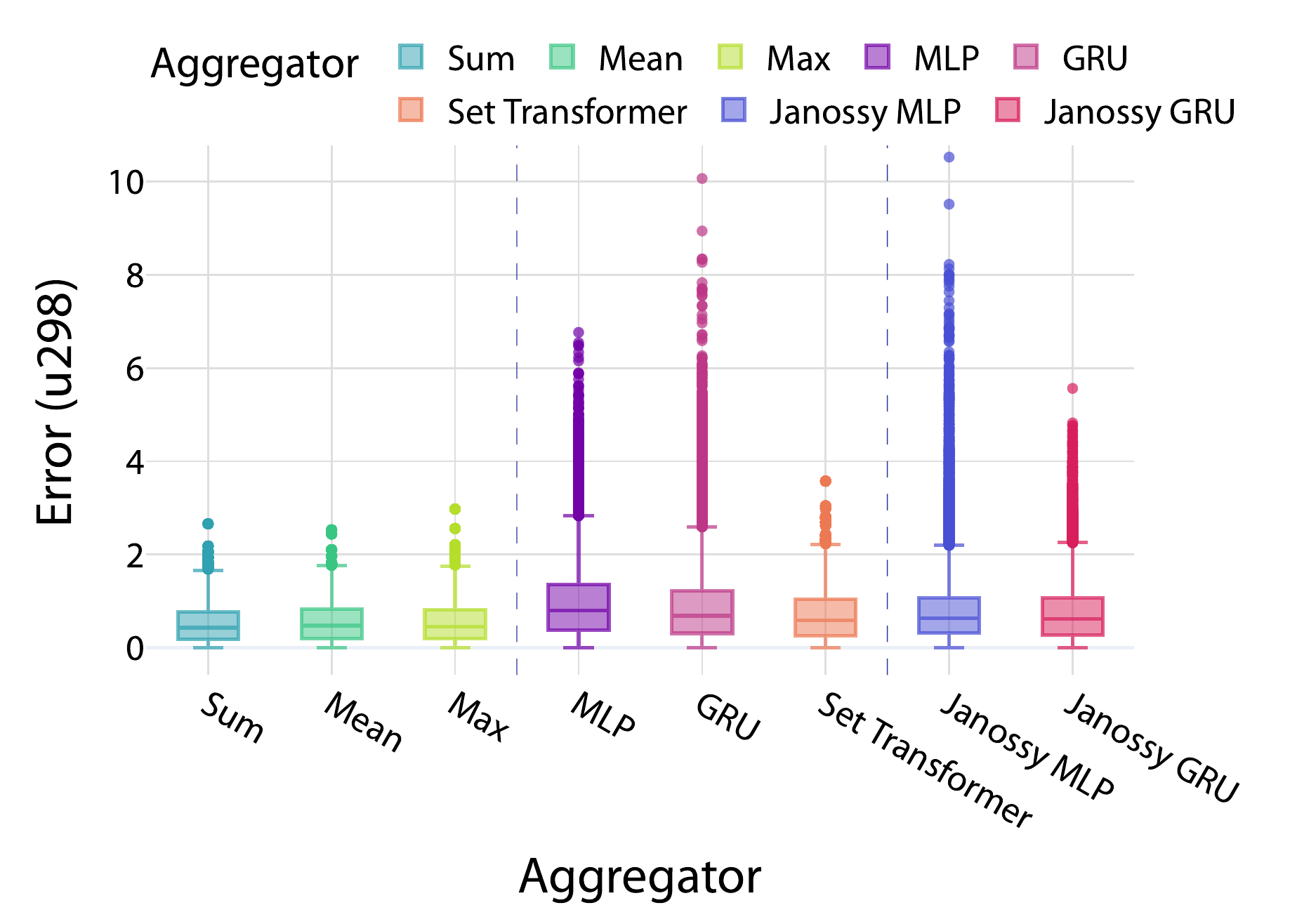}
    \end{subfigure}
    \hfill
    \begin{subfigure}[b]{0.24\textwidth}
        \centering
        % \captionsetup{labelfont=bf, font=small, skip=-1pt}
        % \caption{...} \label{subfigure:random-perm-smiles}
        \includegraphics[width=0.9\textwidth]{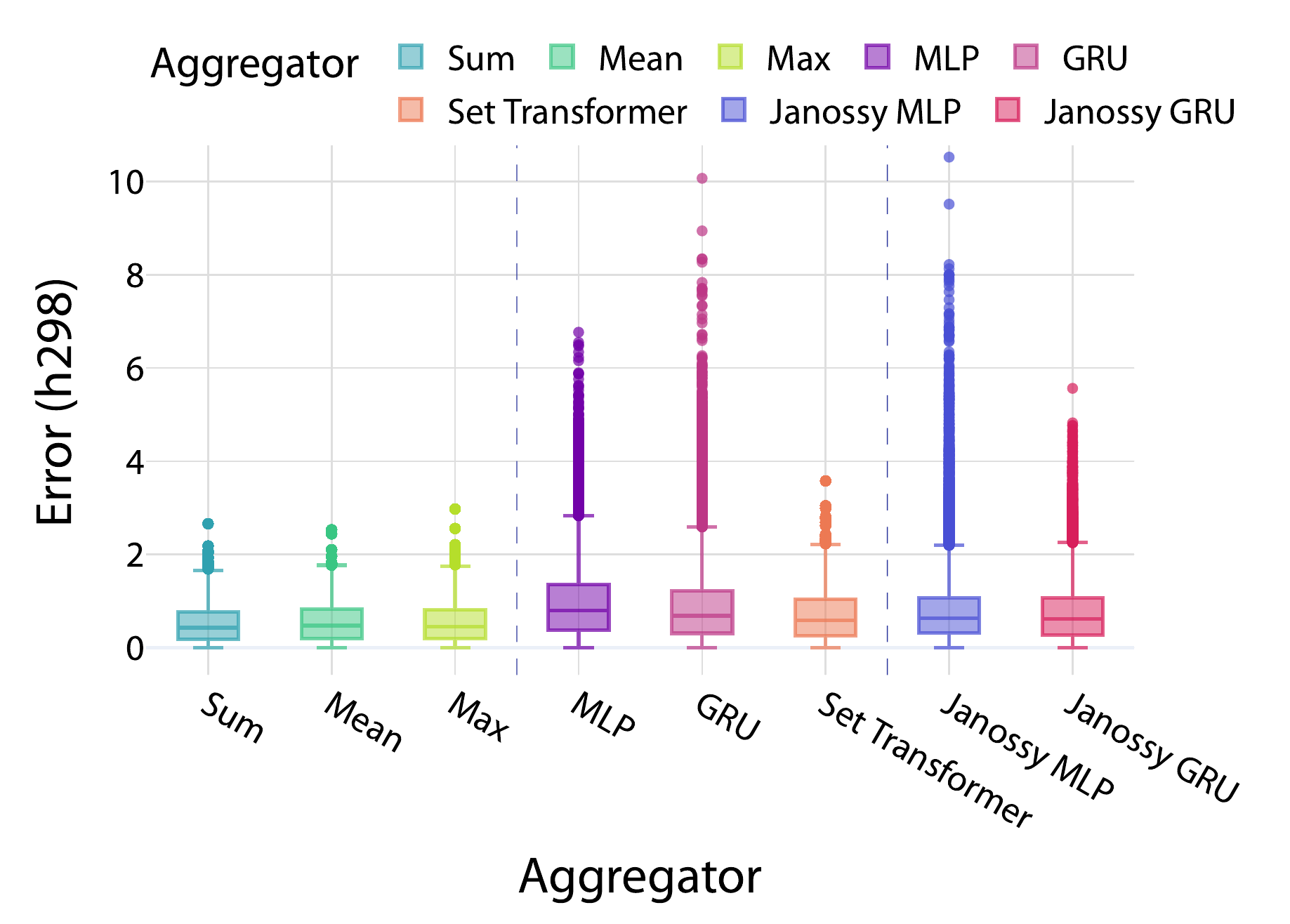}
    \end{subfigure}
    \hfill
    \begin{subfigure}[b]{0.24\textwidth}
        \centering
        % \captionsetup{labelfont=bf, font=small, skip=-1pt}
        % \caption{...} \label{subfigure:random-perm-smiles}
        \includegraphics[width=0.9\textwidth]{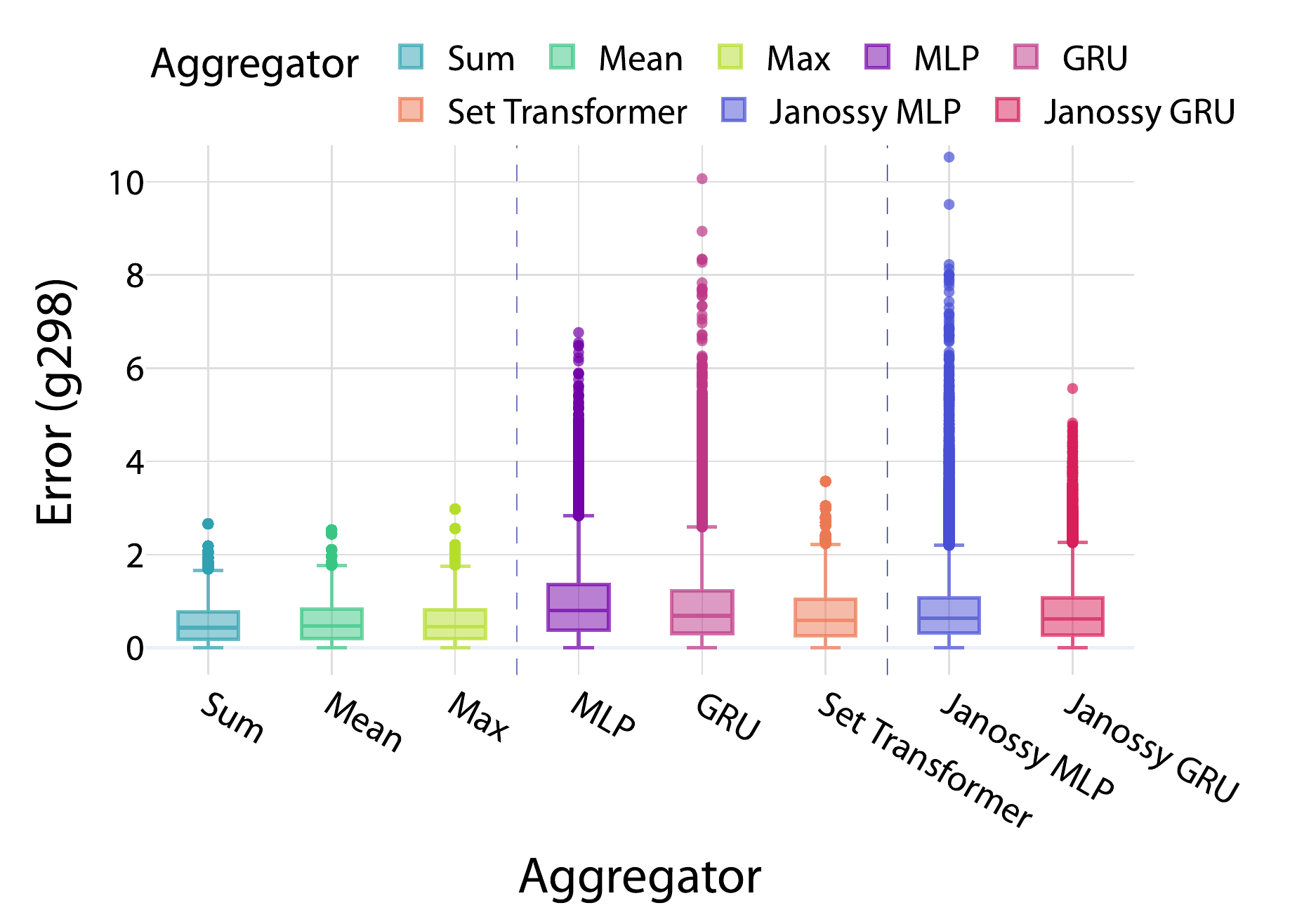}
    \end{subfigure}

\end{figure}
\renewcommand{\familydefault}{\rmdefault}

\subsection{Error for each QM9 prediction task for random \textsc{smiles}}
\begin{figure}[H]
    \captionsetup{skip=4pt, labelfont=bf}
    \caption{A summary of the error distributions for predictions made on random permutations of $50$ randomly selected molecules from the \textsc{qm9} dataset, presented per \textsc{qm9} task ($12$ in total) for the random non-canonical \textsc{smiles} strategy.}
    \label{figure:qm9-perm-smiles-individual}
    \centering

    \begin{subfigure}[b]{0.24\textwidth}
        \centering
        % \captionsetup{labelfont=bf, font=small, skip=-1pt}
        % \caption{...} \label{subfigure:random-perm-smiles}
        \includegraphics[width=0.9\textwidth]{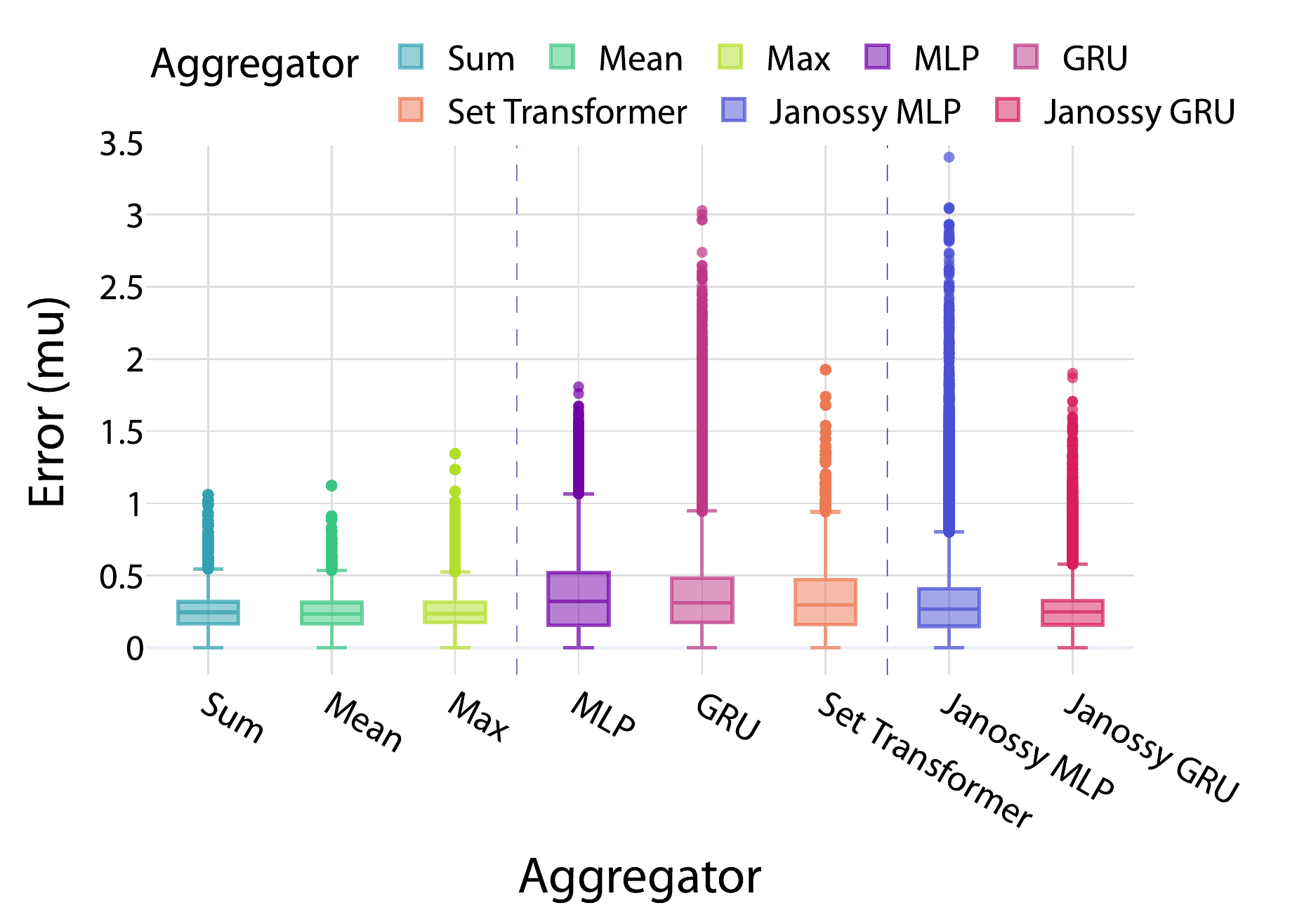}
    \end{subfigure}
    \hfill
    \begin{subfigure}[b]{0.24\textwidth}
        \centering
        % \captionsetup{labelfont=bf, font=small, skip=-1pt}
        % \caption{...} \label{subfigure:random-perm-smiles}
        \includegraphics[width=0.9\textwidth]{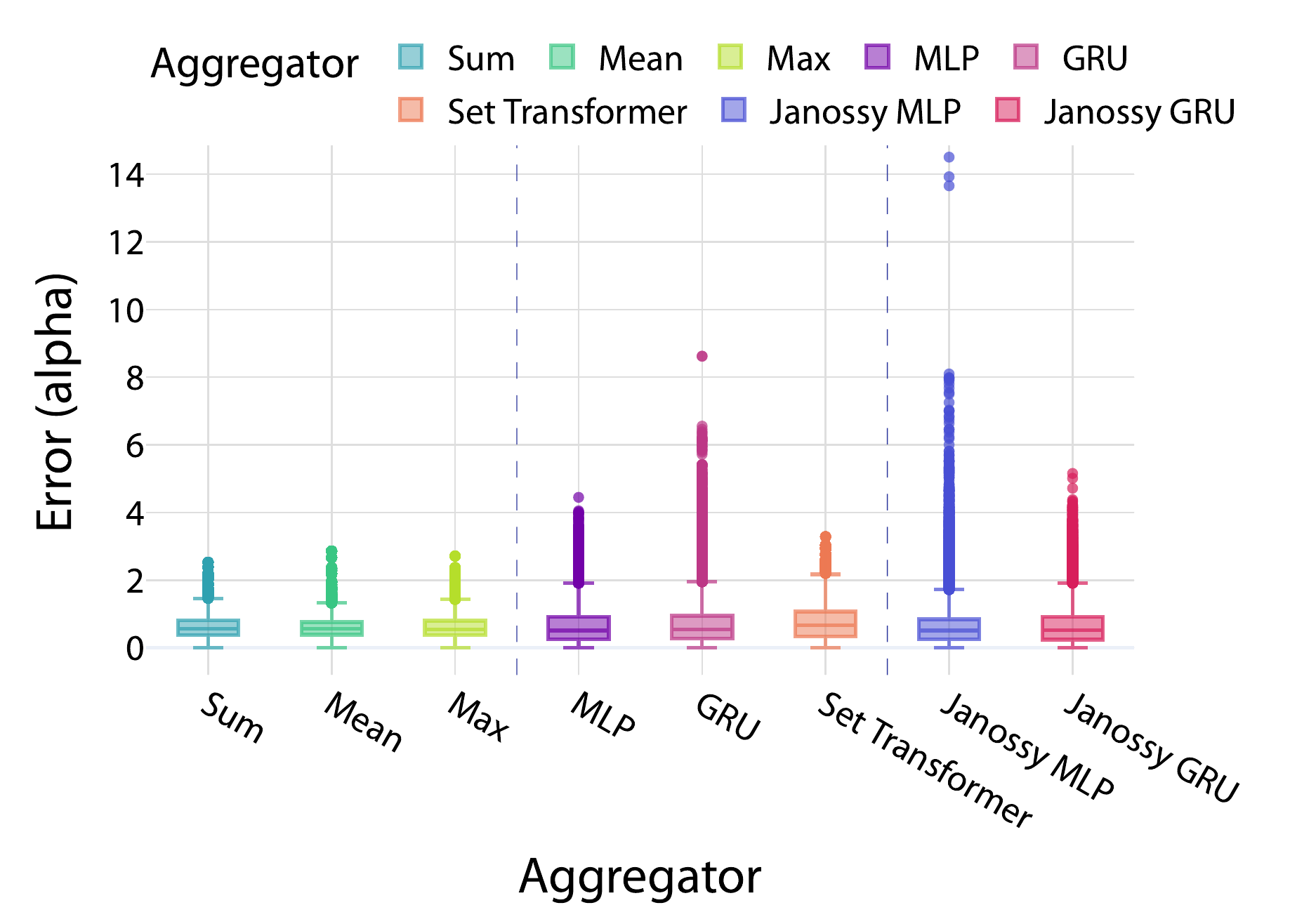}
    \end{subfigure}
    \hfill
    \begin{subfigure}[b]{0.24\textwidth}
        \centering
        % \captionsetup{labelfont=bf, font=small, skip=-1pt}
        % \caption{...} \label{subfigure:random-perm-smiles}
        \includegraphics[width=0.9\textwidth]{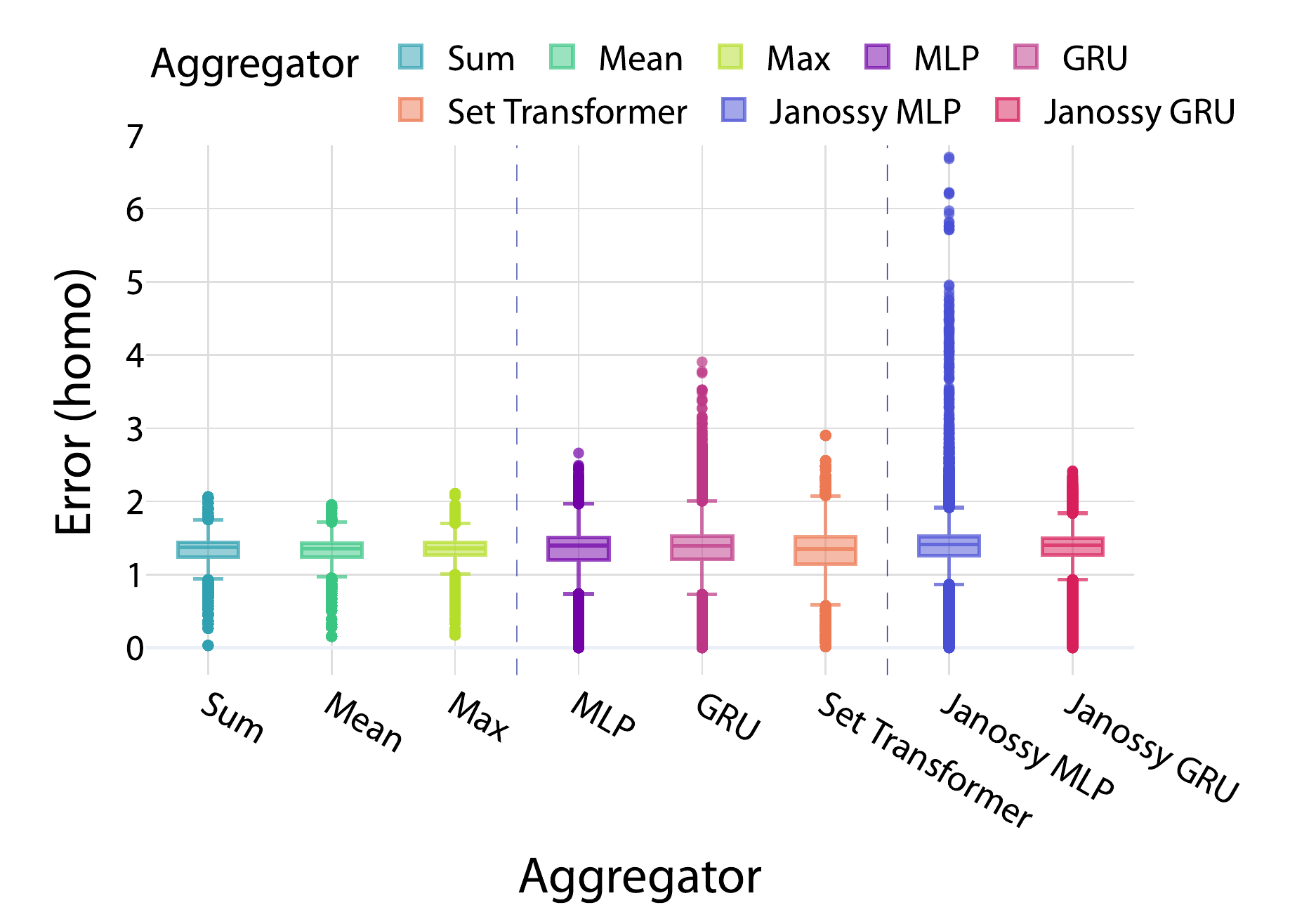}
    \end{subfigure}
    \hfill
    \begin{subfigure}[b]{0.24\textwidth}
        \centering
        % \captionsetup{labelfont=bf, font=small, skip=-1pt}
        % \caption{...} \label{subfigure:random-perm-smiles}
        \includegraphics[width=0.9\textwidth]{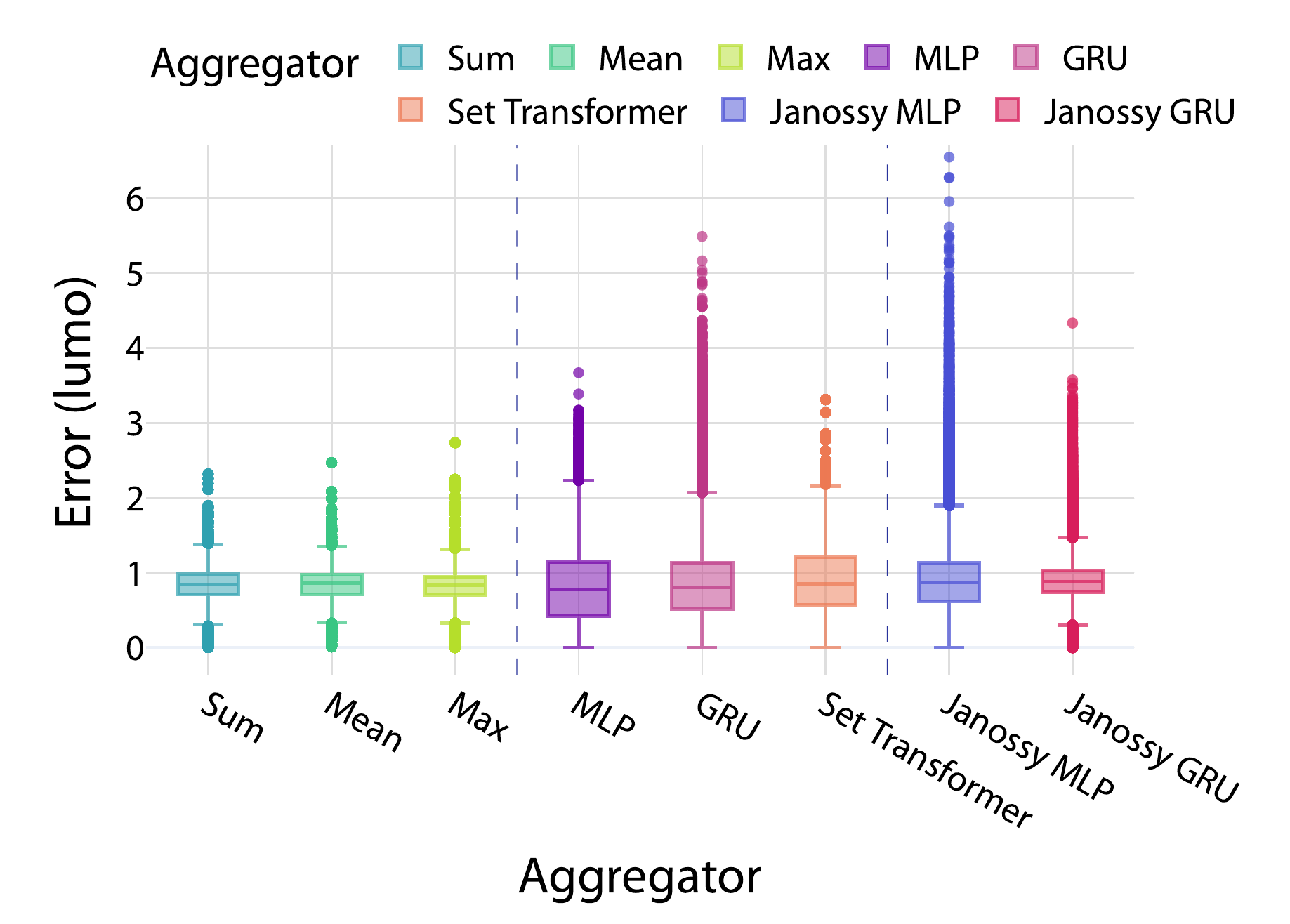}
    \end{subfigure}

    \begin{subfigure}[b]{0.24\textwidth}
        \centering
        % \captionsetup{labelfont=bf, font=small, skip=-1pt}
        % \caption{...} \label{subfigure:random-perm-smiles}
        \includegraphics[width=0.9\textwidth]{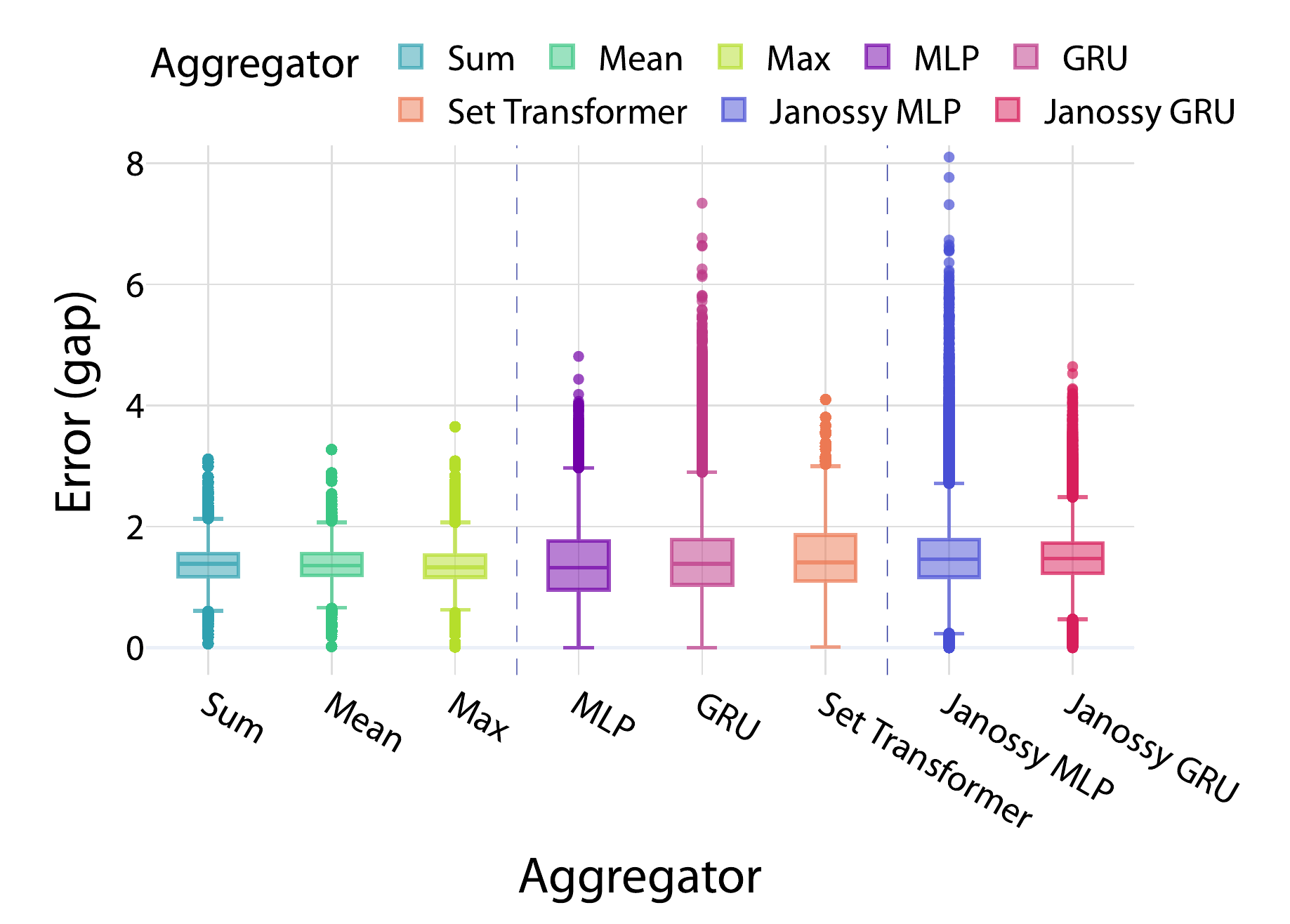}
    \end{subfigure}
    \hfill
    \begin{subfigure}[b]{0.24\textwidth}
        \centering
        % \captionsetup{labelfont=bf, font=small, skip=-1pt}
        % \caption{...} \label{subfigure:random-perm-smiles}
        \includegraphics[width=0.9\textwidth]{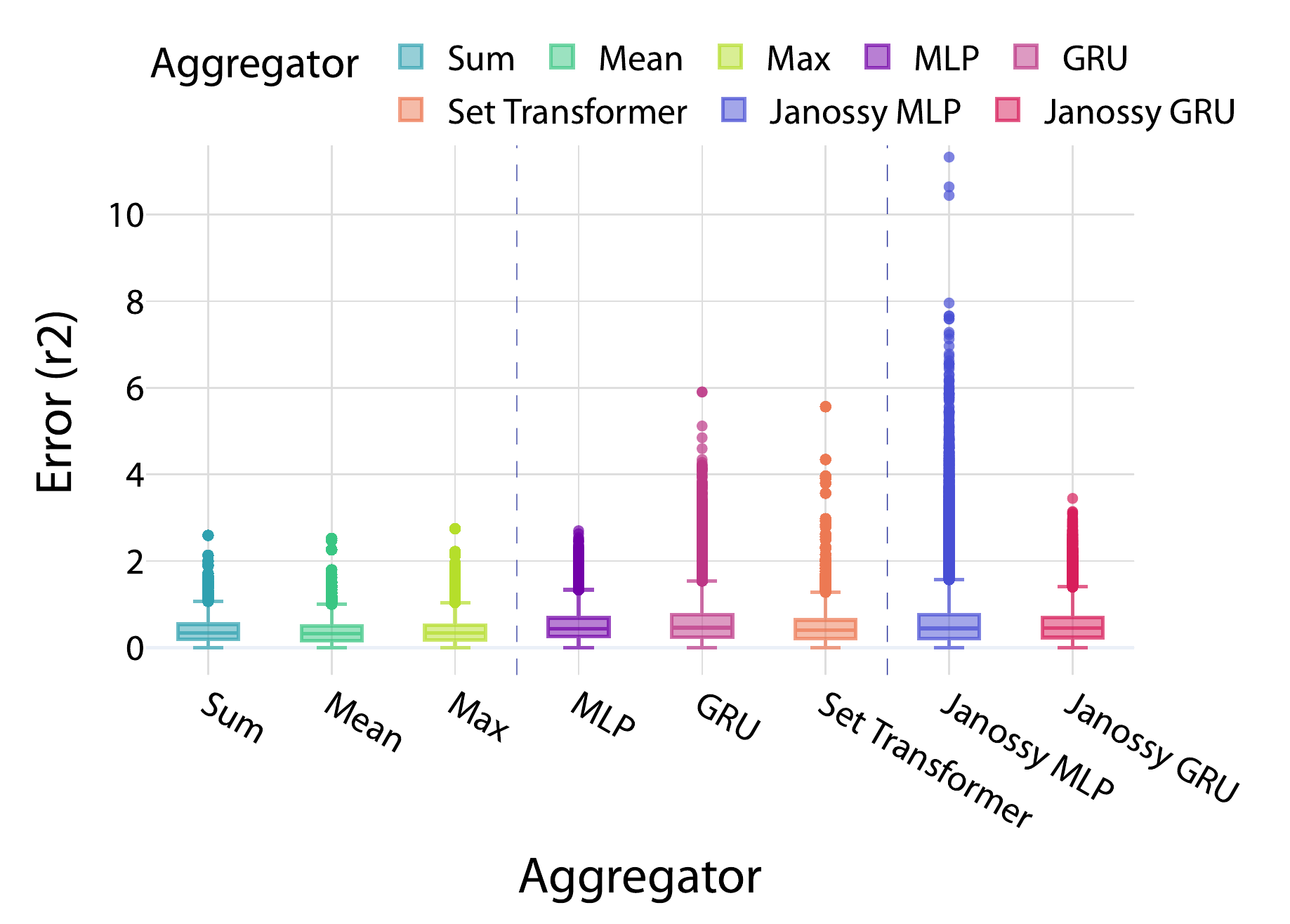}
    \end{subfigure}
    \hfill
    \begin{subfigure}[b]{0.24\textwidth}
        \centering
        % \captionsetup{labelfont=bf, font=small, skip=-1pt}
        % \caption{...} \label{subfigure:random-perm-smiles}
        \includegraphics[width=0.9\textwidth]{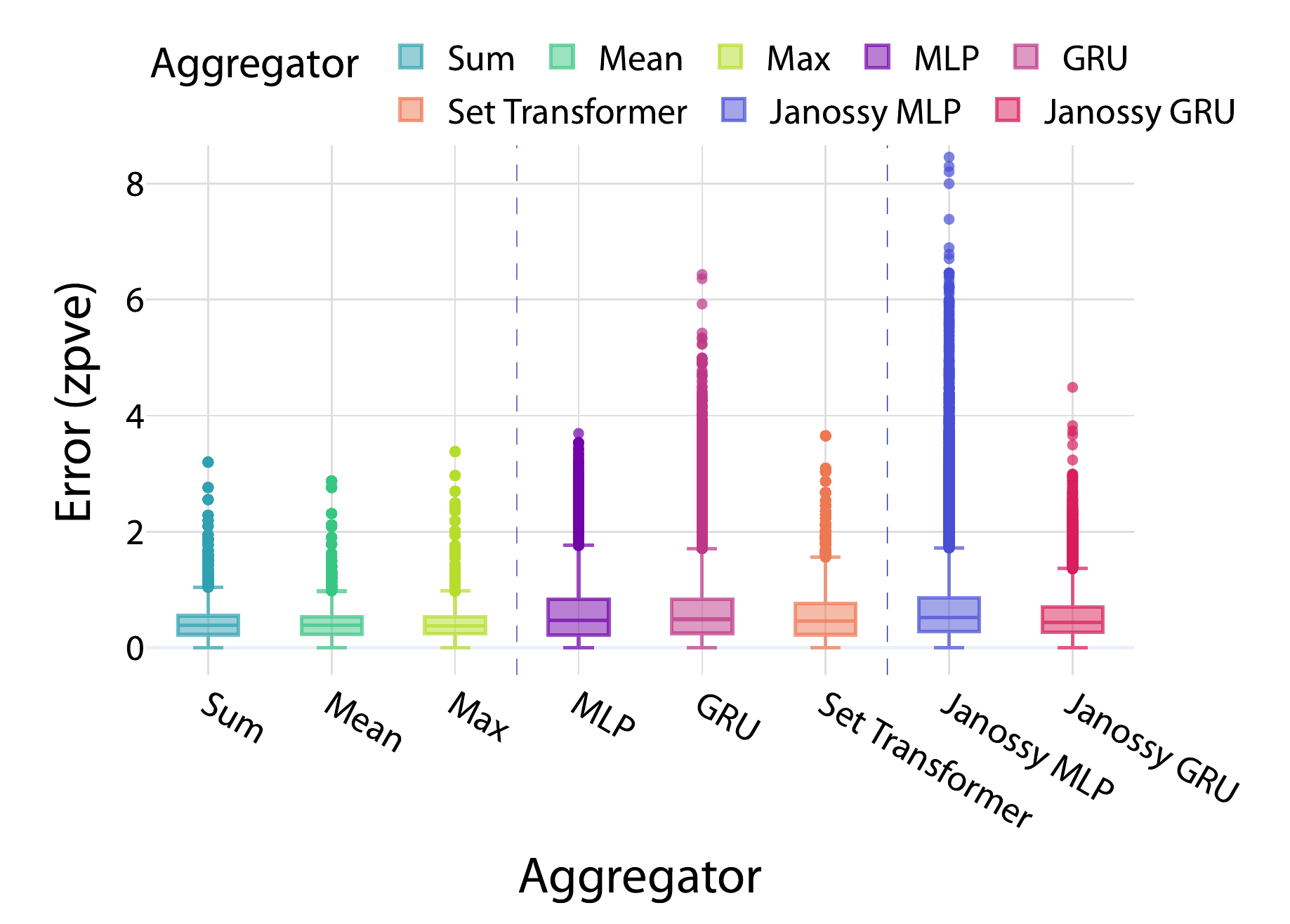}
    \end{subfigure}
    \hfill
    \begin{subfigure}[b]{0.24\textwidth}
        \centering
        % \captionsetup{labelfont=bf, font=small, skip=-1pt}
        % \caption{...} \label{subfigure:random-perm-smiles}
        \includegraphics[width=0.9\textwidth]{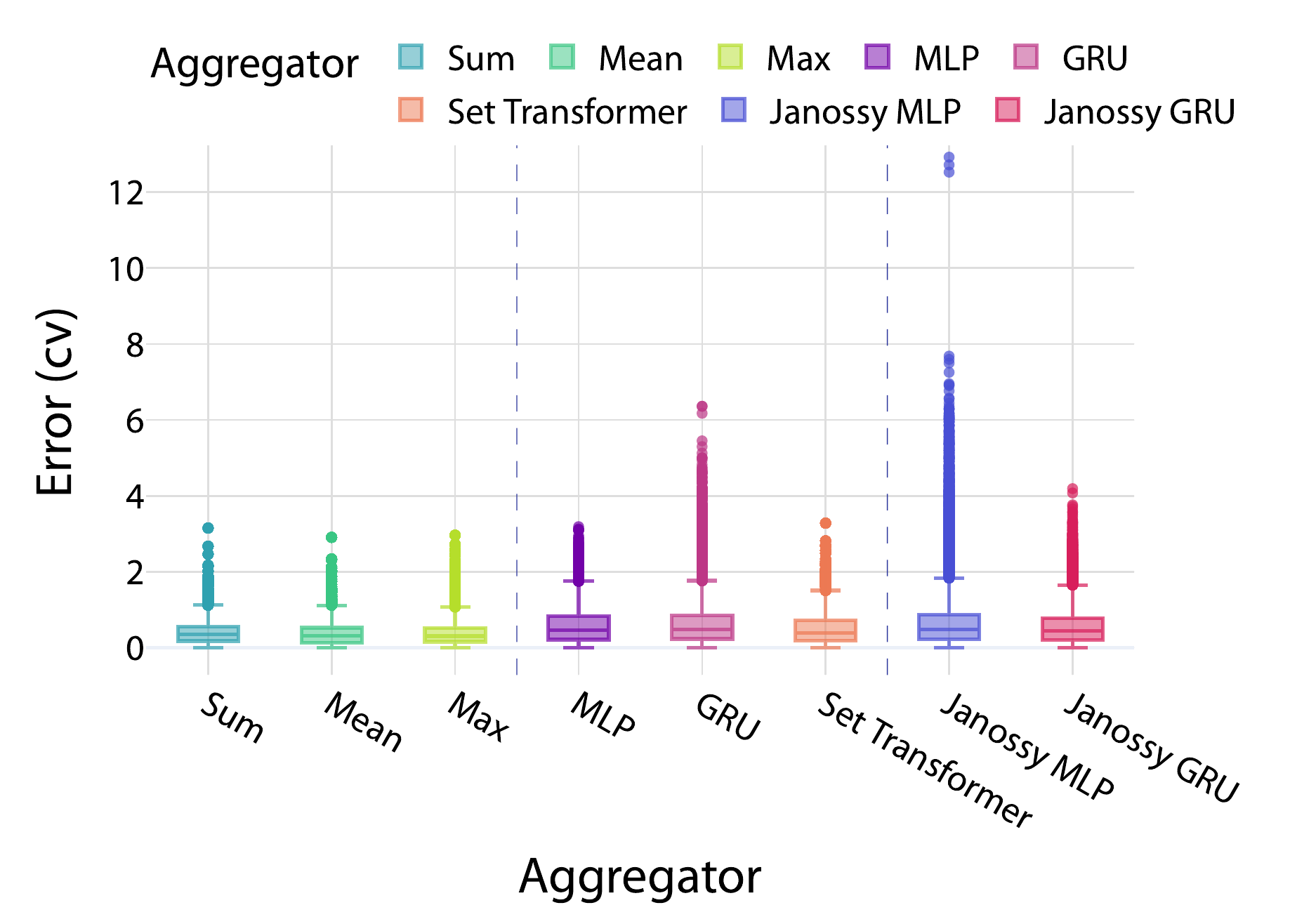}
    \end{subfigure}

    \begin{subfigure}[b]{0.24\textwidth}
        \centering
        % \captionsetup{labelfont=bf, font=small, skip=-1pt}
        % \caption{...} \label{subfigure:random-perm-smiles}
        \includegraphics[width=0.9\textwidth]{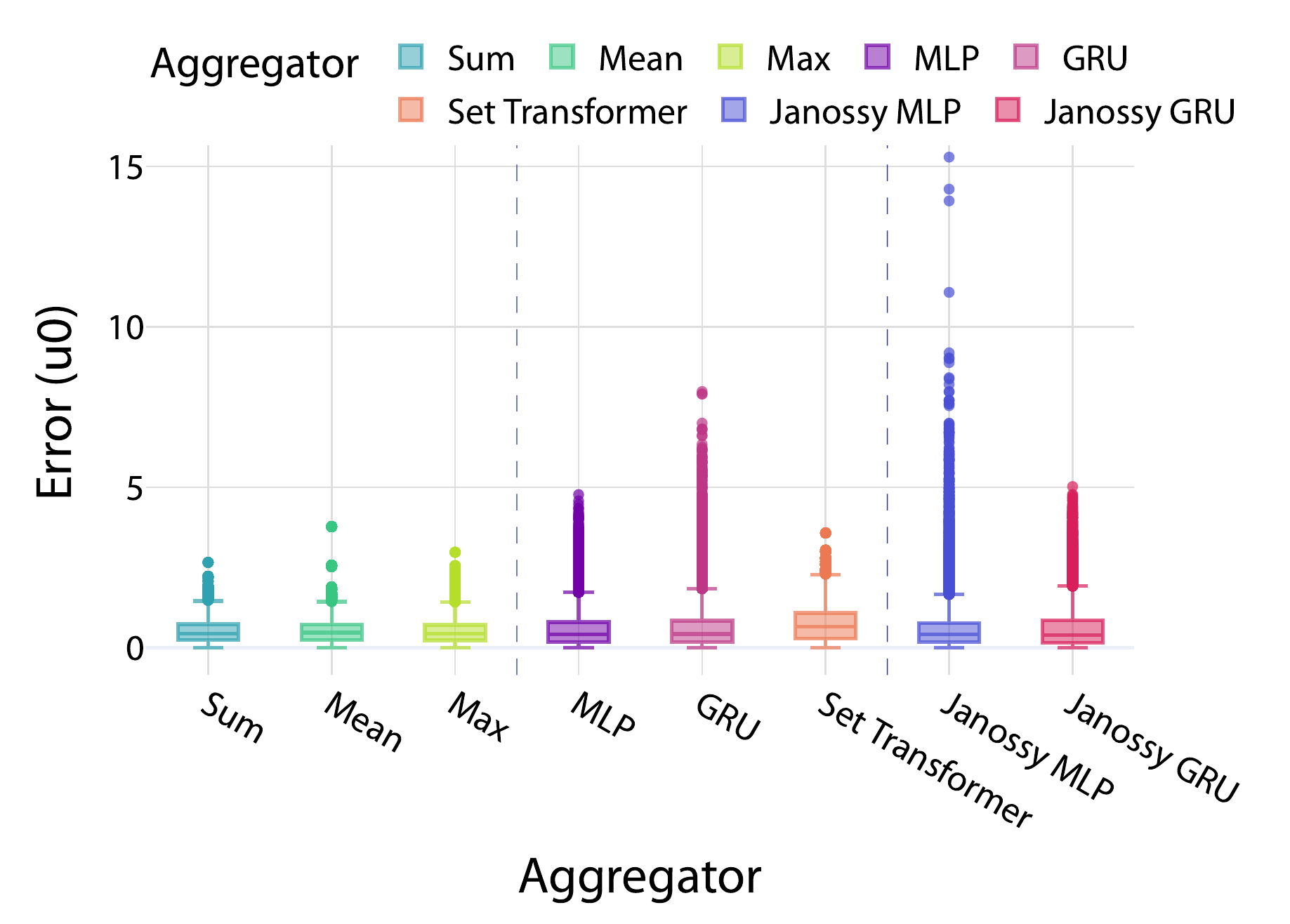}
    \end{subfigure}
    \hfill
    \begin{subfigure}[b]{0.24\textwidth}
        \centering
        % \captionsetup{labelfont=bf, font=small, skip=-1pt}
        % \caption{...} \label{subfigure:random-perm-smiles}
        \includegraphics[width=0.9\textwidth]{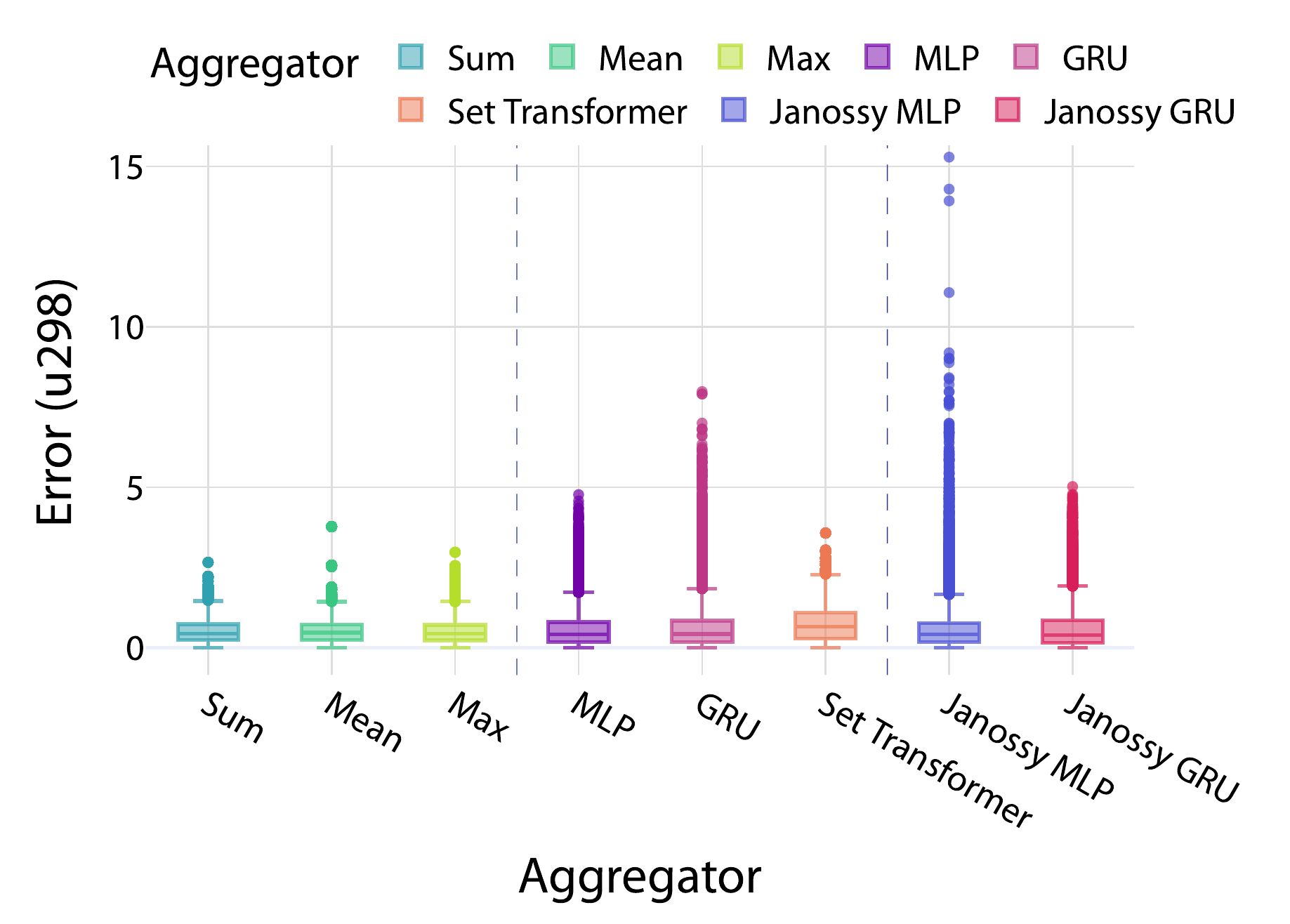}
    \end{subfigure}
    \hfill
    \begin{subfigure}[b]{0.24\textwidth}
        \centering
        % \captionsetup{labelfont=bf, font=small, skip=-1pt}
        % \caption{...} \label{subfigure:random-perm-smiles}
        \includegraphics[width=0.9\textwidth]{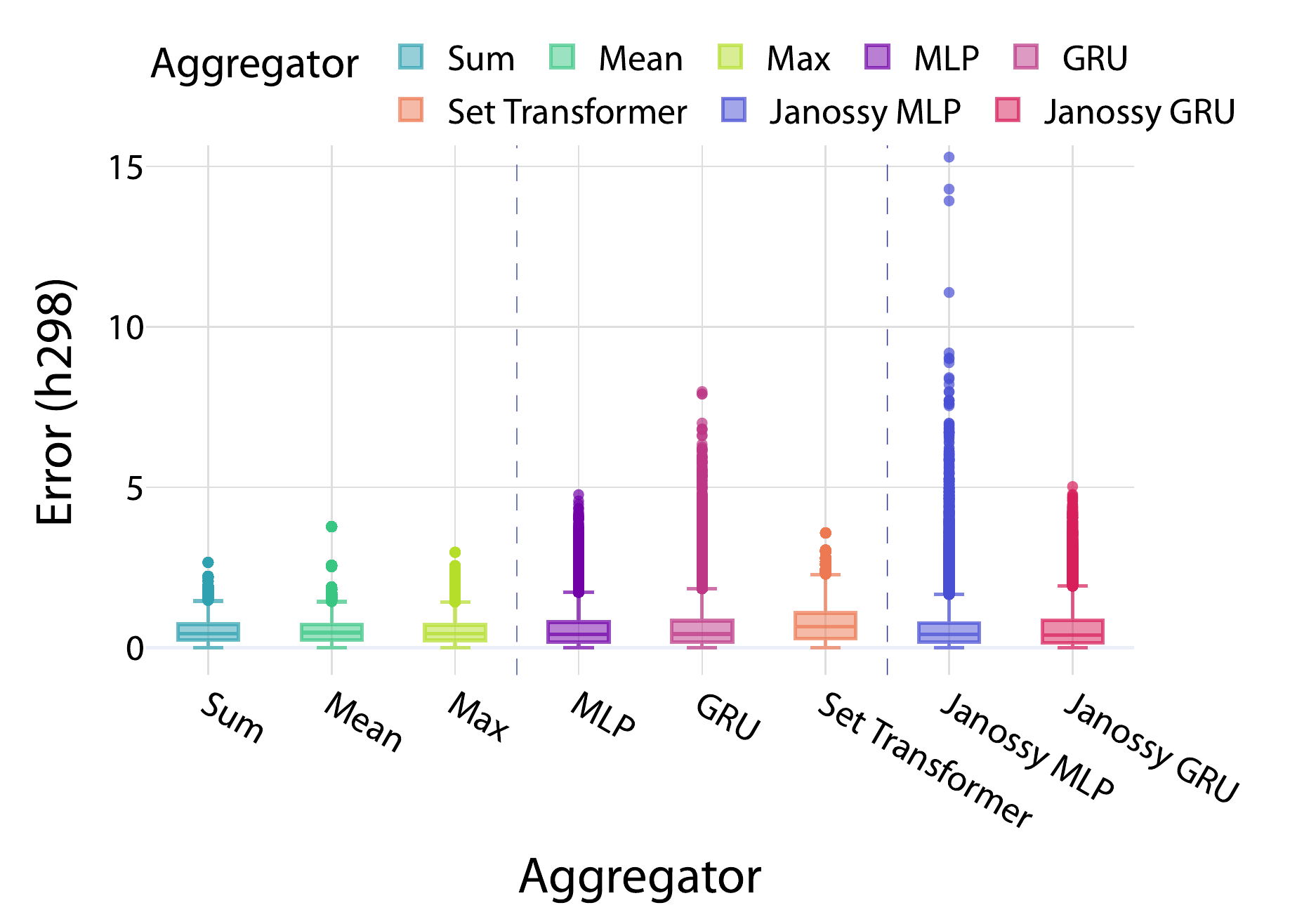}
    \end{subfigure}
    \hfill
    \begin{subfigure}[b]{0.24\textwidth}
        \centering
        % \captionsetup{labelfont=bf, font=small, skip=-1pt}
        % \caption{...} \label{subfigure:random-perm-smiles}
        \includegraphics[width=0.9\textwidth]{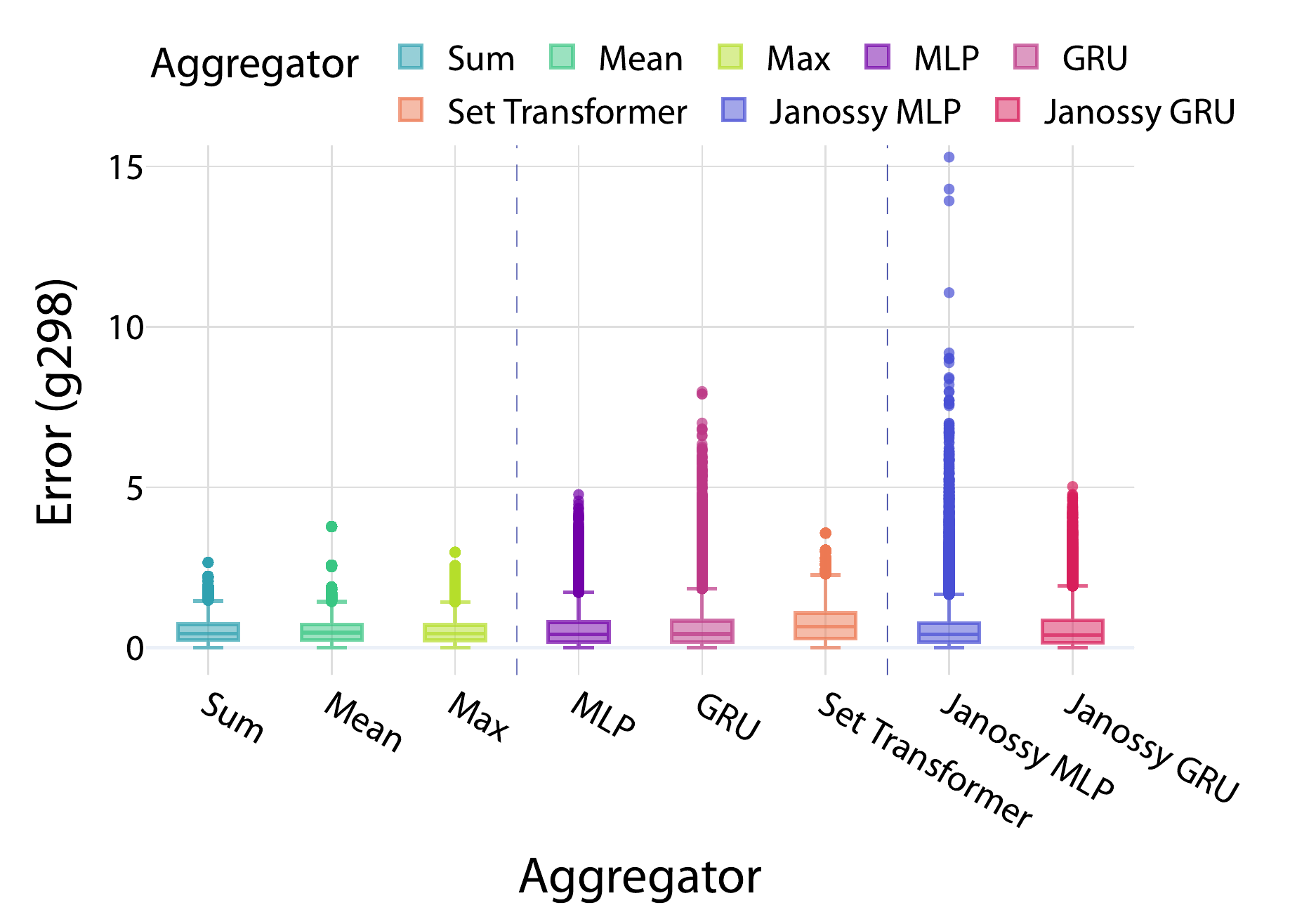}
    \end{subfigure}

\end{figure}
\renewcommand{\familydefault}{\rmdefault}

\clearpage
\section{Distances between graph representations}
\label{section:graph-emb-distance}
\begin{figure}[!h]
    \captionsetup{skip=4pt, labelfont=bf}
    \caption{The Euclidean distance was computed between the initial graph representation (i.e. after the first epoch) and all subsequent epochs for a random molecule of the \textsc{qm9} dataset, for multiple graph convolution types and readouts ($2$-layer \textsc{gnn}s). Generally, models using standard aggregators (sum, mean, max) take a long time to converge ($500$ to $1,000$ epochs), with only minor modifications to the graph representation. The models using neural readouts typically converge in under $100$ epochs and are able to explore a much larger hypothesis space, as indicated by the large distances between the initial and final trained representations.}
    \label{figure:qm9-dist-initial}
    \centering
    \includegraphics[width=1\textwidth]{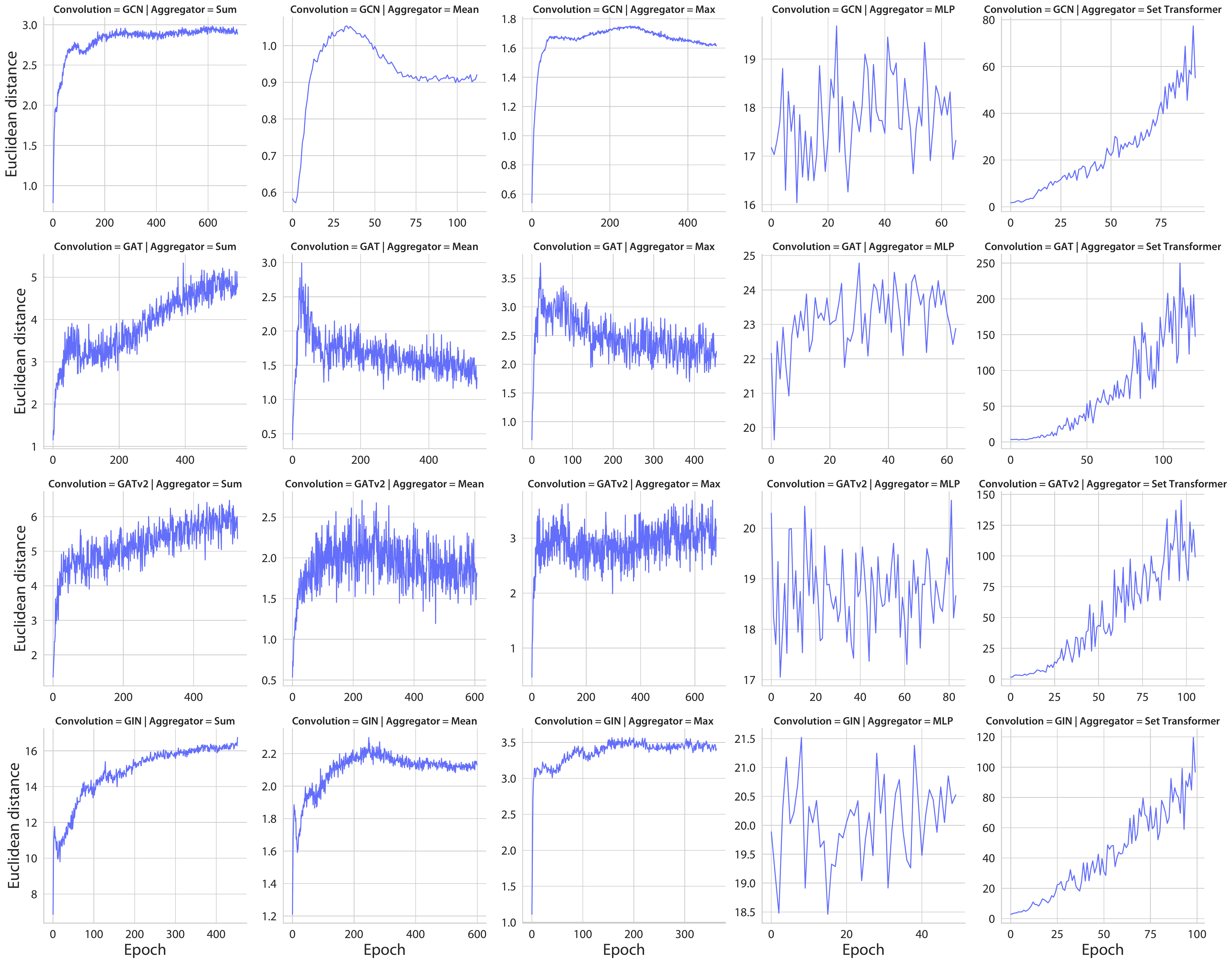}
\end{figure}

\begin{figure}[!h]
    \captionsetup{skip=4pt, labelfont=bf}
    \caption{The Euclidean distance was computed between the graph representations from consecutive epochs for a random molecule of the \textsc{qm9} dataset (same as Appendix \Cref{figure:qm9-dist-initial}), for multiple graph convolution types and readouts ($2$-layer \textsc{gnn}s). Generally, models using standard aggregators (sum, mean, max) take a long time to converge ($500$ to $1,000$ epochs), with only minor modifications to the graph representation. The models using neural readouts typically converge in under $100$ epochs and are able to explore a much larger hypothesis space, as indicated by the large distances between the initial and final trained representations.}
    \label{figure:qm9-dist-e-by-e}
    \centering
    \includegraphics[width=1\textwidth]{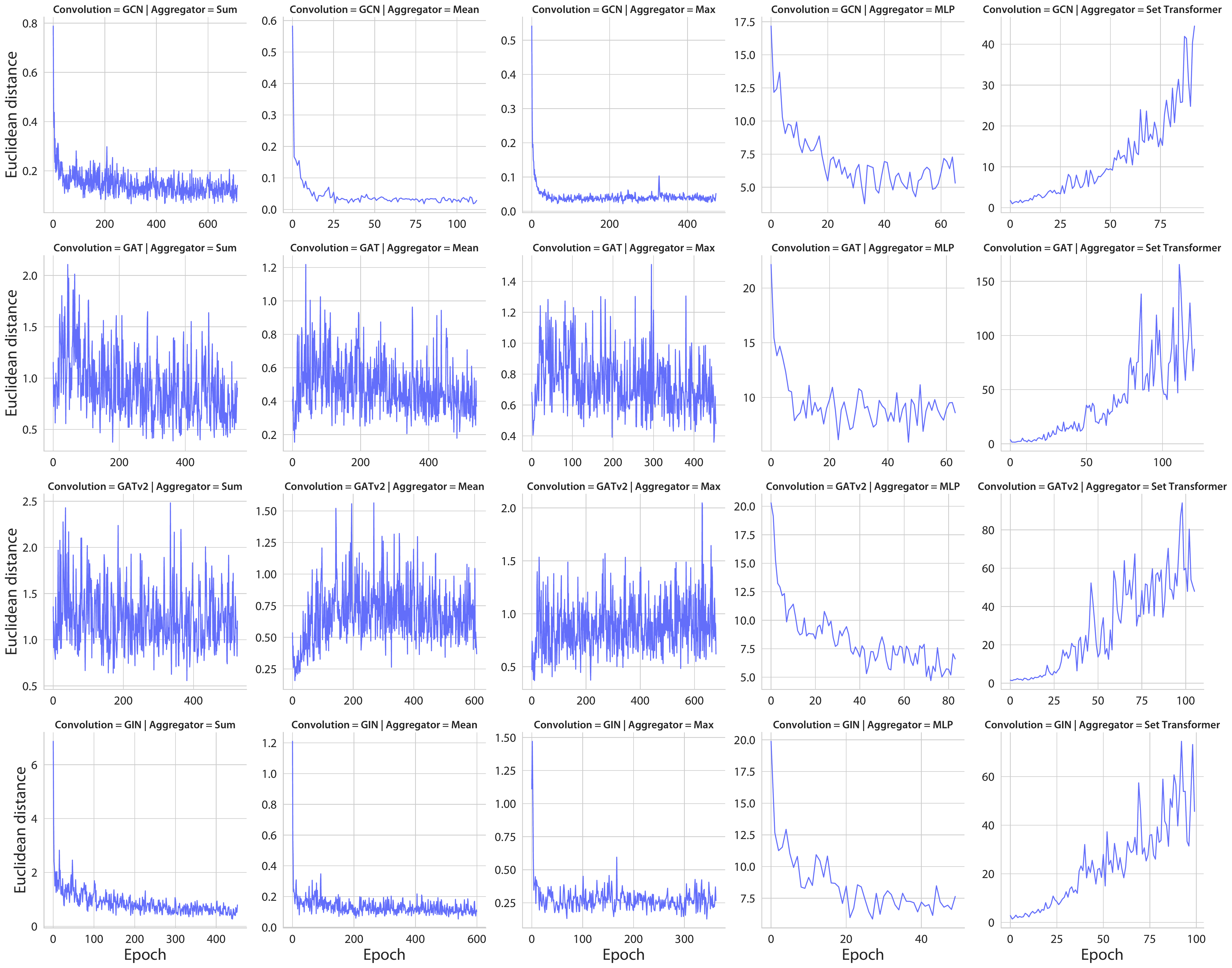}
\end{figure}

\clearpage
\section{Visualization of the learned latent space for different aggregators}
\label{section:vgae-latent}
\subsection*{Guided VGAE architecture}
We adapted our fixed architecture into a variational graph autoencoder (\textsc{vgae}, as introduced by Kipf and Welling \cite{kipf2016variational}). The changes include two additional layers for the $\mu$ and $\sigma$ parameters, as well as the reconstruction and regularization losses (as provided by PyTorch Geometric). The graph embeddings learned by the \textsc{vgae} are fed into a predictor network in an end-to-end fashion, such that the task is supervised.

\renewcommand{\sfdefault}{lmss}
\renewcommand{\familydefault}{\sfdefault}
\begin{table}[h]
    \centering
    \captionsetup{skip=4pt, labelfont=bf, width=1\linewidth}
    \caption{Visualization of the learned latent space of graphs (molecules) for three recently-introduced bio-affinity datasets, using UMAP projections in 3 dimensions. The figure presents a selection of 3 readouts. Angles are chosen to best highlight the 3D space structure. The activity of molecules, as reported in the bioassay, is illustrated according to the color bars.}
    \label{table:latent-space}
    \begin{tabular}{m{1.5cm}cccc}
    \toprule
    \textbf{Dataset} & \textbf{Sum} & \textbf{MLP} & \textbf{Set Transformer} \\ \midrule
    AID1949 &
    \begin{minipage}{.2\textwidth}
        \includegraphics[width=\textwidth]{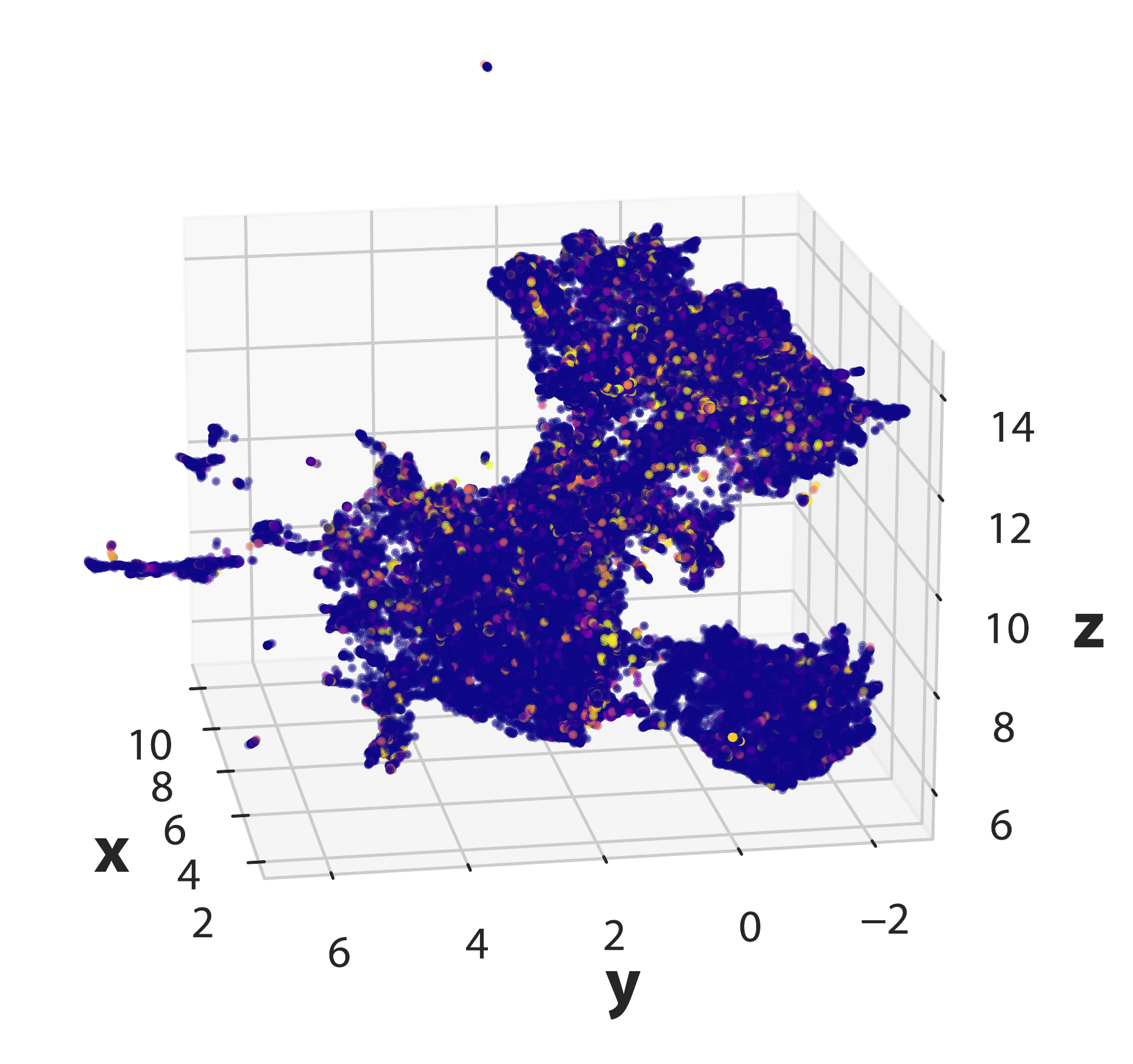}
    \end{minipage}    &
    \begin{minipage}{.2\textwidth}
        \includegraphics[width=\textwidth]{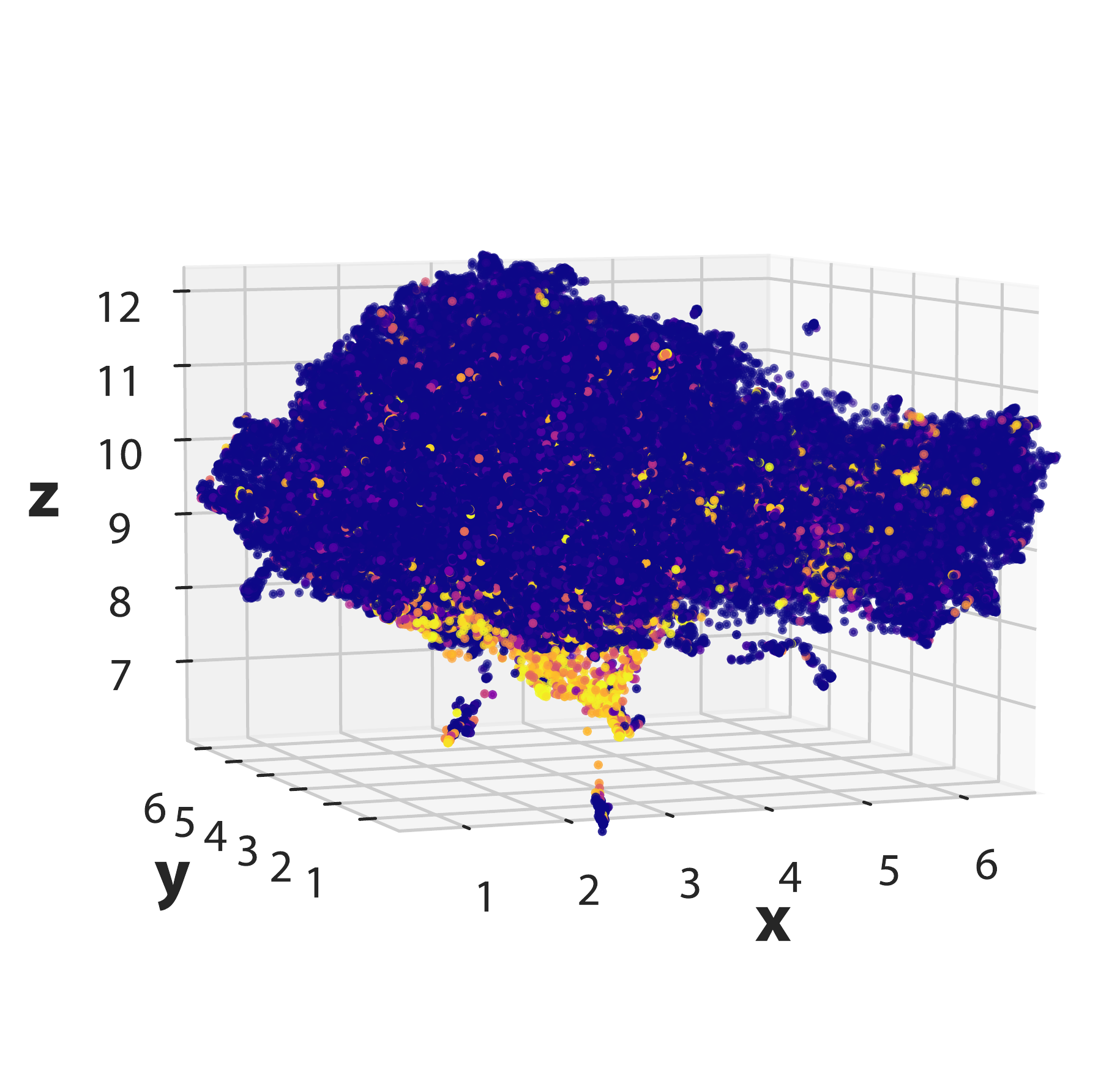}
    \end{minipage}    &
    \begin{minipage}{.24\textwidth}
        \includegraphics[width=\textwidth]{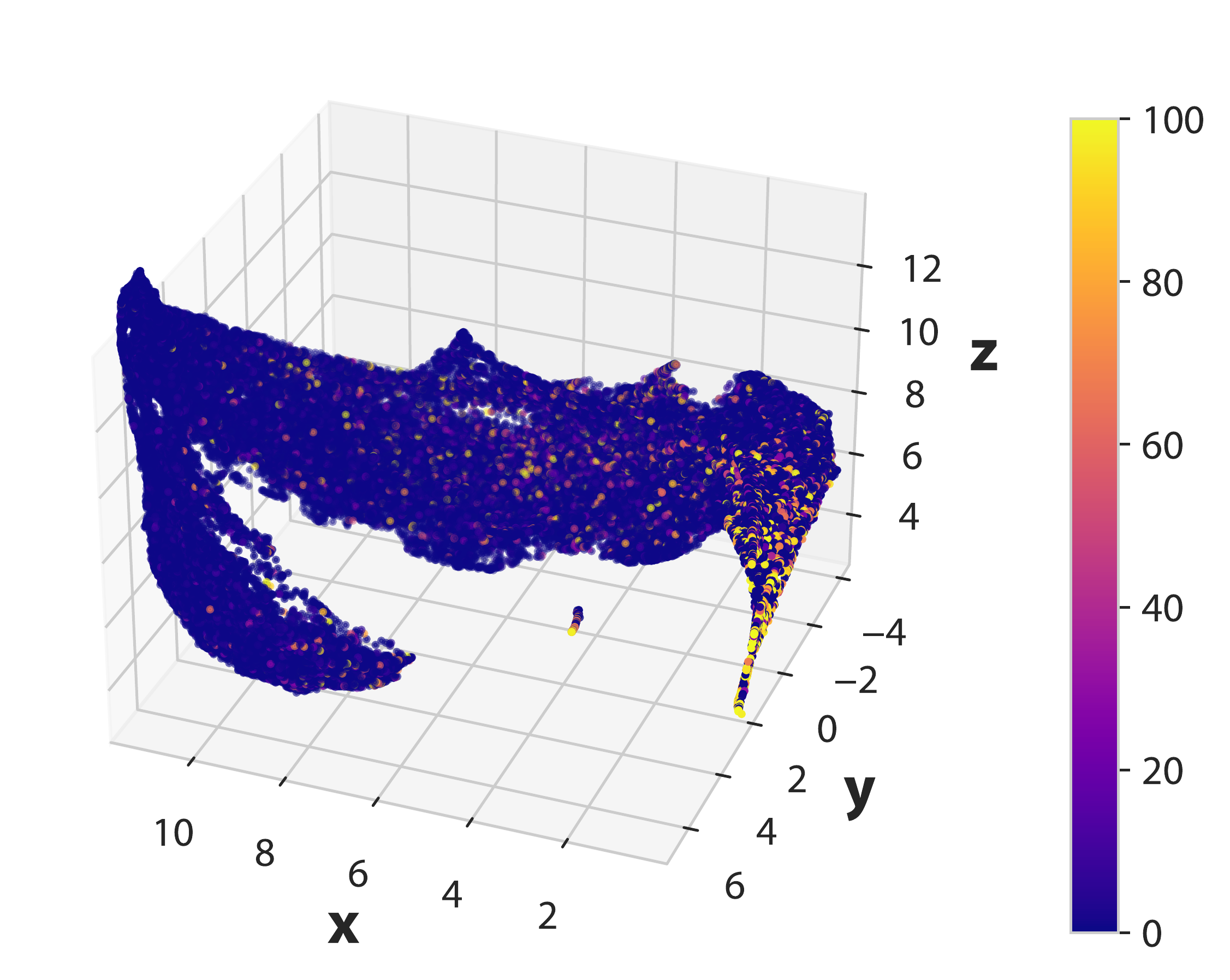}
    \end{minipage}    \\

    AID449762 &
    \begin{minipage}{.2\textwidth}
        \includegraphics[width=\textwidth]{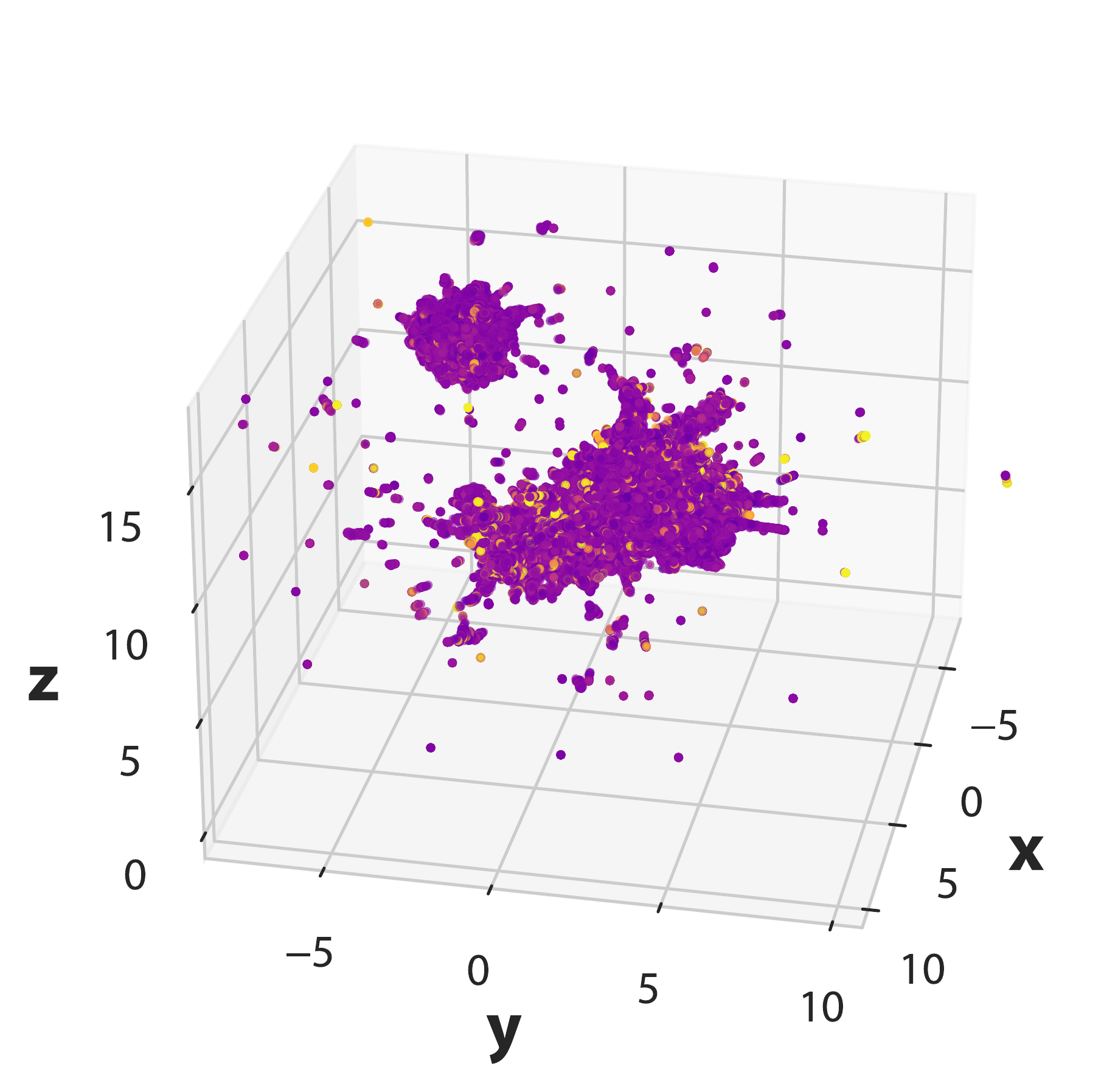}
    \end{minipage}    &
    \begin{minipage}{.2\textwidth}
        \includegraphics[width=\textwidth]{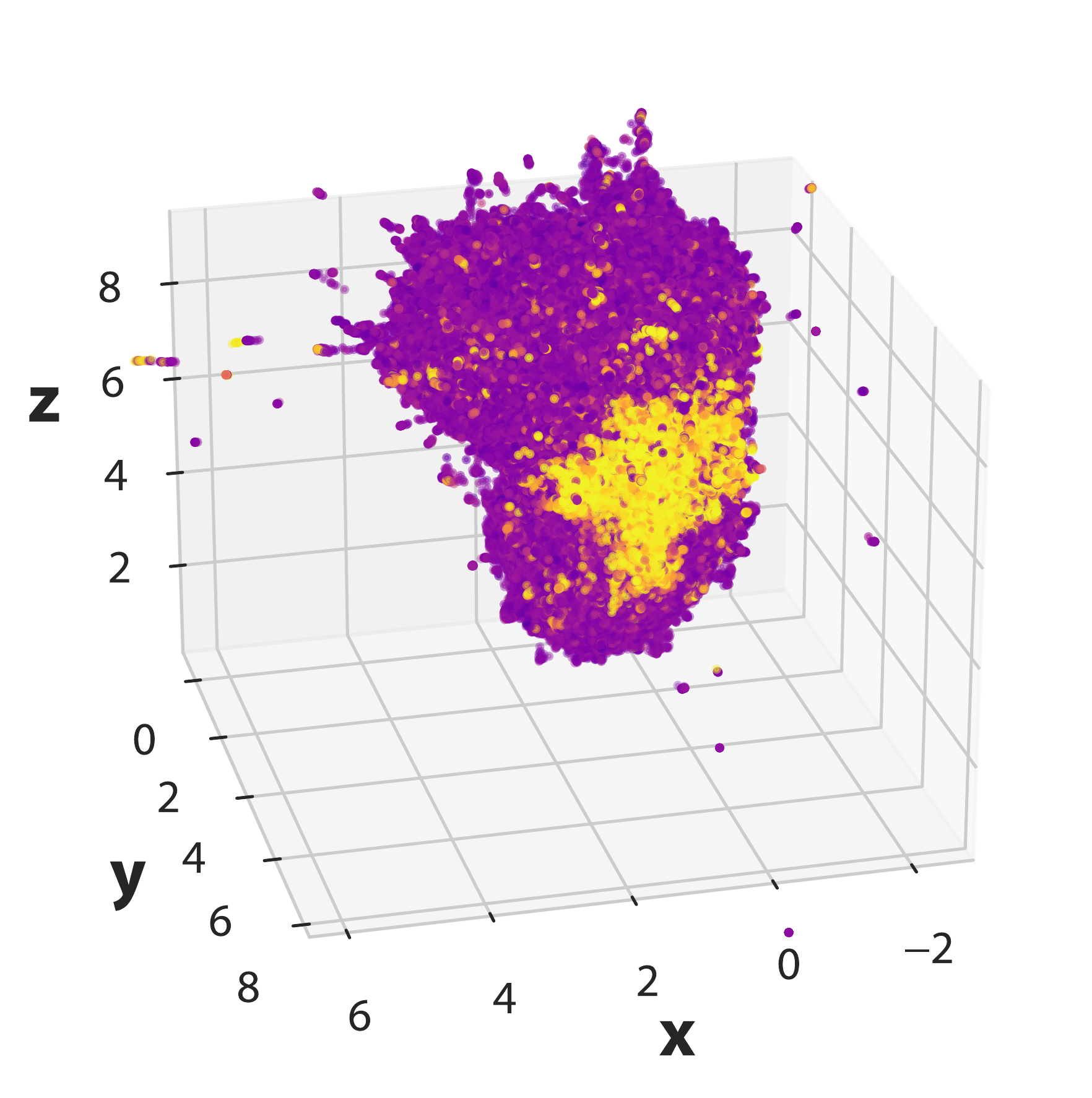}
    \end{minipage}    &
    \begin{minipage}{.24\textwidth}
        \includegraphics[width=\textwidth]{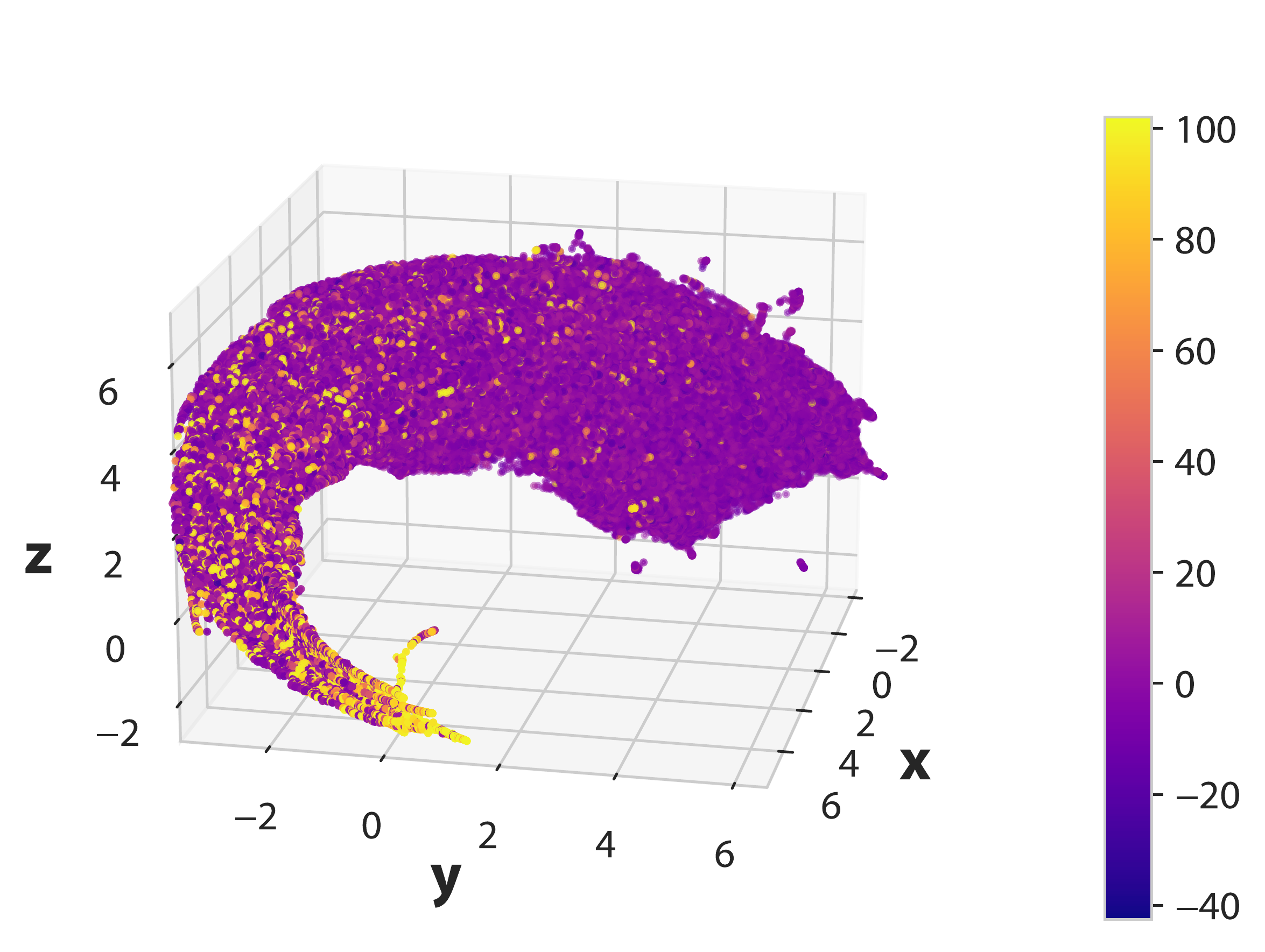}
    \end{minipage} \\

    AID602261 &
    \begin{minipage}{.2\textwidth}
        \includegraphics[width=\textwidth]{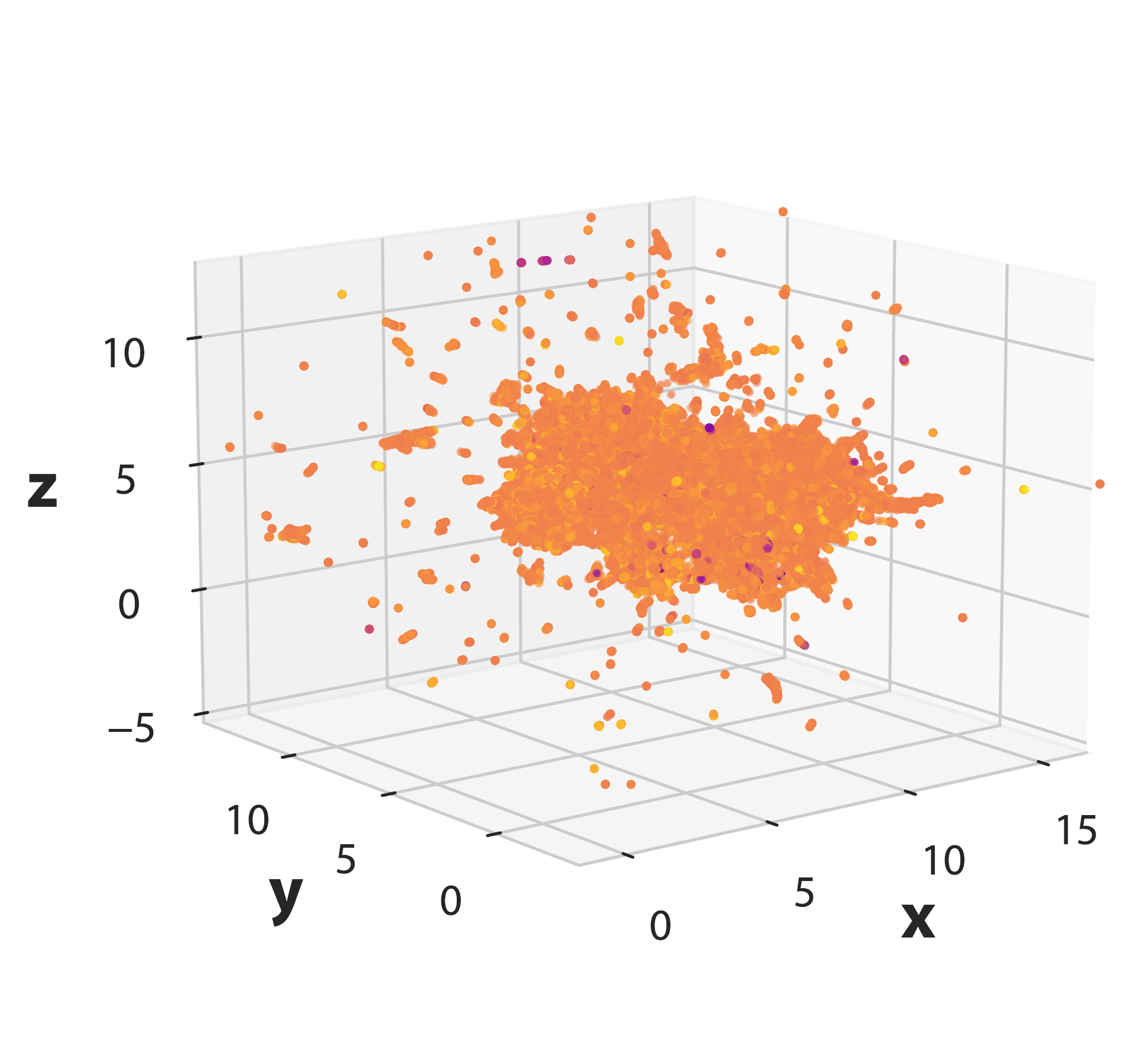}
    \end{minipage}    &
    \begin{minipage}{.2\textwidth}
        \includegraphics[width=\textwidth]{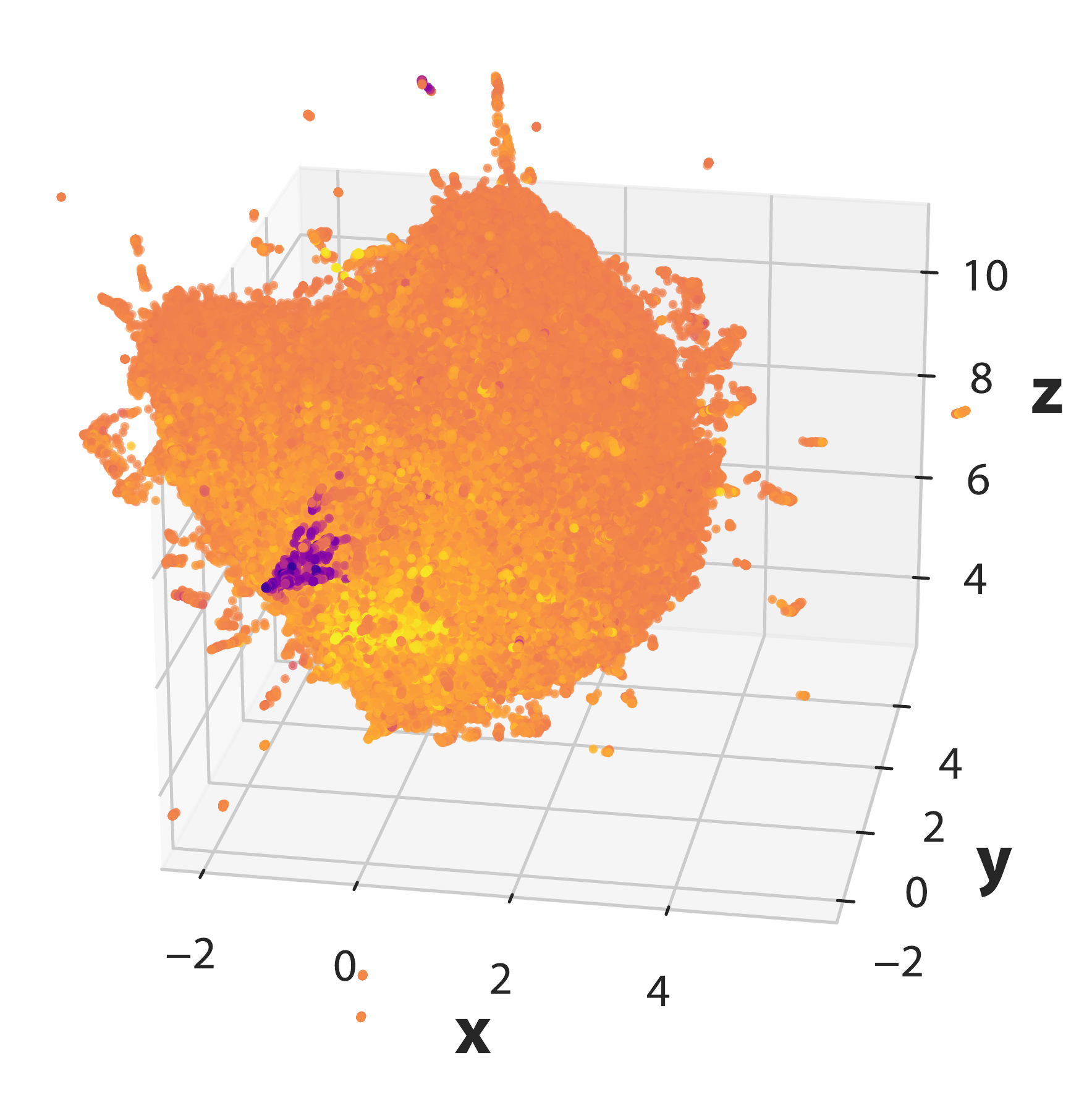}
    \end{minipage}    &
    \begin{minipage}{.245\textwidth}
        \includegraphics[width=\textwidth]{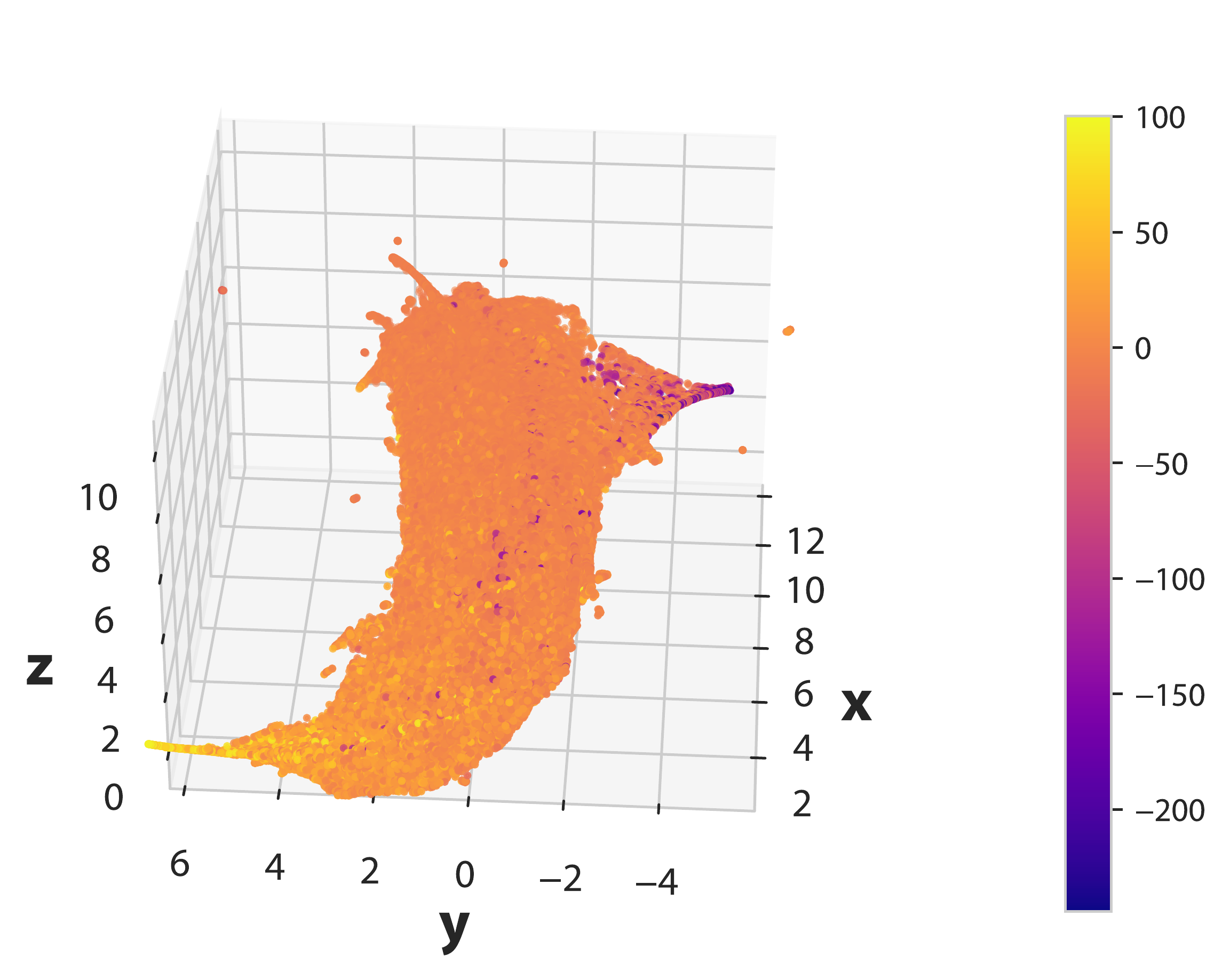}
    \end{minipage} \\

    \bottomrule

    \end{tabular}
\end{table}
\renewcommand{\familydefault}{\rmdefault}

\renewcommand{\sfdefault}{lmss}
\renewcommand{\familydefault}{\sfdefault}
\begin{table}[h]
    \captionsetup{skip=4pt, labelfont=bf, width=1\linewidth}
    \caption{Visualization of the learned latent space of graphs (molecules) for three recently-introduced bio-affinity datasets, using UMAP projections in 3 dimensions. The figure presents the other 3 readouts. Angles are chosen to best highlight the 3D space structure.}
    \label{table:latent-space-supp}
    \begin{tabular}{m{1.5cm}cccc}
    \toprule
    \textbf{Dataset} & \textbf{Mean} & \textbf{Max} & \textbf{GRU} \\ \midrule
    AID1949 &
    \begin{minipage}{.24\textwidth}
        \includegraphics[width=\textwidth]{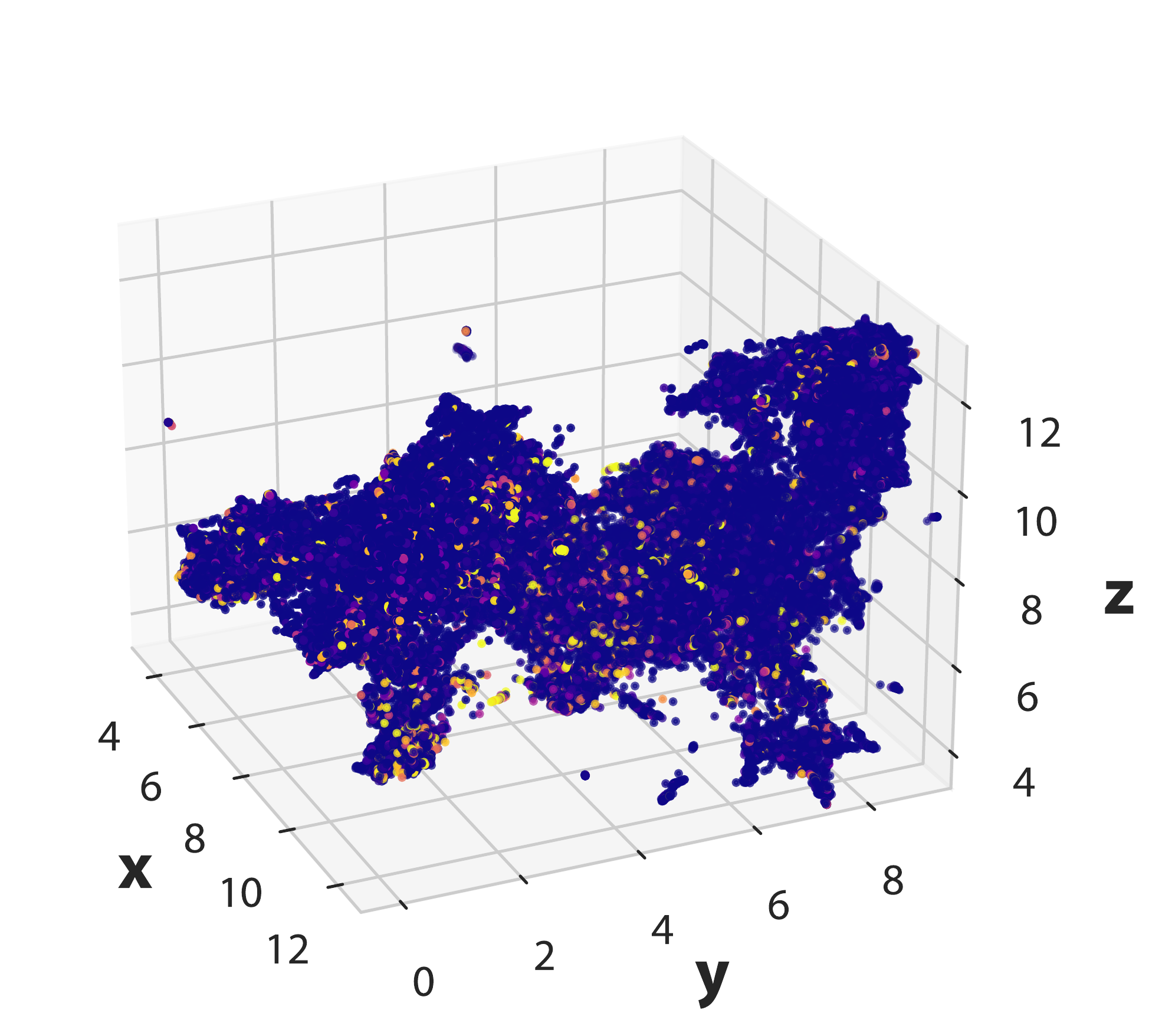}
    \end{minipage}    &
    \begin{minipage}{.24\textwidth}
        \includegraphics[width=\textwidth]{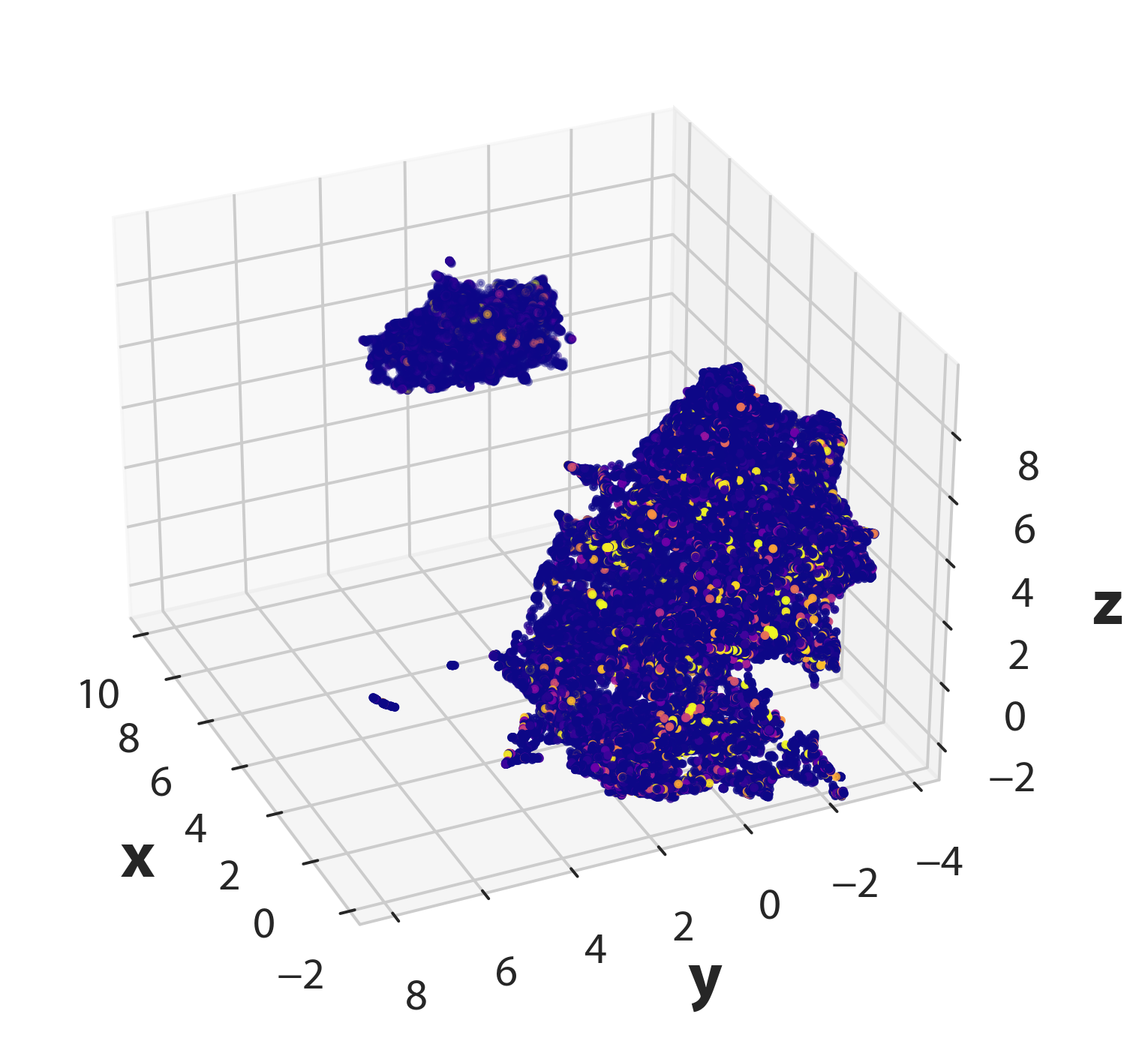}
    \end{minipage}    &
    \begin{minipage}{.28\textwidth}
        \includegraphics[width=\textwidth]{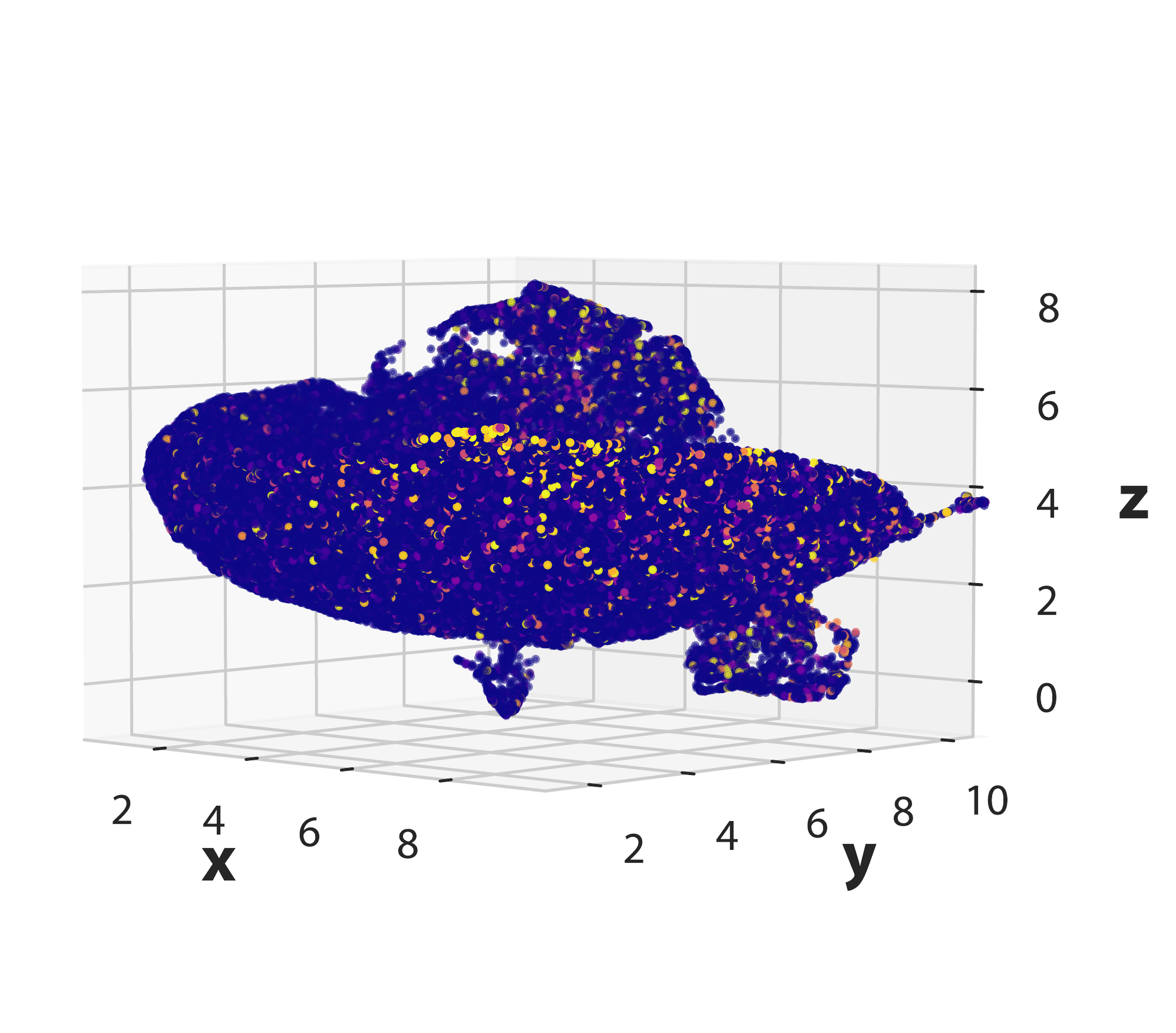}
    \end{minipage}    \\

    AID449762 &
    \begin{minipage}{.24\textwidth}
        \includegraphics[width=\textwidth]{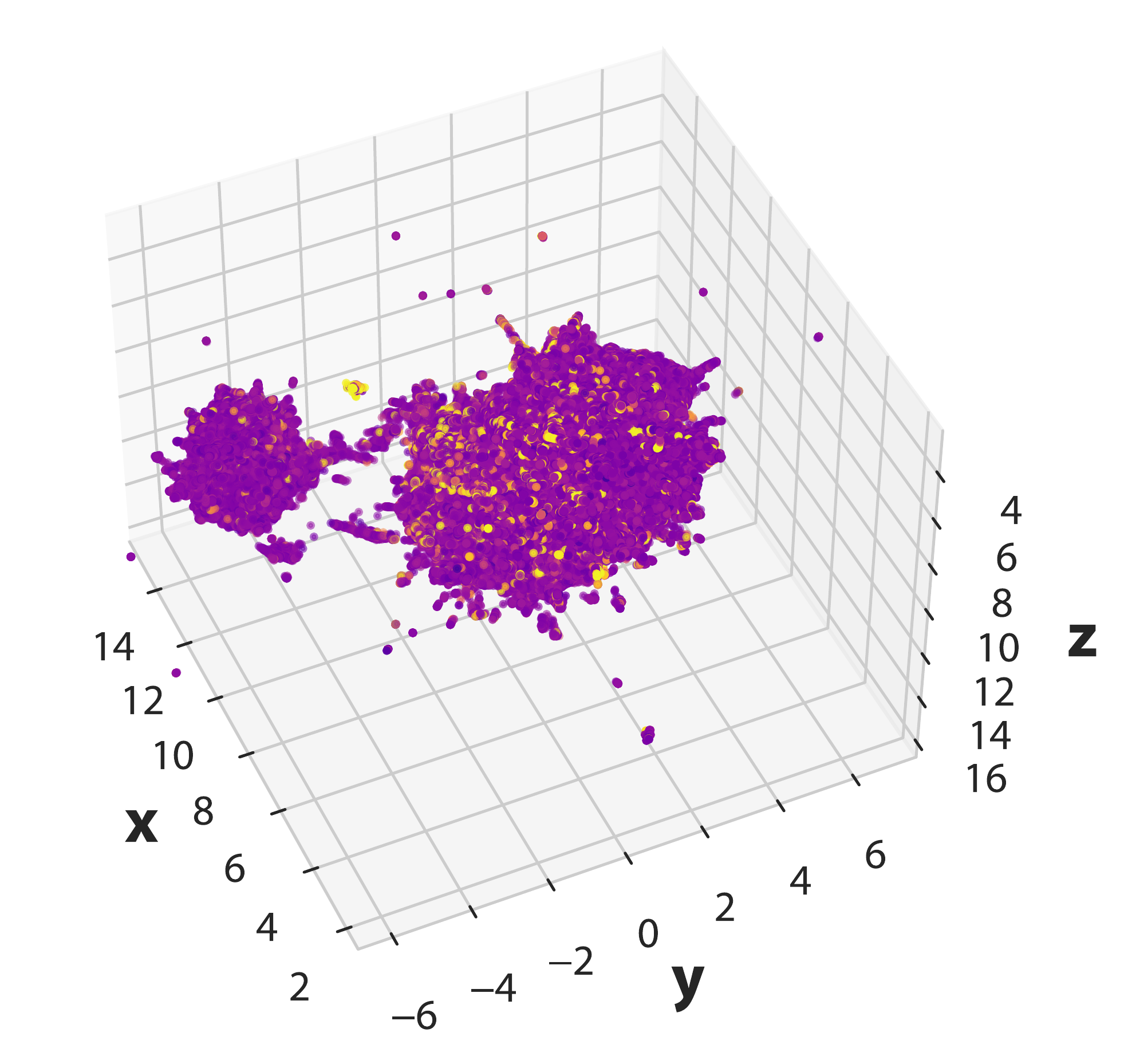}
    \end{minipage}    &
    \begin{minipage}{.24\textwidth}
        \includegraphics[width=\textwidth]{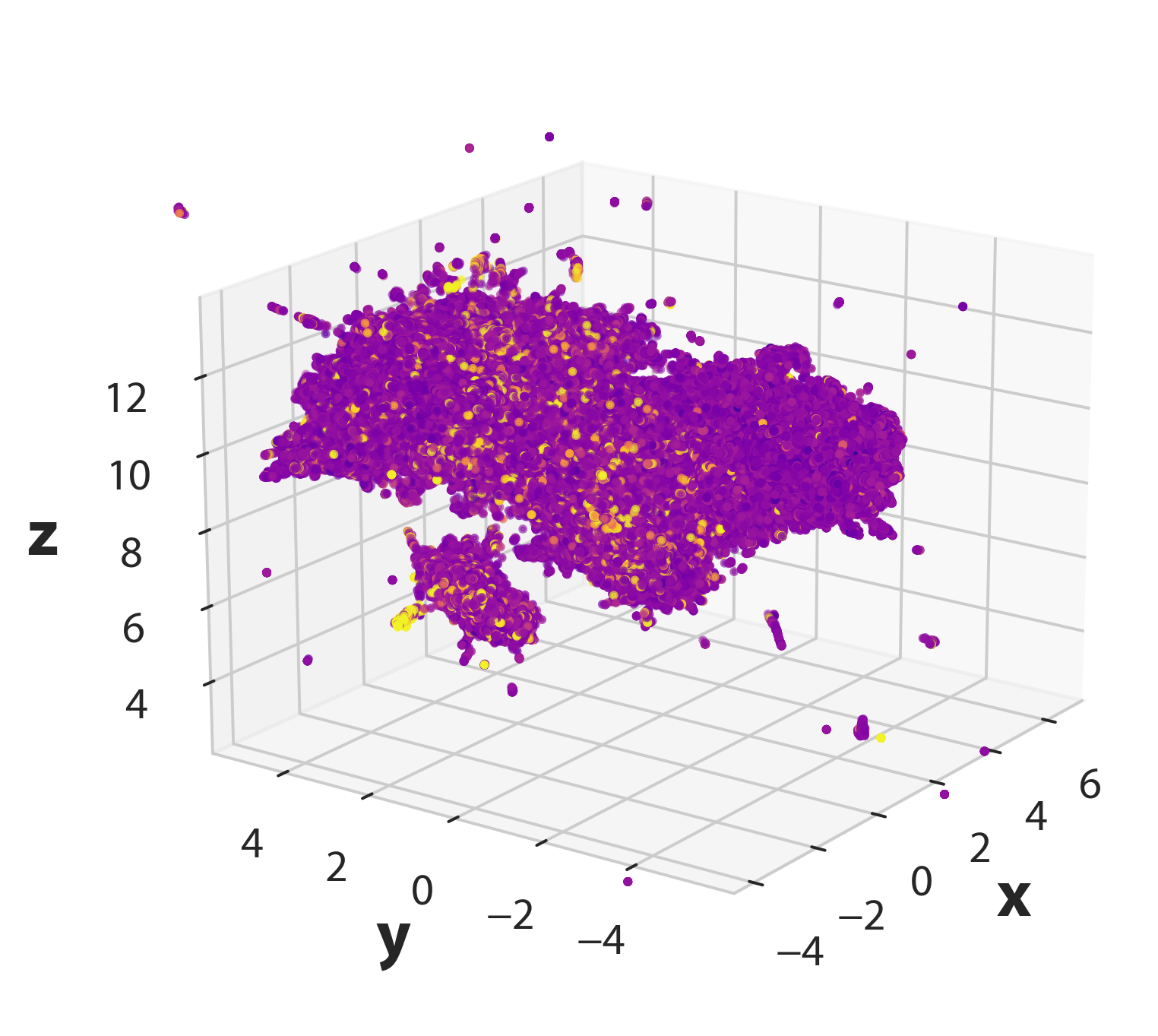}
    \end{minipage}    &
    \begin{minipage}{.24\textwidth}
        \includegraphics[width=\textwidth]{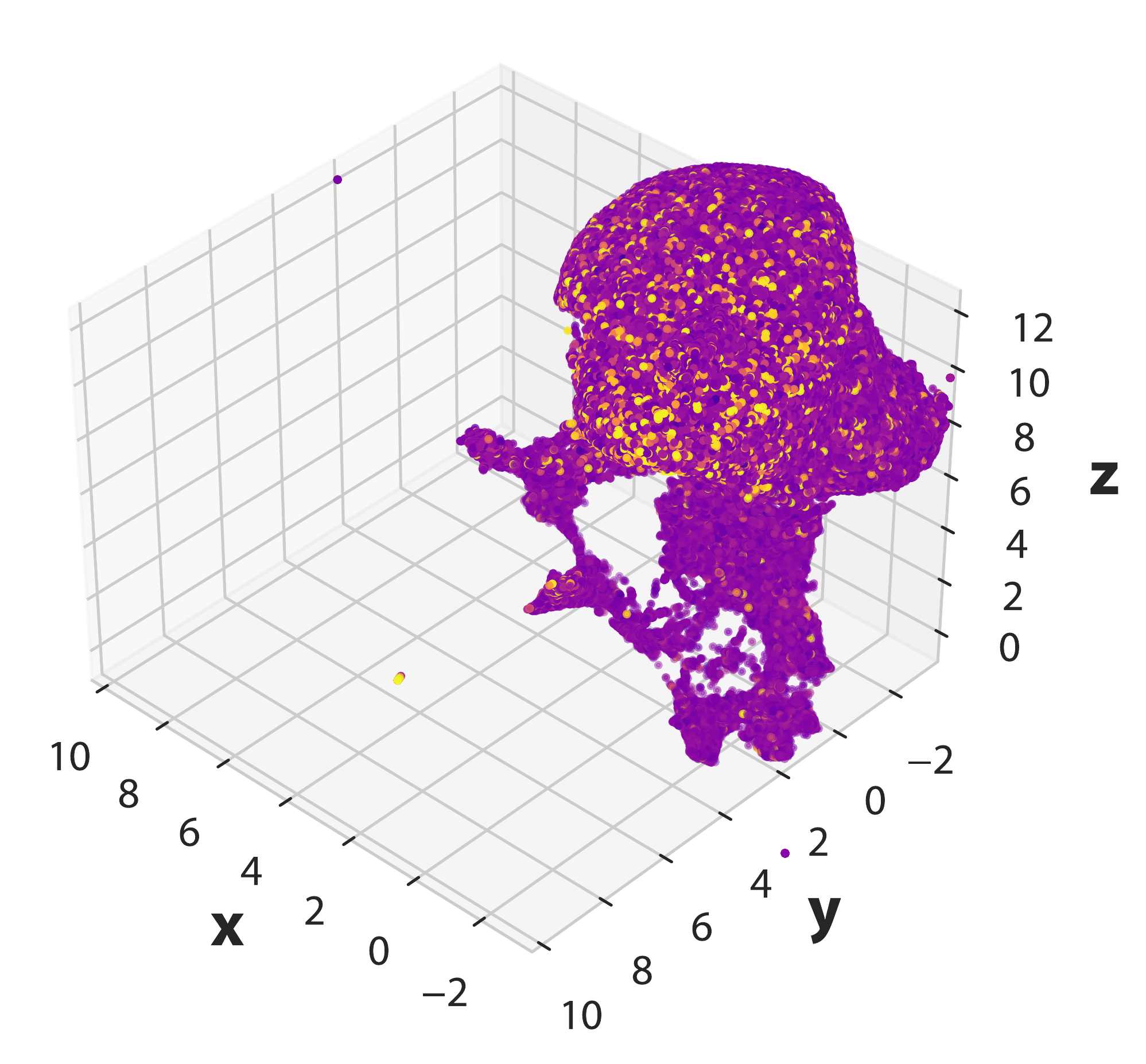}
    \end{minipage} \\

    AID602261 &
    \begin{minipage}{.24\textwidth}
        \includegraphics[width=\textwidth]{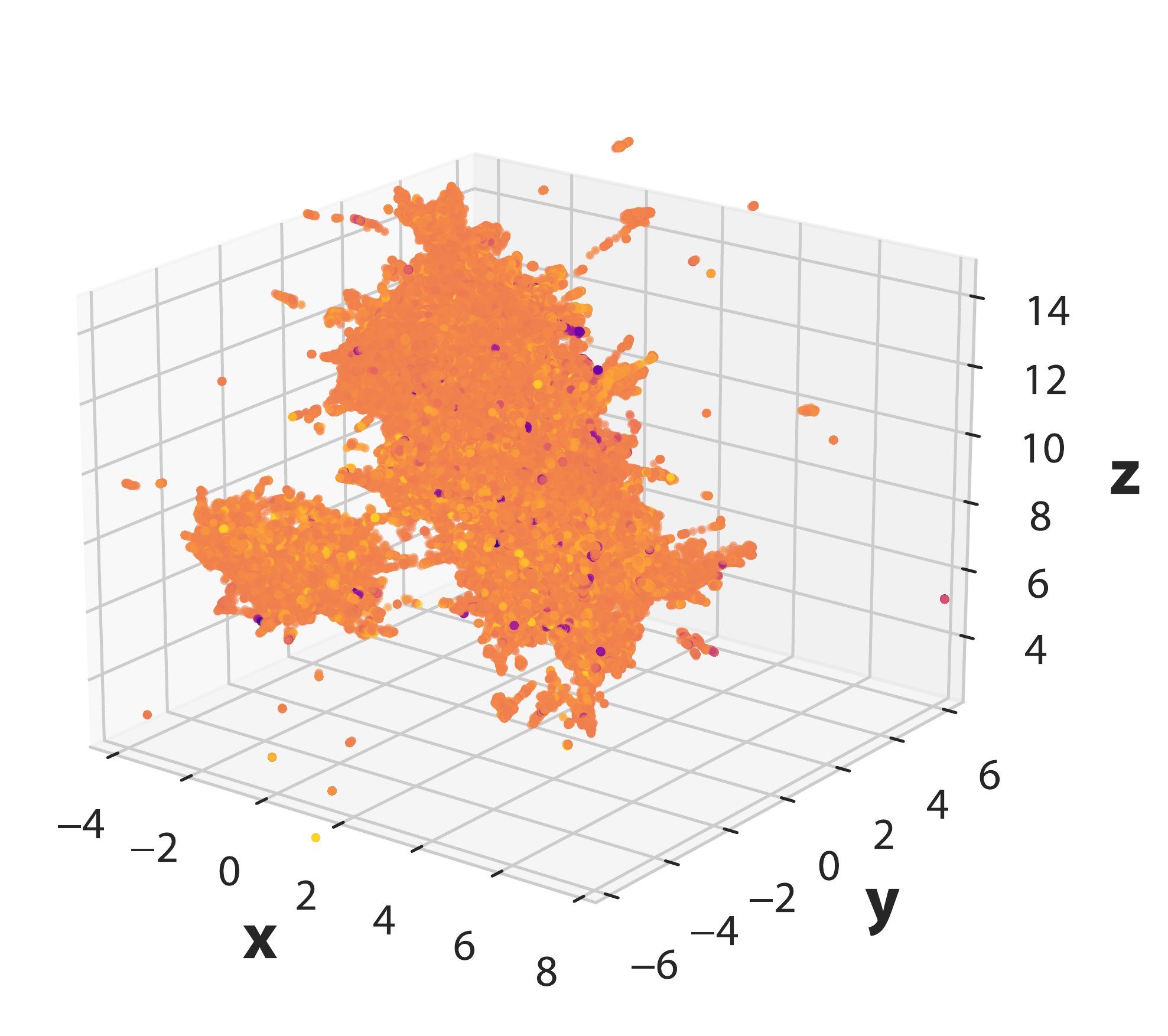}
    \end{minipage}    &
    \begin{minipage}{.24\textwidth}
        \includegraphics[width=\textwidth]{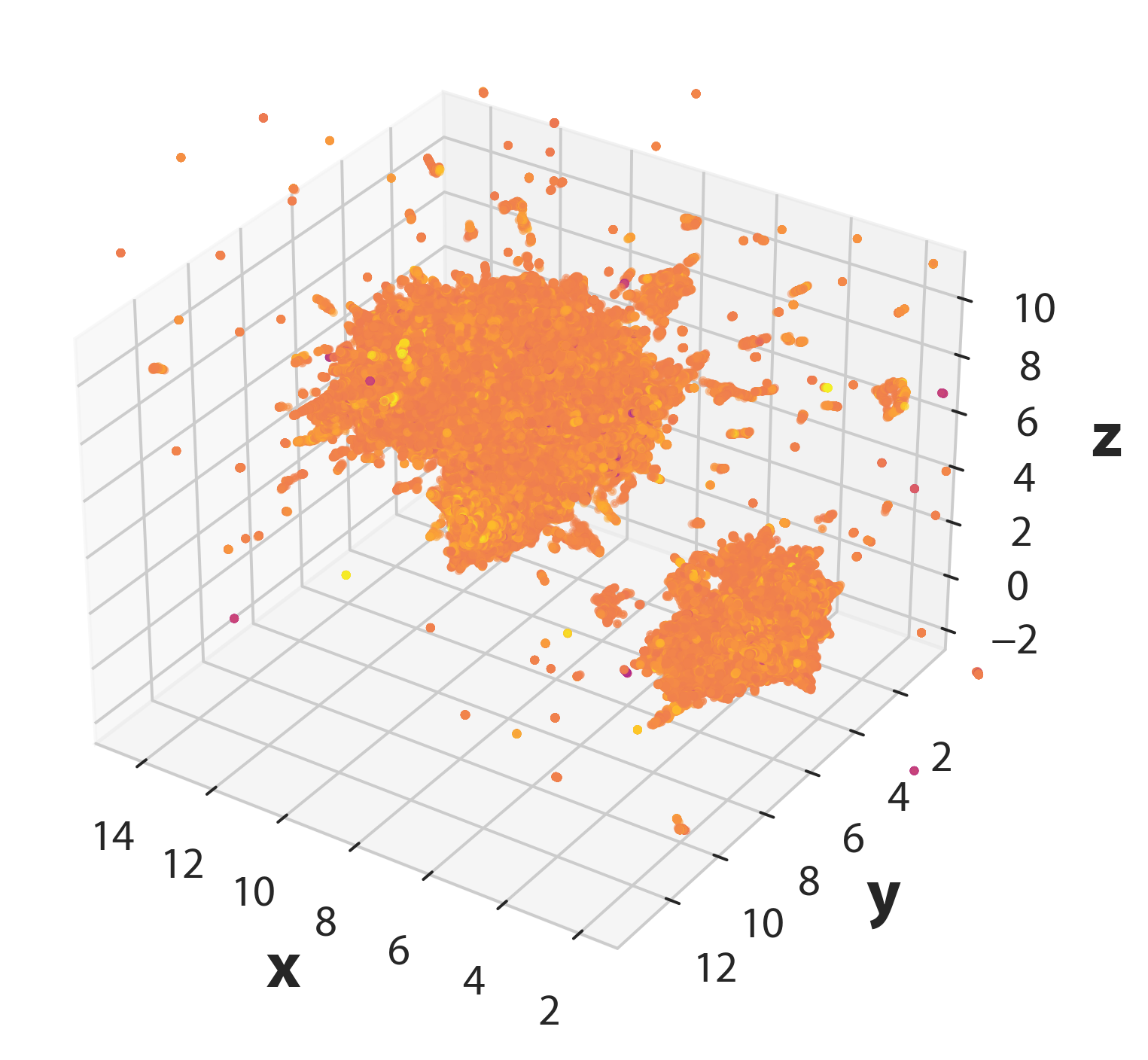}
    \end{minipage}    &
    \begin{minipage}{.24\textwidth}
        \includegraphics[width=\textwidth]{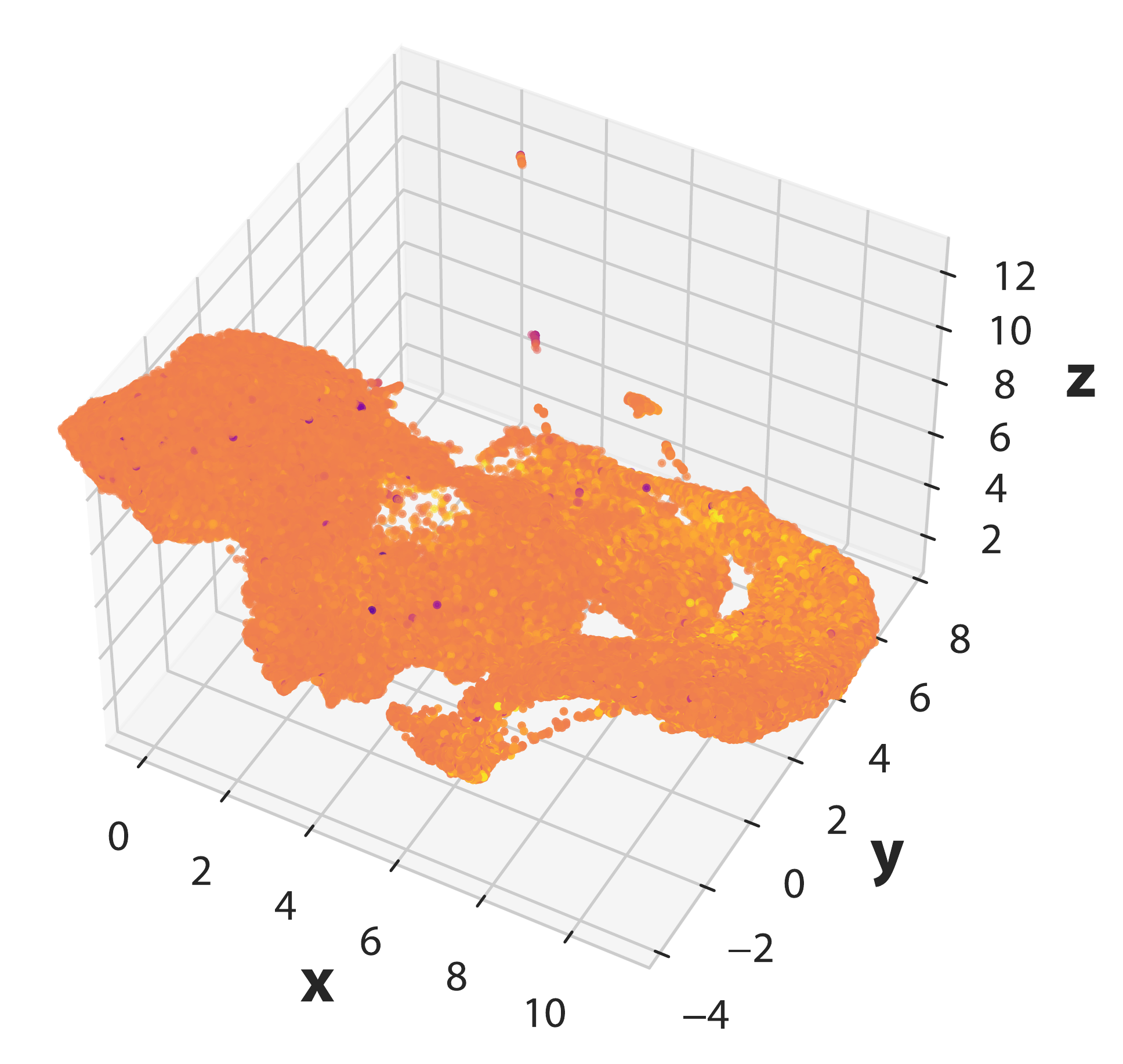}
    \end{minipage} \\

    \bottomrule

    \end{tabular}
\end{table}
\renewcommand{\familydefault}{\rmdefault}

\clearpage
\section{Train losses plotted for the multi-million scale pharma datasets}
\label{sec:AZ-train-losses-plots}
\subsection{VGAE models}
\begin{figure}[htp]
    \captionsetup{skip=4pt, labelfont=bf}
    \caption{Train losses for the \textsc{vgae} models trained on the proprietary dataset with $\approx1$ million molecules.}
    \label{figure:1mil-VGAE-train-loss-supp}
    \centering
    \includegraphics[width=1\textwidth]{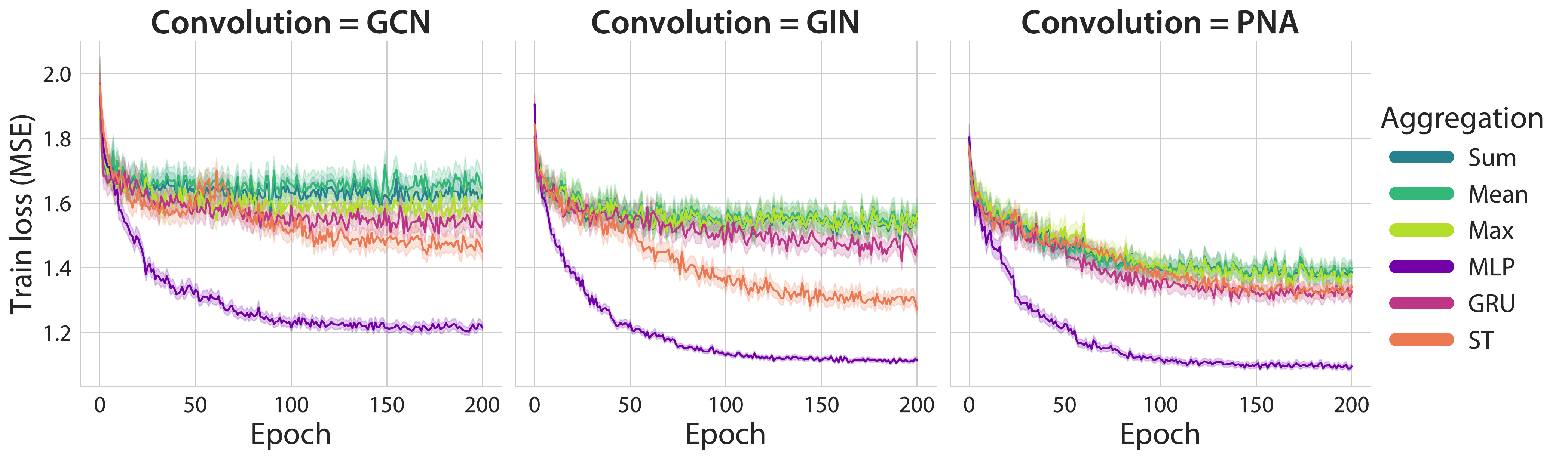}
\end{figure}
\renewcommand{\familydefault}{\rmdefault}

\begin{figure}[htp]
    \captionsetup{skip=4pt, labelfont=bf}
    \caption{Train losses for the \textsc{vgae} models trained on the proprietary dataset with $\approx1.5$ million molecules.}
    \label{figure:1.5mil-VGAE-train-loss-supp}
    \centering
    \includegraphics[width=1\textwidth]{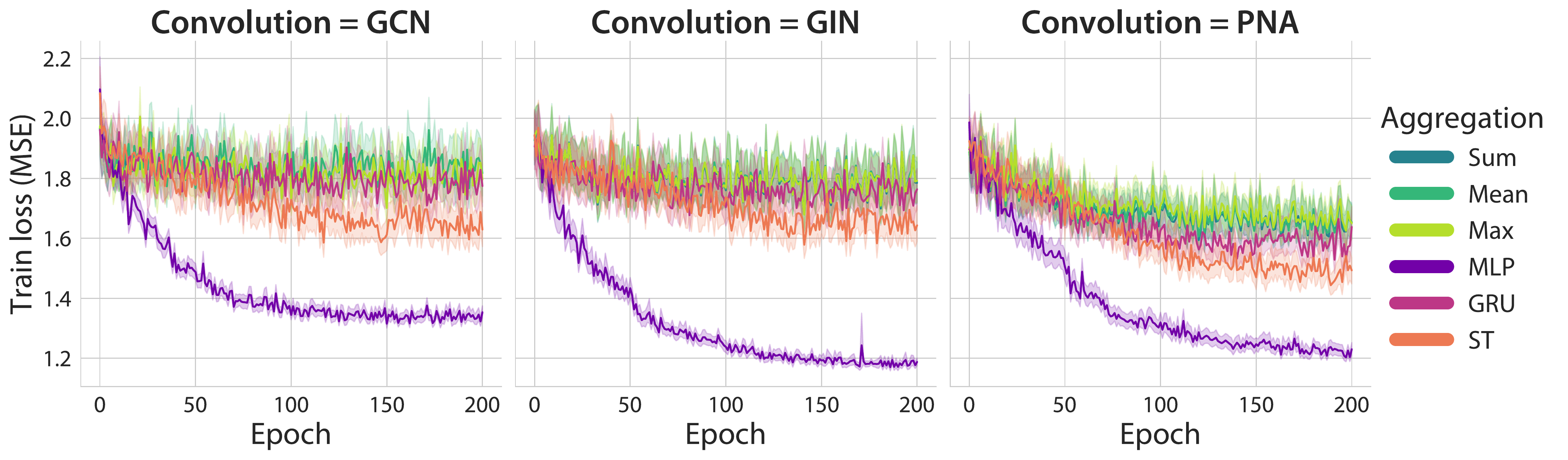}
\end{figure}
\renewcommand{\familydefault}{\rmdefault}

\begin{figure}[htp]
    \captionsetup{skip=4pt, labelfont=bf}
    \caption{Train losses for the \textsc{vgae} models trained on the proprietary dataset with $\approx2$ million molecules.}
    \label{figure:2mil-VGAE-train-loss}
    \centering
    \includegraphics[width=1\textwidth]{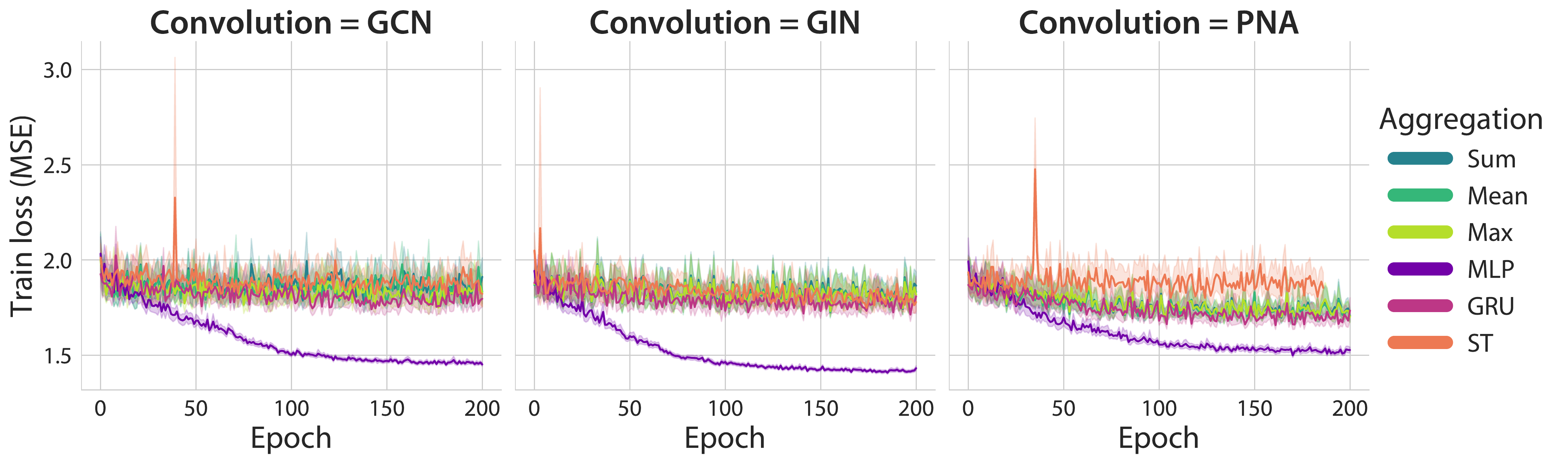}
\end{figure}
\renewcommand{\familydefault}{\rmdefault}

\clearpage
\subsection{GNN models}
\begin{figure}[htp]
    \captionsetup{skip=4pt, labelfont=bf}
    \caption{Train losses for the \textsc{gnn} models trained on the proprietary dataset with $\approx1$ million molecules.}
    \label{figure:1mil-GNN-train-loss}
    \centering
    \includegraphics[width=1\textwidth]{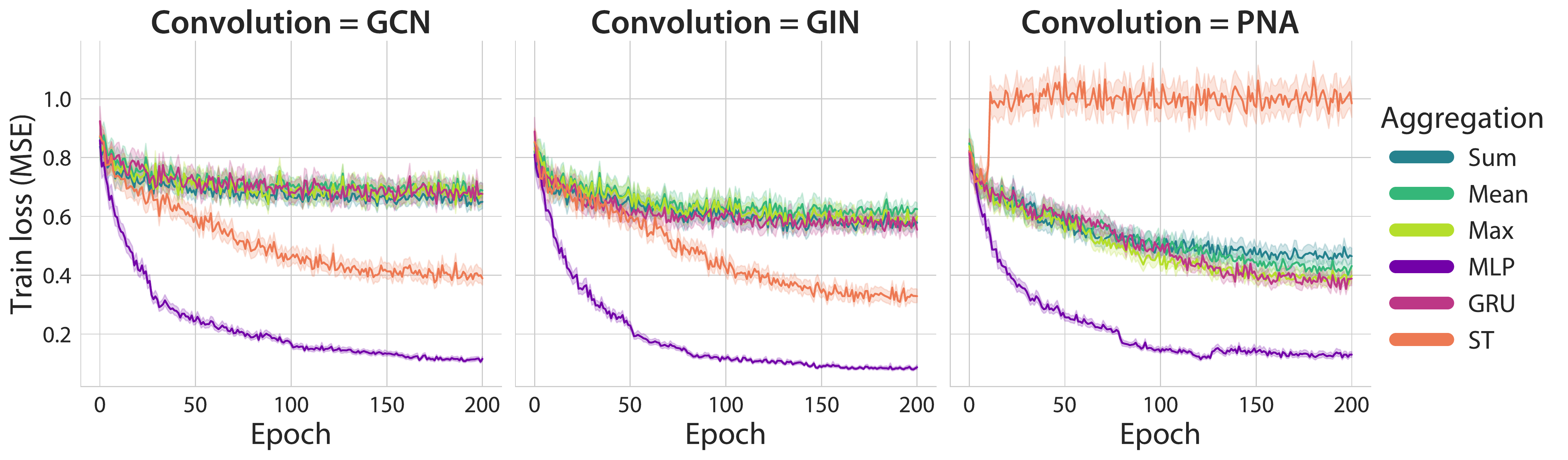}
\end{figure}
\renewcommand{\familydefault}{\rmdefault}

\begin{figure}[htp]
    \captionsetup{skip=4pt, labelfont=bf}
    \caption{Train losses for the \textsc{gnn} models trained on the proprietary dataset with $\approx1.5$ million molecules.}
    \label{figure:1.5mil-GNN-train-loss}
    \centering
    \includegraphics[width=1\textwidth]{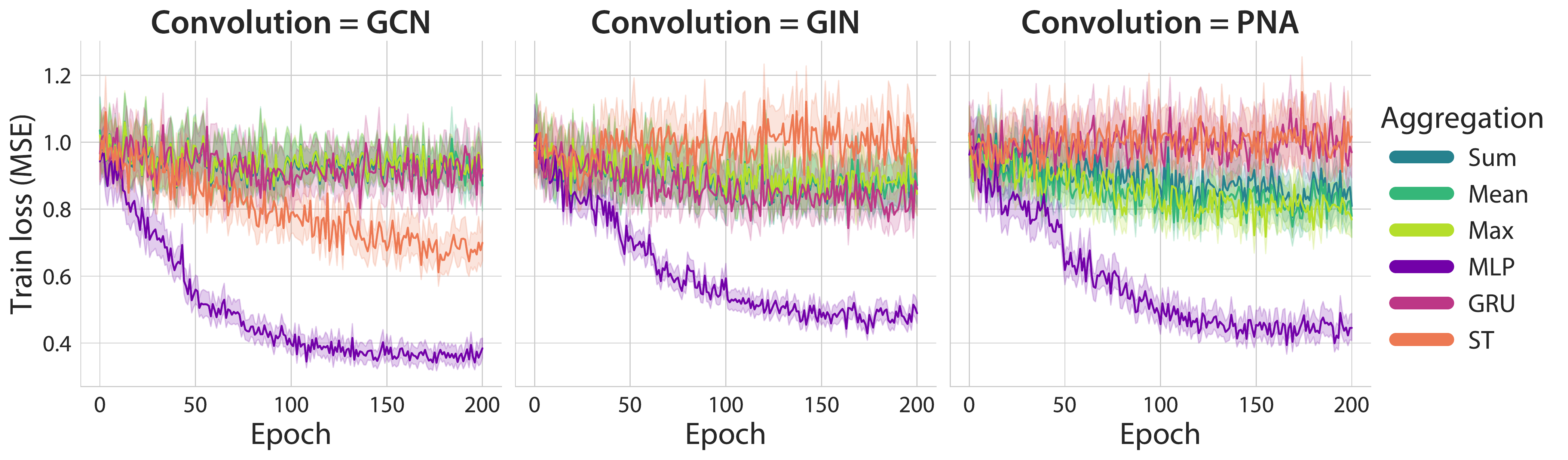}
\end{figure}
\renewcommand{\familydefault}{\rmdefault}

\begin{figure}[htp]
    \captionsetup{skip=4pt, labelfont=bf}
    \caption{Train losses for the \textsc{gnn} models trained on the proprietary dataset with $\approx2$ million molecules.}
    \label{figure:2mil-GNN-train-loss}
    \centering
    \includegraphics[width=1\textwidth]{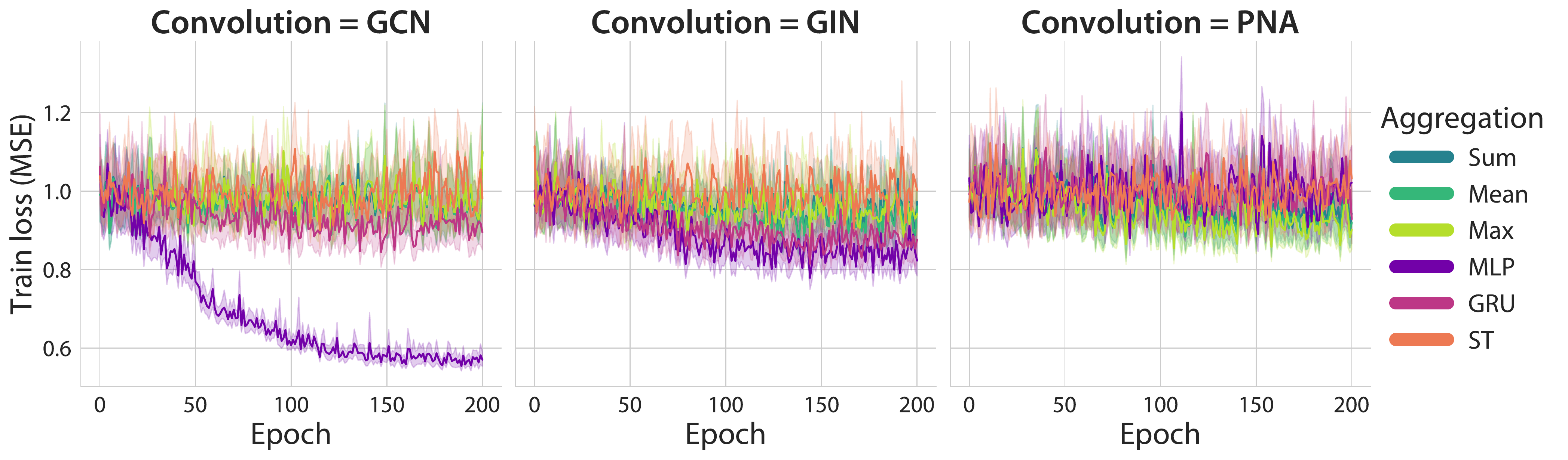}
\end{figure}
\renewcommand{\familydefault}{\rmdefault}

\clearpage
\section{Train losses for varying \textsc{gnn} depths on the proprietary dataset with 1.5 million molecules}
\label{section:train-loss-deeper}
\begin{figure}[H]
    \captionsetup{skip=4pt, labelfont=bf}
    \caption{Train losses (\textsc{mse}) for the \textsc{gnn} and guided \textsc{vgae} models trained on the proprietary dataset with $\approx1.5$ million molecules. The illustration includes representative aggregators for each category (sum, respectively \textsc{mlp}) evaluated with \textsc{gcn} layers across 3 different \textsc{gnn} depths ($2$, $3$, and $4$ layers). The \textsc{vgae} is higher as it includes additional loss terms.}
    \label{figure:1.5mil-GNN-VGAE-deeper}
    \centering
    \includegraphics[width=1\textwidth]{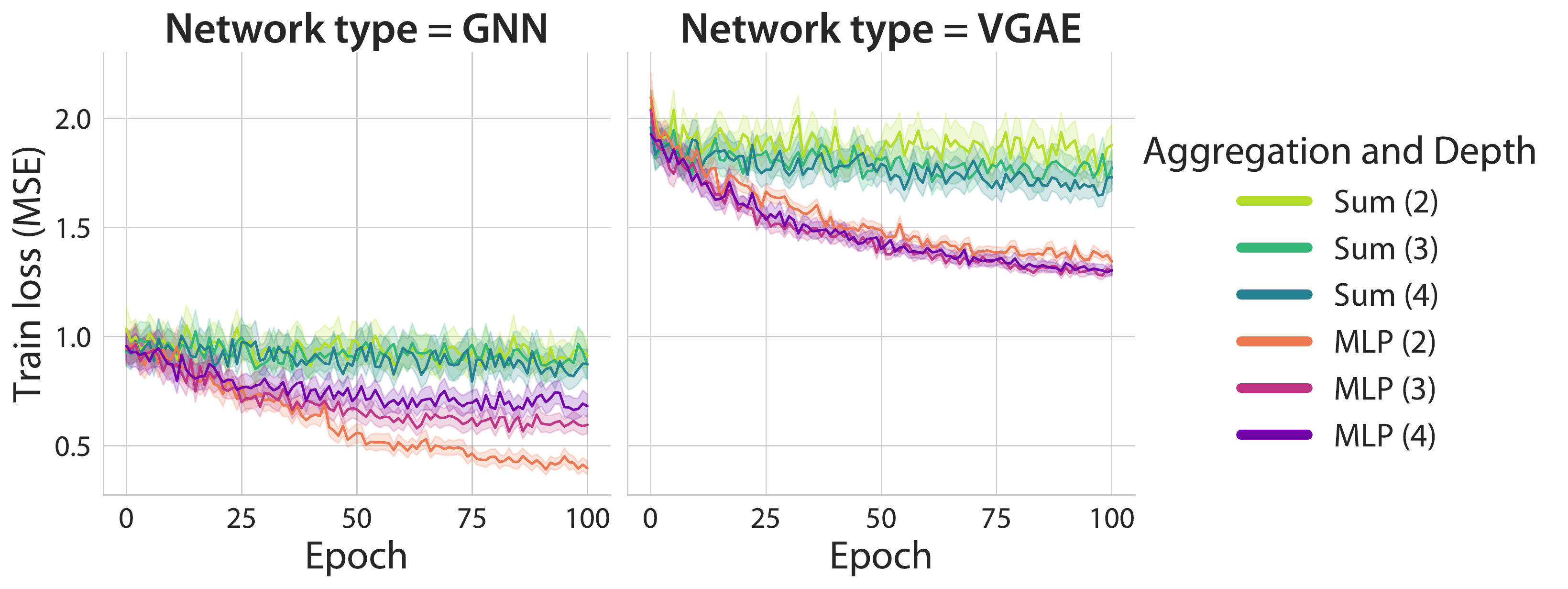}
\end{figure}
\renewcommand{\familydefault}{\rmdefault}

\section{Train metrics for the multi-million scale pharma datasets}
\label{section:train-metric-az-datasets}
\renewcommand{\sfdefault}{lmss}
\renewcommand{\familydefault}{\sfdefault}
\renewcommand{\arraystretch}{1.1}
\begin{table}[htp]
    \centering
    \captionsetup{justification=centering, skip=4pt, labelfont=bf, width=1\linewidth}
    \caption{Train metrics for the VGAE models trained on the proprietary dataset with $\approx$1 million molecules. A p-value of $<\epsilon$ indicates that the number returned by \texttt{scipy} (\texttt{stats.pearsonr}) was below the machine precision (thus, reported as $0$). A value of \textquote{N/A} indicates that it was not possible to compute the metric.}
    \label{table:1mil-VGAE-train-metrics}
    \begin{tabular}{ccclS[table-format=1.2]S[table-format=1.2]c}
        \toprule
        \textbf{Dataset} & \textbf{GNN or VGAE} & \textbf{Convolution} & \multicolumn{1}{c}{\textbf{Aggregator}} & \textbf{MAE} & \textbf{\textbf{\textbf{R}$\bm{^\mathsf{2}}$}} & \textbf{\textbf{\textbf{R}$\bm{^\mathsf{2}}$} p-value} \\
        \midrule
        \multirow{18}{*}[-1.4em]{\makecell{Bio-affinity\\1 mil.}} & \multirow{18}{*}[-1.4em]{VGAE} & \multirow{6}{*}[-0.35em]{GCN} & Sum & 1.1259 & 0.2885 & $<\epsilon$ \\
         &  &  & Mean & 1.1490 & 0.2657 & $<\epsilon$ \\
         &  &  & Max & 1.1126 & 0.3256 & $<\epsilon$ \\ \cmidrule(l){4-7}
         &  &  & \textbf{MLP} & $\bm{\mathsf{0.68}}$ & $\bm{\mathsf{0.78}}$ & \textbf{$<\epsilon$} \\
         &  &  & GRU & 1.0836 & 0.3681 & $<\epsilon$ \\
         &  &  & ST & 0.9961 & 0.4709 & $<\epsilon$ \\ \cmidrule[0.75pt]{3-7}
         &  & \multirow{6}{*}[-0.35em]{GIN} & Sum & 1.0700 & 0.3775 & $<\epsilon$ \\
         &  &  & Mean & 1.0761 & 0.3677 & $<\epsilon$ \\
         &  &  & Max & 1.0719 & 0.3717 & $<\epsilon$ \\ \cmidrule(l){4-7}
         &  &  & \textbf{MLP} & $\bm{\mathsf{0.55}}$ & $\bm{\mathsf{0.86}}$ & \textbf{$<\epsilon$} \\
         &  &  & GRU & 1.0293 & 0.4369 & $<\epsilon$ \\
         &  &  & ST & 0.8351 & 0.6482 & $<\epsilon$ \\ \cmidrule[0.75pt]{3-7}
         &  & \multirow{6}{*}[-0.35em]{PNA} & Sum & 0.9339 & 0.5451 & $<\epsilon$ \\
         &  &  & Mean & 0.9430 & 0.5342 & $<\epsilon$ \\
         &  &  & Max & 1.0260 & 0.5353 & $<\epsilon$ \\ \cmidrule(l){4-7}
         &  &  & \textbf{MLP} & $\bm{\mathsf{0.56}}$ & $\bm{\mathsf{0.86}}$ & \textbf{$<\epsilon$} \\
         &  &  & GRU & 0.8901 & 0.5930 & $<\epsilon$ \\
         &  &  & ST & 0.7550 & 0.7179 & $<\epsilon$ \\ \bottomrule
    \end{tabular}
\end{table}
\renewcommand{\familydefault}{\rmdefault}

\renewcommand{\sfdefault}{lmss}
\renewcommand{\familydefault}{\sfdefault}
\renewcommand{\arraystretch}{1.1}
\begin{table}[htp]
    \centering
    \captionsetup{justification=centering, skip=4pt, labelfont=bf, width=1\linewidth}
    \caption{Train metrics for the GNN models trained on the proprietary dataset with $\approx$1 million molecules. A p-value of $<\epsilon$ indicates that the number returned by \texttt{scipy} (\texttt{stats.pearsonr}) was below the machine precision (thus, reported as $0$). A value of \textquote{N/A} indicates that it was not possible to compute the metric.}
    \label{table:1mil-GNN-train-metrics}
    \begin{tabular}{ccclS[table-format=1.2]S[table-format=1.2]c}
        \toprule
        \textbf{Dataset} & \textbf{GNN or VGAE} & \textbf{Convolution} & \multicolumn{1}{c}{\textbf{Aggregator}} & \textbf{MAE} & \textbf{\textbf{\textbf{R}$\bm{^\mathsf{2}}$}} & \textbf{\textbf{\textbf{R}$\bm{^\mathsf{2}}$} p-value} \\
        \midrule
        \multirow{18}{*}[-1.4em]{\makecell{Bio-affinity\\ 1 mil.}} & \multirow{18}{*}[-1.4em]{GNN} & \multirow{6}{*}[-0.35em]{GCN} & Sum & 1.0960 & 0.3426 & $<\epsilon$ \\
         &  &  & Mean & 1.1205 & 0.3034 & $<\epsilon$ \\
         &  &  & Max & 1.1123 & 0.3172 & $<\epsilon$ \\ \cmidrule(l){4-7}
         &  &  & \textbf{MLP} & $\bm{\mathsf{0.47}}$ & $\bm{\mathsf{0.89}}$ & \textbf{$<\epsilon$} \\
         &  &  & GRU & 1.1107 & 0.3174 & $<\epsilon$ \\
         &  &  & ST & 0.8642 & 0.6017 & $<\epsilon$ \\ \cmidrule[0.75pt]{3-7}
         &  & \multirow{6}{*}[-0.35em]{GIN} & Sum & 1.0324 & 0.4269 & $<\epsilon$ \\
         &  &  & Mean & 1.0626 & 0.3833 & $<\epsilon$ \\
         &  &  & Max & 1.0426 & 0.4118 & $<\epsilon$ \\ \cmidrule(l){4-7}
         &  &  & \textbf{MLP} & $\bm{\mathsf{0.40}}$ & $\bm{\mathsf{0.92}}$ & \textbf{$<\epsilon$} \\
         &  &  & GRU & 1.0367 & 0.4230 & $<\epsilon$ \\
         &  &  & ST & 0.7928 & 0.6633 & $<\epsilon$ \\ \cmidrule[0.75pt]{3-7}
         &  & \multirow{6}{*}[-0.35em]{PNA} & Sum & 0.9423 & 0.5362 & $<\epsilon$ \\
         &  &  & Mean & 0.9011 & 0.5859 & $<\epsilon$ \\
         &  &  & Max & 0.8672 & 0.6145 & $<\epsilon$ \\ \cmidrule(l){4-7}
         &  &  & \textbf{MLP} & $\bm{\mathsf{0.46}}$ & $\bm{\mathsf{0.87}}$ & \textbf{$<\epsilon$} \\
         &  &  & GRU & 0.8461 & 0.6247 & $<\epsilon$ \\
         &  &  & ST & 1.3401 & N/A & N/A \\ \bottomrule
    \end{tabular}
\end{table}
\renewcommand{\familydefault}{\rmdefault}

\renewcommand{\sfdefault}{lmss}
\renewcommand{\familydefault}{\sfdefault}
\renewcommand{\arraystretch}{1.1}
\begin{table}[htp]
    \centering
    \captionsetup{justification=centering, skip=4pt, labelfont=bf, width=1\linewidth}
    \caption{Train metrics for the VGAE models trained on the proprietary dataset with $\approx$1.5 million molecules. A p-value of $<\epsilon$ indicates that the number returned by \texttt{scipy} (\texttt{stats.pearsonr}) was below the machine precision (thus, reported as $0$). A value of \textquote{N/A} indicates that it was not possible to compute the metric.}
    \label{table:1.5mil-VGAE-train-metrics}
    \begin{tabular}{ccclS[table-format=1.2]S[table-format=1.2]c}
        \toprule
        \textbf{Dataset} & \textbf{GNN or VGAE} & \textbf{Convolution} & \multicolumn{1}{c}{\textbf{Aggregator}} & \textbf{MAE} & \textbf{\textbf{\textbf{R}$\bm{^\mathsf{2}}$}} & \textbf{\textbf{\textbf{R}$\bm{^\mathsf{2}}$} p-value} \\
        \midrule
        \multirow{18}{*}[-1.4em]{\makecell{Bio-affinity\\ 1.5 mil.}} & \multirow{18}{*}[-1.4em]{VGAE} & \multirow{6}{*}[-0.35em]{GCN} & Sum & 1.2835 & 0.0748 & $<\epsilon$ \\
         &  &  & Mean & 1.2809 & 0.0515 & $<\epsilon$ \\
         &  &  & Max & 1.3154 & 0.0743 & $<\epsilon$ \\ \cmidrule(l){4-7}
         &  &  & \textbf{MLP} & $\bm{\mathsf{0.93}}$ & $\bm{\mathsf{0.64}}$ & \textbf{$<\epsilon$} \\
         &  &  & GRU & 1.2755 & 0.1026 & $<\epsilon$ \\
         &  &  & ST & 1.1748 & 0.2948 & $<\epsilon$ \\ \cmidrule[0.75pt]{3-7}
         &  & \multirow{6}{*}[-0.35em]{GIN} & Sum & 1.2553 & 0.1104 & $<\epsilon$ \\
         &  &  & Mean & 1.2532 & 0.1001 & $<\epsilon$ \\
         &  &  & Max & 1.3116 & 0.1034 & $<\epsilon$ \\ \cmidrule(l){4-7}
         &  &  & \textbf{MLP} & $\bm{\mathsf{0.78}}$ & $\bm{\mathsf{0.78}}$ & \textbf{$<\epsilon$} \\
         &  &  & GRU & 1.2645 & 0.1333 & $<\epsilon$ \\
         &  &  & ST & 1.1868 & 0.2699 & $<\epsilon$ \\ \cmidrule[0.75pt]{3-7}
         &  & \multirow{6}{*}[-0.35em]{PNA} & Sum & 1.2101 & 0.2597 & $<\epsilon$ \\
         &  &  & Mean & 1.2091 & 0.2582 & $<\epsilon$ \\
         &  &  & Max & 1.3389 & 0.2432 & $<\epsilon$ \\ \cmidrule(l){4-7}
         &  &  & \textbf{MLP} & $\bm{\mathsf{0.83}}$ & $\bm{\mathsf{0.73}}$ & \textbf{$<\epsilon$} \\
         &  &  & GRU & 1.1891 & 0.2993 & $<\epsilon$ \\
         &  &  & ST & 1.0897 & 0.4297 & $<\epsilon$ \\ \bottomrule
    \end{tabular}
\end{table}
\renewcommand{\familydefault}{\rmdefault}

\renewcommand{\sfdefault}{lmss}
\renewcommand{\familydefault}{\sfdefault}
\renewcommand{\arraystretch}{1.1}
\begin{table}[htp]
    \centering
    \captionsetup{justification=centering, skip=4pt, labelfont=bf, width=1\linewidth}
    \caption{Train metrics for the GNN models trained on the proprietary dataset with $\approx$1.5 million molecules. A p-value of $<\epsilon$ indicates that the number returned by \texttt{scipy} (\texttt{stats.pearsonr}) was below the machine precision (thus, reported as $0$). A value of \textquote{N/A} indicates that it was not possible to compute the metric.}
    \label{table:1.5mil-GNN-train-metrics}
    \begin{tabular}{ccclS[table-format=1.2]S[table-format=1.2]c}
        \toprule
        \textbf{Dataset} & \textbf{GNN or VGAE} & \textbf{Convolution} & \multicolumn{1}{c}{\textbf{Aggregator}} & \textbf{MAE} & \textbf{\textbf{\textbf{R}$\bm{^\mathsf{2}}$}} & \textbf{\textbf{\textbf{R}$\bm{^\mathsf{2}}$} p-value} \\
        \midrule
        \multirow{18}{*}[-1.4em]{\makecell{Bio-affinity\\ 1.5 mil.}} & \multirow{18}{*}[-1.4em]{GNN} & \multirow{6}{*}[-0.35em]{GCN} & Sum & 1.2862 & 0.0690 & $<\epsilon$ \\
         &  &  & Mean & 1.2877 & 0.0596 & $<\epsilon$ \\
         &  &  & Max & 1.2877 & 0.0604 & $<\epsilon$ \\ \cmidrule(l){4-7}
         &  &  & \textbf{MLP} & $\bm{\mathsf{0.90}}$ & $\bm{\mathsf{0.64}}$ & \textbf{$<\epsilon$} \\
         &  &  & GRU & 1.2788 & 0.0933 & $<\epsilon$ \\
         &  &  & ST & 1.1546 & 0.3232 & $<\epsilon$ \\ \cmidrule[0.75pt]{3-7}
         &  & \multirow{6}{*}[-0.35em]{GIN} & Sum & 1.2630 & 0.1250 & $<\epsilon$ \\
         &  &  & Mean & 1.2652 & 0.1209 & $<\epsilon$ \\
         &  &  & Max & 1.2688 & 0.1137 & $<\epsilon$ \\ \cmidrule(l){4-7}
         &  &  & \textbf{MLP} & $\bm{\mathsf{1.04}}$ & $\bm{\mathsf{0.50}}$ & \textbf{$<\epsilon$} \\
         &  &  & GRU & 1.2470 & 0.1607 & $<\epsilon$ \\
         &  &  & ST & 1.3003 & N/A & N/A \\ \cmidrule[0.75pt]{3-7}
         &  & \multirow{6}{*}[-0.35em]{PNA} & Sum & 1.2515 & 0.1572 & $<\epsilon$ \\
         &  &  & Mean & 1.2422 & 0.1901 & $<\epsilon$ \\
         &  &  & Max & 1.2258 & 0.1984 & $<\epsilon$ \\ \cmidrule(l){4-7}
         &  &  & \textbf{MLP} & $\bm{\mathsf{0.96}}$ & $\bm{\mathsf{0.56}}$ & \textbf{$<\epsilon$} \\
         &  &  & GRU & 1.3003 & 0.0000 & $\mathsf{4.22}\times\mathsf{10}^{\mathsf{21}}$ \\
         &  &  & ST & 1.3003 & N/A & N/A \\ \bottomrule
    \end{tabular}
\end{table}
\renewcommand{\familydefault}{\rmdefault}

\renewcommand{\sfdefault}{lmss}
\renewcommand{\familydefault}{\sfdefault}
\renewcommand{\arraystretch}{1.1}
\begin{table}[htp]
    \centering
    \captionsetup{justification=centering, skip=4pt, labelfont=bf, width=1\linewidth}
    \caption{Train metrics for the VGAE models trained on the proprietary dataset with $\approx$2 million molecules. A p-value of $<\epsilon$ indicates that the number returned by \texttt{scipy} (\texttt{stats.pearsonr}) was below the machine precision (thus, reported as $0$). A value of \textquote{N/A} indicates that it was not possible to compute the metric.}
    \label{table:2mil-VGAE-train-metrics}
    \begin{tabular}{ccclS[table-format=1.2]S[table-format=1.2]c}
        \toprule
        \textbf{Dataset} & \textbf{GNN or VGAE} & \textbf{Convolution} & \multicolumn{1}{c}{\textbf{Aggregator}} & \textbf{MAE} & \textbf{\textbf{\textbf{R}$\bm{^\mathsf{2}}$}} & \textbf{\textbf{\textbf{R}$\bm{^\mathsf{2}}$} p-value} \\
        \midrule
        \multirow{18}{*}[-1.4em]{\makecell{Bio-affinity\\ 2 mil.}} & \multirow{18}{*}[-1.4em]{VGAE} & \multirow{6}{*}[-0.35em]{GCN} & Sum & 0.8261 & 0.0202 & $<\epsilon$ \\
         &  &  & Mean & 0.8259 & 0.0319 & $<\epsilon$ \\
         &  &  & Max & 0.8353 & 0.0600 & $<\epsilon$ \\ \cmidrule(l){4-7}
         &  &  & \textbf{MLP} &  $\bm{\mathsf{0.63}}$ &  $\bm{\mathsf{0.52}}$ & \textbf{$<\epsilon$} \\
         &  &  & GRU & 0.8238 & 0.0947 & $<\epsilon$ \\
         &  &  & ST & 0.8259 & N/A & N/A \\ \cmidrule[0.75pt]{3-7}
         &  & \multirow{6}{*}[-0.35em]{GIN} & Sum & 0.8227 & 0.0592 & $<\epsilon$ \\
         &  &  & Mean & 0.8236 & 0.0712 & $<\epsilon$ \\
         &  &  & Max & 0.8348 & 0.0756 & $<\epsilon$ \\ \cmidrule(l){4-7}
         &  &  & \textbf{MLP} &  $\bm{\mathsf{0.61}}$ &  $\bm{\mathsf{0.55}}$ & \textbf{$<\epsilon$} \\
         &  &  & GRU & 0.8216 & 0.1120 & $<\epsilon$ \\
         &  &  & ST & 0.8187 & 0.0964 & $<\epsilon$ \\ \cmidrule[0.75pt]{3-7}
         &  & \multirow{6}{*}[-0.35em]{PNA} & Sum & 0.8151 & 0.1483 & $<\epsilon$ \\
         &  &  & Mean & 0.8151 & 0.1523 & $<\epsilon$ \\
         &  &  & Max & 0.8384 & 0.1478 & $<\epsilon$ \\ \cmidrule(l){4-7}
         &  &  & \textbf{MLP} &  $\bm{\mathsf{0.70}}$ &  $\bm{\mathsf{0.42}}$ & \textbf{$<\epsilon$} \\
         &  &  & GRU & 0.8128 & 0.1692 & $<\epsilon$ \\
         &  &  & ST & 0.8259 & N/A & N/A \\ \bottomrule
    \end{tabular}
\end{table}
\renewcommand{\familydefault}{\rmdefault}

\renewcommand{\sfdefault}{lmss}
\renewcommand{\familydefault}{\sfdefault}
\renewcommand{\arraystretch}{1.1}
\begin{table}[htp]
    \centering
    \captionsetup{justification=centering, skip=4pt, labelfont=bf, width=1\linewidth}
    \caption{Train metrics for the GNN models trained on the proprietary dataset with $\approx$2 million molecules. A p-value of $<\epsilon$ indicates that the number returned by \texttt{scipy} (\texttt{stats.pearsonr}) was below the machine precision (thus, reported as $0$). A value of \textquote{N/A} indicates that it was not possible to compute the metric.}
    \label{table:2mil-GNN-train-metrics}
    \begin{tabular}{ccclS[table-format=1.2]S[table-format=1.2]c}
        \toprule
        \textbf{Dataset} & \textbf{GNN or VGAE} & \textbf{Convolution} & \multicolumn{1}{c}{\textbf{Aggregator}} & \textbf{MAE} & \textbf{\textbf{\textbf{R}$\bm{^\mathsf{2}}$}} & \textbf{\textbf{\textbf{R}$\bm{^\mathsf{2}}$} p-value} \\
        \midrule
        \multirow{18}{*}[-1.4em]{\makecell{Bio-affinity\\ 2 mil.}} & \multirow{18}{*}[-1.4em]{GNN} & \multirow{6}{*}[-0.35em]{GCN} & Sum & 0.8257 & 0.0268 & $<\epsilon$ \\
         &  &  & Mean & 0.8255 & 0.0297 & $<\epsilon$ \\
         &  &  & Max & 0.8256 & 0.0225 & $<\epsilon$ \\ \cmidrule(l){4-7}
        %  &  &  & \textbf{Dense} & $\bm{\mathsf{0.6765}}$ & $\bm{\mathsf{0.4288}}$ & \textbf{0} \\
        &  &  & \textbf{MLP} & $\bm{\mathsf{0.68}}$ & $\bm{\mathsf{0.43}}$ & \textbf{$<\epsilon$} \\
         &  &  & GRU & 0.8159 & 0.0910 & $<\epsilon$ \\
         &  &  & ST & 0.8259 & N/A & N/A \\ \cmidrule[0.75pt]{3-7}
         &  & \multirow{6}{*}[-0.35em]{GIN} & Sum & 0.8233 & 0.0484 & $<\epsilon$ \\
         &  &  & Mean & 0.8226 & 0.0643 & $<\epsilon$ \\
         &  &  & Max & 0.8228 & 0.0496 & $<\epsilon$ \\ \cmidrule(l){4-7}
        %  &  &  & \textbf{Dense} & 0.7968 & 0.1551 & 0 \\
         &  &  & \textbf{MLP} & $\bm{\mathsf{0.80}}$ & $\bm{\mathsf{0.16}}$ & \textbf{$<\epsilon$} \\
         &  &  & GRU & 0.8113 & 0.1205 & $<\epsilon$ \\
         &  &  & ST & 0.8259 & N/A & N/A \\ \cmidrule[0.75pt]{3-7}
         &  & \multirow{6}{*}[-0.35em]{PNA} & Sum & 0.8216 & 0.0490 & $<\epsilon$ \\
         &  &  & Mean & 0.8201 & 0.0624 & $<\epsilon$ \\
        %  &  &  & Max & 0.8174 & 0.0749 & 0 \\
         &  &  &  \textbf{Max} & $\bm{\mathsf{0.82}}$ & $\bm{\mathsf{0.07}}$ & $<\epsilon$ \\ \cmidrule(l){4-7}
         &  &  & MLP & 0.8259 & 0.0000 & $\mathsf{1.29}\times\mathsf{10}^{\mathsf{-13}}$ \\
         &  &  & GRU & 0.8260 & 0.0007 & $<\epsilon$ \\
         &  &  & ST & 0.8259 & N/A & N/A \\
        \bottomrule
    \end{tabular}
\end{table}
\renewcommand{\familydefault}{\rmdefault}

\clearpage
\section{Experimental platform}
\label{sec:exp-platform}
We used two different platforms for training all the models discussed in the paper. Firstly, a workstation equipped with an AMD Ryzen 5950X processor with 16 cores and 32 threads, an Nvidia RTX 3090 graphics card with 24GB of VRAM, and 64GB of DDR4 RAM. The used operating system is Ubuntu 21.10, with Python 3.9.9, PyTorch 1.10.1 with CUDA 11.3, PyTorch Geometric 2.0.3, and PyTorch Lightning 1.5.7.

Secondly, we used GPU-enabled systems from the AstraZeneca Scientific Computing Platform, generally equipped with Intel processors, Nvidia Tesla V100 GPUs with either 16GB or 32GB, and as much RAM as needed for the experiments. The cloud systems run CentOS Linux 7, with Python 3.9.7, PyTorch 1.8.2 and CUDA 10.2, PyTorch Geometric 2.0.1, and PyTorch Lightning 1.5.5.

% \clearpage
\section{Statistical significance of dataset attributes for the observed performance}
\label{sec:stat-analysis}
We computed the difference between the best neural aggregator and the best non-neural aggregator for each dataset and for each one of the five random splits ($\Delta \text{R}^2$ for regression datasets and $\Delta \textsc{mcc}$ for classification datasets). Using these resulting metrics, we fitted multiple linear regression models (\texttt{lm} in R) to explain and quantify the relationship between dataset attributes (size, average number of nodes per graph, and others) and the difference in performance.

For the regression datasets, all the dataset attributes from \Cref{section:datasets-summary}, \Cref{table:summary-of-datasets} except the number of tasks were statistically significant at different levels (\Cref{sec:stat-analysis}, \Cref{table:lm-regr-delta-r2}). Most had a positive relationship to the $\Delta \text{R}^2$ except the average number of edges per graph and the number of tasks per dataset. These results suggest that larger regression datasets with more nodes per graph and more node features are likely to see large improvements when using neural aggregators.

For the classification datasets there were no statistically significant dataset attributes (\Cref{sec:stat-analysis}, \Cref{table:lm-regr-cls-mcc}), indicating that the proposed techniques are not particularly tied to characteristics like dataset size or graph properties.

\renewcommand{\familydefault}{\sfdefault}
\renewcommand{\arraystretch}{1.1}
\begin{table}[H]
    \centering
    \captionsetup{justification=centering, skip=4pt, labelfont=bf, width=0.98\linewidth}
    \caption{Summary of the multiple linear regression model that explains the $\Delta \text{R}^2$ in terms of the dataset attributes. $p$-value significance is indicated by the \textquote{Sign.} column (\textquote{***} for < 0.001, \textquote{**} for < 0.01, \textquote{.} for < 0.1).}
    \label{table:lm-regr-delta-r2}
    \begin{tabular}{lcccc} \toprule
     Predictors  & Estimates & 95\% Confidence Interval & p-value & Sign. \\\midrule
    (Intercept) & $-7.06 \times 10^{-01}$     & $-9.88 \times 10^{-01}$ \quad -- \quad $-4.24 \times 10^{-01}$ & $1.33 \times 10^{-05}$ & *** \\
    Size        & $\textcolor{white}{-}1.84 \times 10^{-06}$      & $\textcolor{white}{-}8.99 \times 10^{-07}$ \quad -- \quad $\textcolor{white}{-}2.78 \times 10^{-06}$    & $3.46 \times 10^{-04}$ & ***  \\
    Avg. nodes   & $\textcolor{white}{-}6.32 \times 10^{-02}$     & $\textcolor{white}{-}2.71 \times 10^{-02}$    \quad -- \quad $\textcolor{white}{-}9.94 \times 10^{-02}$    & $1.14 \times 10^{-03}$ & **  \\
    Avg. edges   & $-2.36 \times 10^{-02}$     & $-3.92 \times 10^{-02}$    \quad -- \quad $-8.01 \times 10^{-03}$    & $4.11 \times 10^{-03}$ & **   \\
    Node attr.   & $\textcolor{white}{-}1.83 \times 10^{-02}$      & $\textcolor{white}{-}1.03 \times 10^{-02}$    \quad -- \quad $\textcolor{white}{-}2.63 \times 10^{-02}$    & $4.96 \times 10^{-05}$ & ***  \\
    Num. tasks     & $-8.59 \times 10^{-03}$     & $-1.73\times 10^{-02}$    \quad -- \quad $\textcolor{white}{-}9.44 \times 10^{-05}$    & $5.24 \times 10^{-02}$ & .  \\ \bottomrule
    \end{tabular}
\end{table}
\renewcommand{\familydefault}{\rmdefault}

\renewcommand{\familydefault}{\sfdefault}
\renewcommand{\arraystretch}{1.1}
\begin{table}[H]
    \centering
    \captionsetup{justification=centering, skip=4pt, labelfont=bf, width=0.98\linewidth}
    \caption{Summary of the multiple linear regression model that explains the $\Delta$MCC in terms of the dataset attributes. $p$-value significance is indicated by the \textquote{Sign.} column (\textquote{***} for < 0.001).}
    \label{table:lm-regr-cls-mcc}
    \begin{tabular}{lcccc} \toprule
     Predictors  & Estimates & 95\% Confidence Interval & p-value & Sign. \\\midrule
    (Intercept) & $\textcolor{white}{-}8.83 \times 10^{-02}$     & $\textcolor{white}{-}5.40 \times 10^{-02}$ \quad -- \quad \textcolor{white}{-}$1.23 \times 10^{-01}$ & $1.10 \times 10^{-06}$ & *** \\
    Size        & $-1.60 \times 10^{-07}$      & $-4.94 \times 10^{-07}$ \quad -- \quad \textcolor{white}{-}$1.73 \times 10^{-07}$    & $0.344$ &  \\
    Avg. nodes   & $\textcolor{white}{-}2.10 \times 10^{-04}$     & $-2.10 \times 10^{-04}$    \quad -- \quad \textcolor{white}{-}$6.29 \times 10^{-04}$    & $0.326$ &  \\
    Avg. edges   & $-8.56 \times 10^{-05}$     & $-2.65 \times 10^{-04}$    \quad -- \quad \textcolor{white}{-}$9.43 \times 10^{-05}$    & $0.349$ &  \\
    Node attr.   & $-3.36 \times 10^{-05}$      & $-8.60 \times 10^{-05}$    \quad -- \quad \textcolor{white}{-}$1.88 \times 10^{-05}$    & $0.207$ &  \\
    Num. tasks     & $-3.70 \times 10^{-04}$     & $-1.29 \times 10^{-03}$    \quad -- \quad \textcolor{white}{-}$5.47 \times 10^{-04}$    & $0.426$ &  \\ \bottomrule
    \end{tabular}
\end{table}
\renewcommand{\familydefault}{\rmdefault}

\section{Similar molecules with different representations}
\label{section:similar-molecules-diff-repr}
\begin{figure}[H]
    \centering
    \captionsetup{labelfont=bf}
    \caption{Similar molecules with different adjacency representations. The \textsc{smiles} is provided for both molecules.}
    \label{figure:similar-molecules}
    \begin{subfigure}[t]{0.49\textwidth}
        \centering
        \captionsetup{labelfont=bf, font=small, skip=-1pt}
        \caption{\texttt{NC[C@H](CC(=O)O)c1ccc(Cl)cc1}}
        \includegraphics[width=1\textwidth]{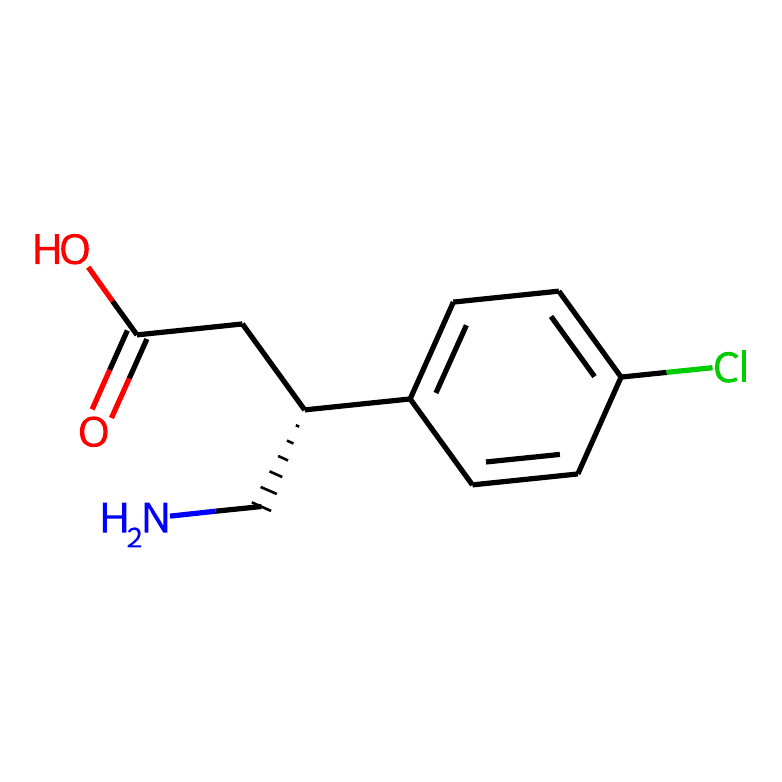}
    \end{subfigure}
    \hfill
    \begin{subfigure}[t]{0.49\textwidth}
        \centering
        \captionsetup{labelfont=bf, font=small, skip=-1pt}
        \caption{\texttt{C1=CC(=CC(=C1)Cl)C(CC(=O)O)CN}}
        \includegraphics[width=0.9\textwidth]{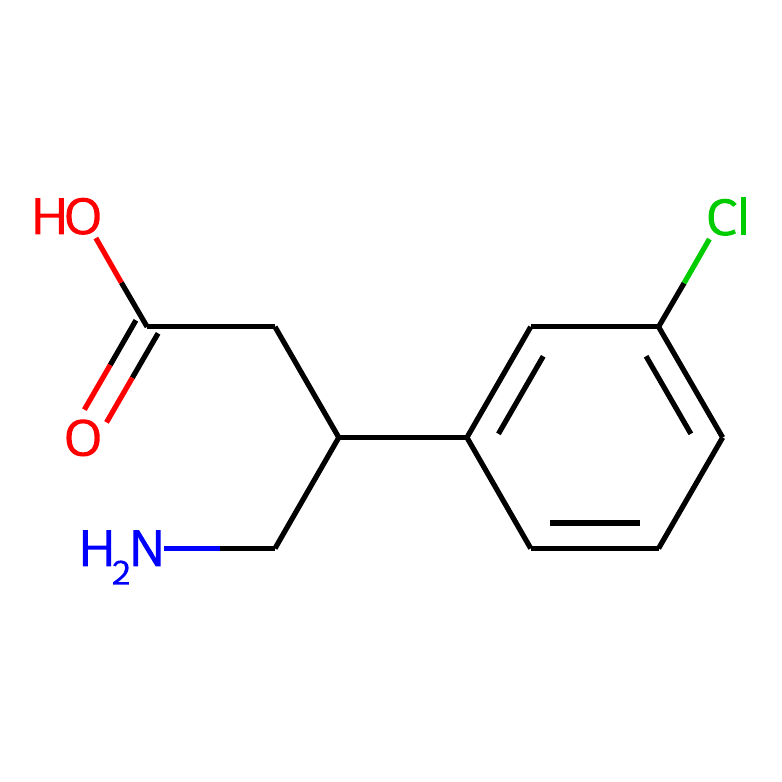}
    \end{subfigure}
\end{figure}

The example illustrates two very similar molecules with greatly different adjacency matrix representations (\Cref{section:similar-molecules-diff-repr}, \Cref{figure:similar-molecules}). Despite the similarity, only $6$ out of $14$ rows are identical in both matrices (rows $1$, $2$, $3$, $4$, $6$, $8$, numbering from $0$). There are no rows which occur in both matrices but in different orders.

\renewcommand{\sfdefault}{lmss}
\renewcommand{\familydefault}{\sfdefault}
\vspace{-1.5cm}
\renewcommand{\arraystretch}{0.95}
\setlength{\arraycolsep}{4.2pt}
$\begin{pmatrix}
    0  &  1  &  0  &  0  &  0  &  0  &  0  &  0  &  0  &  0  &  0  &  0  &  0  &  0 \\
    1  &  0  &  1  &  0  &  0  &  0  &  0  &  0  &  0  &  0  &  0  &  0  &  0  &  0 \\
    0  &  1  &  0  &  1  &  0  &  0  &  0  &  1  &  0  &  0  &  0  &  0  &  0  &  0 \\
    0  &  0  &  1  &  0  &  1  &  0  &  0  &  0  &  0  &  0  &  0  &  0  &  0  &  0 \\
    0  &  0  &  0  &  1  &  0  &  1  &  1  &  0  &  0  &  0  &  0  &  0  &  0  &  0 \\
    0  &  0  &  0  &  0  &  1  &  0  &  0  &  0  &  0  &  0  &  0  &  0  &  0  &  0 \\
    0  &  0  &  0  &  0  &  1  &  0  &  0  &  0  &  0  &  0  &  0  &  0  &  0  &  0 \\
    0  &  0  &  1  &  0  &  0  &  0  &  0  &  0  &  1  &  0  &  0  &  0  &  0  &  1 \\
    0  &  0  &  0  &  0  &  0  &  0  &  0  &  1  &  0  &  1  &  0  &  0  &  0  &  0 \\
    0  &  0  &  0  &  0  &  0  &  0  &  0  &  0  &  1  &  0  &  1  &  0  &  0  &  0 \\
    0  &  0  &  0  &  0  &  0  &  0  &  0  &  0  &  0  &  1  &  0  &  1  &  1  &  0 \\
    0  &  0  &  0  &  0  &  0  &  0  &  0  &  0  &  0  &  0  &  1  &  0  &  0  &  0 \\
    0  &  0  &  0  &  0  &  0  &  0  &  0  &  0  &  0  &  0  &  1  &  0  &  0  &  1 \\
    0  &  0  &  0  &  0  &  0  &  0  &  0  &  1  &  0  &  0  &  0  &  0  &  1  &  0
\end{pmatrix}
\begin{pmatrix}
    0  &  1  &  0  &  0  &  0  &  1  &  0  &  0  &  0  &  0  &  0  &  0  &  0  &  0 \\
    1  &  0  &  1  &  0  &  0  &  0  &  0  &  0  &  0  &  0  &  0  &  0  &  0  &  0 \\
    0  &  1  &  0  &  1  &  0  &  0  &  0  &  1  &  0  &  0  &  0  &  0  &  0  &  0 \\
    0  &  0  &  1  &  0  &  1  &  0  &  0  &  0  &  0  &  0  &  0  &  0  &  0  &  0 \\
    0  &  0  &  0  &  1  &  0  &  1  &  1  &  0  &  0  &  0  &  0  &  0  &  0  &  0 \\
    1  &  0  &  0  &  0  &  1  &  0  &  0  &  0  &  0  &  0  &  0  &  0  &  0  &  0 \\
    0  &  0  &  0  &  0  &  1  &  0  &  0  &  0  &  0  &  0  &  0  &  0  &  0  &  0 \\
    0  &  0  &  1  &  0  &  0  &  0  &  0  &  0  &  1  &  0  &  0  &  0  &  1  &  0 \\
    0  &  0  &  0  &  0  &  0  &  0  &  0  &  1  &  0  &  1  &  0  &  0  &  0  &  0 \\
    0  &  0  &  0  &  0  &  0  &  0  &  0  &  0  &  1  &  0  &  1  &  1  &  0  &  0 \\
    0  &  0  &  0  &  0  &  0  &  0  &  0  &  0  &  0  &  1  &  0  &  0  &  0  &  0 \\
    0  &  0  &  0  &  0  &  0  &  0  &  0  &  0  &  0  &  1  &  0  &  0  &  0  &  0 \\
    0  &  0  &  0  &  0  &  0  &  0  &  0  &  1  &  0  &  0  &  0  &  0  &  0  &  1 \\
    0  &  0  &  0  &  0  &  0  &  0  &  0  &  0  &  0  &  0  &  0  &  0  &  1  &  0
\end{pmatrix}$
\renewcommand{\familydefault}{\rmdefault}

\section{Similar molecules and the expressiveness of aggregators}
\label{section:similar-agg-expr}
\begin{figure}[H]
    \centering
    \captionsetup{labelfont=bf}
    \caption{Example of similar molecules with different properties.}
    \label{figure:hormones}
    \begin{subfigure}[t]{0.49\textwidth}
        \centering
        \captionsetup{labelfont=bf, font=small, skip=-1pt}
        \caption{Estradiol\\\texttt{CC12CCC3C(C1CCC2O)CCC4=C3C=CC(=C4)O}}
        \includegraphics[width=1\textwidth]{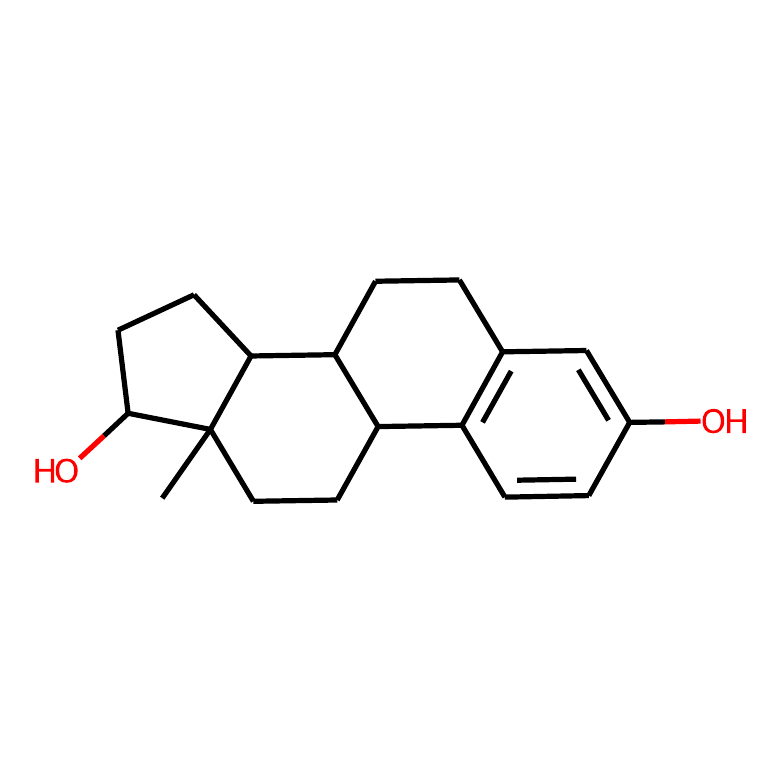}
    \end{subfigure}
    \hfill
    \begin{subfigure}[t]{0.49\textwidth}
        \centering
        \captionsetup{labelfont=bf, font=small, skip=-1pt}
        \caption{Testosterone\\\texttt{CC12CCC3C(C1CCC2O)CCC4=CC(=O)CCC34C}}
        \includegraphics[width=0.9\textwidth]{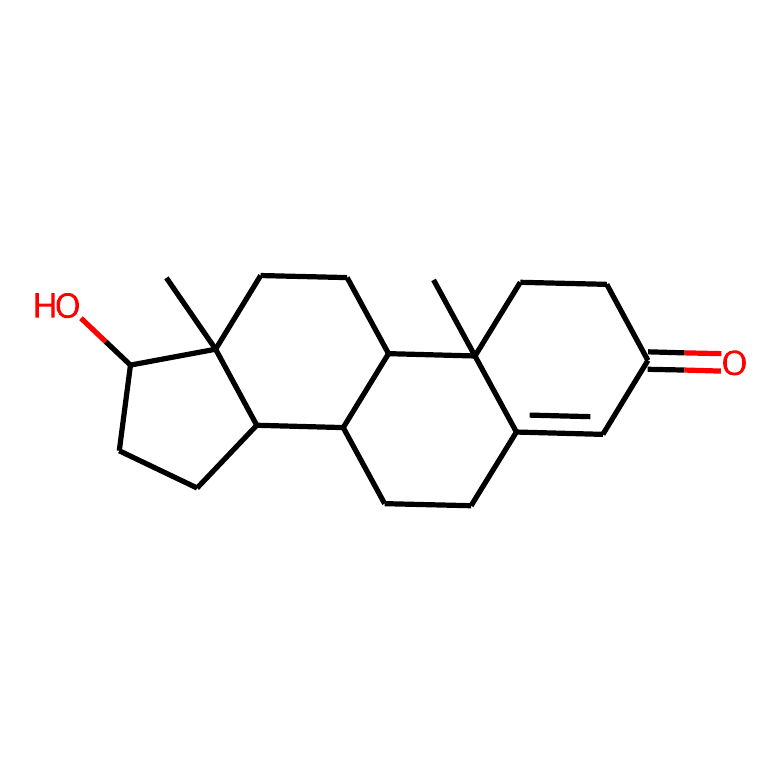}
    \end{subfigure}
\end{figure}

As a proof of concept, we selected two well-known molecules with similar structures, estradiol and testosterone (\Cref{section:similar-agg-expr}, \Cref{figure:hormones}). Despite the similarity, their induced biological effects can be very different. We can easily find several examples where the two compounds have different activity levels (active vs inactive) in high-throughput assays from PubChem, for example AID $588544$, AID $1347036$, AID $624032$, or AID $1259394$. The compound identifiers (CIDs) for the two compounds are $5757$ for estradiol and $6013$ for testosterone.

However, \textsc{gnn}s using sum, mean, or max readouts might find it challenging to discern between the two for bio-affinity predictions tasks. On a toy \textsc{gnn} with a single \textsc{gcn} layer, DeepChem featurisation ($30$ node features), $5$ output features from the \textsc{gcn} layer, and random initialization with a seed of $1$ (\texttt{pytorch\_lightning.seed\_everything(1)}), the three classical aggregators produced extremely close outputs (\Cref{section:similar-agg-expr}, \Cref{table:hormones-node-agg}). Although this is only a toy example, it highlights one possible limitation of the simple, existing readout functions.

\renewcommand{\sfdefault}{lmss}
\renewcommand{\familydefault}{\sfdefault}
\renewcommand{\arraystretch}{1.2}
\begin{table}[H]
    \centering
    \captionsetup{labelfont=bf, skip=5pt}
    \caption{Output of the three simple functions for the two similar molecules.}
    \label{table:hormones-node-agg}
    \begin{tabular}{lrr}
    \toprule
    \textbf{Aggregator} & \multicolumn{2}{c}{\textbf{Molecule}} \\
     & \multicolumn{1}{c}{Estradiol} & \multicolumn{1}{c}{Testosterone} \\ \midrule
    Sum & -39.296 & -40.433 \\
    Mean & -0.393 & -0.385 \\
    Max & 0.829 & 0.762 \\ \bottomrule
    \end{tabular}
\end{table}
\renewcommand{\familydefault}{\rmdefault}

\clearpage
\section{Detailed metrics for all 39 datasets/benchmarks}
\label{section:all-benchmarks-metrics}
The metrics for each layer type, readout, and random split are available in \textbf{Supplementary File 2} (available on GitHub).
\subsection{MoleculeNet regression models}
\renewcommand{\sfdefault}{lmss}
\renewcommand{\familydefault}{\sfdefault}

\begin{table}[!h]
    \captionsetup{justification=centering, skip=4pt, labelfont=bf, width=1.2\linewidth}
    \scriptsize
    \centering
    \caption{Detailed metrics (mean $\pm$ standard deviation) for the MoleculeNet regression datasets. For QM9, any differences in performance compared to other 2-layer models such as those in \Cref{figure:qm9-7-layers} might be due to different GNN hyperparameters, such as the output or intermediate node dimension (QM9-specific experiments generally used larger dimensions).}
    \label{table:MOLNET-REGR}
    \hspace*{-3cm}
\begin{tabular}{@{}clcc@{\hskip 0.5cm}cc@{\hskip 0.5cm}cc@{\hskip 0.5cm}cc@{\hskip 0.5cm}cc@{}}
    \toprule
                         &            & \multicolumn{2}{c@{\hspace{0.5cm}}}{\textbf{GCN}} & \multicolumn{2}{c@{\hspace{0.75cm}}}{\textbf{GAT}} & \multicolumn{2}{c@{\hspace{0.75cm}}}{\textbf{GATv2}} & \multicolumn{2}{c@{\hspace{0.75cm}}}{\textbf{GIN}} & \multicolumn{2}{c@{\hspace{0.25cm}}}{\textbf{PNA}} \\ \cmidrule(lr{0.5cm}){3-4} \cmidrule(r{0.5cm}){5-6} \cmidrule(r{0.5cm}){7-8} \cmidrule(r{0.5cm}){9-10} \cmidrule{11-12}
    Data.              & Agg. & \textbf{MAE}             & \textbf{\textbf{R}$\bm{^\mathsf{2}}$}             & \textbf{MAE}             & \textbf{\textbf{R}$\bm{^\mathsf{2}}$}            & \textbf{MAE}              & \textbf{\textbf{R}$\bm{^\mathsf{2}}$}             & \textbf{MAE}             & \textbf{\textbf{R}$\bm{^\mathsf{2}}$}            & \textbf{MAE}             & \textbf{\textbf{R}$\bm{^\mathsf{2}}$}            \\ \cmidrule(r){1-1} \cmidrule(lr){2-2} \cmidrule(lr){3-3} \cmidrule(lr{0.5cm}){4-4} \cmidrule(r){5-5} \cmidrule(lr{0.5cm}){6-6} \cmidrule(r){7-7} \cmidrule(lr{0.5cm}){8-8} \cmidrule(r){9-9} \cmidrule(lr{0.5cm}){10-10} \cmidrule(r){11-11} \cmidrule(l){12-12}
    \multirow{6}{*}{QM9} & Sum        & 0.74 $\pm$ 0.00     & 0.09 $\pm$ 0.00    & 0.76 $\pm$ 0.00     & 0.05 $\pm$ 0.01    & 0.75 $\pm$ 0.00      & 0.08 $\pm$ 0.01     & 0.71 $\pm$ 0.00     & 0.15 $\pm$ 0.00    & 0.70 $\pm$ 0.00     & 0.17 $\pm$ 0.01    \\
                         & Mean       & 0.73 $\pm$ 0.00     & 0.10 $\pm$ 0.00    & 0.75 $\pm$ 0.01     & 0.07 $\pm$ 0.01    & 0.74 $\pm$ 0.00      & 0.10 $\pm$ 0.00     & 0.72 $\pm$ 0.00     & 0.13 $\pm$ 0.00    & 0.70 $\pm$ 0.00     & 0.16 $\pm$ 0.01    \\
                         & Max        & 0.73 $\pm$ 0.00     & 0.11 $\pm$ 0.00    & 0.74 $\pm$ 0.00     & 0.10 $\pm$ 0.00    & 0.73 $\pm$ 0.00      & 0.11 $\pm$ 0.00     & 0.70 $\pm$ 0.00     & 0.16 $\pm$ 0.00    & 0.68 $\pm$ 0.00     & 0.20 $\pm$ 0.00    \\
\bottomrule
    \end{tabular}
    \hspace*{-3cm}
\end{table}
\renewcommand{\familydefault}{\rmdefault}

\clearpage
\subsection{MoleculeNet classification models}
\renewcommand{\sfdefault}{lmss}
\renewcommand{\familydefault}{\sfdefault}

\begin{table}[!h]
    \captionsetup{justification=centering, skip=4pt, labelfont=bf, width=1.2\linewidth}
    \scriptsize
    \centering
    \caption{Detailed metrics (mean $\pm$ standard deviation) for the MoleculeNet classification datasets.}
    \label{table:MOLNET-CLS}
    \hspace*{-3cm}
    \begin{tabular}{@{}clcc@{\hskip 0.5cm}cc@{\hskip 0.5cm}cc@{\hskip 0.5cm}cc@{\hskip 0.5cm}cc@{}}
    \toprule
                         &            & \multicolumn{2}{c@{\hspace{0.5cm}}}{\textbf{GCN}} & \multicolumn{2}{c@{\hspace{0.75cm}}}{\textbf{GAT}} & \multicolumn{2}{c@{\hspace{0.75cm}}}{\textbf{GATv2}} & \multicolumn{2}{c@{\hspace{0.75cm}}}{\textbf{GIN}} & \multicolumn{2}{c@{\hspace{0.25cm}}}{\textbf{PNA}} \\ \cmidrule(lr{0.5cm}){3-4} \cmidrule(r{0.5cm}){5-6} \cmidrule(r{0.5cm}){7-8} \cmidrule(r{0.5cm}){9-10} \cmidrule{11-12}
    Data.              & Agg. & \textbf{AUROC}             & \textbf{MCC}             & \textbf{AUROC}             & \textbf{MCC}             & \textbf{AUROC}              & \textbf{MCC}              & \textbf{AUROC}             & \textbf{MCC}             & \textbf{AUROC}             & \textbf{MCC}             \\ \cmidrule(r){1-1} \cmidrule(lr){2-2} \cmidrule(lr){3-3} \cmidrule(lr{0.5cm}){4-4} \cmidrule(r){5-5} \cmidrule(lr{0.5cm}){6-6} \cmidrule(r){7-7} \cmidrule(lr{0.5cm}){8-8} \cmidrule(r){9-9} \cmidrule(lr{0.5cm}){10-10} \cmidrule(r){11-11} \cmidrule(l){12-12}
    \multirow{6}{*}{PCBA} &
      Sum &
      0.53 $\pm$ 0.00 &
      0.20 $\pm$ 0.00 &
      0.54 $\pm$ 0.00 &
      0.22 $\pm$ 0.01 &
      0.54 $\pm$ 0.00 &
      0.24 $\pm$ 0.00 &
      0.55 $\pm$ 0.00 &
      0.25 $\pm$ 0.01 &
      0.56 $\pm$ 0.01 &
      0.27 $\pm$ 0.01 \\
     &
      Mean &
      0.53 $\pm$ 0.00 &
      0.19 $\pm$ 0.01 &
      0.53 $\pm$ 0.00 &
      0.20 $\pm$ 0.01 &
      0.54 $\pm$ 0.01 &
      0.22 $\pm$ 0.01 &
      0.55 $\pm$ 0.00 &
      0.24 $\pm$ 0.01 &
      0.56 $\pm$ 0.01 &
      0.27 $\pm$ 0.01 \\
     &
      Max &
      0.53 $\pm$ 0.00 &
      0.20 $\pm$ 0.01 &
      0.53 $\pm$ 0.00 &
      0.20 $\pm$ 0.01 &
      0.53 $\pm$ 0.00 &
      0.21 $\pm$ 0.01 &
      0.55 $\pm$ 0.00 &
      0.24 $\pm$ 0.01 &
      0.57 $\pm$ 0.01 &
      0.28 $\pm$ 0.01 \\
\bottomrule
    \end{tabular}
    \hspace*{-3cm}
\end{table}

\renewcommand{\familydefault}{\rmdefault}

\clearpage
\subsection{Social networks classification models}
\renewcommand{\sfdefault}{lmss}
\renewcommand{\familydefault}{\sfdefault}

\begin{table}[!h]
    \captionsetup{justification=centering, skip=4pt, labelfont=bf, width=1.5\linewidth}
    \centering
    \scriptsize
    \caption{Detailed metrics (mean $\pm$ standard deviation) for the social network classification datasets. \textquote{OOM} stands for out-of-memory (RAM).}
    \label{table:SOCIAL}
    \hspace*{-3cm}
    \begin{tabular}{@{}clcc@{\hskip 0.5cm}cc@{\hskip 0.5cm}cc@{\hskip 0.5cm}cc@{\hskip 0.5cm}cc@{}}
    \toprule
                         &            & \multicolumn{2}{c@{\hspace{0.5cm}}}{\textbf{GCN}} & \multicolumn{2}{c@{\hspace{0.75cm}}}{\textbf{GAT}} & \multicolumn{2}{c@{\hspace{0.75cm}}}{\textbf{GATv2}} & \multicolumn{2}{c@{\hspace{0.75cm}}}{\textbf{GIN}} & \multicolumn{2}{c@{\hspace{0.25cm}}}{\textbf{PNA}} \\ \cmidrule(lr{0.5cm}){3-4} \cmidrule(r{0.5cm}){5-6} \cmidrule(r{0.5cm}){7-8} \cmidrule(r{0.5cm}){9-10} \cmidrule{11-12}
    Data.              & Agg. & \textbf{AUROC}             & \textbf{MCC}             & \textbf{AUROC}             & \textbf{MCC}             & \textbf{AUROC}              & \textbf{MCC}              & \textbf{AUROC}             & \textbf{MCC}             & \textbf{AUROC}             & \textbf{MCC}             \\ \cmidrule(r){1-1} \cmidrule(lr){2-2} \cmidrule(lr){3-3} \cmidrule(lr{0.5cm}){4-4} \cmidrule(r){5-5} \cmidrule(lr{0.5cm}){6-6} \cmidrule(r){7-7} \cmidrule(lr{0.5cm}){8-8} \cmidrule(r){9-9} \cmidrule(lr{0.5cm}){10-10} \cmidrule(r){11-11} \cmidrule(l){12-12}
    \multirow{6}{*}[-0.35em]{\makecell{IMDB-\\BINARY}} & Sum        & 0.74 $\pm$ 0.03 & 0.49 $\pm$ 0.07 & 0.73 $\pm$ 0.02 & 0.47 $\pm$ 0.05 & 0.72 $\pm$ 0.03 & 0.44 $\pm$ 0.06 & 0.74 $\pm$ 0.03 & 0.49 $\pm$ 0.05 & 0.72 $\pm$ 0.03 & 0.45 $\pm$ 0.05  \\
                & Mean       & 0.73 $\pm$ 0.04 & 0.46 $\pm$ 0.08 & 0.72 $\pm$ 0.03 & 0.45 $\pm$ 0.05 & 0.71 $\pm$ 0.03 & 0.42 $\pm$ 0.07 & 0.72 $\pm$ 0.02 & 0.44 $\pm$ 0.03 & 0.72 $\pm$ 0.03 & 0.46 $\pm$ 0.08  \\
                & Max        & 0.72 $\pm$ 0.03 & 0.44 $\pm$ 0.06 & 0.72 $\pm$ 0.04 & 0.44 $\pm$ 0.09 & 0.72 $\pm$ 0.05 & 0.44 $\pm$ 0.11 & 0.74 $\pm$ 0.02 & 0.47 $\pm$ 0.03 & 0.71 $\pm$ 0.02 & 0.43 $\pm$ 0.03  \\ \cmidrule(l){2-12}
                & MLP      & 0.71 $\pm$ 0.03 & 0.42 $\pm$ 0.07 & 0.70 $\pm$ 0.02 & 0.40 $\pm$ 0.04 & 0.71 $\pm$ 0.05 & 0.41 $\pm$ 0.10 & 0.71 $\pm$ 0.03 & 0.42 $\pm$ 0.06 & 0.71 $\pm$ 0.05 & 0.42 $\pm$ 0.09  \\
\bottomrule
    \end{tabular}
    \hspace*{-3cm}
\end{table}

\begin{table}[!h]
    \captionsetup{justification=centering, skip=4pt, labelfont=bf, width=1.05\linewidth}
    \footnotesize
    \centering
    \caption{Detailed metrics (mean $\pm$ sd.) for the REDDIT-MULTI-12K dataset. \textquote{OOM} stands for out-of-memory (RAM). Only the MCC is reported as this is a multi-label task.}
    \label{table:REDDIT-12-MCC}
    \begin{tabular}{@{}clc@{\hskip 0.5cm}c@{\hskip 0.5cm}c@{\hskip 0.5cm}c@{\hskip 0.5cm}c@{}}
    \toprule
    Dataset              &  Aggregator           & \textbf{GCN} & \textbf{GAT} & \textbf{GATv2} & \textbf{GIN} & \textbf{PNA} \\ \midrule
    \multirow{6}{*}[-0.35em]{REDDIT-MULTI-12K} & Sum   & 0.33 $\pm$ 0.01 & 0.29 $\pm$ 0.03 & 0.29 $\pm$ 0.03 & 0.33 $\pm$ 0.02 & OOM  \\
                                  & Mean  & 0.28 $\pm$ 0.01 & 0.22 $\pm$ 0.01 & 0.23 $\pm$ 0.02 & 0.29 $\pm$ 0.01  & OOM  \\
                                  & Max   & 0.28 $\pm$ 0.02 & 0.11 $\pm$ 0.01 & 0.09 $\pm$ 0.02 & 0.27 $\pm$ 0.01  & OOM  \\ \cmidrule(l){2-7}
                                  & MLP & 0.24 $\pm$ 0.01 & 0.25 $\pm$ 0.02 & 0.24 $\pm$ 0.02 & 0.27 $\pm$ 0.03  & OOM  \\
                                  & GRU   & 0.18 $\pm$ 0.03 & 0.26 $\pm$ 0.07 & 0.25 $\pm$ 0.03 & 0.26 $\pm$ 0.02  & OOM  \\
                                  & ST    & 0.36 $\pm$ 0.01 & 0.31 $\pm$ 0.02 & 0.30 $\pm$ 0.01 & 0.31 $\pm$ 0.02  & OOM  \\ \bottomrule
    \end{tabular}
\end{table}

\renewcommand{\familydefault}{\rmdefault}

\clearpage
\subsection{Synthetic classification models}
\renewcommand{\sfdefault}{lmss}
\renewcommand{\familydefault}{\sfdefault}

\begin{table}[!h]
    \captionsetup{justification=centering, skip=4pt, labelfont=bf, width=1.3\linewidth}
    \scriptsize
    \centering
    \caption{Detailed metrics (mean $\pm$ standard deviation) for the SYNTHETIC and SYNTHETICnew datasets.}
    \label{table:SYNTHETIC-BINARY}
    \hspace*{-3cm}
    \begin{tabular}{@{}clcc@{\hskip 0.5cm}cc@{\hskip 0.5cm}cc@{\hskip 0.5cm}cc@{\hskip 0.5cm}cc@{}}
    \toprule
                         &            & \multicolumn{2}{c@{\hspace{0.5cm}}}{\textbf{GCN}} & \multicolumn{2}{c@{\hspace{0.75cm}}}{\textbf{GAT}} & \multicolumn{2}{c@{\hspace{0.75cm}}}{\textbf{GATv2}} & \multicolumn{2}{c@{\hspace{0.75cm}}}{\textbf{GIN}} & \multicolumn{2}{c@{\hspace{0.25cm}}}{\textbf{PNA}} \\ \cmidrule(lr{0.5cm}){3-4} \cmidrule(r{0.5cm}){5-6} \cmidrule(r{0.5cm}){7-8} \cmidrule(r{0.5cm}){9-10} \cmidrule{11-12}
    Dataset              & Agg. & \textbf{AUROC}             & \textbf{MCC}             & \textbf{AUROC}             & \textbf{MCC}             & \textbf{AUROC}              & \textbf{MCC}              & \textbf{AUROC}             & \textbf{MCC}             & \textbf{AUROC}             & \textbf{MCC}             \\ \cmidrule(r){1-1} \cmidrule(lr){2-2} \cmidrule(lr){3-3} \cmidrule(lr{0.5cm}){4-4} \cmidrule(r){5-5} \cmidrule(lr{0.5cm}){6-6} \cmidrule(r){7-7} \cmidrule(lr{0.5cm}){8-8} \cmidrule(r){9-9} \cmidrule(lr{0.5cm}){10-10} \cmidrule(r){11-11} \cmidrule(l){12-12}
    \multirow{6}{*}[-0.35em]{\makecell{SYNTHE-\\TIC}} & Sum   & 0.87 $\pm$ 0.18 & 0.74 $\pm$ 0.36 & 0.99 $\pm$ 0.02 & 0.97 $\pm$ 0.04 & 0.51 $\pm$ 0.01 & 0.03 $\pm$ 0.07 & 0.94 $\pm$ 0.03 & 0.89 $\pm$ 0.06 & 1.00 $\pm$ 0.00 & 1.00 $\pm$ 0.00  \\
                               & Mean  & 0.50 $\pm$ 0.00 & 0.00 $\pm$ 0.00 & 0.68 $\pm$ 0.23 & 0.36 $\pm$ 0.46 & 0.50 $\pm$ 0.00 & 0.00 $\pm$ 0.00 & 0.91 $\pm$ 0.13 & 0.86 $\pm$ 0.22 & 0.60 $\pm$ 0.22 & 0.20 $\pm$ 0.45  \\
                               & Max   & 0.87 $\pm$ 0.21 & 0.75 $\pm$ 0.42 & 0.78 $\pm$ 0.25 & 0.59 $\pm$ 0.47 & 0.50 $\pm$ 0.00 & 0.00 $\pm$ 0.00 & 0.95 $\pm$ 0.04 & 0.91 $\pm$ 0.08 & 0.99 $\pm$ 0.01 & 0.99 $\pm$ 0.03  \\
\end{tabular}
    \hspace*{-3cm}
\end{table}

\begin{table}[!h]
    \captionsetup{justification=centering, skip=4pt, labelfont=bf, width=1\linewidth}
    \footnotesize
    \centering
    \caption{Detailed metrics (mean $\pm$ standard deviation) for the Synthie, TRIANGLES, and COLORS-3 datasets. Only the MCC is reported as this is a multi-label task.}
    \label{table:SYNTHETIC-MULTI}
    \begin{tabular}{@{}clc@{\hskip 0.5cm}c@{\hskip 0.5cm}c@{\hskip 0.5cm}c@{\hskip 0.5cm}c@{}}
    \toprule
    Dataset              &  Aggregator           & \textbf{GCN} & \textbf{GAT} & \textbf{GATv2} & \textbf{GIN} & \textbf{PNA} \\ \midrule
    \multirow{6}{*}[-0.35em]{Synthie} & Sum   & 0.54 $\pm$ 0.06 & 0.52 $\pm$ 0.12 & 0.46 $\pm$ 0.09 & 0.53 $\pm$ 0.11 & 0.95 $\pm$ 0.03  \\
                         & Mean  & 0.45 $\pm$ 0.11 & 0.36 $\pm$ 0.21 & 0.52 $\pm$ 0.11 & 0.58 $\pm$ 0.07 & 0.93 $\pm$ 0.05  \\
                         & Max   & 0.14 $\pm$ 0.04 & 0.24 $\pm$ 0.07 & 0.23 $\pm$ 0.07 & 0.30 $\pm$ 0.14 & 0.59 $\pm$ 0.04  \\ \cmidrule(l){2-7}
                         & MLP & 0.71 $\pm$ 0.08 & 0.72 $\pm$ 0.13 & 0.75 $\pm$ 0.05 & 0.59 $\pm$ 0.08 & 0.93 $\pm$ 0.03  \\
                         & GRU   & 0.75 $\pm$ 0.12 & 0.48 $\pm$ 0.21 & 0.79 $\pm$ 0.11 & 0.72 $\pm$ 0.03 & 0.79 $\pm$ 0.08  \\
                         & ST    & 0.37 $\pm$ 0.23 & 0.61 $\pm$ 0.08 & 0.71 $\pm$ 0.11 & 0.60 $\pm$ 0.09 & 0.88 $\pm$ 0.12 \\ \cmidrule[0.75pt]{1-7}
    \multirow{6}{*}[-0.35em]{TRIANGLES} & Sum   & 0.11 $\pm$ 0.01 & 0.18 $\pm$ 0.01 & 0.17 $\pm$ 0.01 & 0.13 $\pm$ 0.01 & 0.10 $\pm$ 0.04  \\
\bottomrule
    \end{tabular}
\end{table}

\renewcommand{\familydefault}{\rmdefault}

\clearpage
\subsection{Bioinformatics classification models}
\renewcommand{\sfdefault}{lmss}
\renewcommand{\familydefault}{\sfdefault}

\begin{table}[!h]
    \captionsetup{justification=centering, skip=4pt, labelfont=bf, width=1.2\linewidth}
    \centering
    \scriptsize
    \caption{Detailed metrics (mean $\pm$ standard deviation) for the PROTEINS\_full dataset.}
    \label{table:PROTEINS_full}
    \hspace*{-3cm}
    \begin{tabular}{@{}clcc@{\hskip 0.5cm}cc@{\hskip 0.5cm}cc@{\hskip 0.5cm}cc@{\hskip 0.5cm}cc@{}}
        \toprule
                             &            & \multicolumn{2}{c@{\hspace{0.5cm}}}{\textbf{GCN}} & \multicolumn{2}{c@{\hspace{0.75cm}}}{\textbf{GAT}} & \multicolumn{2}{c@{\hspace{0.75cm}}}{\textbf{GATv2}} & \multicolumn{2}{c@{\hspace{0.75cm}}}{\textbf{GIN}} & \multicolumn{2}{c@{\hspace{0.25cm}}}{\textbf{PNA}} \\ \cmidrule(lr{0.5cm}){3-4} \cmidrule(r{0.5cm}){5-6} \cmidrule(r{0.5cm}){7-8} \cmidrule(r{0.5cm}){9-10} \cmidrule{11-12}
        Dataset              & Agg. & \textbf{AUROC}             & \textbf{MCC}             & \textbf{AUROC}             & \textbf{MCC}             & \textbf{AUROC}              & \textbf{MCC}              & \textbf{AUROC}             & \textbf{MCC}             & \textbf{AUROC}             & \textbf{MCC}             \\ \cmidrule(r){1-1} \cmidrule(lr){2-2} \cmidrule(lr){3-3} \cmidrule(lr{0.5cm}){4-4} \cmidrule(r){5-5} \cmidrule(lr{0.5cm}){6-6} \cmidrule(r){7-7} \cmidrule(lr{0.5cm}){8-8} \cmidrule(r){9-9} \cmidrule(lr{0.5cm}){10-10} \cmidrule(r){11-11} \cmidrule(l){12-12}
    \multirow{6}{*}[-0.35em]{\makecell{PROTEINS\_\\full}} & Sum & 0.67 $\pm$ 0.06 & 0.38 $\pm$ 0.13 & 0.52 $\pm$ 0.03 & 0.05 $\pm$ 0.07 & 0.57 $\pm$ 0.09 & 0.14 $\pm$ 0.17 & 0.55 $\pm$ 0.04 & 0.18 $\pm$ 0.12 & 0.64 $\pm$ 0.04 & 0.33 $\pm$ 0.07 \\
     & Mean & 0.69 $\pm$ 0.06 & 0.41 $\pm$ 0.11 & 0.67 $\pm$ 0.06 & 0.35 $\pm$ 0.10 & 0.69 $\pm$ 0.07 & 0.39 $\pm$ 0.15 & 0.52 $\pm$ 0.02 & 0.09 $\pm$ 0.11 & 0.71 $\pm$ 0.06 & 0.47 $\pm$ 0.09 \\
     & Max & 0.69 $\pm$ 0.04 & 0.41 $\pm$ 0.08 & 0.51 $\pm$ 0.02 & 0.06 $\pm$ 0.06 & 0.53 $\pm$ 0.03 & 0.13 $\pm$ 0.10 & 0.57 $\pm$ 0.06 & 0.21 $\pm$ 0.16 & 0.72 $\pm$ 0.02 & 0.44 $\pm$ 0.05 \\ \cmidrule(l){2-12}
     & MLP & 0.62 $\pm$ 0.11 & 0.26 $\pm$ 0.24 & 0.58 $\pm$ 0.12 & 0.17 $\pm$ 0.23 & 0.59 $\pm$ 0.11 & 0.21 $\pm$ 0.22 & 0.66 $\pm$ 0.05 & 0.33 $\pm$ 0.09 & 0.67 $\pm$ 0.06 & 0.35 $\pm$ 0.13 \\
\bottomrule
    \end{tabular}
    \hspace*{-3cm}
\end{table}

\begin{table}[!h]
    \captionsetup{justification=centering, skip=4pt, labelfont=bf, width=1.05\linewidth}
    \footnotesize
    \centering
    \caption{Detailed metrics (mean $\pm$ standard deviation) for the ENZYMES dataset. Only the MCC is reported as this is a multi-label task.}
    \label{table:ENZYMES-MCC}
    \begin{tabular}{@{}clc@{\hskip 0.5cm}c@{\hskip 0.5cm}c@{\hskip 0.5cm}c@{\hskip 0.5cm}c@{}}
    \toprule
    Dataset              &  Aggregator           & \textbf{GCN} & \textbf{GAT} & \textbf{GATv2} & \textbf{GIN} & \textbf{PNA} \\ \midrule
    \multirow{6}{*}[-0.35em]{ENZYMES} & Sum   & 0.16 $\pm$ 0.06 & 0.21 $\pm$ 0.14 & 0.27 $\pm$ 0.05 & 0.50 $\pm$ 0.07 & 0.42 $\pm$ 0.08  \\
    & Mean  & 0.16 $\pm$ 0.09 & 0.24 $\pm$ 0.10 & 0.32 $\pm$ 0.15 & 0.44 $\pm$ 0.06 & 0.39 $\pm$ 0.11  \\
    & Max   & 0.20 $\pm$ 0.10 & 0.28 $\pm$ 0.11 & 0.25 $\pm$ 0.14 & 0.41 $\pm$ 0.05 & 0.40 $\pm$ 0.11  \\ \cmidrule(l){2-7}
    & MLP & 0.47 $\pm$ 0.08 & 0.53 $\pm$ 0.04 & 0.52 $\pm$ 0.07 & 0.48 $\pm$ 0.03 & 0.51 $\pm$ 0.07  \\
    & GRU   & 0.24 $\pm$ 0.07 & 0.32 $\pm$ 0.10 & 0.27 $\pm$ 0.13 & 0.38 $\pm$ 0.09 & 0.17 $\pm$ 0.05  \\
    & ST    & 0.23 $\pm$ 0.12 & 0.25 $\pm$ 0.13 & 0.28 $\pm$ 0.07 & 0.37 $\pm$ 0.04 & 0.29 $\pm$ 0.10 \\ \bottomrule
    \end{tabular}
\end{table}
\renewcommand{\familydefault}{\rmdefault}

\subsection{TUDataset computer vision models}
\renewcommand{\sfdefault}{lmss}
\renewcommand{\familydefault}{\sfdefault}
\begin{table}[!h]
    \captionsetup{justification=centering, skip=4pt, labelfont=bf, width=1.03\linewidth}
    \footnotesize
    \centering
    \caption{Detailed metrics (mean $\pm$ standard deviation) for the TUDataset computer vision datasets. Only the MCC is reported as this is a multi-label task.}
    \label{table:CV-TUD-MCC}
    \begin{tabular}{@{}clc@{\hskip 0.5cm}c@{\hskip 0.5cm}c@{\hskip 0.5cm}c@{\hskip 0.5cm}c@{}}
    \toprule
    Dataset              &  Aggregator           & \textbf{GCN} & \textbf{GAT} & \textbf{GATv2} & \textbf{GIN} & \textbf{PNA} \\ \midrule
    \multirow{6}{*}[-0.35em]{Cuneiform} & Sum                             & 0.51 $\pm$ 0.29 & 0.56 $\pm$ 0.25 & 0.66 $\pm$ 0.09  & 0.73 $\pm$ 0.07 & 0.58 $\pm$ 0.34  \\
    & Mean                            & 0.00 $\pm$ 0.02 & 0.15 $\pm$ 0.25 & -0.05 $\pm$ 0.04 & 0.80 $\pm$ 0.09 & 0.03 $\pm$ 0.06  \\
    & Max                             & 0.26 $\pm$ 0.35 & 0.36 $\pm$ 0.33 & 0.26 $\pm$ 0.35  & 0.77 $\pm$ 0.12 & 0.55 $\pm$ 0.30  \\ \cmidrule(l){2-7}
    & MLP                           & 0.50 $\pm$ 0.13 & 0.58 $\pm$ 0.17 & 0.62 $\pm$ 0.11  & 0.64 $\pm$ 0.09 & 0.64 $\pm$ 0.11  \\
    & GRU                             & 0.18 $\pm$ 0.08 & 0.05 $\pm$ 0.09 & 0.11 $\pm$ 0.15  & 0.51 $\pm$ 0.12 & 0.16 $\pm$ 0.14  \\
    & ST                              & 0.32 $\pm$ 0.19 & 0.21 $\pm$ 0.32 & 0.40 $\pm$ 0.25  & 0.47 $\pm$ 0.15 & 0.40 $\pm$ 0.31 \\ \cmidrule[0.75pt]{1-7}

    \multirow{6}{*}[-0.35em]{COIL-RAG}   & Sum                            & 0.91 $\pm$ 0.01 & 0.94 $\pm$ 0.02 & 0.94 $\pm$ 0.02 & 0.95 $\pm$ 0.02 & 0.90 $\pm$ 0.02  \\
\bottomrule
    \end{tabular}
\end{table}

\renewcommand{\familydefault}{\rmdefault}

\clearpage
\subsection{ZINC regression models}
\renewcommand{\sfdefault}{lmss}
\renewcommand{\familydefault}{\sfdefault}

\begin{table}[!h]
    \captionsetup{justification=centering, skip=4pt, labelfont=bf, width=1\linewidth}
    \centering
    \small
    \caption{Detailed metrics for the ZINC dataset. The results are reported only for the provided train/validation/test splits (no custom random splits).}
    \label{table:ZINC}
    \begin{tabular}{@{}clcc@{\hskip 0.5cm}cc@{\hskip 0.5cm}cc@{\hskip 0.5cm}cc@{\hskip 0.5cm}cc@{}}
        \toprule
                             &            & \multicolumn{2}{c@{\hspace{0.5cm}}}{\textbf{GCN}} & \multicolumn{2}{c@{\hspace{0.75cm}}}{\textbf{GAT}} & \multicolumn{2}{c@{\hspace{0.75cm}}}{\textbf{GATv2}} & \multicolumn{2}{c@{\hspace{0.75cm}}}{\textbf{GIN}} & \multicolumn{2}{c@{\hspace{0.25cm}}}{\textbf{PNA}} \\ \cmidrule(lr{0.5cm}){3-4} \cmidrule(r{0.5cm}){5-6} \cmidrule(r{0.5cm}){7-8} \cmidrule(r{0.5cm}){9-10} \cmidrule{11-12}
        Dataset              & Agg. & \textbf{MAE}             & \textbf{\textbf{R}$\bm{^\mathsf{2}}$}            & \textbf{MAE}             & \textbf{\textbf{R}$\bm{^\mathsf{2}}$}             & \textbf{MAE}              & \textbf{\textbf{R}$\bm{^\mathsf{2}}$}              & \textbf{MAE}             & \textbf{\textbf{R}$\bm{^\mathsf{2}}$}             & \textbf{MAE}             & \textbf{\textbf{R}$\bm{^\mathsf{2}}$}            \\ \cmidrule(r){1-1} \cmidrule(lr){2-2} \cmidrule(lr){3-3} \cmidrule(lr{0.5cm}){4-4} \cmidrule(r){5-5} \cmidrule(lr{0.5cm}){6-6} \cmidrule(r){7-7} \cmidrule(lr{0.5cm}){8-8} \cmidrule(r){9-9} \cmidrule(lr{0.5cm}){10-10} \cmidrule(r){11-11} \cmidrule(l){12-12}
        \multirow{6}{*}[-0.35em]{ZINC}
        & Sum  & 0.80  & 0.59  & 0.86  & 0.59  & 1.06  & 0.47  & 0.51  & 0.80  & 0.36  & 0.87  \\
        & Mean  & 0.74  & 0.63  & 0.90  & 0.54  & 1.26  & 0.43  & 0.57  & 0.77  & 0.39  & 0.86  \\
        & Max  & 0.88  & 0.57  & 1.25  & 0.50  & 1.34  & 0.39  & 0.64  & 0.72  & 0.47  & 0.86 \\ \cmidrule(l){2-12}
        & MLP  & 0.68  & 0.69  & 0.79  & 0.60  & 1.49  & 0.42  & 0.50  & 0.79  & 0.36  & 0.88  \\
        & GRU  & 0.92  & 0.51  & 1.46  & 0.06  & 0.91  & 0.67  & 0.56  & 0.79  & 0.33  & 0.89 \\
        & ST  & 0.72  & 0.64  & 1.04  & 0.47  & 2.39  & 0.42  & 0.53  & 0.79  & 0.43  & 0.85 \\ \bottomrule
    \end{tabular}
\end{table}

\renewcommand{\familydefault}{\rmdefault}

\subsection{QM9 regression models with Janossy neural aggregation}
\renewcommand{\sfdefault}{lmss}
\renewcommand{\familydefault}{\sfdefault}
\begin{table}[!h]
    \captionsetup{justification=centering, skip=4pt, labelfont=bf, width=1.4\linewidth}
    \scriptsize
    \caption{Detailed metrics (mean $\pm$ standard deviation) for the QM9 dataset, including the two Janossy variants. Any differences in performance compared to other 2-layer models such as those in \Cref{figure:qm9-7-layers} might be due to different GNN hyperparameters, such as the output or intermediate node dimension (QM9-specific experiments generally used larger dimensions).}
    \label{table:QM9-Janossy}
    \hspace*{-3cm}
    \begin{tabular}{@{}clcc@{\hskip 0.5cm}cc@{\hskip 0.5cm}cc@{\hskip 0.5cm}cc@{\hskip 0.5cm}cc@{}}
    \toprule
                         &            & \multicolumn{2}{c@{\hspace{0.5cm}}}{\textbf{GCN}} & \multicolumn{2}{c@{\hspace{0.75cm}}}{\textbf{GAT}} & \multicolumn{2}{c@{\hspace{0.75cm}}}{\textbf{GATv2}} & \multicolumn{2}{c@{\hspace{0.75cm}}}{\textbf{GIN}} & \multicolumn{2}{c@{\hspace{0.25cm}}}{\textbf{PNA}} \\ \cmidrule(lr{0.5cm}){3-4} \cmidrule(r{0.5cm}){5-6} \cmidrule(r{0.5cm}){7-8} \cmidrule(r{0.5cm}){9-10} \cmidrule{11-12}
    Data.              & Agg. & \textbf{MAE}             & \textbf{\textbf{R}$\bm{^\mathsf{2}}$}             & \textbf{MAE}             & \textbf{\textbf{R}$\bm{^\mathsf{2}}$}             & \textbf{MAE}              & \textbf{\textbf{R}$\bm{^\mathsf{2}}$}              & \textbf{MAE}             & \textbf{\textbf{R}$\bm{^\mathsf{2}}$}             & \textbf{MAE}             & \textbf{\textbf{R}$\bm{^\mathsf{2}}$}             \\ \cmidrule(lr){2-2} \cmidrule(lr){3-3} \cmidrule(lr{0.5cm}){4-4} \cmidrule(r){5-5} \cmidrule(lr{0.5cm}){6-6} \cmidrule(r){7-7} \cmidrule(lr{0.5cm}){8-8} \cmidrule(r){9-9} \cmidrule(lr{0.5cm}){10-10} \cmidrule(r){11-11} \cmidrule(l){12-12}

    \multirow{8}{*}[-0.35em]{QM9} &
      Sum &
      0.74 $\pm$ 0.00 &
      0.09 $\pm$ 0.00 &
      0.76 $\pm$ 0.00 &
      0.05 $\pm$ 0.01 &
      0.75 $\pm$ 0.00 &
      0.08 $\pm$ 0.01 &
      0.71 $\pm$ 0.00 &
      0.15 $\pm$ 0.00 &
      0.70 $\pm$ 0.00 &
      0.17 $\pm$ 0.01 \\
     &
      Mean &
      0.73 $\pm$ 0.00 &
      0.10 $\pm$ 0.00 &
      0.75 $\pm$ 0.01 &
      0.07 $\pm$ 0.01 &
      0.74 $\pm$ 0.00 &
      0.10 $\pm$ 0.00 &
      0.72 $\pm$ 0.00 &
      0.13 $\pm$ 0.00 &
      0.70 $\pm$ 0.00 &
      0.16 $\pm$ 0.01 \\
     &
      Max &
      0.73 $\pm$ 0.00 &
      0.11 $\pm$ 0.00 &
      0.74 $\pm$ 0.00 &
      0.10 $\pm$ 0.00 &
      0.73 $\pm$ 0.00 &
      0.11 $\pm$ 0.00 &
      0.70 $\pm$ 0.00 &
      0.16 $\pm$ 0.00 &
      0.68 $\pm$ 0.00 &
      0.20 $\pm$ 0.00 \\ \cmidrule(l){2-12}
     &
      MLP &
      0.63 $\pm$ 0.00 &
      0.31 $\pm$ 0.00 &
      0.64 $\pm$ 0.00 &
      0.30 $\pm$ 0.01 &
      0.64 $\pm$ 0.01 &
      0.29 $\pm$ 0.01 &
      0.60 $\pm$ 0.00 &
      0.38 $\pm$ 0.00 &
      0.58 $\pm$ 0.00 &
      0.41 $\pm$ 0.00 \\
     &
      GRU &
      0.62 $\pm$ 0.01 &
      0.34 $\pm$ 0.02 &
      0.62 $\pm$ 0.01 &
      0.34 $\pm$ 0.01 &
      0.61 $\pm$ 0.02 &
      0.35 $\pm$ 0.03 &
      0.60 $\pm$ 0.00 &
      0.38 $\pm$ 0.00 &
      0.60 $\pm$ 0.01 &
      0.37 $\pm$ 0.02 \\
     &
      ST &
      0.60 $\pm$ 0.01 &
      0.38 $\pm$ 0.01 &
      0.63 $\pm$ 0.02 &
      0.30 $\pm$ 0.04 &
      0.62 $\pm$ 0.01 &
      0.32 $\pm$ 0.02 &
      0.59 $\pm$ 0.00 &
      0.39 $\pm$ 0.01 &
      0.57 $\pm$ 0.01 &
      0.44 $\pm$ 0.01 \\ \cmidrule(l){2-12}
     &
      Janossy MLP &
      0.67 $\pm$ 0.02 &
      0.23 $\pm$ 0.04 &
      0.68 $\pm$ 0.01 &
      0.22 $\pm$ 0.03 &
      0.67 $\pm$ 0.01 &
      0.23 $\pm$ 0.02 &
      0.61 $\pm$ 0.01 &
      0.34 $\pm$ 0.02 &
      0.61 $\pm$ 0.01 &
      0.34 $\pm$ 0.02 \\
     &
      Janossy GRU &
      0.67 $\pm$ 0.02 &
      0.22 $\pm$ 0.04 &
      0.67 $\pm$ 0.01 &
      0.23 $\pm$ 0.02 &
      0.71 $\pm$ 0.02 &
      0.15 $\pm$ 0.04 &
      0.66 $\pm$ 0.03 &
      0.25 $\pm$ 0.06 &
      0.63 $\pm$ 0.01 &
      0.32 $\pm$ 0.02 \\ \bottomrule

    \end{tabular}
    \hspace*{-3cm}
\end{table}
\renewcommand{\familydefault}{\rmdefault}

\subsection{MalNetTiny models}
\label{sec:MalNetTiny-MCC}

\renewcommand{\sfdefault}{lmss}
\renewcommand{\familydefault}{\sfdefault}
\begin{table}[!h]
    \captionsetup{justification=centering, skip=4pt, labelfont=bf, width=1.05\linewidth}
    \footnotesize
    \centering
    \caption{Detailed metrics (mean $\pm$ standard deviation) for the MalNetTiny dataset. \textquote{OOM} stands for out-of-memory (RAM). Only the MCC is reported as this is a multi-label task.}
    \label{table:MalNetTiny-MCC}
    \begin{tabular}{@{}clc@{\hskip 0.5cm}c@{\hskip 0.5cm}c@{\hskip 0.5cm}c@{\hskip 0.5cm}c@{}}
    \toprule
    Dataset              &  Aggregator           & \textbf{GCN} & \textbf{GAT} & \textbf{GATv2} & \textbf{GIN} & \textbf{PNA} \\ \midrule
    \multirow{6}{*}[-0.35em]{MalNetTiny} & Sum & 0.81 $\pm$ 0.04 & 0.81 $\pm$ 0.05 & 0.80 $\pm$ 0.02 & 0.87 $\pm$ 0.03 & OOM  \\
                                  & Mean & 0.72 $\pm$ 0.02 & 0.73 $\pm$ 0.02 & 0.75 $\pm$ 0.05 & 0.88 $\pm$ 0.01  & OOM  \\
                                  & Max & 0.81 $\pm$ 0.02 & 0.77 $\pm$ 0.04 & 0.76 $\pm$ 0.04 & 0.89 $\pm$ 0.02  & OOM  \\ \cmidrule(l){2-7}
                                  & MLP & 0.80 $\pm$ 0.02 & 0.81 $\pm$ 0.01 & 0.80 $\pm$ 0.01 & 0.82 $\pm$ 0.02  & OOM  \\
                                  & GRU & 0.68 $\pm$ 0.03 & 0.69 $\pm$ 0.04 & 0.70 $\pm$ 0.02 & 0.70 $\pm$ 0.02  & OOM  \\
                                  & ST & 0.84 $\pm$ 0.03 & 0.82 $\pm$ 0.03 & 0.84 $\pm$ 0.02 & 0.88 $\pm$ 0.03  & OOM  \\ \bottomrule
    \end{tabular}
\end{table}
\renewcommand{\familydefault}{\rmdefault}

\clearpage
\subsection{GNNBenchmark computer vision models}
\label{sec:MNIST-CIFAR10-MCC}

\renewcommand{\sfdefault}{lmss}
\renewcommand{\familydefault}{\sfdefault}

\begin{table}[!h]
    \captionsetup{justification=centering, skip=4pt, labelfont=bf, width=1.1\linewidth}
    \footnotesize
    \centering
    \caption{Detailed metrics for the MNIST and CIFAR10 datasets. All models use the provided train/validation/test splits (no custom splits), which are used in five different runs and aggregated (mean $\pm$ standard deviation). Only the MCC is reported as this is a multi-label task.}
    \label{table:MNIST-CIFAR10-MCC}
    \begin{tabular}{@{}clc@{\hskip 0.5cm}c@{\hskip 0.5cm}c@{\hskip 0.5cm}c@{\hskip 0.5cm}c@{}}
    \toprule
    Dataset & Aggregator & \textbf{GCN} & \textbf{GAT} & \textbf{GATv2} & \textbf{GIN} & \textbf{PNA} \\ \midrule
    \multirow{6}{*}[-0.35em]{MNIST} & Sum & 0.42 $\pm$ 0.01 & 0.35 $\pm$ 0.04 & 0.52 $\pm$ 0.03 & 0.51 $\pm$ 0.01 & 0.74 $\pm$ 0.00 \\
    & Mean & 0.39 $\pm$ 0.01 & 0.29 $\pm$ 0.01 & 0.33 $\pm$ 0.07 & 0.46 $\pm$ 0.01 & 0.72 $\pm$ 0.01 \\
    & Max & 0.25 $\pm$ 0.03 & 0.40 $\pm$ 0.05 & 0.43 $\pm$ 0.17 & 0.43 $\pm$ 0.01 & 0.71 $\pm$ 0.01 \\ \cmidrule(l){2-7}
    & MLP & 0.28 $\pm$ 0.02 & 0.28 $\pm$ 0.01 & 0.26 $\pm$ 0.04 & 0.31 $\pm$ 0.01 & 0.47 $\pm$ 0.03 \\
    & GRU & 0.36 $\pm$ 0.03 & 0.25 $\pm$ 0.04 & 0.37 $\pm$ 0.09 & 0.44 $\pm$ 0.02 & 0.66 $\pm$ 0.02 \\
    & ST & 0.48 $\pm$ 0.01 & 0.60 $\pm$ 0.03 & 0.64 $\pm$ 0.02 & 0.56 $\pm$ 0.01 & 0.77 $\pm$ 0.00 \\ \cmidrule[0.75pt]{1-7}
    \multirow{6}{*}[-0.35em]{CIFAR10} & Sum & 0.28 $\pm$ 0.01 & 0.32 $\pm$ 0.01 & 0.34 $\pm$ 0.03 & 0.31 $\pm$ 0.01 & 0.50 $\pm$ 0.00 \\
    & Mean & 0.27 $\pm$ 0.00 & 0.29 $\pm$ 0.01 & 0.30 $\pm$ 0.01 & 0.30 $\pm$ 0.01 & 0.51 $\pm$ 0.00 \\
    & Max & 0.29 $\pm$ 0.00 & 0.36 $\pm$ 0.00 & 0.35 $\pm$ 0.01 & 0.31 $\pm$ 0.00 & 0.46 $\pm$ 0.01 \\ \cmidrule(l){2-7}
    & MLP & 0.17 $\pm$ 0.00 & 0.25 $\pm$ 0.00 & 0.19 $\pm$ 0.02 & 0.17 $\pm$ 0.00 & 0.29 $\pm$ 0.01 \\
    & GRU & 0.24 $\pm$ 0.01 & 0.31 $\pm$ 0.02 & 0.29 $\pm$ 0.07 & 0.26 $\pm$ 0.01 & 0.45 $\pm$ 0.01 \\
    & ST & 0.32 $\pm$ 0.01 & 0.42 $\pm$ 0.01 & 0.39 $\pm$ 0.01 & 0.35 $\pm$ 0.00 & 0.48 $\pm$ 0.00 \\ \bottomrule
    \end{tabular}
\end{table}
\renewcommand{\familydefault}{\rmdefault}

\clearpage
\subsection{TUDataset small molecules models}
\label{sec:TUD-small-mols}

\renewcommand{\sfdefault}{lmss}
\renewcommand{\familydefault}{\sfdefault}

\begin{table}[!htp]
    \captionsetup{justification=centering, skip=4pt, labelfont=bf, width=1.2\linewidth}
    \centering
    \scriptsize
    \caption{Detailed metrics (mean $\pm$ standard deviation) for the AIDS, FRANKENSTEIN, MUTAG, and Mutagenicity datasets.}
    \label{table:tud-extra}
    \hspace*{-3cm}
    \begin{tabular}{@{}clcc@{\hskip 0.5cm}cc@{\hskip 0.5cm}cc@{\hskip 0.5cm}cc@{\hskip 0.5cm}cc@{}}
        \toprule
        &            & \multicolumn{2}{c@{\hspace{0.5cm}}}{\textbf{GCN}} & \multicolumn{2}{c@{\hspace{0.75cm}}}{\textbf{GAT}} & \multicolumn{2}{c@{\hspace{0.75cm}}}{\textbf{GATv2}} & \multicolumn{2}{c@{\hspace{0.75cm}}}{\textbf{GIN}} & \multicolumn{2}{c@{\hspace{0.25cm}}}{\textbf{PNA}} \\ \cmidrule(lr{0.5cm}){3-4} \cmidrule(r{0.5cm}){5-6} \cmidrule(r{0.5cm}){7-8} \cmidrule(r{0.5cm}){9-10} \cmidrule{11-12}
        Dataset              & Agg. & \textbf{AUROC}             & \textbf{MCC}             & \textbf{AUROC}             & \textbf{MCC}             & \textbf{AUROC}              & \textbf{MCC}              & \textbf{AUROC}             & \textbf{MCC}             & \textbf{AUROC}             & \textbf{MCC}             \\ \cmidrule(r){1-1} \cmidrule(lr){2-2} \cmidrule(lr){3-3} \cmidrule(lr{0.5cm}){4-4} \cmidrule(r){5-5} \cmidrule(lr{0.5cm}){6-6} \cmidrule(r){7-7} \cmidrule(lr{0.5cm}){8-8} \cmidrule(r){9-9} \cmidrule(lr{0.5cm}){10-10} \cmidrule(r){11-11} \cmidrule(l){12-12}
    \multirow{6}{*}[-0.35em]{\makecell{AIDS}} & Sum & 0.97 $\pm$ 0.02 & 0.96 $\pm$ 0.03 & 0.97 $\pm$ 0.03 & 0.95 $\pm$ 0.04 & 0.97 $\pm$ 0.02 & 0.95 $\pm$ 0.02 & 0.96 $\pm$ 0.02 & 0.95 $\pm$ 0.02 & 0.97 $\pm$ 0.02 & 0.96 $\pm$ 0.02 \\
    & Mean & 0.97 $\pm$ 0.02 & 0.95 $\pm$ 0.02 & 0.98 $\pm$ 0.02 & 0.96 $\pm$ 0.03 & 0.97 $\pm$ 0.02 & 0.96 $\pm$ 0.03 & 0.96 $\pm$ 0.02 & 0.95 $\pm$ 0.02 & 0.97 $\pm$ 0.02 & 0.96 $\pm$ 0.02 \\
    & Max & 0.97 $\pm$ 0.02 & 0.96 $\pm$ 0.02 & 0.97 $\pm$ 0.02 & 0.96 $\pm$ 0.02 & 0.97 $\pm$ 0.02 & 0.94 $\pm$ 0.03 & 0.97 $\pm$ 0.01 & 0.95 $\pm$ 0.01 & 0.97 $\pm$ 0.02 & 0.96 $\pm$ 0.03 \\ \cmidrule(l){2-12}
    & MLP & 1.00 $\pm$ 0.00 & 1.00 $\pm$ 0.01 & 1.00 $\pm$ 0.00 & 1.00 $\pm$ 0.01 & 1.00 $\pm$ 0.00 & 1.00 $\pm$ 0.01 & 1.00 $\pm$ 0.00 & 1.00 $\pm$ 0.01 & 1.00 $\pm$ 0.00 & 1.00 $\pm$ 0.01 \\
    & GRU & 1.00 $\pm$ 0.00 & 1.00 $\pm$ 0.01 & 1.00 $\pm$ 0.00 & 1.00 $\pm$ 0.01 & 1.00 $\pm$ 0.00 & 1.00 $\pm$ 0.01 & 1.00 $\pm$ 0.01 & 0.99 $\pm$ 0.01 & 1.00 $\pm$ 0.00 & 0.99 $\pm$ 0.01 \\
    & ST & 0.97 $\pm$ 0.02 & 0.96 $\pm$ 0.03 & 0.98 $\pm$ 0.02 & 0.96 $\pm$ 0.03 & 0.97 $\pm$ 0.02 & 0.95 $\pm$ 0.03 & 0.96 $\pm$ 0.02 & 0.89 $\pm$ 0.09 & 0.98 $\pm$ 0.02 & 0.97 $\pm$ 0.03 \\ \cmidrule[0.75pt]{1-12}
    \multirow{6}{*}[-0.35em]{\makecell{FRANKEN\\STEIN}} & Sum & 0.62 $\pm$ 0.01 & 0.25 $\pm$ 0.02 & 0.60 $\pm$ 0.00 & 0.21 $\pm$ 0.02 & 0.59 $\pm$ 0.02 & 0.19 $\pm$ 0.03 & 0.60 $\pm$ 0.03 & 0.20 $\pm$ 0.06 & 0.65 $\pm$ 0.03 & 0.29 $\pm$ 0.06 \\
    & Mean & 0.60 $\pm$ 0.03 & 0.21 $\pm$ 0.06 & 0.60 $\pm$ 0.02 & 0.20 $\pm$ 0.04 & 0.59 $\pm$ 0.03 & 0.19 $\pm$ 0.05 & 0.61 $\pm$ 0.04 & 0.22 $\pm$ 0.08 & 0.62 $\pm$ 0.03 & 0.25 $\pm$ 0.04 \\
    & Max & 0.60 $\pm$ 0.02 & 0.20 $\pm$ 0.03 & 0.60 $\pm$ 0.02 & 0.20 $\pm$ 0.04 & 0.61 $\pm$ 0.03 & 0.21 $\pm$ 0.06 & 0.62 $\pm$ 0.02 & 0.23 $\pm$ 0.05 & 0.63 $\pm$ 0.03 & 0.25 $\pm$ 0.05 \\ \cmidrule(l){2-12}
    & MLP & 0.61 $\pm$ 0.02 & 0.23 $\pm$ 0.03 & 0.63 $\pm$ 0.02 & 0.27 $\pm$ 0.05 & 0.65 $\pm$ 0.02 & 0.29 $\pm$ 0.05 & 0.65 $\pm$ 0.02 & 0.30 $\pm$ 0.05 & 0.68 $\pm$ 0.01 & 0.37 $\pm$ 0.03 \\
    & GRU & 0.66 $\pm$ 0.03 & 0.32 $\pm$ 0.07 & 0.66 $\pm$ 0.01 & 0.32 $\pm$ 0.02 & 0.65 $\pm$ 0.02 & 0.29 $\pm$ 0.04 & 0.62 $\pm$ 0.03 & 0.23 $\pm$ 0.05 & 0.58 $\pm$ 0.03 & 0.17 $\pm$ 0.07 \\
    & ST & 0.61 $\pm$ 0.03 & 0.22 $\pm$ 0.06 & 0.59 $\pm$ 0.02 & 0.17 $\pm$ 0.04 & 0.58 $\pm$ 0.02 & 0.16 $\pm$ 0.04 & 0.62 $\pm$ 0.02 & 0.24 $\pm$ 0.04 & 0.63 $\pm$ 0.02 & 0.26 $\pm$ 0.04 \\ \cmidrule[0.75pt]{1-12}
    \multirow{6}{*}[-0.35em]{\makecell{MUTAG}} & Sum & 0.65 $\pm$ 0.18 & 0.30 $\pm$ 0.30 & 0.66 $\pm$ 0.17 & 0.33 $\pm$ 0.28 & 0.67 $\pm$ 0.16 & 0.36 $\pm$ 0.26 & 0.90 $\pm$ 0.08 & 0.75 $\pm$ 0.17 & 0.84 $\pm$ 0.13 & 0.67 $\pm$ 0.23 \\
    & Mean & 0.68 $\pm$ 0.19 & 0.32 $\pm$ 0.30 & 0.69 $\pm$ 0.19 & 0.35 $\pm$ 0.32 & 0.70 $\pm$ 0.20 & 0.37 $\pm$ 0.34 & 0.90 $\pm$ 0.07 & 0.78 $\pm$ 0.14 & 0.79 $\pm$ 0.12 & 0.53 $\pm$ 0.19 \\
    & Max & 0.71 $\pm$ 0.23 & 0.38 $\pm$ 0.42 & 0.72 $\pm$ 0.24 & 0.40 $\pm$ 0.43 & 0.65 $\pm$ 0.22 & 0.29 $\pm$ 0.40 & 0.91 $\pm$ 0.07 & 0.78 $\pm$ 0.14 & 0.90 $\pm$ 0.06 & 0.75 $\pm$ 0.18 \\ \cmidrule(l){2-12}
    & MLP & 0.91 $\pm$ 0.12 & 0.84 $\pm$ 0.24 & 0.90 $\pm$ 0.15 & 0.80 $\pm$ 0.30 & 0.84 $\pm$ 0.07 & 0.67 $\pm$ 0.14 & 0.87 $\pm$ 0.13 & 0.71 $\pm$ 0.26 & 0.89 $\pm$ 0.11 & 0.75 $\pm$ 0.22 \\
    & GRU & 0.81 $\pm$ 0.19 & 0.55 $\pm$ 0.32 & 0.78 $\pm$ 0.17 & 0.51 $\pm$ 0.29 & 0.87 $\pm$ 0.06 & 0.71 $\pm$ 0.09 & 0.84 $\pm$ 0.14 & 0.65 $\pm$ 0.26 & 0.88 $\pm$ 0.07 & 0.76 $\pm$ 0.10 \\
    & ST & 0.85 $\pm$ 0.08 & 0.68 $\pm$ 0.17 & 0.85 $\pm$ 0.11 & 0.69 $\pm$ 0.18 & 0.83 $\pm$ 0.12 & 0.69 $\pm$ 0.15 & 0.89 $\pm$ 0.09 & 0.72 $\pm$ 0.19 & 0.83 $\pm$ 0.09 & 0.67 $\pm$ 0.13 \\ \cmidrule[0.75pt]{1-12}
    \multirow{6}{*}[-0.35em]{\makecell{Mutagen\\icity}} & Sum & 0.79 $\pm$ 0.02 & 0.59 $\pm$ 0.03 & 0.76 $\pm$ 0.01 & 0.53 $\pm$ 0.02 & 0.79 $\pm$ 0.01 & 0.57 $\pm$ 0.02 & 0.81 $\pm$ 0.02 & 0.63 $\pm$ 0.04 & 0.82 $\pm$ 0.02 & 0.64 $\pm$ 0.05 \\
    & Mean & 0.78 $\pm$ 0.02 & 0.56 $\pm$ 0.04 & 0.77 $\pm$ 0.01 & 0.55 $\pm$ 0.03 & 0.75 $\pm$ 0.01 & 0.52 $\pm$ 0.03 & 0.81 $\pm$ 0.01 & 0.61 $\pm$ 0.03 & 0.80 $\pm$ 0.03 & 0.60 $\pm$ 0.06 \\
    & Max & 0.79 $\pm$ 0.02 & 0.57 $\pm$ 0.03 & 0.74 $\pm$ 0.01 & 0.50 $\pm$ 0.02 & 0.71 $\pm$ 0.01 & 0.44 $\pm$ 0.03 & 0.83 $\pm$ 0.01 & 0.65 $\pm$ 0.02 & 0.81 $\pm$ 0.02 & 0.62 $\pm$ 0.04 \\ \cmidrule(l){2-12}
    & MLP & 0.73 $\pm$ 0.02 & 0.48 $\pm$ 0.03 & 0.77 $\pm$ 0.02 & 0.53 $\pm$ 0.03 & 0.76 $\pm$ 0.01 & 0.52 $\pm$ 0.02 & 0.79 $\pm$ 0.01 & 0.57 $\pm$ 0.02 & 0.78 $\pm$ 0.01 & 0.56 $\pm$ 0.02 \\
    & GRU & 0.73 $\pm$ 0.01 & 0.47 $\pm$ 0.02 & 0.75 $\pm$ 0.01 & 0.49 $\pm$ 0.02 & 0.74 $\pm$ 0.01 & 0.48 $\pm$ 0.03 & 0.81 $\pm$ 0.02 & 0.61 $\pm$ 0.04 & 0.73 $\pm$ 0.02 & 0.46 $\pm$ 0.04 \\
    & ST & 0.80 $\pm$ 0.02 & 0.59 $\pm$ 0.03 & 0.75 $\pm$ 0.04 & 0.52 $\pm$ 0.06 & 0.78 $\pm$ 0.03 & 0.57 $\pm$ 0.05 & 0.81 $\pm$ 0.01 & 0.63 $\pm$ 0.02 & 0.82 $\pm$ 0.01 & 0.64 $\pm$ 0.02 \\ \cmidrule[0.75pt]{1-12}
    \multirow{6}{*}[-0.35em]{\makecell{YeastH}} & Sum & 0.54 $\pm$ 0.01 & 0.20 $\pm$ 0.02 & 0.54 $\pm$ 0.01 & 0.20 $\pm$ 0.02 & 0.54 $\pm$ 0.00 & 0.21 $\pm$ 0.02 & 0.57 $\pm$ 0.01 & 0.27 $\pm$ 0.03 & 0.58 $\pm$ 0.01 & 0.29 $\pm$ 0.02 \\
    & Mean & 0.52 $\pm$ 0.00 & 0.16 $\pm$ 0.01 & 0.52 $\pm$ 0.00 & 0.17 $\pm$ 0.01 & 0.53 $\pm$ 0.00 & 0.17 $\pm$ 0.01 & 0.55 $\pm$ 0.01 & 0.24 $\pm$ 0.02 & 0.56 $\pm$ 0.01 & 0.25 $\pm$ 0.01 \\
    & Max & 0.53 $\pm$ 0.00 & 0.17 $\pm$ 0.01 & 0.51 $\pm$ 0.00 & 0.14 $\pm$ 0.02 & 0.52 $\pm$ 0.00 & 0.16 $\pm$ 0.01 & 0.57 $\pm$ 0.01 & 0.26 $\pm$ 0.03 & 0.57 $\pm$ 0.01 & 0.25 $\pm$ 0.03 \\ \cmidrule(l){2-12}
    & MLP & 0.60 $\pm$ 0.00 & 0.24 $\pm$ 0.01 & 0.60 $\pm$ 0.01 & 0.24 $\pm$ 0.01 & 0.60 $\pm$ 0.01 & 0.25 $\pm$ 0.02 & 0.61 $\pm$ 0.01 & 0.26 $\pm$ 0.02 & 0.60 $\pm$ 0.01 & 0.24 $\pm$ 0.02 \\
    & GRU & 0.54 $\pm$ 0.02 & 0.17 $\pm$ 0.02 & 0.55 $\pm$ 0.01 & 0.19 $\pm$ 0.01 & 0.54 $\pm$ 0.01 & 0.19 $\pm$ 0.02 & 0.58 $\pm$ 0.02 & 0.26 $\pm$ 0.02 & 0.55 $\pm$ 0.01 & 0.17 $\pm$ 0.01 \\
    & ST & 0.62 $\pm$ 0.00 & 0.31 $\pm$ 0.01 & 0.59 $\pm$ 0.02 & 0.30 $\pm$ 0.01 & 0.60 $\pm$ 0.01 & 0.30 $\pm$ 0.02 & 0.64 $\pm$ 0.01 & 0.34 $\pm$ 0.02 & 0.64 $\pm$ 0.01 & 0.35 $\pm$ 0.02 \\ \bottomrule
    \end{tabular}
    \hspace*{-3cm}
\end{table}
\renewcommand{\familydefault}{\rmdefault}

\renewcommand{\sfdefault}{lmss}
\renewcommand{\familydefault}{\sfdefault}
\begin{table}[!htp]
    \captionsetup{justification=centering, skip=4pt, labelfont=bf, width=1.2\linewidth}
    \centering
    \scriptsize
    \caption{Detailed metrics (mean $\pm$ standard deviation) for the alchemy\_full dataset.}
    \label{table:alchemy}
    \hspace*{-3cm}
    \begin{tabular}{@{}clcc@{\hskip 0.5cm}cc@{\hskip 0.5cm}cc@{\hskip 0.5cm}cc@{\hskip 0.5cm}cc@{}}
        \toprule
        &            & \multicolumn{2}{c@{\hspace{0.5cm}}}{\textbf{GCN}} & \multicolumn{2}{c@{\hspace{0.75cm}}}{\textbf{GAT}} & \multicolumn{2}{c@{\hspace{0.75cm}}}{\textbf{GATv2}} & \multicolumn{2}{c@{\hspace{0.75cm}}}{\textbf{GIN}} & \multicolumn{2}{c@{\hspace{0.25cm}}}{\textbf{PNA}} \\ \cmidrule(lr{0.5cm}){3-4} \cmidrule(r{0.5cm}){5-6} \cmidrule(r{0.5cm}){7-8} \cmidrule(r{0.5cm}){9-10} \cmidrule{11-12}
        Dataset              & Agg. & \textbf{MAE}             & \textbf{\textbf{R}$\bm{^\mathsf{2}}$}              & \textbf{MAE}             & \textbf{\textbf{R}$\bm{^\mathsf{2}}$}              & \textbf{MAE}              & \textbf{\textbf{R}$\bm{^\mathsf{2}}$}               & \textbf{MAE}             & \textbf{\textbf{R}$\bm{^\mathsf{2}}$}              & \textbf{MAE}             & \textbf{\textbf{R}$\bm{^\mathsf{2}}$}              \\ \cmidrule(r){1-1} \cmidrule(lr){2-2} \cmidrule(lr){3-3} \cmidrule(lr{0.5cm}){4-4} \cmidrule(r){5-5} \cmidrule(lr{0.5cm}){6-6} \cmidrule(r){7-7} \cmidrule(lr{0.5cm}){8-8} \cmidrule(r){9-9} \cmidrule(lr{0.5cm}){10-10} \cmidrule(r){11-11} \cmidrule(l){12-12}
    \multirow{6}{*}[-0.35em]{\makecell{alchemy\\\_full}} & Sum & 17.78 $\pm$ 0.48 & 0.99 $\pm$ 0.00 & 46.80 $\pm$ 2.39 & 0.98 $\pm$ 0.00 & 46.00 $\pm$ 4.02 & 0.98 $\pm$ 0.00 & 17.49 $\pm$ 1.04 & 0.99 $\pm$ 0.00 & 10.30 $\pm$ 0.42 & 1.00 $\pm$ 0.00 \\
    & Mean & 28.91 $\pm$ 0.74 & 0.97 $\pm$ 0.00 & 56.22 $\pm$ 2.65 & 0.97 $\pm$ 0.00 & 49.94 $\pm$ 1.14 & 0.97 $\pm$ 0.00 & 41.46 $\pm$ 4.19 & 0.97 $\pm$ 0.00 & 16.71 $\pm$ 0.50 & 0.99 $\pm$ 0.00 \\
    & Max & 18.90 $\pm$ 0.24 & 0.99 $\pm$ 0.00 & 78.62 $\pm$ 9.31 & 0.99 $\pm$ 0.01 & 67.37 $\pm$ 3.42 & 0.99 $\pm$ 0.00 & 31.51 $\pm$ 9.16 & 0.99 $\pm$ 0.00 & 13.80 $\pm$ 0.99 & 1.00 $\pm$ 0.00 \\ \cmidrule(l){2-12}
    & MLP & 12.75 $\pm$ 0.74 & 1.00 $\pm$ 0.00 & 15.87 $\pm$ 1.33 & 1.00 $\pm$ 0.00 & 17.76 $\pm$ 1.88 & 1.00 $\pm$ 0.00 & 20.24 $\pm$ 7.66 & 0.99 $\pm$ 0.00 & 10.14 $\pm$ 1.75 & 1.00 $\pm$ 0.00 \\
    & GRU & 9.25 $\pm$ 0.58 & 1.00 $\pm$ 0.00 & 16.16 $\pm$ 1.07 & 1.00 $\pm$ 0.00 & 15.23 $\pm$ 1.54 & 1.00 $\pm$ 0.00 & 13.35 $\pm$ 0.61 & 0.99 $\pm$ 0.00 & 7.24 $\pm$ 0.40 & 1.00 $\pm$ 0.00 \\
    & ST & 9.58 $\pm$ 0.38 & 1.00 $\pm$ 0.00 & 19.52 $\pm$ 8.84 & 0.99 $\pm$ 0.01 & 15.11 $\pm$ 1.27 & 1.00 $\pm$ 0.00 & 13.83 $\pm$ 1.62 & 1.00 $\pm$ 0.00 & 10.25 $\pm$ 1.19 & 1.00 $\pm$ 0.00 \\ \bottomrule
    \end{tabular}
    \hspace*{-3cm}
\end{table}
\renewcommand{\familydefault}{\rmdefault}

\end{document}